%% file: main.tex
\documentclass{article}

\usepackage{microtype}            
\usepackage{graphicx}
\usepackage{subfigure}
\usepackage{booktabs} % for professional tables
\usepackage{longtable}
\usepackage{lmodern}
\usepackage{rotating}
\usepackage{tabularx}
\usepackage{xltabular}

\usepackage{hyperref}

\usepackage{titletoc}

\usepackage[accepted]{icml2024}

\usepackage{amsmath}
\usepackage{amssymb}
\usepackage{mathtools}
\usepackage{multirow}
\usepackage{longtable}

\usepackage[capitalize,noabbrev]{cleveref}

\usepackage[textsize=tiny]{todonotes}

\icmltitlerunning{Technical Report}

\begin{document}

\twocolumn[
\icmltitle{MMT-Bench: A Comprehensive Multimodal Benchmark for Evaluating Large Vision-Language Models Towards Multitask AGI}



\icmlsetsymbol{equal}{*}
\icmlsetsymbol{corr}{$\dagger$}
% Kaining Ying, Fanqing Meng, Jin Wang, Zhiqian Li, Han Lin, Yue Yang, Hao Zhang, Wenbo Zhang, Yuqi Lin, Shuo Liu, jiayi lei, Quanfeng Lu
% ~Quanfeng_Lu1
% , Peng Gao, Runjian Chen, Peng Xu, Renrui Zhang, Haozhe Zhang, Yali Wang, Yu Qiao, Ping Luo, Kaipeng Zhang, Wenqi Shao
\begin{icmlauthorlist}
\icmlauthor{Kaining Ying}{equal,pjlab}
\icmlauthor{Fanqing Meng}{equal,sjtu,pjlab}
\icmlauthor{Jin Wang}{equal,hku}
\icmlauthor{Zhiqian Li}{pjlab,hku}
\icmlauthor{Han Lin}{sjtu,pjlab}
\icmlauthor{Yue Yang}{sjtu,pjlab}
\icmlauthor{Hao Zhang}{pjlab}
\icmlauthor{Wenbo Zhang}{uoa}
\icmlauthor{Yuqi Lin}{pjlab,zju}
\icmlauthor{Shuo Liu}{pjlab}
\icmlauthor{Jiayi Lei}{pjlab,sjtu}
\icmlauthor{Quanfeng Lu}{pjlab}
\icmlauthor{Runjian Chen}{pjlab,hku}
\icmlauthor{Peng Xu}{pjlab,hku}
\icmlauthor{Renrui Zhang}{pjlab}
\icmlauthor{Haozhe Zhang}{zju}
\icmlauthor{Peng Gao}{pjlab}
\icmlauthor{Yali Wang}{cas}
\icmlauthor{Yu Qiao}{pjlab}
\icmlauthor{Ping Luo}{hku,pjlab}
\icmlauthor{Kaipeng Zhang}{corr,pjlab}
\icmlauthor{Wenqi Shao}{corr,pjlab}

\end{icmlauthorlist}

\icmlaffiliation{pjlab}{Shanghai Artificial Intelligence Laboratory}
\icmlaffiliation{hku}{The University of Hong Kong}
\icmlaffiliation{sjtu}{Shanghai Jiao Tong University}
\icmlaffiliation{cas}{Shenzhen Institutes of Advanced Technology, Chinese Academy of Sciences}
\icmlaffiliation{zju}{Zhejiang University}
\icmlaffiliation{uoa}{The University of Adelaide}
%\icmlaffiliation{xjtu}{Xi'an Jiaotong University}

\icmlcorrespondingauthor{Wenqi Shao}{shaowenqi@pjlab.org.cn}
\icmlcorrespondingauthor{Kaipeng Zhang}{zhangkaipeng@pjlab.org.cn}

\icmlkeywords{Machine Learning, ICML}

\vskip 0.3in
]



\printAffiliationsAndNotice{\icmlEqualContribution} % otherwise use the standard text.

\begin{abstract}
Large Vision-Language Models (LVLMs) show significant strides in general-purpose multimodal applications such as visual dialogue and embodied navigation. However, existing multimodal evaluation benchmarks cover a limited number of multimodal tasks testing rudimentary capabilities, falling short in tracking LVLM development. In this study, we present MMT-Bench, a comprehensive benchmark designed to assess LVLMs across massive multimodal tasks requiring expert knowledge and deliberate visual recognition, localization, reasoning, and planning. MMT-Bench comprises $31,325$ meticulously curated multi-choice visual questions from various multimodal scenarios such as vehicle driving and embodied navigation, covering $32$ core meta-tasks and $162$ subtasks in multimodal understanding. Due to its extensive task coverage, MMT-Bench enables the evaluation of LVLMs using a task map, facilitating the discovery of in- and out-of-domain tasks. Evaluation results involving $30$ LVLMs such as the proprietary GPT-4V, GeminiProVision, and open-sourced InternVL-Chat, underscore the significant challenges posed by MMT-Bench. We anticipate that MMT-Bench will inspire the community to develop next-generation multimodal foundation models aimed at achieving general-purpose multimodal intelligence.

\end{abstract}

\section{Introduction}
\label{sec:intro}

In recent years, Large Vision-Language Models (LVLMs) \cite{internlmxcomposer, gpt4v, liu2023llava} have emerged as powerful tools for advancing artificial intelligence, demonstrating remarkable progress in various domains such as visual dialogue, video analysis and document understanding. Driven by diverse and high-quality instruction fine-tuning data mined from various fields, LVLMs will continue to advance towards multitask AGI \cite{gemini, Qwen-VL}. As pointed out in Levels of AGI \cite{{morris2023levels}}, the breadth (generality) of tasks is a fundamental criterion for different levels of AGI. A multitask AGI model can perform a wide range of tasks across different domains with human-like proficiency, which could revolutionize many fields such as personalized education \cite{latif2023agi} and medical diagnosis \cite{singhal2023large}. Therefore, it is crucial to build a comprehensive evaluation benchmark to track multitask AGI development.

However, evaluating LVLMs significantly lags behind their development \cite{morris2023levels, yue2023mmmu, liu2024convbench}. A line of work attempts to bridge this gap by proposing various multimodal evaluation benchmarks. Examples include LVLM-eHub \cite{xu2023lvlm}, MMBench \cite{liu2023mmbench}, MME \cite{fu2023mme}, and SEED-Bench \cite{li2023seed}, which propose dimensions of multimodal capabilities and corresponding test samples. However, these benchmarks have limited coverage of multimodal tasks while testing rudimentary capabilities like visual recognition and text-scarce OCR. Therefore, they cannot fulfil the requirement of the breadth of tasks \cite{morris2023levels}. Moreover, recent LVLMs continue to excel in these benchmarks. For instance, InternLM-XComposer2 \cite{internlmxcomposer2} achieved $2242.7$/$2800$ and $79.6$/$100$ overall performance on MME and MMBench, respectively. Other works, such as MathVista \cite{lu2023mathvista} and MMMU \cite{mmmu}, focus on discipline knowledge understanding and reasoning but are constrained to visual questions with scientific diagram images, limiting their breadth for benchmarking multitask AGI.

  %For example, LVLM-eHub \cite{xu2023lvlm}, MMBench \cite{liu2023mmbench}, MME \cite{fu2023mme}, and SEED-Bench \cite{li2023seed} propose several dimensions of multimodal capabilities and then collect corresponding test samples. Although these datasets can measure the performance of LVLMs in some specific domains, they cover limited multimodal tasks and thus cannot fulfil the requirement of the breadth of tasks \cite{morris2023levels}. Moreover, these benchmarks only examine rudimentary multimodal capabilities such as visual recognition and text-scarce OCR. The latest LVLMs have consistently excelled in existing evaluation benchmarks. For example, InternLM-XComposer2 \cite{internlmxcomposer2} has achieved $2242.7$/$2800$ and $79.6$/$100$ overall performance on MME and MMBench, respectively. Recent work such as MathVista \cite{lu2023mathvista} and MMMU \cite{mmmu} measure the performance of LVLMs in discipline knowledge understanding and reasoning. However, they only include visual questions with the image type of scientific diagram and examine expert discipline knowledge. Hence, these datasets still fall short in breadth for benchmarking multitask AGI.

\begin{figure*}
    \centering
    \includegraphics[width=0.9\linewidth]{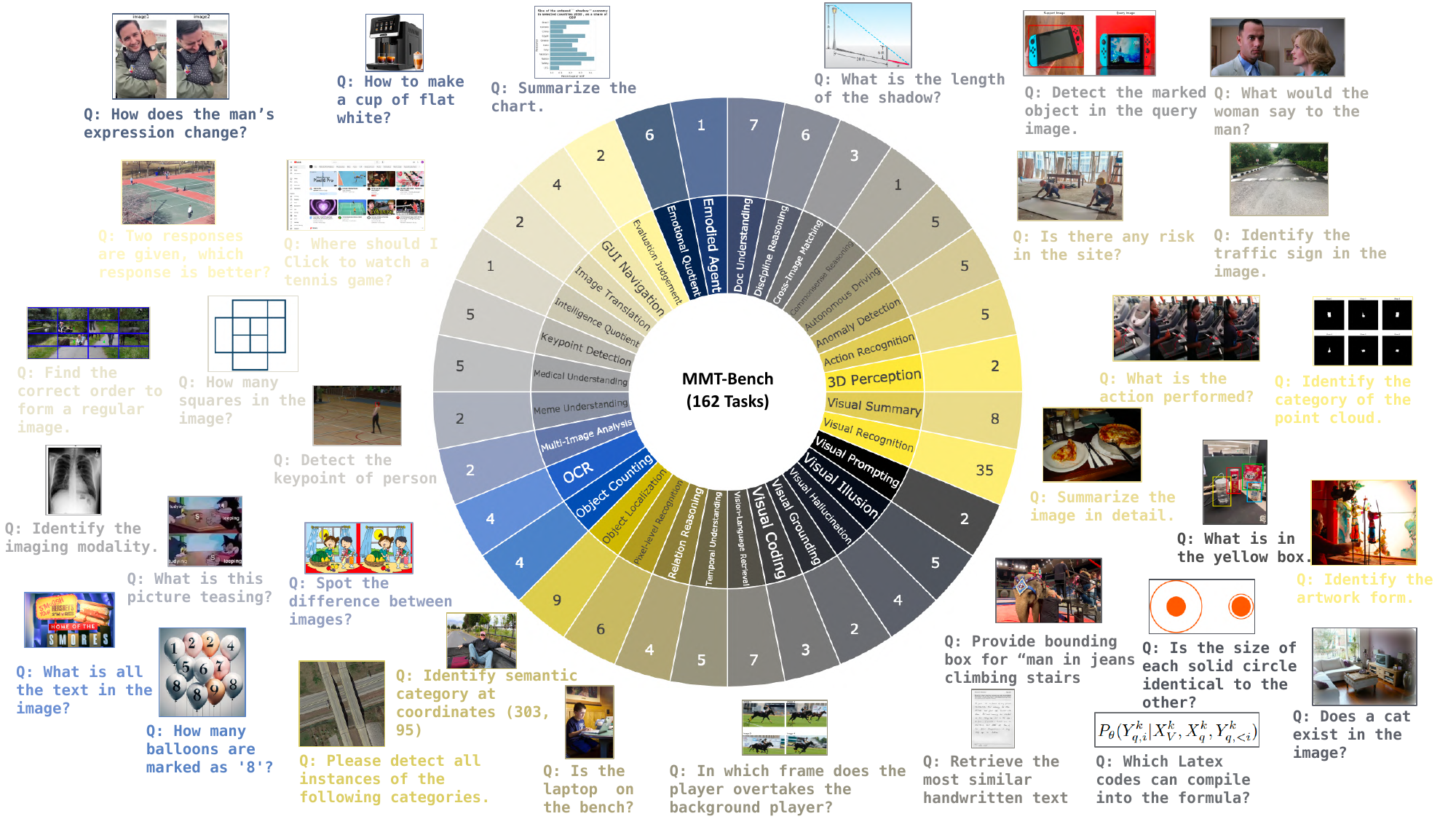}
    \vspace{-0.2in}
    \caption{Visualization of MMT-Bench. Our MMT-Bench consists of $32$ meta-tasks (middle ring) which are decomposed into $162$ subtasks (outer ring). For each meta-task, we denote the number of subtasks in it and illustrate one example of the pair of the image and the question (see task hierarchy in Table~\ref{tab:task_1} to  Table~\ref{tab:task_3} of Appendix). MMT-Bench can be comprehensive enough to evaluate the multitask performance of LVLMs.}
    \label{fig:mmt-bench}
\end{figure*}

\begin{figure*}[t!]
    \centering
    \includegraphics[width=0.9\linewidth]{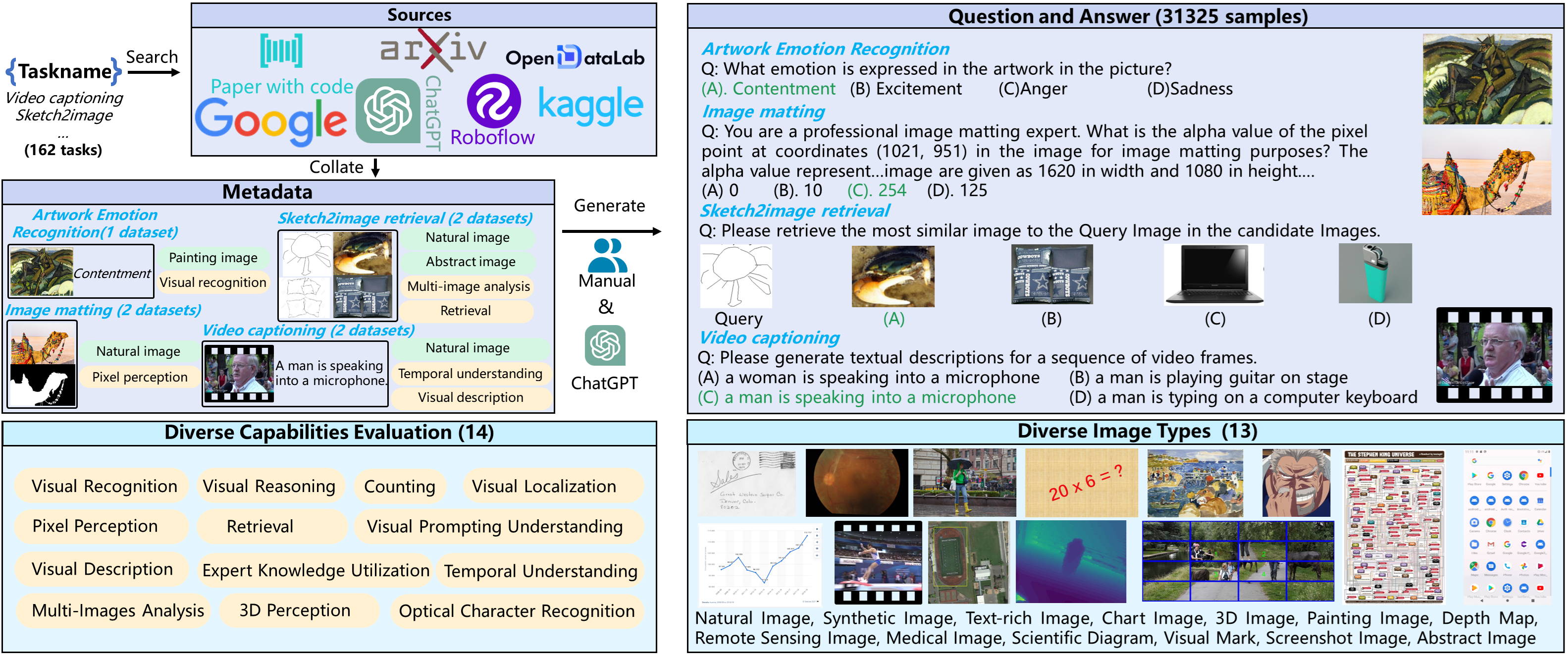}
    \caption{An illustration of our pipeline for data collection. First, given a task name, we retrieve its related datasets from the internet. Then we collate them in a uniform data format - metadata. Finally, we generate questions with choices and answers from metadata using manually designed rules or ChatGPT. Our benchmarks cover capabilities evaluation with diverse image types.}
    \label{fig:collect}
\end{figure*}

To address this challenge, we introduce MMT-Bench, a new benchmark designed to comprehensively assess LVLMs in multimodal multitask understanding. The \textbf{breadth} of MMT-Bench features in three aspects. First, MMT-Bench is meticulously curated and comprises $32$K multi-choice visual questions covering $32$ core meta-tasks and a total of $162$ subtasks (Fig.~\ref{fig:mmt-bench}), which is $8.1$ times larger than MM-Bench \cite{liu2023mmbench}. 
Second, it encompasses $13$ image types such as natural scenes, synthetic images, depth maps, text-rich images, paintings, screenshots, point clouds, medical images, et al. (Fig.~\ref{fig:collect}). Such diversity demands the model to be capable enough to interpret various visual inputs.
Third, MMT-Bench spans multimodal scenarios such as vehicle driving, GUI navigation, and embodied AI, testing $14$ kinds of multimodal capabilities including visual recognition, localization, reasoning, OCR, counting, 3D perception, temporal understanding, et al. (Fig.~\ref{fig:collect}).

We assess $30$ publicly available LVLMs under various input modes for best evaluation performance. Our findings highlight the significant challenges posed by MMT-Bench. For instance, GPT-4V only achieves $62.0$/$100$ and $55.6$/$100$ overall scores across all subtasks and subtasks except for visual recognition tasks, respectively, indicating significant room for improvement towards multitask AGI. Thanks to the extensive coverage of multimodal tasks, MMT-Bench enables the evaluation of LVLMs using a task map. This facilitates the discovery of both in- and out-of-domain tasks, providing valuable insights for multimodal commercial applications and ongoing efforts to enhance LVLMs. We summarize the findings as follows:
\begin{table*}[t!]
    \centering
    \small
    \caption{The comparison between MMT-Bench and existing evaluation benchmarks. MMT-Bench consists of massive samples and multimodal tasks compared with other benchmarks. I, T, V, and P respectively represent image, text, video, and point cloud. }
    %both automatic and GPT evaluation for quantitative assessment and human evaluation in the arena platform which features anonymous randomized pairwise battles.   }
    \label{tab:comparison-LVLM-bench}
    \scalebox{1.0}{%
        \begin{tabular}{c|cccc cc}
            \toprule
            \multirow{2}{*}{Benchmark} &\multicolumn{6}{c}{Data Collection} \\
             \cmidrule{2-7}
            & \# Sample & \# Meta-task & \# Task & \# Modality & Source & Answer Type\\
            \cmidrule{1-1}\cmidrule{2-7}
            % VisWiz \cite{vizwiz} & 32K & 1 & 1 & Annotated & Open & N & N & Y & N\\
            % TextVQA \cite{textvqa} & 45K & 1 & 1 & Annotated & Multi-Choice & N & N & Y & N\\
            SEED-Bench \cite{li2023seed} & 19K &  12 & 12 & I + T + V & Annotated & Multi-Choice \\
            MMBench \cite{liu2023mmbench} & 3K & 2 & 20& I + T  & Repurposed & Multi-Choice \\
            MM-VET \cite{yu2023mm} & 0.2K &  6 & N/A & I + T & Repurposed & Multi-Choice  \\
            MMMU \cite{yue2023mmmu} & 11.5K &  6 & 30 & I + T & Annotated & Multi-Choice/Open  \\
            Tiny LVLM-eHub \cite{shao2023tiny} & 2.1K &  5 & 42& I + T  & Repurposed & Multi-Choice/Open \\
            \cmidrule{1-1}\cmidrule{2-7}
            MMT-Bench & 31K &  32 & 162 & I + T + V + P & Repurposed & Multi-Choice \\
            \bottomrule
        \end{tabular}%
    }
\end{table*}
\begin{itemize}
    \item The open-source model InternVL-chat has taken a leading position in MMT-Bench, surpassing other closed-source models such as QWen-VL-Plus, GPT-4V, and GeminiProVision.
    \item The comprehensive error analyses conducted on $162$ multimodal tasks reveal that top-performing LVLMs such as InternVL-chat, GPT4V, and GeminiProVision are predominantly prone to perception, reasoning, and knowledge errors. 
    %These findings provide valuable insights in guiding future directions for improvement.
    \item The taxonomy analysis shows that current LVLMs perform well in tasks related to visual recognition and description which are in-domain tasks, yet fall short in tasks related to localization and pixel perception which are out-of-domain tasks.
    \item BLIP2 that does not undergo instruction tuning even outperforms most LVLMs that are tuned by millions of instruction-following data, implying that instruction-tuning with data in some tasks even hurts the generalization on other tasks.
    \item Certain tasks show improved performance with specific prompting methods, such as multi-image and coordinate-related tasks, as well as those involving visual referring prompts. However, most models do not exhibit improved performance with visual prompting, suggesting potential areas for future enhancement.
    \item Model performance significantly improves with an
increase in size (7B to 13B) for both LLaVA-v1.5 and LLaVA-
v1.5-Xtuner. Upgrading LLMs, from InternLM to InternLM2,
also enhances the performance of LLaVA.
\end{itemize}

Overall, the contributions of this work are three-fold. i) We build a new evaluation benchmark called MMT-Bench for multimodal multitask comprehension, allowing us to measure the progress on the path to multitask AGI. ii) We evaluate various publicly available LVLMs on MMT-Bench, revealing that current LVLMs including InternVL-Chat, GPT-4V, and GeminiProVision achieve plain performance in multitask intelligence. iii) We present a taskonomy analysis by evaluating LVLMs on a task map built upon MMT-Bench, facilitating the discovery of both in- and out-of-domain tasks relative to current LVLMs. We anticipate that MMT-Bench will inspire the community to push the boundaries of LVLM research and development, driving us closer to the realization of truly intelligent multimodal systems. The MMT-Bench is open-sourced at \url{https://github.com/OpenGVLab/MMT-Bench}.

\section{Related Work}

\textbf{LVLM.} As the Large Language Models (LLMs) continue to garner impressive achievements \cite{Qwen-VL, 2023internlm, llama1,llama2, vicuna, flant5}, academic emphasis is increasingly shifting towards the exploration and development of Large Visual Language Models (LVLMs), to bolster the multimodal understanding and generative capabilities of models. Some notable open-source LVLMs, such as mPLUG-Owl2 \cite{ye2023mplugowl2}, LLaVA \cite{liu2023llava}, and LLaMA-Adapter \cite{gao2023llamaadapterv2,zhang2023llamaadapter}, have adopted LLMs as their backbone, processing visual features through these LLMs, ultimately achieving an innovative integration of text and visuals. In addition, closed-source models like Gemini \cite{gemini} and GPT-4V \cite{yang2023dawn} have demonstrated remarkable results across numerous tasks, making groundbreaking contributions. 
We aim to undertake an in-depth and comprehensive exploration of LVLMs and their capabilities by testing them on massive multimodal tasks.

\textbf{LVLM Evaluation.}
Recently, LVLMs have demonstrated remarkable capabilities to handle many visual-language tasks, which makes previous single-task benchmarks \cite{antol2015vqa,hudson2019gqa,krishna2017visual,lin2014microsoft,marino2019ok} insufficient to provide comprehensive evaluations of current LVLMs.
To this end, current LVLM evaluation benchmarks aimed to provide relatively holistic evaluations for the overall reasoning capabilities of LVLMs, such as OwlEval \cite{ye2023mplug}, LVLM-eHub \cite{xu2023lvlm}, SEED-Bench \cite{li2023seed}, LAMM \cite{yin2023lamm}, MM-Vet \cite{yu2023mm} and MMBench \cite{liu2023mmbench}.
However, these benchmarks only covered a small range of multimodal tasks and vision-language skills, making them not comprehensive enough to asses multitask AGI capabilities.
Besides, recent studies also presented benchmarks of LVLMs which required expert-level domain knowledge, such as Mathvista \cite{lu2023mathvista} and MMMU \cite{mmmu}.
In comparison, our proposed MMT-Bench covers an extensive range of multimodal reasoning capabilities with sufficient test samples from various modalities as shown in Table~ \ref{tab:comparison-LVLM-bench}, which requires expert knowledge and deliberate visual
recognition, localization, reasoning, and planning. Our MMT-Bench poses significant challenges for the current state-of-the-art LVLMs.

\textbf{Multitask Analysis.}
% Present analyses of large multimodal models \cite{yu2023mm, liu2023mmbench, li2023seed}, like the LVLM-ehub \cite{xu2023lvlm}, focus on meta-task performance and limit identification, utilizing a pre-set task division. However, our study extraly offers a fresh perspective, employing a task map \cite{ilharco2022taskvector} derived from the tasks' inherent properties and a model performance map. This comprehensive method calls for a more holistic conclusion.
Characterizing various tasks and establishing inter-task relationships is an effective means for multitask analysis \cite{ilharco2022taskvector, achille2019task2vec, zamir2018taskonomy, wallace2021can}, with wide applications in areas such as meta-learning and transfer learning. A substantial amount of research has been conducted in Taskonomy \cite{zamir2018taskonomy}. It utilizes transfer learning to model the structure of the space of visual tasks, thereby harnessing the interconnections among visual tasks to avoid redundancy in learning. Task2Vec \cite{achille2019task2vec} extracts fisher information as task vectors, which is used in meta-learning. In our paper, thanks to the vast amount of task data collected, we evaluate LVLMs on a task map and conclude challenging tasks for the current LVLMs.

\section{MMT-Bench}\label{sec:mmtbench}
In this section, we describe how to build the task hierarchy in Sec.~\ref{sec:task} and the pipeline of data collection in Sec.~\ref{sec:data_collection}.
\subsection{Tasks}
\label{sec:task}
\textbf{Hierarchical Task Structure.} We utilize a hierarchical structure to include as more as multimodal tasks to build the MMT-Bench. First, all co-authors come up with meta-tasks for multimodal understanding by brainstorming. We then collect $32$ meta-tasks by deduplication and filtering for important tasks as depicted in Fig. \ref{fig:mmt-bench}. Second, we decompose each meta-task into several subtasks. The subtask is kept in the MMT-Bench by three criteria. i) Whether the subtask examines the basic multimodal capability. ii) Whether the subtask challenges the current LVLMs. iii) Whether the test sample for the subtask can be publicly accessible. After selection, MMT-Bench comprises $162$ sub-tasks, which is $3.8$ times larger than TinyLVLM-eHub which previously contained the most tasks \cite{shao2023tiny}. The detailed comparison between MMT-Bench and previous benchmarks is provided in Table~\ref{tab:comparison-LVLM-bench}. We also present the whole hierarchical structure in Table~\ref{tab:task_1} of the Appendix.

\subsection{Data Collection}
\label{sec:data_collection}
We design an efficient pipeline (see Fig. \ref{fig:collect}) to construct multi-choice visual questions evaluation data for each subtask and the data collection is completed by dozens of co-authors specializing in artificial intelligence.

\textbf{Datasets Search.}
We conduct comprehensive searches for related datasets using various sources such as Google, Paper With Code, Kaggle, and ChatGPT, based on the name of the subtask. After downloading the datasets, we meticulously assess their suitability for evaluating the subtask, ensuring usability and relevance. While most tasks have multiple datasets available, a few may only have one dataset publicly accessible.
% Each subtask includes more than one dataset to ensure data diversity.

\textbf{Metadata Construction.}
We define a uniform format, the metadata, to collate downloaded datasets. It enables the further generation of visual questions and answers. Each sample of metadata consists of images and meta-information. The meta-information (see Fig. \ref{fig:collect}) includes the necessary information to generate questions and answers for the evaluation and also includes manual annotations of required capabilities and the type of visual prompt (i.e., input image). For evaluation efficiency, in each task, we keep the maximum number of samples at 200 by random sampling, and each dataset comprises the same number of samples.

\textbf{Question and Answer Generation.}
For each subtask, we generate multi-choice (maximum eight choices depending on the task) visual questions with choices and answers from their metadata. Specifically, depending on a specific task, we manually design rules or use ChatGPT with well-designed prompts for efficient and high-quality generation. For example, in sketch2image retrieval, we use the corresponding image as a ground-truth answer and generate other choices by randomly sampling other images from metadata. In video captioning, we use ChatGPT to write confused wrong choices.

\textbf{Dataset Statistics.} MMT-Bench comprises $31,325$ meticulously curated multi-choice questions with $13$ input image types such as natural scenes, synthetic images, text-rich images, medical images, et al. (see Fig. \ref{fig:collect}), covering $32$ core meta-tasks and $162$ subtasks for multitask multimodal understanding.
Compared to previous LVLMs benchmarks \cite{mmmu,xu2023lvlm} addressing limited image types and skills, questions in MMT-Bench span diverse multimodal scenarios such as GUI navigation and document understanding, testing $14$ kinds of capabilities including visual recognition, localization, reasoning, OCR, counting, 3D perception, temporal understanding, et al., as shown in Fig. \ref{fig:collect}. These features ensure that MMT-Bench meets the requirement of task breadth for evaluating multitask AGI.

% \usepackage{graphicx}
\begin{table*}[t!]
\centering
\caption{
Quantitative results for 30 LVLMs across 32 meta-tasks are summarized, with $\overline{R}$ representing the average rank. Accuracy is the metric, and the Overall score is computed across all subtasks, excluding visual recognition (VR) as denoted by $*$. 
The maximum value of each meta-task is bolded. Meta-tasks are abbreviated for brevity, with full terms in Sec.~\ref{sec:abbrev} of the appendix.}
\label{tab:overall-results}
\resizebox{0.8\textwidth}{!}{%
\begin{tabular}{l|ll|llllllllllllllll}
\toprule
Model & Overall & $\overline{R}$ & VR & Loc & OCR & Count & HLN & IR & 3D & VC & VG & DU & AR & PLP & I2IT & RR & IQT & Emo \\
 & Overall$^*$ & $\overline{R^*}$ & VI & MemU & VPU & AND & KD & VCR & IEJ & MIA & CIM & TU & VP & MedU & AUD & DKR & EA & GN \\
  \midrule
Frequency Guess & 31.7 & 26.1 & 30.0 & 28.2 & 30.4 & 28.2 & 43.4 & 29.9 & 26.5 & 28.2 & 29.1 & 37.6 & 30.0 & 29.4 & 30.8 & 33.5 & 18.0 & 30.1 \\
 & 32.2 & 25.9 & 52.1 & 32.8 & 29.3 & 44.4 & 33.7 & 27.0 & 30.0 & 46.5 & 28.5 & 29.1 & 29.5 & 30.9 & 29.7 & 29.4 & 28.0 & 29.0 \\
 \midrule
Random Guess & 28.5 & 30.0 & 27.1 & 28.1 & 27.2 & 25.0 & 41.6 & 24.3 & 25.5 & 25.0 & 24.8 & 30.3 & 25.4 & 26.6 & 21.2 & 33.4 & 10.5 & 25.4 \\
 & 28.9 & 29.9 & 50.8 & 25.5 & 31.4 & 36.5 & 32.2 & 28.0 & 25.0 & 48.5 & 26.8 & 27.0 & 28.8 & 27.8 & 26.8 & 25.4 & 27.5 & 24.4 \\
 \midrule
\midrule
InternVL-Chat-v1.2-34B & \textbf{63.4} & \textbf{5.7} & 81.3 & 59.4 & 60.5 & \textbf{66.4} & \textbf{82.4} & 56.3 & 45.5 & 82.3 & 49.4 & 68.3 & 52.6 & 37.4 & 32.8 & 55.0 & 84.0 & 48.7 \\
 & \textbf{58.2} & \textbf{5.7} & \textbf{61.5} & 62.5 & 58.2 & 57.0 & 62.2 & 76.0 & 31.0 & \textbf{82.8} & 56.8 & 45.2 & 41.8 & 71.8 & 57.8 & 49.4 & 74.5 & 41.2 \\
 \midrule
Qwen-VL-Plus & 62.3 & 6.7 & 82.6 & 55.3 & 65.6 & 61.1 & 69.9 & 40.7 & 46.5 & \textbf{86.5} & 43.6 & \textbf{77.3} & 53.4 & 43.1 & \textbf{37.8} & 53.0 & \textbf{84.5} & 41.6 \\
 & 56.6 & 6.8 & 50.3 & 61.0 & \textbf{67.5} & 58.8 & 55.3 & 76.5 & 31.8 & 81.5 & 61.3 & \textbf{45.5} & 33.7 & 73.3 & 59.5 & 46.8 & \textbf{85.0} & 32.6 \\
 \midrule
GPT-4V & 62.0 & 8.3 & \textbf{85.3} & 55.6 & \textbf{68.0} & 51.6 & 69.6 & 44.9 & 42.0 & 80.3 & 25.0 & 69.8 & 47.7 & \textbf{48.2} & 31.8 & 52.5 & 80.0 & 45.1 \\
 & 55.5 & 8.6 & 47.9 & 61.0 & 60.2 & 51.4 & 53.6 & 73.0 & 43.4 & 70.2 & 55.2 & 44.6 & \textbf{53.3} & 74.0 & 55.6 & \textbf{53.4} & 80.9 & 39.7 \\
 \midrule
GeminiProVision & 61.6 & 8.3 & 84.7 & 43.6 & 59.5 & 56.4 & 65.9 & \textbf{68.4} & 45.2 & 80.1 & 33.0 & 71.6 & \textbf{57.4} & 40.3 & 31.5 & 58.5 & 11.0 & \textbf{55.2} \\
 & 55.1 & 8.5 & 47.5 & 75.8 & 50.9 & 47.4 & 49.5 & \textbf{86.5} & 35.0 & 70.2 & 33.3 & 40.5 & 46.0 & \textbf{82.6} & \textbf{59.5} & 49.2 & 74.5 & 33.4 \\
 \midrule
LLaVA-NEXT-34B & 60.8 & 7.5 & 76.7 & \textbf{61.0} & 64.1 & 66.3 & 70.1 & 38.8 & 48.5 & 85.9 & \textbf{56.2} & 69.1 & 50.6 & 41.9 & 22.8 & 54.9 & 76.5 & 50.3 \\
 & 56.3 & 7.5 & 57.8 & 55.5 & 57.2 & \textbf{61.2} & \textbf{62.7} & 75.0 & 22.2 & 77.8 & 43.0 & 45.4 & 40.2 & 61.9 & 55.1 & 48.1 & 80.0 & \textbf{41.4} \\
 \midrule
XComposer2 & 55.7 & 11.7 & 75.3 & 47.9 & 43.9 & 51.0 & 69.5 & 32.4 & 40.5 & 73.7 & 42.6 & 62.0 & 46.3 & 43.9 & 31.5 & 50.5 & 8.0 & 53.6 \\
 & 50.0 & 11.7 & 52.6 & 71.2 & 56.1 & 56.2 & 41.5 & 83.0 & \textbf{43.8} & 80.8 & 61.2 & 36.6 & 36.3 & 53.5 & 48.8 & 43.8 & 50.5 & 29.4 \\
 \midrule
BLIP2 & 54.8 & 12.8 & 75.1 & 54.1 & 48.1 & 29.8 & 66.1 & 27.4 & 47.8 & 78.7 & 33.5 & 43.0 & 51.1 & 46.1 & 28.2 & 53.0 & 14.0 & 43.1 \\
 & 49.1 & 12.8 & 55.6 & 76.2 & 39.8 & 43.7 & 60.2 & 77.0 & 29.8 & 62.8 & \textbf{73.0} & 42.7 & 43.2 & 60.1 & 44.6 & 37.0 & 80.5 & 33.4 \\
 \midrule
Yi-VL-34B & 54.2 & 14.3 & 74.6 & 47.0 & 58.0 & 59.4 & 65.8 & 28.8 & 38.8 & 74.0 & 41.5 & 56.4 & 40.4 & 38.4 & 19.5 & 51.7 & 68.5 & 39.7 \\
 & 48.6 & 14.3 & 51.3 & 56.2 & 61.2 & 52.4 & 49.5 & 71.5 & 25.5 & 66.0 & 48.0 & 39.2 & 32.0 & 59.6 & 48.2 & 44.3 & 57.0 & 32.4 \\
 \midrule
Monkey-Chat & 53.4 & 15.5 & 79.0 & 40.1 & 51.0 & 43.6 & 63.1 & 26.8 & 46.5 & 68.9 & 27.5 & 51.1 & 49.3 & 32.2 & 29.5 & \textbf{61.8} & 11.0 & 45.1 \\
 & 46.0 & 15.8 & 55.3 & 69.5 & 43.6 & 44.6 & 36.3 & 85.5 & 26.0 & 58.8 & 61.7 & 36.8 & 33.3 & 68.0 & 43.6 & 38.1 & 46.0 & 29.8 \\
 \midrule
DeepSeek-VL-7B & 53.2 & 15.0 & 75.6 & 42.0 & 61.1 & 44.5 & 60.6 & 30.5 & 47.2 & 69.1 & 38.4 & 51.9 & 44.8 & 38.3 & 23.5 & 48.8 & 37.0 & 43.8 \\
 & 46.5 & 15.2 & 47.7 & 59.8 & 53.5 & 45.4 & 41.0 & 41.0 & 38.8 & 35.0 & 67.2 & 33.1 & 30.7 & 69.7 & 48.8 & 36.4 & 67.5 & 36.8 \\
 \midrule
Yi-VL-6B & 53.2 & 14.7 & 73.5 & 49.4 & 53.1 & 56.2 & 63.9 & 26.0 & 43.5 & 63.4 & 42.1 & 55.2 & 43.8 & 35.3 & 26.8 & 48.8 & 47.0 & 46.1 \\
 & 47.5 & 14.5 & 55.8 & 54.5 & 49.2 & 53.0 & 51.8 & 65.5 & 34.2 & 52.0 & 43.3 & 37.6 & 37.0 & 60.6 & 46.9 & 40.2 & 48.0 & 34.8 \\
 \midrule
LLaVA-NEXT-13B & 53.0 & 15.0 & 74.0 & 35.6 & 51.8 & 59.2 & 63.6 & 32.7 & \textbf{50.0} & 75.0 & 44.6 & 53.6 & 46.5 & 34.0 & 26.2 & 50.0 & 50.0 & 44.5 \\
 & 46.8 & 14.9 & 57.5 & 55.0 & 32.2 & 49.6 & 38.9 & 47.0 & 18.0 & 36.5 & 59.8 & 38.9 & 22.5 & 55.8 & 55.7 & 38.5 & 70.0 & 41.0 \\
 \midrule
TransCore-M & 52.7 & 13.1 & 73.6 & 40.5 & 50.4 & 54.5 & 71.9 & 27.5 & 45.0 & 75.6 & 35.1 & 45.3 & 46.9 & 38.3 & 25.0 & 53.2 & 15.0 & 46.3 \\
 & 46.9 & 12.9 & 55.6 & \textbf{76.8} & 51.9 & 43.7 & 38.6 & 85.5 & 34.2 & 52.8 & 65.8 & 29.7 & 28.8 & 61.1 & 46.5 & 38.4 & 39.5 & 35.6 \\
 \midrule
QWen-VL-Chat & 52.5 & 16.0 & 77.5 & 33.7 & 46.9 & 46.7 & 63.9 & 27.5 & 45.0 & 73.0 & 26.5 & 51.5 & 50.9 & 32.7 & 30.5 & 57.4 & 13.5 & 45.4 \\
 & 45.4 & 16.3 & 50.9 & 74.2 & 42.4 & 40.2 & 35.9 & 86.0 & 30.0 & 49.2 & 58.3 & 37.3 & 30.8 & 67.1 & 45.4 & 35.6 & 55.0 & 30.2 \\
 \midrule
Claude3V-Haiku & 52.2 & 17.7 & 74.3 & 44.8 & 54.4 & 51.1 & 63.6 & 34.6 & 38.2 & 67.6 & 26.9 & 69.8 & 46.2 & 35.5 & 22.8 & 50.0 & 59.5 & 35.2 \\
 & 46.4 & 17.7 & 42.9 & 53.8 & 43.2 & 41.2 & 53.3 & 70.5 & 31.5 & 34.8 & 52.5 & 35.9 & 34.2 & 62.7 & 34.1 & 40.4 & 54.5 & 35.1 \\
 \midrule
XComposer & 52.1 & 17.1 & 75.4 & 40.4 & 44.1 & 39.9 & 66.5 & 49.7 & 47.0 & 72.1 & 27.2 & 36.6 & 47.9 & 39.6 & 24.5 & 50.2 & 14.0 & 45.9 \\
 & 45.6 & 17.3 & 53.4 & 63.8 & 40.6 & 43.4 & 42.3 & 78.0 & 29.0 & 66.2 & 52.3 & 33.1 & 28.3 & 55.6 & 40.8 & 39.3 & 38.5 & 34.2 \\
 \midrule
mPLUG-Owl2 & 52.0 & 17.3 & 76.5 & 45.8 & 44.5 & 47.6 & 63.4 & 27.6 & 45.2 & 66.6 & 33.0 & 42.4 & 45.2 & 41.6 & 25.5 & 52.0 & 18.0 & 42.0 \\
 & 45.0 & 17.5 & 58.5 & 59.0 & 40.1 & 49.4 & 32.9 & 85.5 & 30.0 & 55.0 & 57.7 & 31.9 & 27.3 & 63.4 & 45.5 & 38.1 & 35.0 & 27.8 \\
 \midrule
RBDash-v1-13B & 51.8 & 15.7 & 72.2 & 42.2 & 53.6 & 51.6 & 66.6 & 26.3 & 40.8 & 75.5 & 36.9 & 48.1 & 47.1 & 38.3 & 22.5 & 55.9 & 14.0 & 43.4 \\
 & 46.1 & 15.3 & 57.1 & 67.5 & 51.4 & 45.7 & 33.2 & 78.0 & 39.0 & 32.0 & 64.2 & 31.6 & 25.5 & 59.3 & 46.3 & 38.1 & 53.5 & 32.4 \\
 \midrule
LLaVA-v1.5-13B & 51.7 & 15.3 & 73.8 & 38.8 & 51.8 & 55.1 & 65.8 & 27.2 & 39.8 & 70.4 & 37.4 & 45.7 & 46.6 & 37.6 & 28.0 & 58.2 & 13.5 & 45.3 \\
 & 45.7 & 15.2 & 58.1 & 66.0 & 43.9 & 48.3 & 31.4 & 79.0 & 35.8 & 28.5 & 62.5 & 33.3 & 27.5 & 58.6 & 46.6 & 39.4 & 40.5 & 37.5 \\
 \midrule
CogVLM-Chat & 51.6 & 17.5 & 77.7 & 24.7 & 48.5 & 49.8 & 66.0 & 26.1 & 42.2 & 69.8 & 28.8 & 49.1 & 46.3 & 33.2 & 23.8 & 61.6 & 14.0 & 50.3 \\
 & 44.2 & 17.9 & 52.4 & 75.5 & 39.8 & 43.4 & 28.2 & 82.0 & 28.0 & 70.8 & 45.8 & 35.5 & 28.3 & 65.9 & 44.9 & 36.9 & 48.0 & 29.9 \\
 \midrule
ShareGPT4V-7B & 51.5 & 16.4 & 74.2 & 36.0 & 47.8 & 50.9 & 62.4 & 27.8 & 45.2 & 71.6 & 35.4 & 47.9 & 46.2 & 39.2 & 21.8 & 59.8 & 14.0 & 44.3 \\
 & 45.1 & 16.4 & 54.5 & 70.5 & 47.1 & 48.2 & 26.3 & 83.0 & 27.8 & 38.0 & 64.3 & 32.1 & 30.0 & 60.8 & 46.1 & 38.9 & 42.0 & 28.9 \\
 \midrule
LLaVA-NEXT-7B & 51.1 & 18.1 & 73.3 & 29.5 & 52.0 & 56.8 & 59.9 & 28.7 & 43.2 & 69.8 & 37.0 & 49.7 & 47.9 & 32.6 & 22.8 & 49.0 & 47.5 & 48.1 \\
 & 44.6 & 18.0 & 57.8 & 54.0 & 38.5 & 44.3 & 34.6 & 42.5 & 18.8 & 32.5 & 67.8 & 39.1 & 23.3 & 55.5 & 53.5 & 37.0 & 65.0 & 31.6 \\
 \midrule
LLaVA-v1.5-13B-XTuner & 51.1 & 16.8 & 72.5 & 40.7 & 46.8 & 54.1 & 66.5 & 26.4 & 47.5 & 68.8 & 35.6 & 47.0 & 44.2 & 38.3 & 26.0 & 52.4 & 14.0 & 51.0 \\
 & 45.1 & 16.5 & 54.4 & 66.5 & 47.9 & 52.0 & 28.8 & 82.0 & 39.2 & 37.0 & 56.8 & 28.3 & 28.3 & 49.1 & 44.4 & 37.3 & 33.5 & 40.9 \\
 \midrule
LLaVA-InternLM2-7B & 50.8 & 17.5 & 73.3 & 38.9 & 49.5 & 51.8 & 67.8 & 27.7 & 49.5 & 66.4 & 36.9 & 37.7 & 43.7 & 35.1 & 14.2 & 58.0 & 0.0 & 51.1 \\
 & 44.4 & 17.4 & 52.3 & 62.5 & 45.1 & 57.2 & 35.2 & 83.0 & 34.2 & 55.8 & 58.2 & 26.8 & 18.5 & 57.8 & 45.1 & 33.7 & 35.5 & 35.2 \\
 \midrule
LLaVA-v1.5-7B-XTuner & 50.2 & 19.5 & 72.5 & 41.1 & 46.0 & 49.9 & 62.1 & 26.0 & 45.5 & 66.4 & 35.3 & 42.8 & 45.8 & 42.5 & 25.5 & 53.9 & 11.5 & 44.2 \\
 & 43.9 & 19.3 & 60.1 & 56.5 & 42.6 & 47.2 & 28.4 & 80.5 & 32.2 & 41.2 & 63.2 & 29.9 & 24.2 & 52.5 & 43.4 & 37.2 & 32.0 & 30.5 \\
 \midrule
SharedCaptioner & 49.9 & 19.6 & 72.8 & 41.8 & 47.8 & 46.2 & 63.1 & 27.0 & 44.2 & 61.9 & 27.0 & 39.5 & 46.7 & 33.5 & 25.0 & 59.5 & 14.5 & 39.9 \\
 & 43.2 & 19.5 & 55.1 & 53.8 & 45.4 & 38.3 & 33.6 & 82.5 & 20.2 & 57.8 & 56.8 & 32.6 & 28.7 & 59.4 & 44.7 & 38.4 & 45.0 & 29.6 \\
 \midrule
LLaVA-InternLM-7B & 49.7 & 19.6 & 70.1 & 38.7 & 47.6 & 46.0 & 62.0 & 25.5 & 42.0 & 65.0 & 26.5 & 43.9 & 45.6 & 38.3 & 25.0 & 52.4 & 14.0 & 47.0 \\
 & 43.9 & 19.3 & 57.5 & 58.2 & 45.6 & 46.5 & 33.2 & 75.5 & 33.0 & 57.0 & 59.7 & 28.0 & 27.3 & 52.0 & 42.2 & 38.1 & 46.5 & 37.6 \\
 \midrule
LLaVA-v1.5-7B & 49.5 & 20.3 & 72.8 & 34.3 & 45.0 & 47.5 & 61.6 & 26.1 & 44.8 & 68.1 & 34.0 & 40.8 & 46.6 & 36.0 & 22.2 & 58.0 & 12.5 & 42.5 \\
 & 43.1 & 20.3 & 57.6 & 70.5 & 33.3 & 49.1 & 31.6 & 81.0 & 27.8 & 37.5 & 62.3 & 31.7 & 27.5 & 56.8 & 45.1 & 35.6 & 42.5 & 20.4 \\
 \midrule
LLaMA-Adapter-v2-7B & 40.4 & 27.5 & 62.3 & 32.5 & 35.0 & 30.1 & 46.5 & 24.1 & 33.8 & 34.8 & 25.2 & 30.2 & 43.9 & 33.1 & 18.2 & 44.9 & 11.0 & 36.0 \\
 & 34.1 & 27.4 & 36.4 & 40.5 & 33.8 & 30.4 & 34.9 & 71.0 & 33.2 & 42.2 & 35.8 & 31.1 & 25.8 & 52.0 & 29.1 & 32.0 & 25.0 & 29.9 \\
 \midrule
VisualGLM-6B & 38.6 & 27.1 & 55.0 & 33.1 & 33.8 & 31.1 & 39.2 & 26.0 & 36.8 & 40.5 & 31.1 & 39.1 & 39.2 & 32.4 & 26.8 & 43.8 & 14.0 & 33.1 \\
 & 33.9 & 27.0 & 28.9 & 44.8 & 27.1 & 34.5 & 35.2 & 65.0 & 28.0 & 35.8 & 48.2 & 30.8 & 23.5 & 44.0 & 26.2 & 29.6 & 37.5 & 21.1 \\
 \bottomrule
\end{tabular}%
}
\end{table*}

\section{Experiments}
In this section, we conduct a comprehensive evaluation of 30 LVLMs on the MMT-Bench. Sec.~\ref{sec:evaluation_deatils} presents the selected LVLMs zoo and the evaluation methods. The quantitative evaluation of each meta-task is provided in Sec.~\ref{sec:overall_evaluation}. We present the analysis of specific tasks with different prompt methods in Sec. \ref{sec:specific}. Furthermore, we give an error analysis of three representative LVLMs in Sec.~\ref{sec:error}.

% MMT-Bench includes tasks that require diverse prompt forms, including those for multi-image, coordinate-related, and visual-mark-related questions. To better understand these aspects, we conduct ablation experiments, as detailed in Sections \ref{sec:multi_image}, \ref{sec:coordinates}, and \ref{sec:visual_mark}.

\subsection{Evaluation Details}
\label{sec:evaluation_deatils}

\textbf{Selected LVLMs.} 
For completeness, we test 30 representative LVLMs varying in parameters, vision encoders (InternVL \cite{chen2023internvl}, EVA-CLIP-ViT \cite{EVA-CLIP}, CLIP-ViT \cite{CLIP}), and LLMs (QWen \cite{Qwen-VL}, InternLM \cite{2023internlm}, LLaMA \cite{llama1, llama2}, Vicuna \cite{vicuna}, Flan-T5 \cite{flant5}). For details, see Appendix \ref{lvlm_info}.

\textbf{Evaluation Methods.} 
In MMT-Bench, samples are in a multi-choice format, e.g., `What is this? Options: (\textit{A}) Dog (\textit{B}) Cat'. To extract the choice from LVLMs' responses, we follow OpenCompass' protocol \cite{2023opencompass}: 1) Check if the response includes option letters (\textit{A}/\textit{B}); 2) Check for option content (`dog'/`cat'); 3) Use ChatGPT for extraction. If these steps fail, we set the model selection as option letter \textit{Z} to avoid random assignment \cite{mmmu}. 
Accuracy is the primary metric. 
% Experiments run on a node with 8 NVIDIA A100 GPUs, using Transformer library v4.33.1 and PyTorch v2.0.1.

\subsection{Overall Evaluation}
\label{sec:overall_evaluation}
% This section offers a comparative analysis of LVLMs on the MMT-Bench, summarized in Table~\ref{tab:overall-results}, alongside \textit{Random Choice} and \textit{Frequent Choice} baselines—the former selects answers at random, while the latter chooses the most common option in each meta-task. Here we only analyze each meta-task, omitting the specific sub-tasks that are analyzed in detail in the Appendix~\ref{}. 
% As described in Sec.~\ref{sec:specific}, certain tasks respond better to specific prompt forms, and for these, we report scores under optimal settings in Table~\ref{tab:overall-results}.
% We summarize our key findings as follows: 
This section evaluates LVLMs on MMT-Bench alongside \textit{Random Choice} and \textit{Frequent Choice} baselines. We report the overall score for all meta-tasks as well as the best performance on each meta-task in Table~\ref{tab:overall-results}. The detailed results of each sub-task are provided in the Sec.~\ref{sec:all_results} of the Appendix. Various prompt settings for all tasks are investigated. We summarize the key findings as follows.
%with the effects of various prompting strategies discussed in Sec.~\ref{sec:specific} Key findings follow:

\textbf{i) The Comprehensive Challenge of MMT-Bench:} The benchmark poses significant challenges, with even advanced models like InternVL-Chat, GPT-4V and GeminiProVision achieving just 63.4\%, 62.0\% and 61.6\% accuracy, respectively, indicating substantial room for improvement. Notably, removing its strongest area, Visual Recognition (VR), where it scores 84.7\%, GeminiProVision's overall performance drops to 55.1\%, below satisfactory. The varied task dimensions of the MMT-Bench demand wide-ranging capabilities for optimal performance, emphasizing the benchmark's extensive and rigorous criteria. 
\textbf{ii) The comparison between Open-source LVLMs and close-source LVLMs.} 
The performance of most open-source models lags behind that of closed-source models.
However, leading open-sourced LVLM InternVL-Chat-V1.2-34B have demonstrated remarkable performance, outperforming sophisticated proprietary models such as GPT-4V and GeminiProVision in overall accuracy. 
% While XComposer2, a smaller 7B parameter open-source model, lags behind GeminiProVision by 5.9 points in the overall score.
This achievement suggests that by scaling model size, optimizing training regimes, and leveraging diverse high-quality data, open-sourced LVLMs can rival and even exceed the capabilities of advanced proprietary models. It brings a sense of pride to the open-source community and paves the way for more high-performance yet cost-effective solutions in academia and industry.
% Leading open-source LVLMs XComposer2 with only 7B parameters reach an accuracy of $55.7\%$, which, although trailing behind GeminiProVision by $6$\% accuracy, has managed to outperform it in certain individual tasks such as localization (Loc).
% This achievement suggests that open-source LVLMs can potentially rival or even surpass the performance of sophisticated proprietary models by integrating diverse and high-quality pre-training data.
%
% \textbf{ii) The Comparision between Open-source LVLMs and GeminiProVision.} 
% Leading open-source LVLMs XComposer2 with only 7B parameters reach an accuracy of $55.7\%$, which, although trailing behind GeminiProVision by $6$\% accuracy, has managed to outperform it in certain individual tasks such as localization (Loc).
% This achievement suggests that open-source LVLMs can potentially rival or even surpass the performance of sophisticated proprietary models by integrating diverse and high-quality pre-training data.
%
% XComposer2, a leading 7B-parameter open-source LVLM, achieves $55.7\%$ accuracy, $5.9$ points behind GeminiProVision. However, it outperforms in tasks like localization and Visual Prompt Understanding, etc. This highlights XComposer2's potential and suggests that with high-quality data, as shown in \cite{internlmxcomposer2}, open-source LVLMs could rival or surpass proprietary models.
%
\textbf{iii) The Influence of LLMs and Model Scaling.} 
As shown in Table~\ref{tab:overall-results}, model performance significantly improves with an increase in size (7B to 13B) for both llava-v1.5 and llava-v1.5-tuner. Upgrading LLMs, from internlm to internLM2, also enhances the performance of LLaVA, suggesting that larger or improved LLMs boost multi-task performance, with unchanged training data and visual encoders.
\textbf{iv) Model Performance across Different Meta-Tasks.} 
Most LVLMs excel in Visual Recognition (VR) tasks and Visual Captioning (VC), highlighting the ability of LVLMs to recognize `what' an object is and describe the content shown in the image. However, for fine-grained perception tasks (localization, pixel-level perception, etc) or complex reasoning tasks (image evaluation judgment), most LVLMs struggle.
\textbf{v) BLIP2 impresses in open-source models without instruction-following training, outdoing LLaVA models trained with extensive instruction-following data.} Although instruction-tuned models can give responses aligning better with human preference than BLIP2 in open-set QA on some tasks \cite{liu2023llava}, they perform worse than BLIP2 in close-set settings in MMT-Bench. This reflects MMT-Bench's multi-task challenges and hints at using the taxonomy of MMT-Bench to expand the dataset in supervised fine-tuning for future advancement.

\subsection{Specific Task and Prompt Methods Analysis}
\label{sec:specific}
In this section, we evaluate specific tasks using different prompts for LVLMs. 

\begin{figure}[tb!]
\centering
\includegraphics[width=1.0\linewidth]{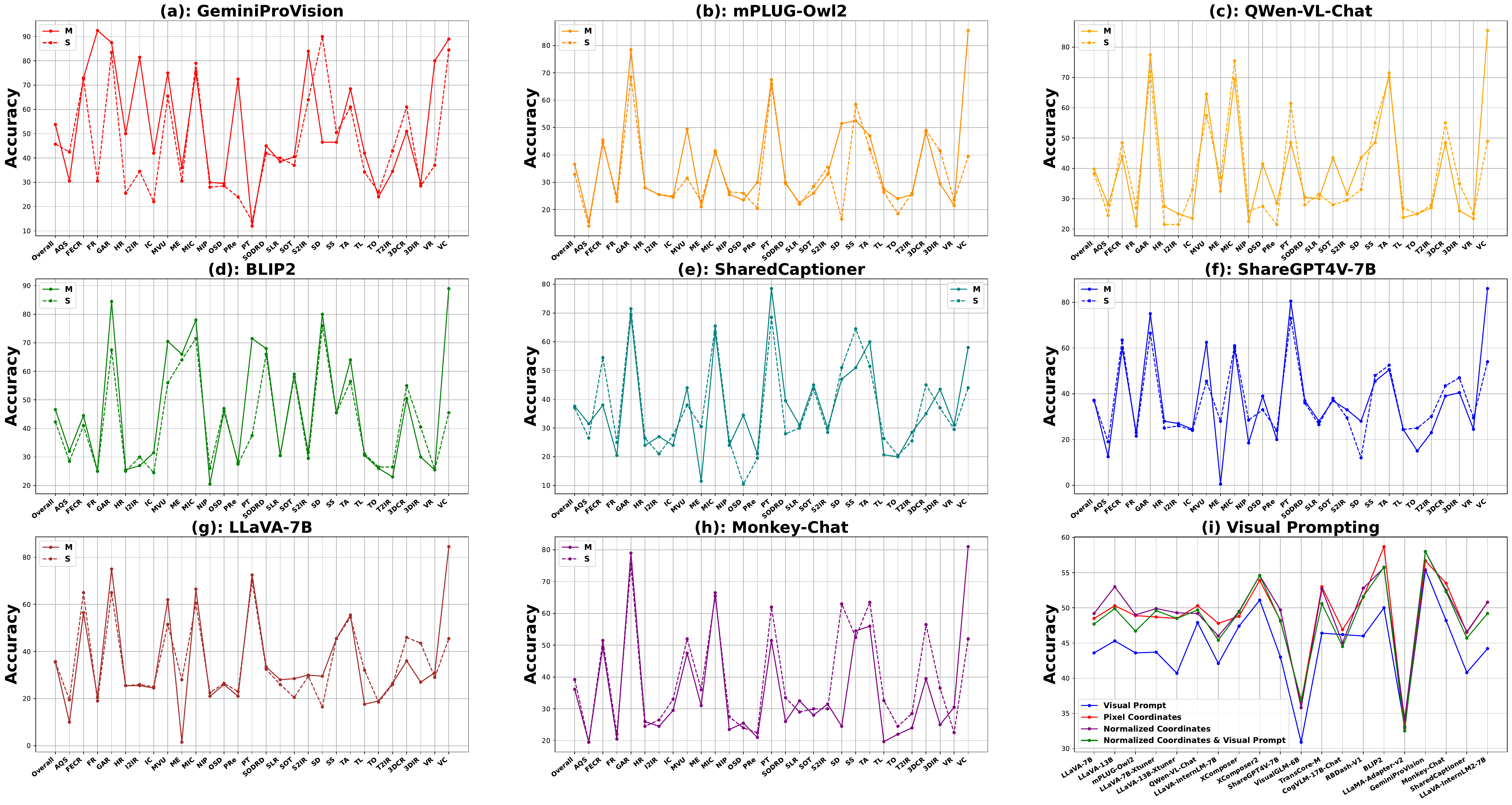} 
\vspace{-0.2in}
\caption{(a)-(h): Comparing the performance of LVLMs between settings of multiple-images prompt (denoted as \textbf{M}) and single-image prompt (denoted as \textbf{S}). Please check Appendix \ref{appendix:multi_image} for the full task names of task name abbreviations. (i): Comparison of different prompting methods for visual referring prompting-related tasks. Here we select 14 subtasks from the MMT-Bench. We only report the average accuracy here. Zoom in for better view.} 
\label{fig:ms-results-and-visual-prompting}
\end{figure}
% \textbf{Multi-Image }
\textbf{Prompting LVLMs with multi-images \textit{vs} single-image.} Here we explore the effects of exploiting multi-image prompts and single-image prompts on the performance of LVLMs.
To this end, we summarized $28$ tasks in our MMT-Bench, which usually require multiple images as input, such as image retrieval and video captioning.
% We then manually merged multiple image into one image for preparing the single-image prompt.
For multi-images prompting, we first evaluated LVLMs which are inherently designed to support multiple images as input (dubbed \textit{Multi-Images LVLMs}), including mPLUG-Owl2, QWen-VL-chat and Gemini-Pro-Vision.
Besides, we also assessed LVLMs which mainly learned on single-image prompts (dubbed \textit{Single-Image LVLMs}) for more comprehensive comparisons, including BLIP2, SharedCaptioner, ShareGPT4V-7B, Monkey and LLaVA-v1.5-7B.
Following previous studies \cite{instructblip,li2023mvbench}, we input each image individually to Single-Image LVLMs and concatenated all output visual embeddings before feeding into LLMs. 
The designed multi-image prompts for Multi-Images LVLMs and Single-Image LVLMs are summarized in Appendix Sec.\ref{appendix:multi_image}.
As for single-image prompting, we manually combine multiple images into one image and feed it into LVLMs (see examples in Fig. \ref{fig:mmt-bench}).

The detailed performance comparisons are presented in Fig.~\ref{fig:ms-results-and-visual-prompting}(a)-(h). We have several observations:
i) Multi-images tasks posed significant challenges to current LVLMs, where the best accuracy achieved by GeminiProVision is only $53.8$. 
ii) For Multi-Images LVLMs, providing multiple images as prompts instead of a single image boosted the overall performance on these tasks, demonstrating their capabilities to extract beneficial information from multiple images. 
For instance, for the task of face retrieval (\textbf{FR}), the performance of GeminiProVision increased from $30.5$ to $92.5$ when providing multiple images as visual prompts.
iii) For Single-Image LVLMs, multi-image prompts also help improve the overall performance of most models, except for Monkey. 
To our surprise, BLIP2 achieved significant performance gain when switching to a multi-image prompt setting, especially on tasks like general action recognition (\textbf{GAR}) and video captioning (\textbf{VC}).
These results highlight the potential of LVLMs to learn more robust unified representations of multiple modalities.

% \subsection{Pixel Coordinates \textit{vs.} Normalized Coordinates}
%\textbf{Detection Tasks with Different Coordinate Formats.} In Fig.~\ref{fig:coord}, we analyze the performance across $19$ detection-related tasks spanning Localization, Pixel-level Perception, and Visual Captioning, comparing outcomes under two different coordinate formats. Notably, GeminiProVision lags behind top open-source LVLMs like BLIP2 and XComposer2, which have been extensively trained with detection data. The preference for normalized coordinates among most models is attributed to their use in the training instruction templates.

% \label{sec:coordinates}

% \begin{figure}[tb!]
% \centering
% \includegraphics[width=0.9\linewidth]{figures/subplotk.pdf} 
% \caption{(a) Comparison of different coordinates format for the detection task. Here we select 19 subtasks from the MMT-Bench; (b) Comparison of different prompting methods for visual referring prompting related tasks. Here we select 14 subtasks from the MMT-Bench. We only report the average accuracy here.} 
% \label{fig:coord}
% \end{figure}

% % \subsection{Text prompt \textit{vs.} Visual prompt}
% \label{sec:visual_mark}

% \begin{figure}[tb!]
% \centering
% \includegraphics[width=0.8\linewidth]{visual_mark.pdf} 
% \vspace{-0.12in}
% \caption{Comparison of different prompting methods for visual referring prompting related task. Here we select 14 subtasks from the MMT-Bench and report the averaged accuracy.} 
% \label{fig:visual_mark}
% \end{figure}

\textbf{Most LVLMs Show Poor Generalization in Visual Referring Prompting.}
Visual referring prompting is an impressive prompting technique that entails direct image edits like drawing bounding boxes or masks to guide LVLMs to focus on specific regions \cite{gpt4v}.
We select $14$ tasks (see Sec.~\ref{sec:visual_mark}) involving visual referring prompting to explore the influence of different prompting methods on the final results. 
We compared three additional settings: using text prompts for bounding boxes in normalized ([0,1]) and pixel ([0, h or w]) formats, and combining visual and text prompts.
As depicted in Fig.~\ref{fig:ms-results-and-visual-prompting}(i), visual prompting (blue curve) significantly lags behind other settings, a disparity mainly attributed to the lack of visual prompting data in most LVLMs during the Supervised Fine-Tuning (SFT) stage. 
% We can see that visual prompting (blue curve) significantly trails behind other settings. Additionally, as indicated in Table \ref{tab:overall-results}, the best-performing model in the task of visual prompt understanding achieves only a 56.1\% accuracy. This disparity is primarily attributed to the insufficient visual prompting data during the SFT stage for most LVLMs.

\subsection{Error Analysis}
\label{sec:error}

\begin{figure}[tb!]
\centering
\includegraphics[width=0.9\linewidth]{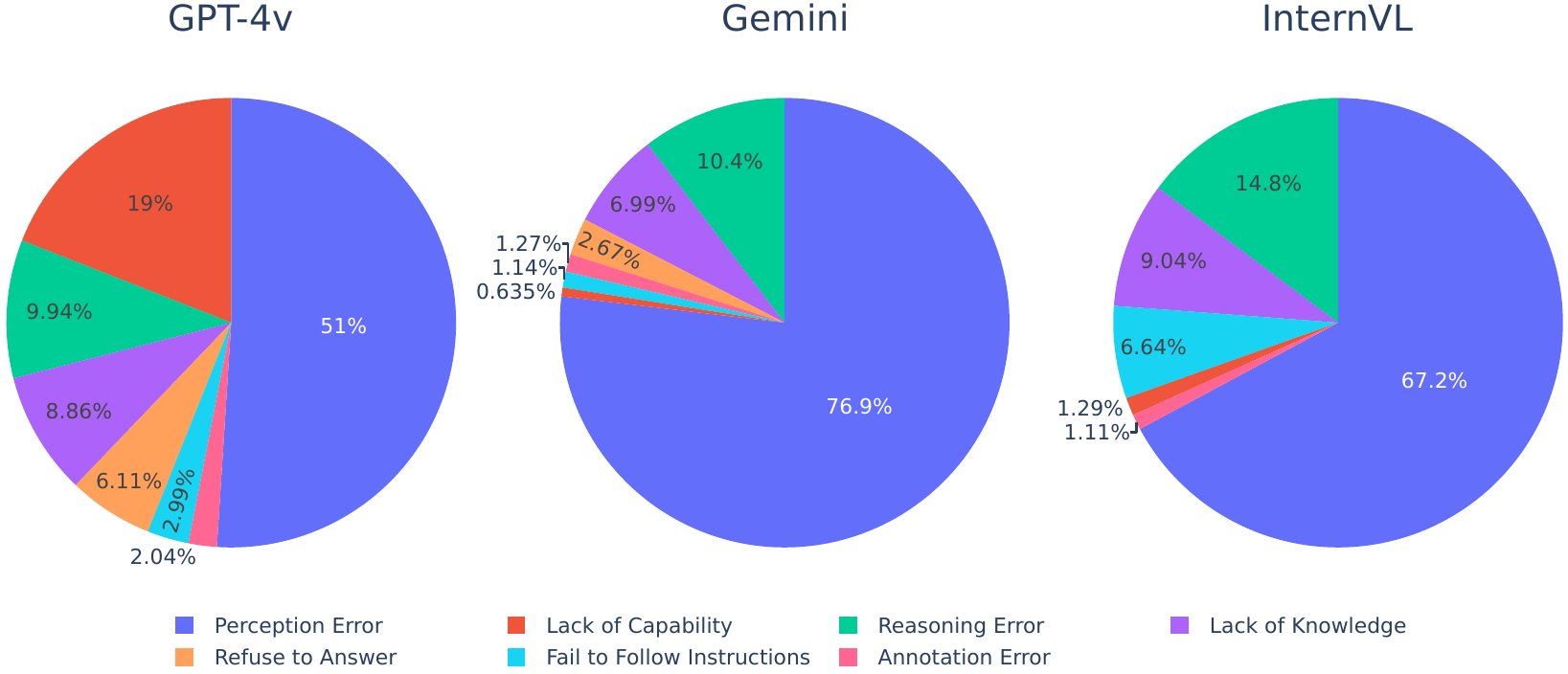} 
\caption{Distribution of error types for GPT-4V, GeminiProVision and InternVL-Chat-V1.2.} 
\label{fig:error_analysis}
\end{figure}
% To analyze the error patterns of the three models (GPT-4V, GeminiProVision, and InternVL-Chat-V1.2), we employed a strategic sampling approach. For each subtask, we randomly selected up to 5 questions where the models provided incorrect answers. This sampling method aimed to ensure a representative and diverse set of error cases across different subtasks.
% To analyze the error distribution of LVLMs on MMT-Bench, we examined three LVLMs: GPT-4V, GeminiProVision, and InternVL-Chat-V1.2 (InternVL). Specifically, for each model's incorrectly answered samples, we randomly selected up to 5 questions based on subtask categorization. We then assigned these error samples to co-authors with expertise in the specific tasks, who carefully analyzed the underlying reasons for the models' mistakes. This process yielded the error distribution presented in Fig~\ref{fig:error_analysis}.
To analyze the error distribution of LVLMs on the MMT-Bench, we examined three LVLMs: GPT-4V, GeminiProVision, and InternVL-Chat-V1.2 (InternVL).
Specifically, we randomly selected up to 5 incorrectly answered questions per subtask for each model. Task-specific experts among the co-authors then analyzed these error samples to identify the underlying reasons for the mistakes, yielding the error distribution presented in Fig.~\ref{fig:error_analysis}. For definitions and case studies of these six error types, please refer to Sec.~\ref{sec:case_study} in the appendix.

As shown in Fig~\ref{fig:error_analysis}, perception error stands out as the most common type of error across all models, with GPT-4V exhibiting a significantly lower perception error rate (51\%) compared to GeminiProVision (76.9\%) and InternVL (67.2\%), indicating its superior performance in perception tasks. Reasoning error emerges as the second most prevalent error type, with InternVL having the highest reasoning error rate (14.8\%), followed by GeminiProVision (10.4\%) and GPT-4V (9.94\%), highlighting the challenges all models face in complex reasoning tasks.

Additionally, the proportion of lack of knowledge errors is similar across the three models, ranging from 6.99\% to 9.0\%. It suggests that insufficient knowledge is a common issue. However, GPT-4V has notably higher error rates in lack of capability (19\%) and Refusing to Answer (6.11\%) compared with the other models, which may be attributed to its more honest approach in acknowledging its limitations and refusing to answer certain questions.

InternVL stands out for its high error rate in failing to follow instructions (6.64\%), significantly surpassing GPT-4V (2.99\%) and GeminiProVision (1.14\%), indicating its struggle in comprehending and executing instructions effectively. On the other hand, annotation error contributes the least to the overall error distribution, implying that the quality of data annotation is high and has a minimal impact on model performance.

To enhance the performance of these large language models, future improvements should focus on addressing the specific error types identified. By targeting perception and reasoning capabilities, tackling the lack of knowledge, and refining the ability to follow instructions, developers can work towards creating more accurate and reliable language models. GPT-4V's honest approach to its limitations also highlights the importance of transparency in AI systems, which can be further explored and incorporated into future model designs.

\section{Taskonomy Analysis}
Thanks to the extensive coverage of tasks in the MMT-Bench, we can evaluate the multimodal performance of LVLMs on a task map. In this way, the roles of different tasks in multimodal capability can be systematically interpreted by analyzing relationships between tasks in the map.
%\subsection{Evaluating on the Task Map}

\begin{figure}[tb!]
\centering
\includegraphics[width=0.75\linewidth]{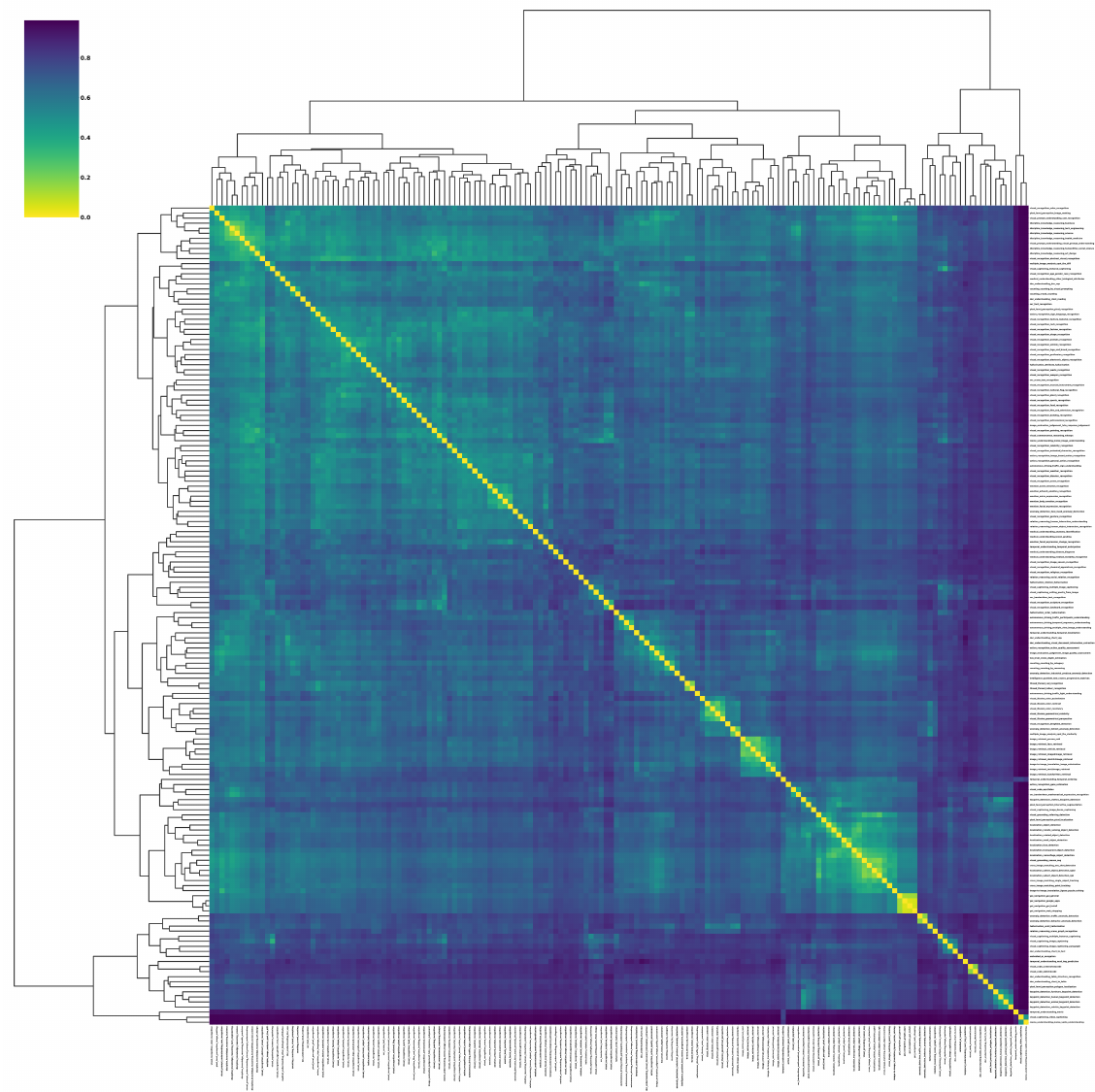} 
\caption{Visualization of task maps and hierarchical clustering with task map. Please zoom in for better visualizations.} 
\label{fig:taskmap}
\end{figure}
\subsection{Analytical Tools}\label{sec:analytical-tools}

\textbf{Task map.} 
To investigate the relationships between subtasks, we quantify each subtask as a task vector by following \cite{ilharco2022taskvector}. Formally, a task vector is defined by the weight variation between the weight fine-tuned on task data $D^t$ and the initial weight $W_0$ of a probing model, as given by $V^t = \arg\min_W \mathcal{L}(W|D^t) - W_0$
%\begin{equation}\label{eq:taskvec}
%    V^t = \arg\min_W \mathcal{L}(W|D^t) - W_0 
%\end{equation}
where the subscript $t$ denotes the task and $\mathcal{L}$ is the task loss. Three steps are adopted to obtain $V^t$. First, we use pre-trained QwenVL-Chat as the probing model because QwenVL-Chat achieves good results on most subtasks, which helps acquire promising task vectors. Second, we construct task data $D^t$ by adapting all multi-choice VQA samples into the instruction-following data for each subtask. Third, unlike TaskVec \cite{ilharco2022taskvector} that finetunes the whole model, we finetune QwenVL-Chat for $3$ epochs using LoRA fine-tuning \cite{hu2021lora} for all $162$ subtasks, which reduces the length of task vector from $9.6$B to $3.5$M and consumes less storage resources.
With task vector, a task map can be constructed as $\mathcal{G}=\{G^{st}\}_{s,t=1}^T$ where $G^{st}=1-\cos(V^s, V^t)$ denotes the cosine distance between task $s$ and $t$ and $T=162$ denoted the total number of subtasks. By definition, we know that $0\leq G^{st}\leq 2$. 

%Inspired by the methodology outlined in XX, we utilize the model weight vector variations of QwenVL-Chat post Lora fine-tuning on designated tasks as task vectors. Specifically, we have a total of 162 task vectors. Each task undergoes Lora fine-tuning for 3 epochs, using a dataset consistent with the evaluation dataset. Subsequently, we employ the cosine value of the angles between task vectors as a measure of their distance to construct a task map. Based on this, we perform hierarchical clustering of the task vectors, with detailed results depicted in fig. \ref{fig:taskmap}.

\begin{table}[!t]
\centering
\caption{The relationship between task distance threshold $\delta$ (normalized by the maximum task distance on the task map) and the consistency of LVLMs performance ranking $\tau_\delta$.  We see that LVLMs have a more consistent performance ranking when two tasks get closer to each other.}
\scalebox{1.0}{
\begin{tabular}{lccccc}
\toprule
$\delta$ & 1 & $\frac{1}{2}$ & $\frac{1}{4}$ & $\frac{1}{6}$ & $\frac{1}{8}$ \\
\midrule
$\tau_\delta$  & 0.29 & 0.31 & 0.32 & 0.41 & 0.60 \\
\bottomrule
\end{tabular}
}
\label{tab:taskmap}
% \vspace{-0.2in}
\end{table}
\textbf{Ranking correlation: Kendall's tau $\mathbf{\tau}$.}
To quantitatively evaluate LVLMs on a task map, we use the metric of Kendall's tau ${\tau}$ to measure the ranking correlation between performance sequences of LVLMs on different subtasks. 
% The intuition is that model $A$ would be superior to model $B$ if model $A$ performs better than model $B$ on task $s$ when task distance $G^{st}$ is small. 
The intuition is that model $A$ would be superior to model $B$ on task $t$ if model $A$ performs better than model $B$ on task $s$ when task distance $G^{st}$ is small. 
The Kendall's tau ${\tau}$ is defined as $\tau^{st} = \frac{2}{M(M-1)} \sum_{1\leq m<n\leq M}\mathrm{sign}((P^s_m - P^s_n)(P^t_m - P^t_n))$
%For completeness, we present the definition of Kendall's tau ${\tau}$ as follows,
%
%\begin{equation}\label{eq:kendalltau}
% \tau^{st} = \frac{2}{M(M-1)} \sum_{1\leq m<n\leq M}\mathrm{sign}((P^s_m - P^s_n)(P^t_m - P^t_n))  
%\end{equation}
where $P^s_m$ denotes the performance of model $m$ on task $s$ and $M$ is the number of LVLMs. The function $\mathrm{sign(\cdot)}$ returns $-1$ if the argument is negative and $1$ otherwise. When $\tau^{st}=1$, LVLMs have completely consistent performance ranking on task $s$ and $t$.

\subsection{Findings on Task Map}\label{sec:taskmap-findings}
\textbf{LVLMs obtain a more consistent performance ranking on tasks closer to each other.} 
%
%We perform hierarchical clustering of the task vectors, with detailed results depicted in fig. \ref{fig:taskmap}.
%
We assess whether LVLMs achieve consistent performance on two tasks close to each other. To measure this consistency, we employ the Kendall tau metric as introduced in Sec. \ref{sec:analytical-tools}. Specifically, we consider all subtask pairs in which two tasks are closer to each other and calculate their average Kendall's tau ${\tau}$, which can be given by 
    $\tau_\delta = \frac{1}{T}\sum_{s=1}^T\frac{1}{|\Delta_s|}\sum_{t\in \Delta_s}\tau^{st}$ where $ \Delta_s=\{t:G^{st}\leq\delta\}
$ 
and $\delta$ is a threshold used to control the proximity between two tasks. 
%This threshold is determined by calculating the ratio of the maximum distance in the task map to a given constant. 
As shown in Table~\ref{tab:taskmap}, as the threshold $\delta$ decreases, the task distance becomes smaller, and $\tau_\delta$ increases. This suggests that LVLMs obtain a more consistent performance ranking on tasks closer to each other. 
% Hence, the performance of LVLMs on a new task could be predicted if the new task is close to one of the subtasks in MMT-Bench.
Hence, the performance of LVLMs on a new task can be predicted if it is close to one of the MMT-Bench subtasks.

%Firstly, as demonstrated by fig. \ref{fig:taskmap}, subtasks belonging to a  meta-task tend to be closely located (such as the four subtasks of the GUI navigation), which intuitively aligns with expectations. Subsequently, we quantitatively validate a fundamental property of the task map to verify its rationality as implemented by our approach. Specifically, models are expected to exhibit similar performances on tasks that are proximate on the task map \cite{xx}. Consequently, the performance ranking of models should remain consistent among similar tasks. To measure this consistency, we employ the Kendall tau metric.

%Our approach initially identifies the most similar task for each given task from the task map. However, considering that the most similar tasks might still be distant, we only calculate the Kendall tau for pairs of tasks whose distance is below a certain threshold. This threshold is determined by calculating the ratio of the maximum distance in the task map to a given constant. As shown in table \ref{tab:taskmap}, as the constant value increases, the threshold decreases, meaning the distance between tasks becomes smaller, and the average Kendall tau increases. This suggests that as tasks become more closely situated, the consistency in model performance rankings improves.

\textbf{Out-of-Domain (OoD) tasks discovery.}
The OoD tasks mean tasks that the current model struggles to handle. Discovering OoD tasks can provide insights for future evaluation efforts and the development of stronger LVLMs. Since model performance on different tasks is related to task distances, we hypothesize that OoD tasks would be grouped in local regions on the task map. Therefore, we conduct hierarchical clustering on the task map to find OoD tasks. Specifically, $162$ subtasks are grouped into $12$ clusters as shown in Fig. \ref{fig:taskmap}. We use two criteria to identify clusters containing OoD tasks. First, LVLMs would achieve poor performance on OoD tasks. In this regard, we calculate the average multimodal performance within each task cluster over all LVLM models. Second, the performance of LVLMs on OoD tasks would be inconsistent with the overall multimodal score in Table~\ref{tab:overall-results} because LVLMs with competitive overall scores would even fail to solve OoD tasks. Hence, we calculate the average ranking correlation $\tau$ within each cluster. 
We present these statistics in Table~\ref{tab:taskmapcluster} and provide a detailed analysis with the clustering results in Appendix \ref{sec:task_map}.
 
%We utilize the task map to analyze tasks that the current model struggles to handle effectively, thereby providing insights for subsequent model design and evaluation efforts. As illustrated, we conduct hierarchical clustering on the task map and, upon selecting a cluster count of 12, extract tasks from each category. We then calculate the Kendall tau between the average model performance in these tasks and the overall model performance, as shown in table \ref{tab:taskmapcluster}. This is contrasted with the model performance, depicted in Fig. \ref{fig:modelperformace}, where tasks are colour-coded to represent different categories, sequentially from the first to the twelfth category. In most cases, we observe that if the average Kendall tau is relatively low (for instance, less than 0.5), the model's performance on these tasks is typically subpar. It's important to note that this finding is intuitive, as a model's relative performance order tends to fluctuate on more challenging tasks, often misaligned with the overall performance.

We can see that clusters $8$, $9$, and $11$ achieve low multimodal accuracy and ranking correlation $\tau$. In sec \ref{sec:overall_evaluation}, we find that the model struggles with handling fine-grained visual tasks, such as detection. Through the analysis of these clusters, we similarly find that current multimodal large models cannot perform fine-grained visual cognition and understanding of positional and spatial relationships, such as localization and detection tasks. Moreover, they exhibit poor performance in tasks related to new data structures or types of images, showing a lack of proficiency in handling tasks related to GUI and special data structures like tables.

\begin{table}[!t]
\centering
\caption{The number of tasks within each cluster after hierarchical clustering, and the Kendall's tau $\tau$ between the average performance of the model on these tasks and the overall performance of the model.}
\scalebox{0.58}{
\begin{tabular}{lcccccccccccc}
\toprule
Cluster & 1 & 2 & 3 & 4 & 5 & 6 & 7 & 8 & 9 & 10 & 11 & 12 \\
\midrule
\# Tasks  & 11 & 53 & 16 & 16 & 9 & 8 & 7 & 16 & 4 & 9 & 10 & 3\\
\midrule
$\tau$  & 0.54 & 0.73 & 0.57 & 0.48 & -0.05 & 0.62 & 0.63 & 0.34 & 0.12 & 0.57 & 0.38 & 0.59\\
\midrule
Acc  & 40.4 & 64.7 & 61.9 & 39.9 & 55.9 & 30.0 & 33.1 & 40.2 & 31.4 & 61.2 & 33.2 & 50.7\\
\bottomrule
\end{tabular}
}
\label{tab:taskmapcluster}
% \vspace{-0.2in}
\end{table}

\textbf{In-domain tasks discovery.}
In-domain tasks are tasks that most current multimodal large models can handle correctly. Discovering in-domain tasks guides the commercial application of LVLMs in specific scenarios. 
Different from OoD tasks, we identify in-domain tasks by looking for clusters with large ranking correlation $\tau$ and high multimodal accuracy.
%
%Similarly, we can identify the in-domain tasks that most current multimodal large models can handle correctly, signifying the general capability of these models to accomplish these tasks. As shown in table \ref{tab:taskmapcluster}, we select clusters with high Kendall tau values and analyze the common characteristics of tasks within these clusters. We arrive at the following conclusions:
%
From Table~\ref{tab:taskmapcluster}, we can see that clusters $2$, $3$, and $10$ achieve relatively high accuracy and large ranking correlation $\tau$. We observe that current multimodal large models possess strong high-level visual comprehension capabilities, enabling them to effectively handle visual recognition tasks, even when dealing with specialized images such as medical images, which is also found in sec \ref{sec:overall_evaluation}. Moreover, they benefit from the powerful LLMs to accurately describe images.
We provide a detailed analysis along with the clustering results in Appendix \ref{sec:task_map}.
% \begin{itemize}

%   \item Cluster 2 mainly comprises visual recognition tasks, which require the model to possess certain high-level visual capabilities, yet these tasks are relatively simple. Examining Table \ref{tab:overall-results} and Fig. \ref{fig:modelperformace}, we observe that the model's performance within this cluster is generally good. This validates that the current multimodal large models possess fundamental abilities for visual-semantic understanding, allowing them to fulfil recognition tasks.

%   \item Cluster 3 mainly includes visual recognition tasks as well, yet extends to cover sophisticated visual understanding tasks that require primary specialist knowledge, such as medicine and emotion. Within this cluster, the model demonstrates large $\tau$ and high accuracy, suggesting that current multimodal models pay attention to tasks necessitating the infusion of domain-specific knowledge, beyond just natural images. This implies a certain ability to handle problems in specialized fields.

%   \item In Cluster 10, LVLMs achieve good performance on tasks related to the visual description of the image. It indicates that current large multimodal models can describe the image well. It would stem from the fact that these models are typically tuned by massive image-text pairs. %Thus, when it comes to generating these kinds of textual outputs, the models possess strong prior knowledge.
% \end{itemize}

%\vspace{+2mm}
\section{Conclusion and Discussion}
In this work, we introduce MMT-Bench, a comprehensive benchmark designed to evaluate LVLMs in multimodal multitask understanding. The breadth of MMT-Bench is highlighted by its meticulously curated dataset of $31,325$ multi-choice questions covering $162$ multimodal tasks. Our evaluation reveal significant challenges for current LVLMs posed by our MMT-Bench. We present a taskonomy analysis of LVLMs on the task map, allowing us to predict the performance of a new task. Our goal with MMT-Bench is to measure the progress on the path to multitask AGI. We shall acknowledge that MMT-Bench may not be sufficient as a standard for determining whether multitask AGI has been achieved, as it is impossible to include all multimodal tasks. However, we believe that it should be necessary for a multitask AGI to achieve strong performance on MMT-Bench. We will continue to expand the task set of MMT-Bench. We believe that MMT-Bench will inspire further research and development in LVLMs, bringing us closer to the realization of truly intelligent multimodal systems.

\textbf{Broader Impact.} 
The development and widespread adoption of MMT-Bench as a benchmark for evaluating large vision-language models (LVLMs) have the potential to significantly impact the field of artificial intelligence. While MMT-Bench offers valuable insights and guidance for advancing LVLM research, it is important to consider its broader impact, including ethical considerations and potential societal consequences.

One potential positive impact of MMT-Bench is its role in driving advancements in LVLM technology, leading to improved performance and capabilities in various multimodal tasks. This could benefit numerous applications, such as visual dialogue, video analysis, and document understanding, ultimately enhancing user experiences and productivity. 

However, it is crucial to recognize and address potential negative impacts as well. One of the primary limitations of MMT-Bench is its reliance on curated data, which may inadvertently introduce biases based on the sources and methodologies used for data collection. For example, the performance of each meta-task is obtained by taking the average over all subtasks, which may lead to biased assessment because meta-tasks comprise different numbers of subtasks. Moreover, the selection of tasks and subtasks in MMT-Bench may only partially capture the diversity of real-world scenarios, leading to a limited understanding of LVLMs' capabilities across different domains and populations. Furthermore, the data collection process might disproportionately represent certain demographics or contexts, which can lead to biased evaluations of LVLMs' performance.

The other concern is that the benchmark's emphasis on performance metrics such as overall scores and task-specific accuracies may oversimplify the evaluation process and obscure nuanced differences in LVLMs' performance. This could mask disparities in model performance across demographic groups or domains, contributing to the perpetuation of biases and inequities in AI systems. We are dedicated to collecting as many multimodal tasks as possible into our MMT-Bench for unbiased evaluation.

%MMT-Bench may inadvertently perpetuate biases present in training data, leading to unfair or discriminatory outcomes in AI systems. Additionally, the reliance on MMT-Bench as a primary evaluation metric could narrow research focus and limit diversity and innovation within the AI community.
\nocite{langley00}

\bibliography{main}
\bibliographystyle{icml2024}

%%%%%%%%%%%%%%%%%%%%%%%%%%%%%%%%%%%%%%%%%%%%%%%%%%%%%%%%%%%%%%%%%%%%%%%%%%%%%%%
%%%%%%%%%%%%%%%%%%%%%%%%%%%%%%%%%%%%%%%%%%%%%%%%%%%%%%%%%%%%%%%%%%%%%%%%%%%%%%%
% APPENDIX
%%%%%%%%%%%%%%%%%%%%%%%%%%%%%%%%%%%%%%%%%%%%%%%%%%%%%%%%%%%%%%%%%%%%%%%%%%%%%%%
%%%%%%%%%%%%%%%%%%%%%%%%%%%%%%%%%%%%%%%%%%%%%%%%%%%%%%%%%%%%%%%%%%%%%%%%%%%%%%%
\clearpage
%\tableofcontents 
%\appendixpage  
%\appendixtoc   

\newpage
\appendix
\onecolumn
%\tableofcontents 

\renewcommand{\thetable}{A\arabic{table}}
\renewcommand{\thefigure}{A\arabic{figure}}
\renewcommand{\theequation}{A\arabic{equation}}
\renewcommand{\theHtable}{A\thetable}
\renewcommand{\theHfigure}{A\thefigure}

\setcounter{table}{0}
\setcounter{figure}{0}

In this appendix, we provide further details as follows:
\begin{itemize}
\item Sec. \ref{sec:task_map}: Presents hierarchical clustering and more analyses on the task map constructed from our MMT-Bench.
\item {Sec. \ref{sec:task-hierarchy}: Includes
details on sample size, visual input types, and capabilities of LVLMs evaluated for each subtask}.
\item Sec. \ref{sec:abbrev}: Enumerates task abbreviations used throughout the paper.
\item Sec. \ref{sec:more-exp-detail}: Presents detailed model configurations and experimental details in multi-images and visual prompting.
\item Sec. \ref{sec:pixel-coor}: Compares the performance on tasks involving pixel coordinates and normalized coordinates.
\item Sec. \ref{sec:analy-img-type-cap}: Compares the performance of LVLMs on different image types and multimodal capabilities.
\item Sec. \ref{sec:case_study}: Illustrates error cases of GPT-4V, GeminiProVision, and InternVL-Chat on $32$ meta-tasks in MMT-Bench.
\item Sec. \ref{sec:ocr}: Gives the comparison of MMT-Bench with Other Benchmarks on OCR-Related Tasks.
\item Sec. \ref{sec:bench_build}: Presents some Ddtails about the benchmark construction.
\item Sec. \ref{sec:opencompass_protocol}:  Discusses the openCompass protocol used in MMT-Bench and other alternatives.
\item Sec. \ref{sec:resource}:  Gives the computaional resources used in evaluation.

\item Sec. \ref{sec:all_results}: Provides the detailed performance of 30 models across all 162 subtasks on MMT-Bench.

\end{itemize}
% % Reset depth to add sections and subsections to ToC
% \addtocontents{toc}{\protect\setcounter{tocdepth}{3}}

% % Setting colorlinks=black just for the table of contents
% \hypersetup{linkcolor=black}

% \tableofcontents % Lists only the appendix sections and subsections


% You can have as much text here as you want. The main body must be at most $8$ pages long.
% For the final version, one more page can be added.
% If you want, you can use an appendix like this one.  

% The $\mathtt{\backslash onecolumn}$ command above can be kept in place if you prefer a one-column appendix, or can be removed if you prefer a two-column appendix.  Apart from this possible change, the style (font size, spacing, margins, page numbering, etc.) should be kept the same as the main body.

\section{Task Map}
\label{sec:task_map}

We perform hierarchical clustering on the taskmap, as shown in Fig.~\ref{fig:taskmap}. When selecting the number of clustering clusters as $12$, we analyze the clustering results of the task map and the model performance on the corresponding tasks. Here, we list the names of the tasks within each cluster in Table~\ref{tab:long}.

\begin{figure*}[h]
\centering
\includegraphics[width=0.99\linewidth]{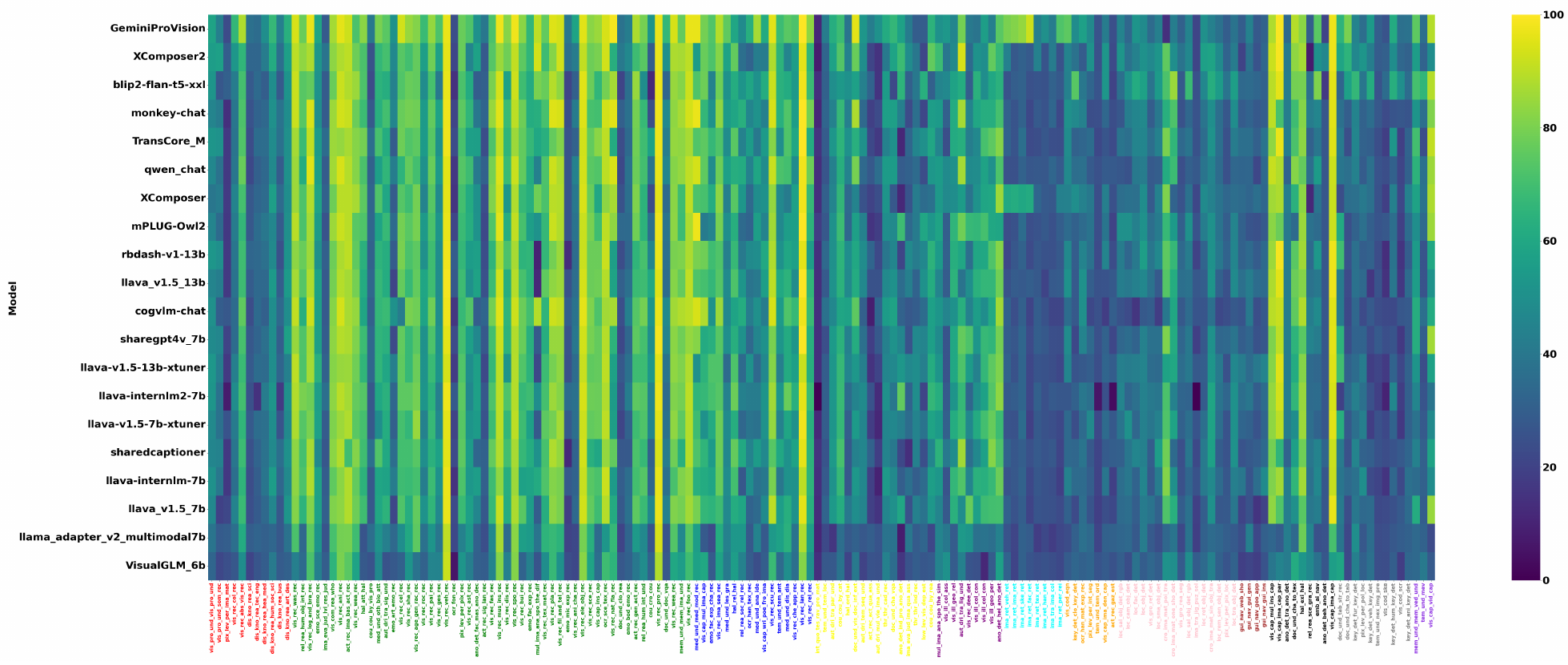} 
\caption{Visualization of model performance on different tasks. Different colours signify the respective categories formed after clustering, arranged from left to right, starting from the first category through to the twelfth. Please zoom in for better visualizations.} 
\label{fig:modelperformace}
\end{figure*}

\textbf{Out-of-Domain (OoD) tasks discovery.} We can see that clusters $8$, $9$, and $11$ achieve low multimodal accuracy and ranking correlation $\tau$. From these clusters, we find that current multimodal large models lack the ability to perform fine-grained visual cognition and understanding of positional and spatial relationships, such as localization and detection tasks. Moreover, they exhibit poor performance in tasks related to new data structures or types of images, showing a lack of proficiency in handling tasks related to GUI and special data structures like tables.

% We summarize the commonalities of these task clusters as below.
% %Therefore, we propose that tasks within categories with a lower average Kendall tau are the ones where the model currently faces difficulties. We summarize the commonalities of some categories of tasks which have relatively low Kendall tau as below (Cluster: 5, 6, 8, 9, 11):

\begin{itemize}

  %\item Cluster 5 primarily encompasses visual illusion tasks, akin to how LLMs are prone to text-based illusions. Similarly, large multimodal models exhibit significant issues with visual illusions, highlighting a critical area for improvement.
  % \item The sixth cluster is predominantly focused on image retrieval tasks, suggesting that current large multimodal models are relatively weak in extracting image features. There is a clear need to enhance the capability of visual representation in these models.
  \item Cluster $8$ mainly involves detection, tracking, and localization tasks, all of which are related to the localization of objects within images. This indicates that current large multimodal models lack fine-grained visual cognition and understanding of positional and spatial relationships.
  \item Tasks in cluster $9$ are centered around GUI navigation, a novel task type requiring strong visual understanding, object localization, and expert knowledge in operating mobile devices \cite{yang2023appagent}. This suggests that current large multimodal models need further optimization for GUI-related tasks.
  \item Apart from detection and localization tasks, cluster $11$ also includes tasks involving the recognition of special images or their conversion into structured text. The former requires models to possess spatial cognition and fine-grained visual capabilities, while the latter demands robust OCR abilities and extensive knowledge (such as understanding and outputting the basic structure of code or tables). Our testing LVLMs currently fall short in this aspect.
\end{itemize}

\textbf{In-Domain tasks discovery.} From Table~\ref{tab:taskmapcluster}, we can see that clusters $2$, $3$, and $10$ achieve relatively high accuracy and large ranking correlation $\tau$. We observe that current multimodal large models possess strong high-level visual comprehension capabilities, enabling them to effectively handle visual recognition tasks, even when dealing with specialized images such as medical images. Moreover, they benefit from the powerful LLMs to accurately describe images.

\begin{itemize}

  \item Cluster $2$ mainly comprises visual recognition tasks, which require the model to possess certain high-level visual capabilities, yet these tasks are relatively simple. Examining Table~\ref{tab:overall-results} and Fig. \ref{fig:modelperformace}, we observe that the model's performance within this cluster is generally good. This validates that the current multimodal large models possess fundamental abilities for visual-semantic understanding, allowing them to fulfil recognition tasks.

  \item Cluster $3$ mainly includes visual recognition tasks as well, yet extends to cover sophisticated visual understanding tasks that require primary specialist knowledge, such as medicine and emotion. Within this cluster, the model demonstrates large $\tau$ and high accuracy, suggesting that current multimodal models pay attention to tasks necessitating the infusion of domain-specific knowledge, beyond just natural images. This implies a certain ability to handle problems in specialized fields.

  \item In Cluster $10$, LVLMs achieve good performance on tasks related to the visual description of the image. It indicates that current large multimodal models can describe the image well. It would stem from the fact that these models are typically tuned by massive image-text pairs. %Thus, when it comes to generating these kinds of textual outputs, the models possess strong prior knowledge.
\end{itemize}

\section{Hierarchical Structure of MMT-Bench}\label{sec:task-hierarchy}
% 我们在表5到7中展示了MMT-Bench中所有的32个meta-tasks，总共包含162个子任务。在表中我们同时展示了每一个子任务的sample num，visual input type 以及考察LVLM对应的能力
In Table~\ref{tab:task_1} to Table~\ref{tab:task_3}, we present all $32$ meta-tasks from MMT-Bench, encompassing a total of $162$ subtasks. These tables include details on sample size, visual input types, and capabilities of LVLMs evaluated for each subtask.
\begin{table*}[ht!]
\centering
\caption{The Abbreviations of terms mentioned in this paper and their corresponding full terms.}
\label{tab:abbreviation}
\resizebox{0.9\textwidth}{!}{%
\begin{tabular}{l|l|l|l}
\toprule
Abbreviation & Full Term & Abbreviation & Full Term \\
\midrule
\multicolumn{4}{c}{Meta-Task} \\
\midrule
VR & Visual Recognition & VI & Visual Illusion \\
Loc & Localization & MemU & Meme Understanding \\
OCR & OCR & VPU & Visual Prompt Understanding \\
Count & Counting & AND & Anomaly Detection \\
HLN & Hallucination & KD & Keypoint Detection \\
IR & Image Retrieval & VCR & Visual Commonsense Reasoning \\
3D & 3D & IEJ & Image Evaluation Judgement \\
VC & Visual Captioning & MIA & Multiple Image Analysis \\
VG & Visual Grounding & CIM & Cross Image Matching \\
DU & Doc Understanding & TU & Temporal Understanding \\
AR & Action Recognition & VCo & Visual Code \\
PLP & Pixel Level Perception & MedU & Medical Understanding \\
I2IT & Image-to-image Translation & AUD & Autonomous Driving \\
RR & Relation Reasoning & DKR & Discipline Knowledge Reasoning \\
IQT & Intelligence Quotient Test & EA & Embodied AI \\
Emo & Emotion & GN & GUI Navigation \\
\midrule
\multicolumn{4}{c}{Subtask} \\
\midrule
AQS & Action Quality Assessment & SODRD & Salient Object Detection RGBD \\
FECR & Facial Expression Change Recognition & SLR & Sign Language Recognition \\
FR & Face Retrieval & SOT & Single Object Tracking \\
GAR & General Action Recognition & S2IR & Sketch2image Retrieval \\
HR & Handwritten Retrieval & SD & Spot the Diff \\
I2IR & Image2image Retrieval & SS & Spot the Similarity \\
IC & Image Colorization & TA & Temporal Anticipation \\
MVU & Meme Video Understanding & TL & Temporal Localization \\
ME & MEVIS & TO & Temporal Ordering \\
MIC & Multiple Image Captioning & T2IR & Text2image Retrieval \\
NIP & Next Image Prediction & 3DCR & 3D CAD Recognition \\
OSD & One-shot Detection & 3DIR & 3D Indoor Recognition \\
PRe & Person Reid & VR & Vehicle Retrieval \\
PT & Point Tracking & VC & Video Captioning \\
\bottomrule
\end{tabular}}
\end{table*}

\section{Task Abbreviations}
\label{sec:abbrev}
Given the extensive number of tasks and models tested within the benchmark, we employ abbreviations to condense the manuscript. The abbreviations used throughout the paper are shown in Table~\ref{tab:abbreviation}.

\section{More Experimental Details}\label{sec:more-exp-detail}
\subsection{LVLMs Model Details}
\label{lvlm_info}
Table~\ref{tab:lvlm} summarizes the LVLMs information used in this paper, including the corresponding parameter sizes, visual encoders, and LLMs. Note that we use follow OpenCompass’ protocol \cite{2023opencompass} to conduct the evaluation process. The inference time varies with different models. For instance, the smaller LLaVA-v1.5-7B \cite{liu2023improvedllava} model takes only $12$ minutes to complete the evaluation using 8 GPUs, while the larger InternVL-Chat-V1.2-34B model \cite{chen2023internvl} requires $79$ minutes and around 80GB of memory. Our open-source codebase supports multi-GPU distributed inference, effectively accelerating the inference process.
\subsection{Multi-Images Prompt Experimental Details}
\label{appendix:multi_image}
In terms of the $28$ tasks requiring multiple images as input, please see Table~\ref{tab:multi-image-prompt-part1}-\ref{tab:multi-image-prompt-part4} for the specific task names given task abbreviations.
Besides, we also present the designed prompt examples for Single-Image LVLMs and Multi-Images LVLMs in Table~\ref{tab:multi-image-prompt-part1}-\ref{tab:multi-image-prompt-part4} for reference.
\subsection{Visual Referring Prompting Experimental Details}
\label{sec:visual_mark}
In Section~\ref{sec:specific}, we explore the differential efficacy of visual prompting compared to alternative prompting strategies across a spectrum of 14 distinct tasks. These encompass human interaction understanding, social relation recognition, human-object interaction recognition, animal keypoint detection, vehicle keypoint detection, human keypoint detection, clothes keypoint detection, scene text recognition, interactive segmentation, instance captioning, multiple instance captioning, one-shot detection, single object tracking, and counting by visual prompting.

\section{Pixel Coordinates \textit{vs.} Normalized Coordinates} \label{sec:pixel-coor}
\begin{figure}[tb!]
\centering
\includegraphics[width=0.8\linewidth]{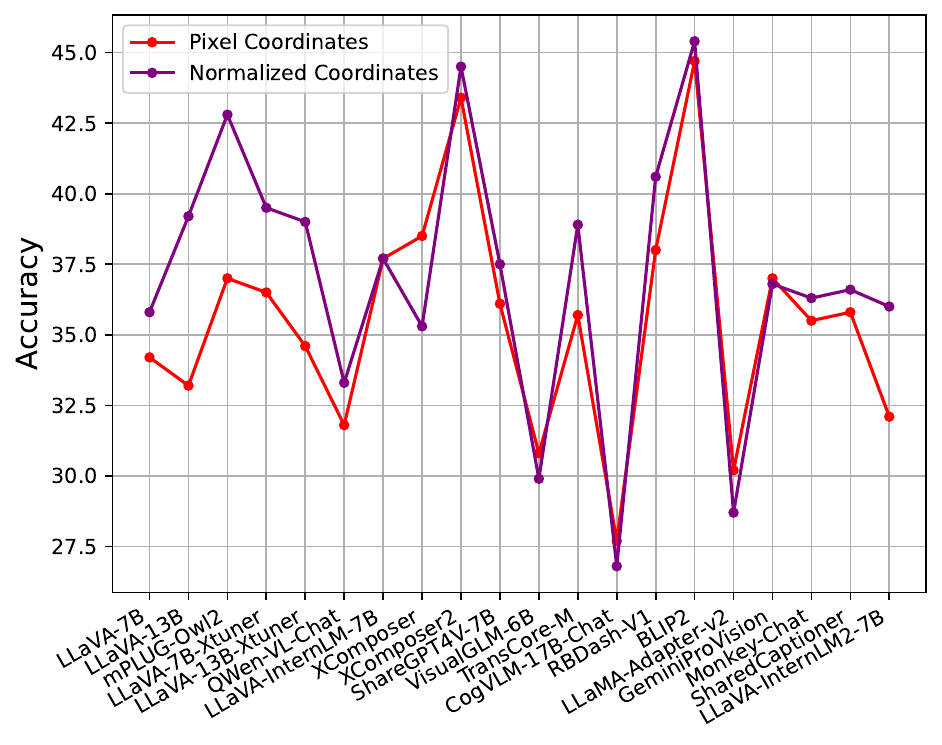} 
\caption{Comparison of coordinate formats for detection tasks across 19 MMT-Bench subtasks, reporting average accuracy.}
\label{fig:coord}
\end{figure}

% In Fig.~\ref{fig:coord}, we analyze the performance across $19$ detection-related tasks spanning Localization, Pixel-level Perception, and Visual Captioning, comparing outcomes under two different coordinate formats. Notably, GeminiProVision lags behind top open-source LVLMs like BLIP2 and XComposer2, which have been extensively trained with detection data. The preference for normalized coordinates among most models is attributed to their use in the training instruction templates.
In Fig.~\ref{fig:coord}, we analyze the performance across $19$ detection-related tasks, specifically point tracking, image matting, pixel recognition, polygon localization, pixel localization, depth estimation, MEVIS, remote sensing object detection, rotated object detection, small object detection, camouflage object detection, salient object detection in RGB-D, transparent object detection, face detection, object detection, salient object detection in RGB, referring detection, reason segmentation, and image dense captioning. These tasks span Localization, Pixel-level Perception, and Visual Captioning, comparing outcomes under two different coordinate formats. Notably, GeminiProVision lags behind top open-source LVLMs like BLIP2 and XComposer2, which have been extensively trained with detection data. The preference for normalized coordinates among most models is attributed to their use in the training instruction templates.

\section{Analysis on Images Types and Capabilities}\label{sec:analy-img-type-cap}
\begin{figure*}
    \centering
    \includegraphics[width=0.8\linewidth]{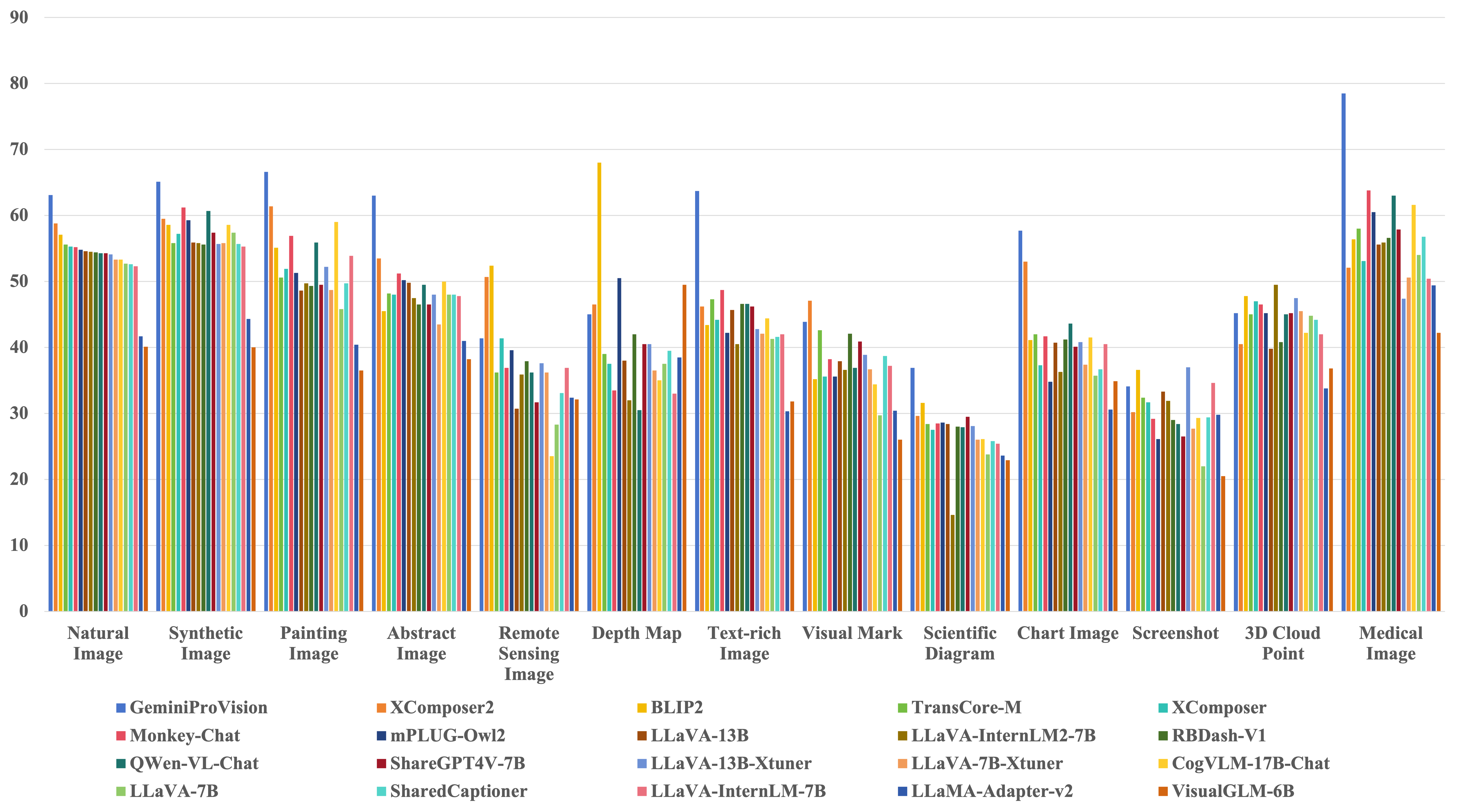}
    \caption{The performance of 20 LVLMs across 13 types of visual input.}
    \label{fig:image_type}
\end{figure*}

\textbf{Performance with Different Visual Types.}
We compare the performance of 20 LVLMs across 13 types of visual input in Fig.~\ref{fig:image_type}. 
Most LVLMs struggle with Scientific Diagrams due to task difficulty, as many, including Scientific and "Raven's Progressive Matrices," require complex reasoning, a capability current LVLMs do not possess well.
%Across most types, GeminiProVision consistently outperforms the other models by a huge margin.

\begin{figure*}
    \centering
    \includegraphics[width=0.8\linewidth]{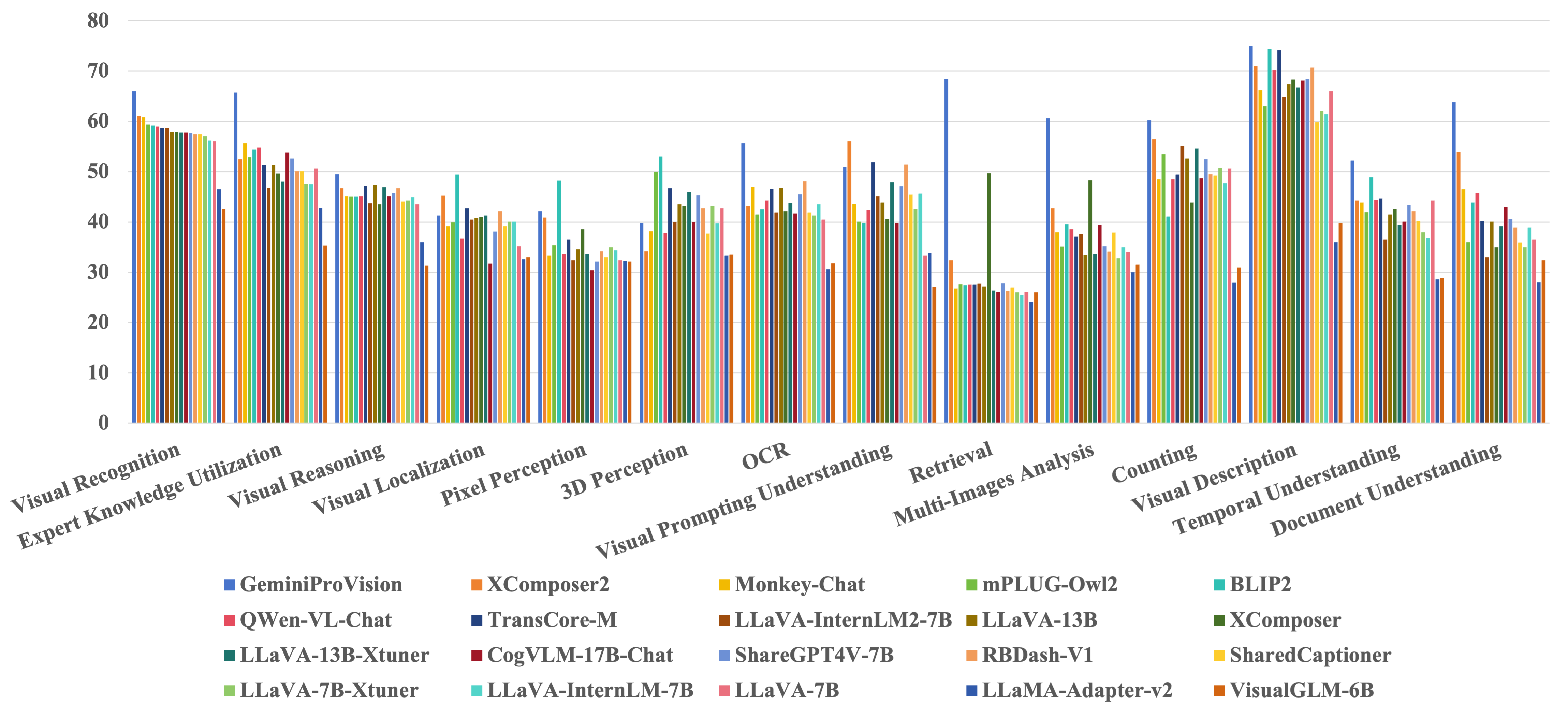}
    \caption{The performance of 20 LVLMs across 14 capabilities.}
    \label{fig:capability}
\end{figure*}

\textbf{Performance Across Multimodal Capabilities.}
We also compare the performance of $20$ LVLMs across $14$ types of visual input in Fig.~\ref{fig:capability}. As we can see, GeminiProVision once again exhibits strong superiority across most capabilities, especially in retrieval and multi-image analysis (involving the recognition and matching of multiple images), vastly outperforming other open-source LVLMs. This superiority stems from GeminiProVision's support for multi-image mode and its powerful generalization abilities, guiding the future direction of open-source models towards the focus on multi-image and video understanding.

\begin{xltabular}{\textwidth}{>{\hsize=0.4\hsize}l|>{\hsize=1.6\hsize}X|>{\hsize=0.4\hsize}X}
\caption{Details of task clustering on the task map of our MMT-Bench.} \label{tab:long} \\
\toprule \multicolumn{1}{c}{\textbf{Meta-Task}} & \multicolumn{1}{|c|}{\textbf{Subtask}} & \multicolumn{1}{c}{\textbf{\# subtasks}} \\ \midrule 
\endfirsthead

\multicolumn{3}{c}%
{\tablename\ \thetable{} -- continued from previous page} \\
\toprule \multicolumn{1}{c}{\textbf{Meta-Task}} & \multicolumn{1}{|c|}{\textbf{Subtask}} & \multicolumn{1}{c}{\textbf{\# subtasks}} \\ \midrule 
\endhead

%\hline \multicolumn{3}{|r|}{{Continued on next page}} \\
\bottomrule
\endfoot

\bottomrule
\endlastfoot

\multicolumn{3}{c}{Cluster ID: 1} \\ 
\midrule
Visual Prompt Understanding  & Visual Prompt Understanding, Som (Set-of-marks) Recognition & 2 \\
\midrule
 Pixel Level Perception & Image Matting &  1\\
 \midrule
 Visual Recognition & Color Recognition, Abstract Visual Recognition & 2\\
 \midrule
 Discipline Knowledge Reasoning & Science, Tech Engineering, Health Medicine, Humanities Social Science, Business, Art Design & 6\\
 \midrule
 \multicolumn{3}{c}{Cluster ID: 2} \\ 
 \midrule
 Visual Recognition & Waste recognition, Logo and Brand Recognition, Animals Recognition, Weapon Recognition,
Celebrity Recognition, Shape Recognition, Age Gender Race Recognition, Rock Recognition,
Painting Recognition, Gesture Recognition, Vehicle Recognition, Astronomical Recognition,
Fashion Recognition, Musical Instrument Recognition, Disaster Recognition, Sports Recognition,
Building Recognition, Texture Material Recognition, Plant Recognition, Film and Television Recognition,
Animated Character Recognition, Electronic Object Recognition, Scene Recognition,
National Flag Recognition, Profession Recognition, Weather Recognition, Food Recognition & 27 \\
\midrule
Relation Reasoning & Human Object Interaction Recognition, Human Interaction Understanding & 2\\
\midrule
Action Recognition & Image-based Action Recognition, Sign Language Recognition, General Action Recognition & 4 \\
\midrule
Emotion & Scene Emotion Recognition, Artwork Emotion Recognition, Facial Expression Recognition,
Micro Expression Recognition, Body Emotion Recognition & 5 \\
\midrule
Image Evaluation Judgement & Lvlm Response Judgement & 1\\
\midrule
Visual Commonsense Reasoning & WhoopsVQA & 1\\
\midrule
Hallucination & Attribute Hallucination & 1\\
\midrule
Counting&
Counting by Visual Prompting, Crowd Counting & 2 \\
\midrule
Medical Understanding&
Other Biological Attributes & 1\\
\midrule
Autonomous Driving&
Traffic Sign Understanding & 1 \\
\midrule
OCR&
Font Recognition, Scene Text Recognition & 2\\
\midrule
Pixel Level Perception&
Pixel Recognition & 1\\
\midrule
Anomaly Detection&
Face Mask Anomaly Detection & 1\\
\midrule
Multiple Image Analysis&
Spot the Diff & 1\\
\midrule
Visual Captioning&
Instance Captioning & 1\\
\midrule
Doc Understanding&
Clock Reading, Doc VQA & 2\\
\midrule
Meme Understanding&
Meme Image Understanding & 1\\
 \midrule
 \multicolumn{3}{c}{Cluster ID: 3} \\ 
 \midrule
Medical Understanding& Medical Modality Recognition, Lesion Grading, Disease DiagnoseAnatomy Identification& 3\\ 
 \midrule
Visual Captioning& Multiple Image Captioning, Writing Poetry from Image& 2\\ 
 \midrule
Emotion& Facial Expression Change Recognition& 1\\ 
 \midrule
Visual Recognition& Image Season Recognition, Sculpture Recognition, Chemical Apparatus Recognition, Landmark Recognition, Religious Recognition& 5\\ 
 \midrule
Hallucination& Relation Hallucination& 1\\ 
 \midrule
Relation reasoning& Social Relation Recognition& 1\\ 
 \midrule
OCR& Handwritten Text Recognition& 1\\ 
 \midrule
Temporal Understanding& Temporal Anticipation& 1\\ 
 \midrule
 \midrule
 \multicolumn{3}{c}{Cluster ID: 4} \\ 
 \midrule
Intelligence Quotient Test& Ravens Progressive Matrices& 1\\ 
 \midrule
Temporal Understanding& Temporal Localization& 1\\ 
 \midrule
Autonomous Driving& Traffic Participants Understanding, Temporal Sequence Understanding, Multiple View Image Understanding& 3\\ 
 \midrule
Counting& Counting by Category, Counting by Reasoning& 2\\ 
 \midrule
Hallucination& Order Hallucination& 1\\ 
 \midrule
Doc Understanding& Visual Document Information Extraction, Chart VQA& 2\\ 
 \midrule
Action Recognition& Action Quality Assessment,& 2\\ 
 \midrule
3D& 3D Cad Recognition, 3D indoor recognition& 2\\ 
 \midrule
Anomaly Detection& Industrial Produce Anomaly Detection& 1\\ 
 \midrule
Image Evaluation Judgement& Image Quality Assessment& 1\\ 
 \midrule
Low Level Vision& Depth Estimation& 1\\

 \midrule
 \multicolumn{3}{c}{Cluster ID: 5} \\ 
 \midrule
 Multiple Image Analysis& Spot the Similarity& 1\\ 
 \midrule
Visual Illusion& Color Assimilation, Geometrical Relativity, Color Constancy, Color Contrast, Geometrical Perspective& 5\\ 
 \midrule
Autonomous Driving& Traffic Light Understanding& 1\\ 
 \midrule
Visual Recognition& Deepfake Detection& 1\\ 
 \midrule
Anomaly Detection& Helmet Anomaly Detection& 1\\

  \midrule
 \multicolumn{3}{c}{Cluster ID: 6} \\ 
 \midrule\\
 
 Image Retrieval& Vehicle Retrieval, Image2image Retrieval, Sketch2image Retrieval, Face Retrieval, Text2image Retrieval, Handwritten Retrieval, Person Reid& 7\\ 
 \midrule
Image-to-image translation& Image Colorization& 1\\

  \midrule
 \multicolumn{3}{c}{Cluster ID: 7} \\ 
 \midrule\\
 Visual Code& Eqn2latex,& 2\\ 
 \midrule
Keypoint Detection& Clothes Keypoint Detection& 1\\ 
 \midrule
OCR& Handwritten Math Expression recognition& 1\\ 
 \midrule
Pixel Level Perception& Interactive Segmentation& 1\\ 
 \midrule
Temporal Understanding& Temporal Ordering& 1\\ 
 \midrule
Visual Captioning& Image Dense Captioning& 1\\ 
 \midrule
Action Recognition& Gaze Estimation& 1\\

  \midrule
 \multicolumn{3}{c}{Cluster ID: 8} \\ 
 \midrule\\
 
 Localization& Salient Object Detection RGB, Camouflage Object Detection, Face Detection, Object Detection, Small Object Detection, Salient Object Detection RGBD, Rotated Object Detection, Remote Sensing Object Detection, Transparent Object Detection& 9\\ 
 \midrule
Visual Grounding& Referring Detection, Reason Seg& 2\\ 
 \midrule
Cross Image Matching& Point Tracking, One Shot Detection,& 3\\ 
 \midrule
Image-to-image Translation& Jigsaw Puzzle Solving& 1\\ 
 \midrule
Cross Image Catching& Single Object Tracking& 1\\ 
 \midrule
Pixel Level Perception& Pixel Localization& 1\\

  \midrule
 \multicolumn{3}{c}{Cluster ID: 9} \\ 
 \midrule\\
 GUI Navigation&
Web Shopping, GUI General, Google Apps, GUI Install & 4\\
 
  \midrule
 \multicolumn{3}{c}{Cluster ID: 10} \\ 
 \midrule\\
 
 Visual Captioning& Multiple Instance Captioning, Image Captioning Paragraph, Image Captioning& 3\\ 
 \midrule
Anomaly Detection& Traffic Anomaly Detection& 1\\ 
 \midrule
Doc Understanding& Chart to text& 1\\ 
 \midrule
Hallucination& Exist Hallucination& 1\\ 
 \midrule
Relation Reasoning& Scene Graph Recognition& 1\\ 
 \midrule
Embodied AI& Navigation& 1\\ 
 \midrule
Anomaly Detection& Behavior Anomaly Detection& 1\\ 
 
  \midrule
 \multicolumn{3}{c}{Cluster ID: 11} \\ 
 \midrule
 Doc Understanding& Table Structure Recognition, Chart to Table& 2\\ 
 \midrule
Keypoint Detection& Furniture Keypoint Detection, Vehicle Keypoint Detection, Human Keypoint Detection, Animal Keypoint Detection& 4\\ 
 \midrule
Pixel Level Perception& Polygon Localization,& 2\\ 
 \midrule
Temporal Understanding& Next Image Prediction& 1\\ 
 \midrule
Visual Code& Sketch2code, Screenshot2code& 2\\

  \midrule
 \multicolumn{3}{c}{Cluster ID: 12} \\ 
 \midrule

Meme Understanding& Meme Video Understanding& 1\\ 
 \midrule
Temporal Understanding& Mevis& 1\\ 
 \midrule
Visual Captioning& Video Captioning& 1\\ 

 \bottomrule
\end{xltabular}

% \usepackage{graphicx}
\begin{table*}[]
\centering
\caption{MMT-Bench subtask details (part 1): including sample number, visual input types, and evaluated LVLM capabilities.}
\label{tab:task_1}
\resizebox{0.8\textwidth}{!}{%
\begin{tabular}{l|lll}
\toprule
\multicolumn{1}{l|}{Subtask Name} & \multicolumn{1}{l|}{Sample Num} & \multicolumn{1}{l|}{Visual Input Type} & Capability \\
\midrule
\multicolumn{4}{c}{\textbf{Visual Grounding}} \\ \midrule
\multicolumn{1}{l|}{Reason Seg} & \multicolumn{1}{l|}{196} & \multicolumn{1}{l|}{Natural Image} & Visual Reasoning,Visual Localization \\ \midrule
\multicolumn{1}{l|}{Referring Detection} & \multicolumn{1}{l|}{200} & \multicolumn{1}{l|}{Natural Image} & Visual Localization \\ \midrule
\multicolumn{4}{c}{\textbf{Doc Understanding}} \\ \midrule
\multicolumn{1}{l|}{Doc Vqa} & \multicolumn{1}{l|}{200} & \multicolumn{1}{l|}{Text-rich Image} & Document Understanding \\ \midrule
\multicolumn{1}{l|}{Visual Document Information Extraction} & \multicolumn{1}{l|}{200} & \multicolumn{1}{l|}{Text-rich Image} & Document Understanding \\ \midrule
\multicolumn{1}{l|}{Chart To Text} & \multicolumn{1}{l|}{200} & \multicolumn{1}{l|}{Chart Image} & Document Understanding \\ \midrule
\multicolumn{1}{l|}{Chart To Table} & \multicolumn{1}{l|}{200} & \multicolumn{1}{l|}{Chart Image} & Document Understanding \\ \midrule
\multicolumn{1}{l|}{Clock Reading} & \multicolumn{1}{l|}{200} & \multicolumn{1}{l|}{Abstract Image} & Visual Recognition,Document Understanding \\ \midrule
\multicolumn{1}{l|}{Chart Vqa} & \multicolumn{1}{l|}{200} & \multicolumn{1}{l|}{Chart Image} & Document Understanding \\ \midrule
\multicolumn{1}{l|}{Table Structure Recognition} & \multicolumn{1}{l|}{46} & \multicolumn{1}{l|}{Chart Image} & Document Understanding \\ \midrule
\multicolumn{4}{c}{\textbf{Action Recognition}} \\ \midrule
\multicolumn{1}{l|}{Gaze Estimation} & \multicolumn{1}{l|}{200} & \multicolumn{1}{l|}{Natural Image} & Visual Recognition,Visual Localization,Pixel Perception \\ \midrule
\multicolumn{1}{l|}{Image Based Action Recognition} & \multicolumn{1}{l|}{200} & \multicolumn{1}{l|}{Natural Image} & Visual Recognition \\ \midrule
\multicolumn{1}{l|}{General Action Recognition} & \multicolumn{1}{l|}{200} & \multicolumn{1}{l|}{Natural Image} & Visual Recognition,Multi-Images Analysis \\ \midrule
\multicolumn{1}{l|}{Action Quality Assessment} & \multicolumn{1}{l|}{200} & \multicolumn{1}{l|}{Natural Image} & Visual Recognition,Multi-Images Analysis,Expert Knowledge Utilization \\ \midrule
\multicolumn{1}{l|}{Sign Language Recognition} & \multicolumn{1}{l|}{200} & \multicolumn{1}{l|}{Natural Image} & Visual Recognition,Multi-Images Analysis \\ \midrule
\multicolumn{4}{c}{\textbf{Localization}} \\ \midrule
\multicolumn{1}{l|}{Remote Sensing Object Detection} & \multicolumn{1}{l|}{200} & \multicolumn{1}{l|}{Remote Sensing Image} & Visual Recognition,Visual Localization \\ \midrule
\multicolumn{1}{l|}{Rotated Object Detection} & \multicolumn{1}{l|}{90} & \multicolumn{1}{l|}{Remote Sensing Image} & Visual Recognition,Visual Localization \\ \midrule
\multicolumn{1}{l|}{Small Object Detection} & \multicolumn{1}{l|}{200} & \multicolumn{1}{l|}{Natural Image} & Visual Recognition,Visual Localization \\ \midrule
\multicolumn{1}{l|}{Camouflage Object Detection} & \multicolumn{1}{l|}{200} & \multicolumn{1}{l|}{Natural Image} & Visual Recognition,Visual Localization \\ \midrule
\multicolumn{1}{l|}{Salient Object Detection Rgbd} & \multicolumn{1}{l|}{200} & \multicolumn{1}{l|}{Natural Image,Depth Map} & Visual Localization \\ \midrule
\multicolumn{1}{l|}{Transparent Object Detection} & \multicolumn{1}{l|}{200} & \multicolumn{1}{l|}{Natural Image} & Visual Recognition,Visual Localization \\ \midrule
\multicolumn{1}{l|}{Face Detection} & \multicolumn{1}{l|}{200} & \multicolumn{1}{l|}{Natural Image} & Visual Recognition,Visual Localization \\ \midrule
\multicolumn{1}{l|}{Object Detection} & \multicolumn{1}{l|}{200} & \multicolumn{1}{l|}{Natural Image} & Visual Recognition,Visual Localization \\ \midrule
\multicolumn{1}{l|}{Salient Object Detection Rgb} & \multicolumn{1}{l|}{200} & \multicolumn{1}{l|}{Natural Image} & Visual Localization \\ \midrule
\multicolumn{4}{c}{\textbf{Visual Recognition}} \\ \midrule
\multicolumn{1}{l|}{Deepfake Detection} & \multicolumn{1}{l|}{200} & \multicolumn{1}{l|}{Natural Image,Synthetic Image} & Visual Recognition,Visual Reasoning,Expert Knowledge Utilization \\ \midrule
\multicolumn{1}{l|}{Weather Recognition} & \multicolumn{1}{l|}{194} & \multicolumn{1}{l|}{Natural Image} & Visual Recognition \\ \midrule
\multicolumn{1}{l|}{Image Season Recognition} & \multicolumn{1}{l|}{200} & \multicolumn{1}{l|}{Natural Image} & Visual Recognition \\ \midrule
\multicolumn{1}{l|}{Gesture Recognition} & \multicolumn{1}{l|}{200} & \multicolumn{1}{l|}{Natural Image} & Visual Recognition \\ \midrule
\multicolumn{1}{l|}{Muscial Instrument Recognition} & \multicolumn{1}{l|}{200} & \multicolumn{1}{l|}{Natural Image} & Visual Recognition \\ \midrule
\multicolumn{1}{l|}{Food Recognition} & \multicolumn{1}{l|}{200} & \multicolumn{1}{l|}{Natural Image} & Visual Recognition \\ \midrule
\multicolumn{1}{l|}{Landmark Recognition} & \multicolumn{1}{l|}{50} & \multicolumn{1}{l|}{Natural Image} & Visual Recognition,Expert Knowledge Utilization \\ \midrule
\multicolumn{1}{l|}{Scene Recognition} & \multicolumn{1}{l|}{200} & \multicolumn{1}{l|}{Natural Image} & Visual Recognition \\ \midrule
\multicolumn{1}{l|}{Animals Recognition} & \multicolumn{1}{l|}{200} & \multicolumn{1}{l|}{Natural Image} & Visual Recognition \\ \midrule
\multicolumn{1}{l|}{Chemical Apparatusn Recognition} & \multicolumn{1}{l|}{200} & \multicolumn{1}{l|}{Natural Image} & Visual Recognition \\ \midrule
\multicolumn{1}{l|}{Rock Recognition} & \multicolumn{1}{l|}{200} & \multicolumn{1}{l|}{Natural Image} & Visual Recognition,Expert Knowledge Utilization \\ \midrule
\multicolumn{1}{l|}{Fashion Recognition} & \multicolumn{1}{l|}{200} & \multicolumn{1}{l|}{Natural Image} & Visual Recognition \\ \midrule
\multicolumn{1}{l|}{Logo And Brand Recognition} & \multicolumn{1}{l|}{200} & \multicolumn{1}{l|}{Natural Image} & Visual Recognition \\ \midrule
\multicolumn{1}{l|}{Astronomical Recognition} & \multicolumn{1}{l|}{94} & \multicolumn{1}{l|}{Natural Image} & Visual Recognition,Expert Knowledge Utilization \\ \midrule
\multicolumn{1}{l|}{Painting Recognition} & \multicolumn{1}{l|}{200} & \multicolumn{1}{l|}{Painting Image} & Visual Recognition,Expert Knowledge Utilization \\ \midrule
\multicolumn{1}{l|}{Color Recognition} & \multicolumn{1}{l|}{200} & \multicolumn{1}{l|}{Synthetic Image} & Visual Recognition \\ \midrule
\multicolumn{1}{l|}{Plant Recognition} & \multicolumn{1}{l|}{200} & \multicolumn{1}{l|}{Natural Image} & Visual Recognition \\ \midrule
\multicolumn{1}{l|}{Shape Recognition} & \multicolumn{1}{l|}{200} & \multicolumn{1}{l|}{Synthetic Image} & Visual Recognition \\ \midrule
\multicolumn{1}{l|}{Profession Recognition} & \multicolumn{1}{l|}{200} & \multicolumn{1}{l|}{Natural Image} & Visual Recognition \\ \midrule
\multicolumn{1}{l|}{Building Recognition} & \multicolumn{1}{l|}{200} & \multicolumn{1}{l|}{Natural Image} & Visual Recognition,Expert Knowledge Utilization \\ \midrule
\multicolumn{1}{l|}{Electronic Object Recognition} & \multicolumn{1}{l|}{200} & \multicolumn{1}{l|}{Natural Image} & Visual Recognition \\ \midrule
\multicolumn{1}{l|}{Sports Recognition} & \multicolumn{1}{l|}{200} & \multicolumn{1}{l|}{Natural Image} & Visual Recognition \\ \midrule
\multicolumn{1}{l|}{Disaster Recognition} & \multicolumn{1}{l|}{200} & \multicolumn{1}{l|}{Natural Image} & Visual Recognition \\ \midrule
\multicolumn{1}{l|}{Celebrity Recognition} & \multicolumn{1}{l|}{200} & \multicolumn{1}{l|}{Natural Image} & Visual Recognition \\ \midrule
\multicolumn{1}{l|}{Vehicle Recognition} & \multicolumn{1}{l|}{200} & \multicolumn{1}{l|}{Natural Image} & Visual Recognition \\ \midrule
\multicolumn{1}{l|}{National Flag Recognition} & \multicolumn{1}{l|}{200} & \multicolumn{1}{l|}{Synthetic Image} & Visual Recognition \\ \midrule
\multicolumn{1}{l|}{Abstract Visual Recognition} & \multicolumn{1}{l|}{200} & \multicolumn{1}{l|}{Abstract Image} & Visual Recognition \\ \midrule
\multicolumn{1}{l|}{Animated Character Recognition} & \multicolumn{1}{l|}{200} & \multicolumn{1}{l|}{Synthetic Image} & Visual Recognition \\ \midrule
\multicolumn{1}{l|}{Texture Material Recognition} & \multicolumn{1}{l|}{200} & \multicolumn{1}{l|}{Natural Image} & Visual Recognition \\ \midrule
\multicolumn{1}{l|}{Film And Television Recognition} & \multicolumn{1}{l|}{200} & \multicolumn{1}{l|}{Synthetic Image} & Visual Recognition,Expert Knowledge Utilization \\ \midrule
\multicolumn{1}{l|}{Sculpture Recognition} & \multicolumn{1}{l|}{50} & \multicolumn{1}{l|}{Natural Image} & Visual Recognition,Expert Knowledge Utilization \\ \midrule
\multicolumn{1}{l|}{Age Gender Race Recognition} & \multicolumn{1}{l|}{200} & \multicolumn{1}{l|}{Natural Image} & Visual Recognition \\ \midrule
\multicolumn{1}{l|}{Weapon Recognition} & \multicolumn{1}{l|}{200} & \multicolumn{1}{l|}{Natural Image} & Visual Recognition \\ \midrule
\multicolumn{1}{l|}{Religious Recognition} & \multicolumn{1}{l|}{200} & \multicolumn{1}{l|}{Natural Image,Synthetic Image} & Visual Recognition,Expert Knowledge Utilization \\ \midrule
\multicolumn{1}{l|}{Waste Recognition} & \multicolumn{1}{l|}{200} & \multicolumn{1}{l|}{Natural Image} & Visual Recognition,Expert Knowledge Utilization \\ \bottomrule
\end{tabular}%
}
\end{table*}

% \usepackage{graphicx}
\begin{table*}[]
\centering
\caption{MMT-Bench subtask details (part 2): including sample number, visual input types, and evaluated LVLM capabilities.}
\label{tab:task_2}
\resizebox{0.9\textwidth}{!}{%
\begin{tabular}{llll}
\toprule
\multicolumn{1}{l|}{Subtask Name} & \multicolumn{1}{l|}{Sample Num} & \multicolumn{1}{l|}{Visual Input Type} & Capability \\ \midrule
\multicolumn{4}{c}{\textbf{Gui Navigation}} \\ \midrule
\multicolumn{1}{l|}{Gui General} & \multicolumn{1}{l|}{200} & \multicolumn{1}{l|}{Screenshot Image} & Visual Reasoning,Visual Localization \\ \midrule
\multicolumn{1}{l|}{Google Apps} & \multicolumn{1}{l|}{200} & \multicolumn{1}{l|}{Screenshot Image} & Visual Reasoning,Visual Localization \\ \midrule
\multicolumn{1}{l|}{Web Shopping} & \multicolumn{1}{l|}{200} & \multicolumn{1}{l|}{Screenshot Image} & Visual Reasoning,Visual Localization \\ \midrule
\multicolumn{1}{l|}{Gui Install} & \multicolumn{1}{l|}{200} & \multicolumn{1}{l|}{Screenshot Image} & Visual Reasoning,Visual Localization \\ \midrule
\multicolumn{4}{c}{\textbf{OCR}} \\ \midrule
\multicolumn{1}{l|}{Font Recognition} & \multicolumn{1}{l|}{200} & \multicolumn{1}{l|}{Text-rich Image} & OCR \\ \midrule
\multicolumn{1}{l|}{Handwritten Text Recognition} & \multicolumn{1}{l|}{100} & \multicolumn{1}{l|}{Text-rich Image} & OCR \\ \midrule
\multicolumn{1}{l|}{Handwritten Mathematical Expression Recognition} & \multicolumn{1}{l|}{100} & \multicolumn{1}{l|}{Text-rich Image} & OCR \\ \midrule
\multicolumn{1}{l|}{Scene Text Recognition} & \multicolumn{1}{l|}{200} & \multicolumn{1}{l|}{Natural Image,Text-rich Image} & OCR \\ \midrule
\multicolumn{4}{c}{\textbf{Image-to-image Translation}} \\ \midrule
\multicolumn{1}{l|}{Jigsaw Puzzle Solving} & \multicolumn{1}{l|}{200} & \multicolumn{1}{l|}{Natural Image} & Visual Recognition,Visual Reasoning \\ \midrule
\multicolumn{1}{l|}{Image Colorization} & \multicolumn{1}{l|}{200} & \multicolumn{1}{l|}{Natural Image} & Pixel Perception \\ \midrule
\multicolumn{4}{c}{\textbf{Temporal Understanding}} \\ \midrule
\multicolumn{1}{l|}{Next Img Prediction} & \multicolumn{1}{l|}{200} & \multicolumn{1}{l|}{Visual Mark} & Temporal Understanding \\ \midrule
\multicolumn{1}{l|}{Mevis} & \multicolumn{1}{l|}{200} & \multicolumn{1}{l|}{Natural Image} & Temporal Understanding \\ \midrule
\multicolumn{1}{l|}{Temporal Anticipation} & \multicolumn{1}{l|}{200} & \multicolumn{1}{l|}{Natural Image} & Temporal Understanding \\ \midrule
\multicolumn{1}{l|}{Temporal Ordering} & \multicolumn{1}{l|}{200} & \multicolumn{1}{l|}{Natural Image} & Temporal Understanding \\ \midrule
\multicolumn{1}{l|}{Temporal Localization} & \multicolumn{1}{l|}{193} & \multicolumn{1}{l|}{Natural Image} & Temporal Understanding \\ \midrule
\multicolumn{4}{c}{\textbf{Relation Reasoning}} \\ \midrule
\multicolumn{1}{l|}{Social Relation Recognition} & \multicolumn{1}{l|}{200} & \multicolumn{1}{l|}{Natural Image} & Visual Recognition,Visual Reasoning \\ \midrule
\multicolumn{1}{l|}{Human Object Interaction Recognition} & \multicolumn{1}{l|}{200} & \multicolumn{1}{l|}{Natural Image} & Visual Recognition,Visual Reasoning \\ \midrule
\multicolumn{1}{l|}{Scene Graph Recognition} & \multicolumn{1}{l|}{200} & \multicolumn{1}{l|}{Natural Image} & Visual Recognition,Visual Reasoning \\ \midrule
\multicolumn{1}{l|}{Human Interaction Understanding} & \multicolumn{1}{l|}{200} & \multicolumn{1}{l|}{Natural Image} & Visual Recognition,Visual Reasoning \\ \midrule
\multicolumn{4}{c}{\textbf{Discipline Knowledge Reasoning}} \\ \midrule
\multicolumn{1}{l|}{Science} & \multicolumn{1}{l|}{127} & \multicolumn{1}{l|}{Scientific Diagram} & Visual Reasoning,Expert Knowledge Utilization \\ \midrule
\multicolumn{1}{l|}{Health Medicine} & \multicolumn{1}{l|}{140} & \multicolumn{1}{l|}{Natural Image,Chart Image,Medical Image} & Visual Reasoning,Expert Knowledge Utilization \\ \midrule
\multicolumn{1}{l|}{Art Design} & \multicolumn{1}{l|}{110} & \multicolumn{1}{l|}{Synthetic Image,Text-rich Image,Painting Image} & Visual Reasoning,Expert Knowledge Utilization \\ \midrule
\multicolumn{1}{l|}{Humanitites Social Science} & \multicolumn{1}{l|}{112} & \multicolumn{1}{l|}{Synthetic Image,Painting Image} & Visual Reasoning,Expert Knowledge Utilization \\ \midrule
\multicolumn{1}{l|}{Tech Engineering} & \multicolumn{1}{l|}{182} & \multicolumn{1}{l|}{Chart Image,Scientific Diagram} & Visual Reasoning,Expert Knowledge Utilization \\ \midrule
\multicolumn{1}{l|}{Business} & \multicolumn{1}{l|}{120} & \multicolumn{1}{l|}{Text-rich Image,Chart Image} & Visual Reasoning,Expert Knowledge Utilization \\ \midrule
\multicolumn{4}{c}{\textbf{Intelligence Quotient Test}} \\ \midrule
\multicolumn{1}{l|}{Ravens Progressive Matrices} & \multicolumn{1}{l|}{200} & \multicolumn{1}{l|}{Scientific Diagram} & Visual Reasoning,Expert Knowledge Utilization \\ \midrule
\multicolumn{4}{c}{\textbf{Embodied AI}} \\ \midrule
\multicolumn{1}{l|}{Navigation} & \multicolumn{1}{l|}{200} & \multicolumn{1}{l|}{Synthetic Image} & Visual Reasoning,Expert Knowledge Utilization \\ \midrule
\multicolumn{4}{c}{\textbf{Emotion}} \\ \midrule
\multicolumn{1}{l|}{Facail Expression Change Recognition} & \multicolumn{1}{l|}{200} & \multicolumn{1}{l|}{Natural Image} & Visual Recognition,Temporal Understanding \\ \midrule
\multicolumn{1}{l|}{Scene Emotion Recognition} & \multicolumn{1}{l|}{200} & \multicolumn{1}{l|}{Natural Image} & Visual Recognition \\ \midrule
\multicolumn{1}{l|}{Micro Expression Recognition} & \multicolumn{1}{l|}{200} & \multicolumn{1}{l|}{Natural Image} & Visual Recognition \\ \midrule
\multicolumn{1}{l|}{Artwork Emotion Recognition} & \multicolumn{1}{l|}{200} & \multicolumn{1}{l|}{Painting Image} & Visual Recognition \\ \midrule
\multicolumn{1}{l|}{Body Emotion Recognition} & \multicolumn{1}{l|}{200} & \multicolumn{1}{l|}{Natural Image} & Visual Recognition \\ \midrule
\multicolumn{1}{l|}{Facial Expression Recognition} & \multicolumn{1}{l|}{200} & \multicolumn{1}{l|}{Natural Image} & Visual Recognition \\ \midrule
\multicolumn{4}{c}{\textbf{Visual Illusion}} \\ \midrule
\multicolumn{1}{l|}{Color Constancy} & \multicolumn{1}{l|}{72} & \multicolumn{1}{l|}{Synthetic Image} & Visual Recognition,Visual Reasoning \\ \midrule
\multicolumn{1}{l|}{Color Assimilation} & \multicolumn{1}{l|}{200} & \multicolumn{1}{l|}{Synthetic Image} & Visual Recognition,Visual Reasoning \\ \midrule
\multicolumn{1}{l|}{Geometrical Relativity} & \multicolumn{1}{l|}{200} & \multicolumn{1}{l|}{Synthetic Image} & Visual Recognition,Visual Reasoning \\ \midrule
\multicolumn{1}{l|}{Geometrical Perspective} & \multicolumn{1}{l|}{120} & \multicolumn{1}{l|}{Synthetic Image} & Visual Recognition,Visual Reasoning \\ \midrule
\multicolumn{1}{l|}{Color Contrast} & \multicolumn{1}{l|}{200} & \multicolumn{1}{l|}{Synthetic Image} & Visual Recognition,Visual Reasoning \\ \midrule
\multicolumn{4}{c}{\textbf{Meme Understanding}} \\ \midrule
\multicolumn{1}{l|}{Meme Vedio Understanding} & \multicolumn{1}{l|}{200} & \multicolumn{1}{l|}{Natural Image} & Visual Description \\ \midrule
\multicolumn{1}{l|}{Meme Image Understanding} & \multicolumn{1}{l|}{200} & \multicolumn{1}{l|}{Synthetic Image} & Visual Description \\ \midrule
\multicolumn{4}{c}{\textbf{Counting}} \\ \midrule
\multicolumn{1}{l|}{Counting By Visual Prompting} & \multicolumn{1}{l|}{200} & \multicolumn{1}{l|}{Natural Image} & Visual Recognition,Counting \\ \midrule
\multicolumn{1}{l|}{Counting By Category} & \multicolumn{1}{l|}{800} & \multicolumn{1}{l|}{Natural Image} & Visual Recognition,Counting \\ \midrule
\multicolumn{1}{l|}{Crowd Counting} & \multicolumn{1}{l|}{200} & \multicolumn{1}{l|}{Natural Image} & Visual Recognition,Counting \\ \midrule
\multicolumn{1}{l|}{Counting By Reasoning} & \multicolumn{1}{l|}{200} & \multicolumn{1}{l|}{Natural Image} & Visual Recognition,Counting \\ \midrule
\multicolumn{4}{c}{\textbf{Hallucination}} \\ \midrule
\multicolumn{1}{l|}{Order Hallucination} & \multicolumn{1}{l|}{200} & \multicolumn{1}{l|}{Natural Image} & Visual Recognition,Visual Reasoning,Visual Description \\ \midrule
\multicolumn{1}{l|}{Relation Hallucination} & \multicolumn{1}{l|}{200} & \multicolumn{1}{l|}{Natural Image} & Visual Recognition,Visual Reasoning,Visual Description \\ \midrule
\multicolumn{1}{l|}{Attribute Hallucination} & \multicolumn{1}{l|}{200} & \multicolumn{1}{l|}{Natural Image} & Visual Recognition,Visual Reasoning,Visual Description \\ \midrule
\multicolumn{1}{l|}{Exist Hallucination} & \multicolumn{1}{l|}{200} & \multicolumn{1}{l|}{Natural Image} & Visual Recognition,Visual Reasoning \\ \midrule
\multicolumn{4}{c}{\textbf{Image Retrieval}} \\ \midrule
\multicolumn{1}{l|}{Person Reid} & \multicolumn{1}{l|}{200} & \multicolumn{1}{l|}{Natural Image} & Retrieval,Multi-Images Analysis \\ \midrule
\multicolumn{1}{l|}{Sketch2image Retrieval} & \multicolumn{1}{l|}{200} & \multicolumn{1}{l|}{Natural Image,Text-rich Image} & Retrieval,Multi-Images Analysis \\ \midrule
\multicolumn{1}{l|}{Face Retrieval} & \multicolumn{1}{l|}{200} & \multicolumn{1}{l|}{Natural Image} & Retrieval,Multi-Images Analysis \\ \midrule
\multicolumn{1}{l|}{Handwritten Retrieval} & \multicolumn{1}{l|}{200} & \multicolumn{1}{l|}{Text-rich Image} & Retrieval,OCR,Multi-Images Analysis \\ \midrule
\multicolumn{1}{l|}{Vehicle Retrieval} & \multicolumn{1}{l|}{200} & \multicolumn{1}{l|}{Natural Image} & Retrieval,Multi-Images Analysis \\ \midrule
\multicolumn{1}{l|}{Image2image Retrieval} & \multicolumn{1}{l|}{200} & \multicolumn{1}{l|}{Natural Image} & Retrieval,Multi-Images Analysis \\ \midrule
\multicolumn{1}{l|}{Text2image Retrieval} & \multicolumn{1}{l|}{200} & \multicolumn{1}{l|}{Natural Image} & Retrieval,Multi-Images Analysis \\ \midrule
\multicolumn{4}{c}{\textbf{Visual Prompt Understanding}} \\ \midrule
\multicolumn{1}{l|}{Som Recognition} & \multicolumn{1}{l|}{199} & \multicolumn{1}{l|}{Natural Image,Visual Mark} & Visual Recognition,Visual Reasoning,Visual Localization,Visual Prompting Understanding \\ \midrule
\multicolumn{1}{l|}{Visual Prompt Understanding} & \multicolumn{1}{l|}{200} & \multicolumn{1}{l|}{Natural Image,Visual Mark} & Visual Recognition,Visual Reasoning,Visual Localization,Visual Prompting Understanding \\ \bottomrule
\end{tabular}%
}
\end{table*}

% \usepackage{graphicx}
\begin{table*}[]
\centering
\caption{MMT-Bench subtask details (part 3): including sample number, visual input types, and evaluated LVLM capabilities.}
\label{tab:task_3}
\resizebox{0.8\textwidth}{!}{%
\begin{tabular}{|llll|}
\toprule
\multicolumn{1}{l|}{Subtask Name} & \multicolumn{1}{l|}{Sample Num} & \multicolumn{1}{l|}{Visual Input Type} & Capability \\ \midrule
\multicolumn{4}{c}{\textbf{Anomaly Detection}} \\ \midrule
\multicolumn{1}{l|}{Industrial Produce Anomaly Detection} & \multicolumn{1}{l|}{200} & \multicolumn{1}{l|}{Natural Image} & Visual Recognition,Counting \\ \midrule
\multicolumn{1}{l|}{Face Mask Anomaly Dectection} & \multicolumn{1}{l|}{200} & \multicolumn{1}{l|}{Natural Image} & Visual Recognition \\ \midrule
\multicolumn{1}{l|}{Helmet Anomaly Detection} & \multicolumn{1}{l|}{200} & \multicolumn{1}{l|}{Natural Image} & Visual Recognition,Visual Localization \\ \midrule
\multicolumn{1}{l|}{Behavior Anomaly Detection} & \multicolumn{1}{l|}{200} & \multicolumn{1}{l|}{Natural Image} & Visual Recognition,Multi-Images Analysis \\ \midrule
\multicolumn{1}{l|}{Traffic Anomaly Detection} & \multicolumn{1}{l|}{200} & \multicolumn{1}{l|}{Natural Image} & Visual Recognition \\ \midrule
\multicolumn{4}{c}{\textbf{Keypoint Detection}} \\ \midrule
\multicolumn{1}{l|}{Furniture Keypoint Detection} & \multicolumn{1}{l|}{200} & \multicolumn{1}{l|}{Natural Image} & Visual Recognition,Visual Localization,Pixel Perception \\ \midrule
\multicolumn{1}{l|}{Human Keypoint Detection} & \multicolumn{1}{l|}{200} & \multicolumn{1}{l|}{Natural Image} & Visual Recognition,Visual Localization,Pixel Perception \\ \midrule
\multicolumn{1}{l|}{Clothes Keypoint Detection} & \multicolumn{1}{l|}{200} & \multicolumn{1}{l|}{Natural Image} & Visual Recognition,Visual Localization,Pixel Perception \\ \midrule
\multicolumn{1}{l|}{Animal Keypoint Detection} & \multicolumn{1}{l|}{200} & \multicolumn{1}{l|}{Natural Image} & Visual Recognition,Visual Localization,Pixel Perception \\ \midrule
\multicolumn{1}{l|}{Vehicle Keypoint Detection} & \multicolumn{1}{l|}{92} & \multicolumn{1}{l|}{Natural Image} & Visual Recognition,Visual Localization,Pixel Perception \\ \midrule
\multicolumn{4}{c}{\textbf{Visual Commonsense Reasoning}} \\ \midrule
\multicolumn{1}{l|}{Whoops} & \multicolumn{1}{l|}{200} & \multicolumn{1}{l|}{Synthetic Image} & Visual Recognition,Visual Reasoning \\ \midrule
\multicolumn{4}{c}{\textbf{Visual Code}} \\ \midrule
\multicolumn{1}{l|}{Eqn2latex} & \multicolumn{1}{l|}{200} & \multicolumn{1}{l|}{Text-rich Image,Scientific Diagram} & OCR,Document Understanding,Expert Knowledge Utilization \\ \midrule
\multicolumn{1}{l|}{Screenshot2code} & \multicolumn{1}{l|}{200} & \multicolumn{1}{l|}{Screenshot Image} & Document Understanding,Expert Knowledge Utilization \\ \midrule
\multicolumn{1}{l|}{Sketch2code} & \multicolumn{1}{l|}{200} & \multicolumn{1}{l|}{Scientific Diagram} & Document Understanding,Expert Knowledge Utilization \\ \midrule
\multicolumn{4}{c}{\textbf{Image Evaluation Judgement}} \\ \midrule
\multicolumn{1}{l|}{Image Quality Assessment} & \multicolumn{1}{l|}{200} & \multicolumn{1}{l|}{Natural Image} & Visual Reasoning \\ \midrule
\multicolumn{1}{l|}{Lvlm Response Judgement} & \multicolumn{1}{l|}{200} & \multicolumn{1}{l|}{Synthetic Image,Chart Image} & Visual Reasoning \\ \midrule
\multicolumn{4}{c}{\textbf{Pixel Level Perception}} \\ \midrule
\multicolumn{1}{l|}{Polygon Localization} & \multicolumn{1}{l|}{200} & \multicolumn{1}{l|}{Natural Image} & Visual Recognition,Visual Localization,Pixel Perception \\ \midrule
\multicolumn{1}{l|}{Interactive Segmentation} & \multicolumn{1}{l|}{141} & \multicolumn{1}{l|}{Natural Image} & Visual Localization,Pixel Perception \\ \midrule
\multicolumn{1}{l|}{Depth Estimation} & \multicolumn{1}{l|}{200} & \multicolumn{1}{l|}{Natural Image} & Pixel Perception,3D Perception \\ \midrule
\multicolumn{1}{l|}{Pixel Recognition} & \multicolumn{1}{l|}{200} & \multicolumn{1}{l|}{Natural Image} & Visual Recognition,Pixel Perception \\ \midrule
\multicolumn{1}{l|}{Pixel Localization} & \multicolumn{1}{l|}{200} & \multicolumn{1}{l|}{Natural Image} & Visual Recognition,Visual Localization,Pixel Perception \\ \midrule
\multicolumn{1}{l|}{Image Matting} & \multicolumn{1}{l|}{200} & \multicolumn{1}{l|}{Natural Image} & Pixel Perception \\ \midrule
\multicolumn{4}{c}{\textbf{Multiple Image Analysis}} \\ \midrule
\multicolumn{1}{l|}{Spot The Similarity} & \multicolumn{1}{l|}{200} & \multicolumn{1}{l|}{Natural Image,Synthetic Image} & Multi-Images Analysis \\ \midrule
\multicolumn{1}{l|}{Spot The Diff} & \multicolumn{1}{l|}{200} & \multicolumn{1}{l|}{Natural Image} & Multi-Images Analysis \\ \midrule
\multicolumn{4}{c}{\textbf{3D}} \\ \midrule
\multicolumn{1}{l|}{3D Cad Recognition} & \multicolumn{1}{l|}{200} & \multicolumn{1}{l|}{3d Image} & Multi-Images Analysis,3D Perception \\ \midrule
\multicolumn{1}{l|}{3D Indoor Recognition} & \multicolumn{1}{l|}{200} & \multicolumn{1}{l|}{3d Image} & Multi-Images Analysis,3D Perception \\ \midrule
\multicolumn{4}{c}{\textbf{Medical Understanding}} \\ \midrule
\multicolumn{1}{l|}{Anatomy Identification} & \multicolumn{1}{l|}{200} & \multicolumn{1}{l|}{Medical Image} & Visual Recognition,Expert Knowledge Utilization \\ \midrule
\multicolumn{1}{l|}{Medical Modality Recognition} & \multicolumn{1}{l|}{200} & \multicolumn{1}{l|}{Medical Image} & Visual Recognition,Expert Knowledge Utilization \\ \midrule
\multicolumn{1}{l|}{Other Biological Attributes} & \multicolumn{1}{l|}{200} & \multicolumn{1}{l|}{Medical Image} & Visual Recognition,Expert Knowledge Utilization \\ \midrule
\multicolumn{1}{l|}{Disease Diagnose} & \multicolumn{1}{l|}{200} & \multicolumn{1}{l|}{Medical Image} & Visual Recognition,Expert Knowledge Utilization \\ \midrule
\multicolumn{1}{l|}{Lesion Grading} & \multicolumn{1}{l|}{200} & \multicolumn{1}{l|}{Medical Image} & Visual Recognition,Expert Knowledge Utilization \\ \midrule
\multicolumn{4}{c}{\textbf{Cross Image Matching}} \\ \midrule
\multicolumn{1}{l|}{One Shot Detection} & \multicolumn{1}{l|}{200} & \multicolumn{1}{l|}{Natural Image} & Visual Localization \\ \midrule
\multicolumn{1}{l|}{Point Tracking} & \multicolumn{1}{l|}{200} & \multicolumn{1}{l|}{Natural Image} & Visual Localization \\ \midrule
\multicolumn{1}{l|}{Single Object Tracking} & \multicolumn{1}{l|}{200} & \multicolumn{1}{l|}{Natural Image} & Visual Localization \\ \midrule
\multicolumn{4}{c}{\textbf{Visual Captioning}} \\ \midrule
\multicolumn{1}{l|}{Video Captioning} & \multicolumn{1}{l|}{200} & \multicolumn{1}{l|}{Natural Image} & Visual Description,Temporal Understanding \\ \midrule
\multicolumn{1}{l|}{Image Captioning Paragraph} & \multicolumn{1}{l|}{200} & \multicolumn{1}{l|}{Natural Image} & Visual Description \\ \midrule
\multicolumn{1}{l|}{Image Captioning} & \multicolumn{1}{l|}{200} & \multicolumn{1}{l|}{Natural Image} & Visual Description \\ \midrule
\multicolumn{1}{l|}{Instance Captioning} & \multicolumn{1}{l|}{200} & \multicolumn{1}{l|}{Natural Image} & Visual Description \\ \midrule
\multicolumn{1}{l|}{Image Dense Captioning} & \multicolumn{1}{l|}{197} & \multicolumn{1}{l|}{Natural Image} & Visual Description \\ \midrule
\multicolumn{1}{l|}{Multiple Instance Captioning} & \multicolumn{1}{l|}{200} & \multicolumn{1}{l|}{Natural Image} & Visual Description \\ \midrule
\multicolumn{1}{l|}{Multiple Image Captioning} & \multicolumn{1}{l|}{200} & \multicolumn{1}{l|}{Natural Image} & Visual Description,Multi-Images Analysis \\ \midrule
\multicolumn{1}{l|}{Writing Poetry From Image} & \multicolumn{1}{l|}{200} & \multicolumn{1}{l|}{Natural Image,Text-rich Image} & Visual Description \\ \midrule
\multicolumn{4}{c}{\textbf{Autonomous Driving}} \\ \midrule
\multicolumn{1}{l|}{Traffic Participants Understanding} & \multicolumn{1}{l|}{200} & \multicolumn{1}{l|}{Natural Image} & Counting \\ \midrule
\multicolumn{1}{l|}{Multiple View Image Understanding} & \multicolumn{1}{l|}{200} & \multicolumn{1}{l|}{Natural Image} & Visual Reasoning,Multi-Images Analysis,Counting \\ \midrule
\multicolumn{1}{l|}{Traffic Sign Understanding} & \multicolumn{1}{l|}{200} & \multicolumn{1}{l|}{Natural Image} & Visual Reasoning,Expert Knowledge Utilization \\ \midrule
\multicolumn{1}{l|}{Temporal Sequence Understanding} & \multicolumn{1}{l|}{200} & \multicolumn{1}{l|}{Natural Image} & Visual Reasoning,Temporal Understanding \\ \midrule
\multicolumn{1}{l|}{Traffic Light Understanding} & \multicolumn{1}{l|}{200} & \multicolumn{1}{l|}{Natural Image} & Visual Recognition \\ \bottomrule
\end{tabular}%
}
\end{table*}

\begin{table*}[!ht]
    \centering
    \caption{Model architecture of $30$ LVLMs evaluated on MMT-Bench.}
    \label{tab:lvlm}
    \begin{tabular}{llll}
        \toprule
        Models & Parameters & Vision Encoder & LLM \\ 
        \midrule
        GPT-4V \cite{gpt4v} & - & - & - \\ 
        GeminiProVision \cite{gemini} & - & - & - \\ 
        QWen-VL-Plus \cite{qwen} & - & - & - \\ 
        Claude3V-Haiku \cite{Claude2023} & - & - & - \\ 
        LLaVA-Next-34B \cite{llava_next} &34.8B & CLIP ViT-L/14 & Nous-Hermes-2-Yi-34B \\ 
        LLaVA-Next-13B \cite{llava_next} &13.4B & CLIP ViT-L/14 & Vicuna-v1.5-13B \\ 
        LLaVA-Next-7B \cite{llava_next} &7.1B & CLIP ViT-L/14 & Vicuna-v1.5-7B \\ 
        Yi-VL-34B \cite{ai2024yi} &34.6B & CLIP ViT-H/14 & Nous-Hermes-2-Yi-34B \\ 
        Yi-VL-6B \cite{ai2024yi} & 6.6B & CLIP ViT-H/14 & Yi-6B \\ 
        InternVL-Chat-V1.2 \cite{chen2023internvl} & 40B & InternViT-6B & Nous-Hermes-2-Yi-34B \\ 
        DeepSeek-VL-7B \cite{lu2024deepseekvl} & 7.3B & SAM-B \& SigLIP-L & DeekSeek-7B \\  
        Monkey \cite{li2023monkey} & 9.8B & CLIP-ViT-BigHuge & Qwen-7B \\ 
        XComposer \cite{internlmxcomposer} & 8B & EVA-CLIP-G & InternLM-7B \\ 
        XComposer2 \cite{internlmxcomposer2}& 7B & CLIP ViT-L/14 & InternLM2-7B \\ 
        ShareGPT4V \cite{chen2023sharegpt4v}& 7.2B & CLIP ViT-L/14 & Vicuna-v1.5-7B \\ 
        SharedCaptioner \cite{chen2023sharegpt4v} & 8B & EVA-G & InternLM-7B \\ 
        mPLUG-Owl2 \cite{ye2023mplugowl2}& 8.2B & CLIP ViT-L/14 & LLaMA2-7B \\ 
        LLaVA-v1.5-7B \cite{liu2023llava, liu2023improvedllava}& 7.2B & CLIP ViT-L/14 & Vicuna-v1.5-7B \\ 
        LLaVA-v1.5-13B \cite{liu2023llava, liu2023improvedllava}& 13.4B & CLIP ViT-L/14 & Vicuna-v1.5-13B \\ 
        LLaVA-InternLM2-7B \cite{2023xtuner} & 8.1B & CLIP ViT-L/14 & InternLM2-7B \\ 
        LLaVA-InternLM-7B \cite{2023xtuner} & 7.6B & CLIP ViT-L/14 & InternLM-7B \\ 
        LLaVA-v1.5-7B-Xtuner \cite{2023xtuner} & 7.2B & CLIP ViT-L/14 & Vicuna-v1.5-7B \\ 
        LLaVA-v1.5-13B-Xtuner \cite{2023xtuner} & 13.4B & CLIP ViT-L/14 & Vicuna-v1.5-13B \\ 
        LLaMA-Adapter-v2 \cite{gao2023llamaadapterv2}& 7B & CLIP-ViT-L/14 & LLaMA-7B \\ 
        VisualGLM \cite{ding2021cogview}& 8B & EVA-CLIP & ChatGLM-6B \\ 
        CogVLM \cite{wang2023cogvlm}& 17B & EVA-CLIP-E & Vicuna-v1.5-7B \\ 
        TransCore-M \cite{transcore-m}& 13.4B & CLIP ViT/L-14 & PCITransGPT-13B \\ 
        RBDash-v1 \cite{RBDash}& 13.4B & CLIP ViT-L/14 & Vicuna-v1.5-13B \\ 
        BLIP2 \cite{li2023blip2} & 12.1B & EVA-CLIP ViT-G/14 & Flan-T5-XXL \\ 
        QWenVL \cite{Qwen-VL} & 9.6B & CLIP ViT-G/16 & QWen-7B \\
        
        \bottomrule
    \end{tabular}
\end{table*}

%%%%%%%%%%%%%%%%%%%%%%%%%%%%%%%%%%%%%%%%%%%%%%%%%%%%%%%%%%%%%%%%%%%%%%%%%%%%%%%
%%%%%%%%%%%%%%%%%%%%%%%%%%%%%%%%%%%%%%%%%%%%%%%%%%%%%%%%%%%%%%%%%%%%%%%%%%%%%%%

% \begin{table}[!t]
% \centering
% \caption{Key statistics of the MMT-Bench.}
% \begin{tabular}{ll}
% Statistics & Number \\ \hline
% Total Questions & 31325 \\
% Total Meta-tasks & 36 \\
% Total Tasks & 162 \\
% Modality & \multicolumn{1}{r}{Image\&Text\&Video\&Point cloud} \\ \hline
% Tasks with Multiple Images & 28 \\ 
% Total Questions with Multiple Images& 5593 \\
% \hline
% Average question length &  35.50\\
% Average option length &  3.08\\
% \hline
% \end{tabular}
% \end{table}

\begin{table*}[]
\caption{Abbreviations for tasks requiring multiple images as inputs (part one). Here we also present the designed prompt examples we used for Single-Image LVLMs and Multi-Images LVLMs.}
\label{tab:multi-image-prompt-part1}
{
\linespread{1.0} 
\setlength{\tabcolsep}{1pt}
\scriptsize
\begin{tabular}{ccllllllllllllllllllllllllllllllllllll}
\multicolumn{1}{c|}{\multirow{2}{*}{\textbf{Task Abbreviation}}} & \multicolumn{1}{c|}{\multirow{2}{*}{\textbf{Task Name}}} & \multicolumn{1}{l|}{\multirow{2}{*}{\textbf{Prompt Example for Single   Image LVLMs}}} & \multirow{2}{*}{\textbf{Prompt example for   Multiple Image LVLMs}} &  &  &  &  &  &  &  &  &  &  &  &  &  &  &  &  &  &  &  &  &  &  &  &  &  &  &  &  &  &  &  &  &  &  \\
\multicolumn{1}{c|}{} & \multicolumn{1}{c|}{} & \multicolumn{1}{l|}{} &  &  &  &  &  &  &  &  &  &  &  &  &  &  &  &  &  &  &  &  &  &  &  &  &  &  &  &  &  &  &  &  &  &  &  \\ \cline{1-4}
\multicolumn{1}{c|}{\textbf{AQS}} & \multicolumn{1}{c|}{\begin{tabular}[c]{@{}c@{}}action quality \\ assessment\end{tabular}} & \multicolumn{1}{l|}{\begin{tabular}[c]{@{}l@{}}Question: \textless{}image\textgreater{}\textless{}image\textgreater{}\textless{}image\textgreater{}\textless{}image\textgreater\\ What is   the most probable action quality assessment \\ number obtained by the person in   the video?\\      Options:\\      A. 35.99\\      B. 28.0\\      C. 11.27\\      D. 44.98\end{tabular}} & \begin{tabular}[c]{@{}l@{}}Question: \textless{}image\textgreater{}\textless{}image\textgreater{}\textless{}image\textgreater{}\textless{}image\textgreater\\ What is   the most probable action quality assessment \\ number obtained by the person in   the video?\\      Options:\\      A. 35.99\\      B. 28.0\\      C. 11.27\\      D. 44.98\end{tabular} &  &  &  &  &  &  &  &  &  &  &  &  &  &  &  &  &  &  &  &  &  &  &  &  &  &  &  &  &  &  &  &  &  &  \\ \cline{1-4}
\multicolumn{1}{c|}{\textbf{FECR}} & \multicolumn{1}{c|}{\begin{tabular}[c]{@{}c@{}}facail expression \\ change recognition\end{tabular}} & \multicolumn{1}{l|}{\begin{tabular}[c]{@{}l@{}}Question: \textless{}image\textgreater{}\textless{}image\textgreater{}What is the change \\ of expression from   the first image to the second image?\\      Options:\\      A. disgust to happy \\      B. happy to sadness \\      C. anger to surprise \\      D. disgust to fear\end{tabular}} & \begin{tabular}[c]{@{}l@{}}Question: What is the change of expression from \\ Image 1: \textless{}image\textgreater to   Image 2: \textless{}image\textgreater{}?\\      Options:\\      A. disgust to happy \\      B. happy to sadness \\      C. anger to surprise \\      D. disgust to fear\end{tabular} &  &  &  &  &  &  &  &  &  &  &  &  &  &  &  &  &  &  &  &  &  &  &  &  &  &  &  &  &  &  &  &  &  &  \\ \cline{1-4}
\multicolumn{1}{c|}{\textbf{FR}} & \multicolumn{1}{c|}{\begin{tabular}[c]{@{}c@{}}face \\ retrieval\end{tabular}} & \multicolumn{1}{l|}{\begin{tabular}[c]{@{}l@{}}Question:   \textless{}image\textgreater{}\textless{}image\textgreater{}\textless{}image\textgreater{}\textless{}image\textgreater\\ \textless{}image\textgreater{}Please retrieve the most similar person to the query \\ in the candidates.  The first   image is the query image and \\ the remaining images are candidates from   Candidate 1 to \\ Candidate 4.\\      Options: \\      A. Candidate 1 \\      B. Candidate 2 \\      C. Candidate 3 \\      D. Candidate 4\end{tabular}} & \begin{tabular}[c]{@{}l@{}}Question: Please retrieve the most similar person to the query:   \\ \textless{}image\textgreater in the candidates:  Candidate 1: \textless{}image\textgreater{}, \\ Candidate 2:   \textless{}image\textgreater{}, Candidate 3: \textless{}image\textgreater{}, \\ Candidate 4:   \textless{}image\textgreater{}.\\      Options: \\      A. Candidate 1 \\      B. Candidate 2 \\      C. Candidate 3 \\      D. Candidate 4\end{tabular} &  &  &  &  &  &  &  &  &  &  &  &  &  &  &  &  &  &  &  &  &  &  &  &  &  &  &  &  &  &  &  &  &  &  \\ \cline{1-4}
\multicolumn{1}{c|}{\textbf{GAR}} & \multicolumn{1}{c|}{\begin{tabular}[c]{@{}c@{}}general action \\ recognition\end{tabular}} & \multicolumn{1}{l|}{\begin{tabular}[c]{@{}l@{}}Question: \textless{}image\textgreater{}\textless{}image\textgreater{}\textless{}image\textgreater{}\textless{}image\textgreater\\ What is   the action performed by the person in the video?\\      Options: \\      A. rock scissors paper \\      B. sword fighting \\      C. fencing \\      D. balloon blowing\end{tabular}} & \begin{tabular}[c]{@{}l@{}}Question: \textless{}image\textgreater{}\textless{}image\textgreater{}\textless{}image\textgreater{}\textless{}image\textgreater\\ What is   the action performed by the person in the video?\\      Options: \\      A. rock scissors paper \\      B. sword fighting \\      C. fencing \\      D. balloon blowing\end{tabular} &  &  &  &  &  &  &  &  &  &  &  &  &  &  &  &  &  &  &  &  &  &  &  &  &  &  &  &  &  &  &  &  &  &  \\ \cline{1-4}
\multicolumn{1}{c|}{\textbf{HR}} & \multicolumn{1}{c|}{\begin{tabular}[c]{@{}c@{}}handwritten \\ retrieval\end{tabular}} & \multicolumn{1}{l|}{\begin{tabular}[c]{@{}l@{}}Question:   \textless{}image\textgreater{}\textless{}image\textgreater{}\textless{}image\textgreater{}\textless{}image\textgreater\\ \textless{}image\textgreater{}Please  retrieve the most similar handwritten \\ text snapshot to the query in the  candidates. \\    The first image is the query image and the remaining \\ images are   candidates from Candidate 1 to Candidate 4.\\      Options: \\      A. Candidate 1 \\      B. Candidate 2 \\      C. Candidate 3 \\      D. Candidate 4\end{tabular}} & \begin{tabular}[c]{@{}l@{}}Question: Please retrieve the most similar handwritten text \\ snapshot to the   query: \textless{}image\textgreater in the candidates: \\    Candidate 1: \textless{}image\textgreater{},   Candidate 2: \textless{}image\textgreater{}, \\ Candidate 3: \textless{}image\textgreater{}, Candidate 4:   \textless{}image\textgreater{}.\\      Options: \\      A. Candidate 1 \\      B. Candidate 2 \\      C. Candidate 3 \\      D. Candidate 4\end{tabular} &  &  &  &  &  &  &  &  &  &  &  &  &  &  &  &  &  &  &  &  &  &  &  &  &  &  &  &  &  &  &  &  &  &  \\ \cline{1-4}
\multicolumn{1}{c|}{\textbf{I2IR}} & \multicolumn{1}{c|}{\begin{tabular}[c]{@{}c@{}}image2image \\ retrieval\end{tabular}} & \multicolumn{1}{l|}{\begin{tabular}[c]{@{}l@{}}Question:   \textless{}image\textgreater{}\textless{}image\textgreater{}\textless{}image\textgreater{}\textless{}image\textgreater\\ \textless{}image\textgreater{}Please   retrieve the most similar scene to the query\\  in the candidates. The first   image is the query image \\ and the remaining images are candidates from   Candidate 1 \\ to Candidate 4.\\      Options: \\      A. Candidate 1\\       B. Candidate 2 \\      C. Candidate 3 \\      D. Candidate 4\end{tabular}} & \begin{tabular}[c]{@{}l@{}}Question: Please retrieve the   most similar scene to the \\ query: \textless{}image\textgreater in the candidates: Candidate   1: \textless{}image\textgreater{}, \\ Candidate 2: \textless{}image\textgreater{}, Candidate 3:   \textless{}image\textgreater{}, \\ Candidate 4: \textless{}image\textgreater{}.\\      Options: \\      A. Candidate 1\\       B. Candidate 2 \\      C. Candidate 3 \\      D. Candidate 4\end{tabular} &  &  &  &  &  &  &  &  &  &  &  &  &  &  &  &  &  &  &  &  &  &  &  &  &  &  &  &  &  &  &  &  &  &  \\ \cline{1-4}
\multicolumn{1}{c|}{\textbf{IC}} & \multicolumn{1}{c|}{\begin{tabular}[c]{@{}c@{}}image \\ colorization\end{tabular}} & \multicolumn{1}{l|}{\begin{tabular}[c]{@{}l@{}}Question: \textless{}image\textgreater{}\textless{}image\textgreater{}\textless{}image\textgreater{}\textless{}image\textgreater\\   The following images are candidates from Candidate 1 \\ to Candidate 4,  which are from the same picture \\ consisting of four styles: grayscale, original, warm,  and sepia. \\ Which one is the original picture?\\      Options: \\      A. Candidate 1 \\      B. Candidate 2 \\      C. Candidate 3 \\      D. Candidate 4\end{tabular}} & \begin{tabular}[c]{@{}l@{}}Question: The following images: Candidate 1: \textless{}image\textgreater{}, \\ Candidate 2:   \textless{}image\textgreater{}, Candidate 3: \textless{}image\textgreater{}, \\ Candidate 4: \textless{}image\textgreater{},   are from the same picture, \\    which consists of four styles: grayscale,   original, \\ warm, and sepia. Which one is the original picture?\\      Options: \\      A. Candidate 1 \\      B. Candidate 2 \\      C. Candidate 3 \\      D. Candidate 4\end{tabular} &  &  &  &  &  &  &  &  &  &  &  &  &  &  &  &  &  &  &  &  &  &  &  &  &  &  &  &  &  &  &  &  &  &  \\ \cline{1-4}
\multicolumn{1}{c|}{\textbf{MVU}} & \multicolumn{1}{c|}{\begin{tabular}[c]{@{}c@{}}meme video \\ understanding\end{tabular}} & \multicolumn{1}{l|}{\begin{tabular}[c]{@{}l@{}}Question: \textless{}image\textgreater{}\textless{}image\textgreater{}\textless{}image\textgreater{}\textless{}image\textgreater\\   Please   generate a description for this meme\\      Options: \\      A. From beneath the toilet door panel, a hand is reaching \\           out with an   upward-facing palm to receive chopsticks \\            and a spoon from someone outside.   \\      B. The hand is asking for help to get out of the bathroom. \\      C. The hand is actually reaching out for a handshake. \\      D. A person is handing over toilet paper instead of \\ chopsticks and a spoon.\end{tabular}} & \begin{tabular}[c]{@{}l@{}}Question: \textless{}image\textgreater{}\textless{}image\textgreater{}\textless{}image\textgreater{}\textless{}image\textgreater\\ Please   generate a description for this meme\\      Options: \\      A. From beneath the toilet door panel, a hand is \\ reaching out with an   upward-facing palm to receive \\ chopsticks and a spoon from someone outside.   \\      B. The hand is asking for help to get out of the bathroom. \\      C. The hand is actually reaching out for a handshake. \\      D. A person is handing over toilet paper instead of \\ chopsticks and a spoon.\end{tabular} &  &  &  &  &  &  &  &  &  &  &  &  &  &  &  &  &  &  &  &  &  &  &  &  &  &  &  &  &  &  &  &  &  &  \\ \cline{1-4}
\end{tabular}
}
\end{table*}

\begin{table*}[]
\caption{Abbreviations for tasks requiring multiple images as inputs (part two). Here we also present the designed prompt examples we used for Single-Image LVLMs and Multi-Images LVLMs.}
\label{tab:multi-image-prompt-part2}
{
\linespread{1.0} 
\setlength{\tabcolsep}{1pt}
\scriptsize
\begin{tabular}{ccllllllllllllllllllllllllllllllllllll}
\multicolumn{1}{c|}{\multirow{2}{*}{\textbf{Task Abbreviation}}} & \multicolumn{1}{c|}{\multirow{2}{*}{\textbf{Task Name}}} & \multicolumn{1}{l|}{\multirow{2}{*}{\textbf{Prompt Example for Single   Image LVLMs}}} & \multirow{2}{*}{\textbf{Prompt example for   Multiple Image LVLMs}} &  &  &  &  &  &  &  &  &  &  &  &  &  &  &  &  &  &  &  &  &  &  &  &  &  &  &  &  &  &  &  &  &  &  \\
\multicolumn{1}{c|}{} & \multicolumn{1}{c|}{} & \multicolumn{1}{l|}{} &  &  &  &  &  &  &  &  &  &  &  &  &  &  &  &  &  &  &  &  &  &  &  &  &  &  &  &  &  &  &  &  &  &  &  \\ \cline{1-4}
\multicolumn{1}{c|}{\textbf{ME}} & \multicolumn{1}{c|}{mevis} & \multicolumn{1}{l|}{\begin{tabular}[c]{@{}l@{}}Question:   \textless{}image\textgreater{}\textless{}image\textgreater{}\textless{}image\textgreater{}\textless{}image\textgreater\\ \textless{}image\textgreater{}\textless{}image\textgreater{}\textless{}image\textgreater{}\textless{}image\textgreater{}\textless{}image\textgreater\\ I   have provided several frames from a video, and \\ I will also provide a caption.   Provide the output for \\ the detected area in the format {[}x, y, w, h{]}. \\ This   format represents the bounding box,  \\ where {[}x, y, w, h{]} are the coordinates  of   the top-left \\ corner of the bounding box, as well as its width and height.   \\ Note that the width of the input image is 1920 and \\ the height is 945.\\      CAPTION: little girl feeding rabbit\\      Options: \\      A. {[}70, 0, 993, 1007{]} \\      B. {[}203, 0, 1011, 944{]} \\      C. {[}70, 0, 1011, 944{]} \\      D. {[}196, 38, 652, 277{]}\end{tabular}} & \begin{tabular}[c]{@{}l@{}}Question:   \textless{}image\textgreater{}\textless{}image\textgreater{}\textless{}image\textgreater{}\textless{}image\textgreater\\ \textless{}image\textgreater{}\textless{}image\textgreater{}\textless{}image\textgreater{}\textless{}image\textgreater{}\textless{}image\textgreater\\ I   have provided several frames from a video, and I will also \\ provide a caption.   Provide the output for the detected area \\ in the format {[}x, y, w, h{]}. This   format represents the bounding box, \\   where {[}x, y, w, h{]} are the coordinates of   the top-left corner \\ of the bounding box, as well as its width and height.   \\   Note that the width of the input image is 1920 and the height is 945.\\      CAPTION: little girl feeding rabbit\\      Options: \\      A. {[}70, 0, 993, 1007{]} \\      B. {[}203, 0, 1011, 944{]} \\      C. {[}70, 0, 1011, 944{]} \\      D. {[}196, 38, 652, 277{]}\end{tabular} &  &  &  &  &  &  &  &  &  &  &  &  &  &  &  &  &  &  &  &  &  &  &  &  &  &  &  &  &  &  &  &  &  &  \\ \cline{1-4}
\multicolumn{1}{c|}{\textbf{MIC}} & \multicolumn{1}{c|}{\begin{tabular}[c]{@{}c@{}}multiple image \\ captioning\end{tabular}} & \multicolumn{1}{l|}{\begin{tabular}[c]{@{}l@{}}Question:   \textless{}image\textgreater{}\textless{}image\textgreater{}\textless{}image\textgreater{}\textless{}image\textgreater\\ \textless{}image\textgreater{}Describe   this set of images briefly.\\      Options: \\      A. I took a cab to return to the hotel \\      B. the front of the mall was somewhat crowded .\\         i ran past them and took the   escalator down .\\         after shopping for a few hours , i returned to the street .\\         i   tried to catch a cab but a bush blocked me .\\         i decided to just walk back to my   hotel . \\      C. the mall was empty and I took the stairs up \\      D. I quickly caught a bus to my hotel\end{tabular}} & \begin{tabular}[c]{@{}l@{}}Question: Describe this set of images:   \\ \textless{}image\textgreater{}\textless{}image\textgreater{}\textless{}image\textgreater{}\textless{}image\textgreater{}\textless{}image\textgreater   briefly.\\      Options: \\      A. I took a cab to return to the hotel \\      B. the front of the mall was somewhat crowded .\\ i ran past them and took the   escalator down .\\ after shopping for a few hours , i returned to the street .\\ i   tried to catch a cab but a bush blocked me .\\ i decided to just walk back to my   hotel . \\      C. the mall was empty and I took the stairs up \\      D. I quickly caught a bus to my hotel\end{tabular} &  &  &  &  &  &  &  &  &  &  &  &  &  &  &  &  &  &  &  &  &  &  &  &  &  &  &  &  &  &  &  &  &  &  \\ \cline{1-4} \cline{38-38} 
\multicolumn{1}{c|}{\textbf{NIP}} & \multicolumn{1}{c|}{\begin{tabular}[c]{@{}c@{}}next img \\ prediction\end{tabular}} & \multicolumn{1}{l|}{\begin{tabular}[c]{@{}l@{}}Question:   \textless{}image\textgreater{}\textless{}image\textgreater{}\textless{}image\textgreater{}\textless{}image\textgreater\\     \textless{}image\textgreater{}Please   predict the last 10 frames in the \\      candidates of the video based on the first   10 frames of \\     the input video.  Note that the order is from left to right. \\       The   first four images are candidates from Candidate 1 \\     to Candidate 4 and the last   image shows the first 10 frames \\     of the input video.\\      Options: \\      A. Candidate 1: last 10 frames \\      B. Candidate 2: last 10 frames \\      C. Candidate 3: last 10 frames \\      D. Candidate 4: last 10 frames\end{tabular}} & \begin{tabular}[c]{@{}l@{}}Question: Please predict the last 10 frames in the \\ candidates: Candidate 1:   \textless{}image\textgreater{}, Candidate 2: \textless{}image\textgreater{}, \\ Candidate 3: \textless{}image\textgreater{},   Candidate 4: \textless{}image\textgreater{}, of the video: \\    based on the first 10 frames of   the input video: \textless{}image\textgreater{}. \\ Note that the order is from left to   right\\      Options: \\      A. Candidate 1: last 10 frames \\      B. Candidate 2: last 10 frames \\      C. Candidate 3: last 10 frames \\      D. Candidate 4: last 10 frames\end{tabular} &  &  &  &  &  &  &  &  &  &  &  &  &  &  &  &  &  &  &  &  &  &  &  &  &  &  &  &  &  &  &  &  &  &  \\ \cline{1-4}
\multicolumn{1}{c|}{\textbf{OSD}} & \multicolumn{1}{c|}{\begin{tabular}[c]{@{}c@{}}one shot \\ detection\end{tabular}} & \multicolumn{1}{l|}{\begin{tabular}[c]{@{}l@{}}Question: \textless{}image\textgreater{}\textless{}image\textgreater{}According to the prompts in the   \\    Support Image (marked in red), please detect the corresponding \\    object in the   Query Image.  The first image is the Support Image \\     and the second image is the   Query Image. \\    Provide the output for the object in the format {[}x, y, w, h{]}.   \\    This format represents the bounding box, where {[}x, y, w, h{]} are the  \\     coordinates of the top-left corner of the bounding box, \\     as well as its width and height. \\    Note that the width of the input RGB image is 224 \\ and the height   is 224.\\      Options: \\      A. {[}0, 0, 511, 2{]} \\      B. {[}0, 0, 426, 1{]}\\      C. {[}1, 1, 511, 2{]} \\      D. {[}0, 0, 499, 2{]}\end{tabular}} & \begin{tabular}[c]{@{}l@{}}Question: According to the prompts in the Support Image \\ (marked in red):   \textless{}image\textgreater{}, please detect the corresponding\\  object in the Query Image:   \textless{}image\textgreater{}. \\    Provide the output for the object in the format {[}x, y, w,   h{]}. \\ This format represents the bounding box, \\    where {[}x, y, w, h{]} are the   coordinates of the top-left corner \\ of the bounding box, as well as its width   and height. \\    Note that the width of the input RGB image is 224 and \\ the height   is 224.\\      Options: \\      A. {[}0, 0, 511, 2{]} \\      B. {[}0, 0, 426, 1{]}\\      C. {[}1, 1, 511, 2{]} \\      D. {[}0, 0, 499, 2{]}\end{tabular} &  &  &  &  &  &  &  &  &  &  &  &  &  &  &  &  &  &  &  &  &  &  &  &  &  &  &  &  &  &  &  &  &  &  \\ \cline{1-4}
\multicolumn{1}{c|}{\textbf{PRe}} & \multicolumn{1}{c|}{person reid} & \multicolumn{1}{l|}{\begin{tabular}[c]{@{}l@{}}Question:   \textless{}image\textgreater{}\textless{}image\textgreater{}\textless{}image\textgreater{}\textless{}image\textgreater\\    \textless{}image\textgreater{}Please   retrieve the most similar person to \\   the query in the candidates. The first   image is the query \\   image and the remaining images are candidates from   \\   Candidate 1 to Candidate 4.\\      Options: \\      A. Candidate 1 \\      B. Candidate 2 \\      C. Candidate 3 \\      D. Candidate 4\end{tabular}} & \begin{tabular}[c]{@{}l@{}}Question: Please retrieve the most similar person to the \\ query:   \textless{}image\textgreater in the candidates: Candidate 1: \textless{}image\textgreater{}, \\ Candidate 2:   \textless{}image\textgreater{}, Candidate 3: \textless{}image\textgreater{}, \\ Candidate 4:   \textless{}image\textgreater{}.\\      Options: \\      A. Candidate 1 \\      B. Candidate 2 \\      C. Candidate 3 \\      D. Candidate 4\end{tabular} &  &  &  &  &  &  &  &  &  &  &  &  &  &  &  &  &  &  &  &  &  &  &  &  &  &  &  &  &  &  &  &  &  &  \\ \cline{1-4}
\multicolumn{1}{c|}{\textbf{PT}} & \multicolumn{1}{c|}{point tracking} & \multicolumn{1}{l|}{\begin{tabular}[c]{@{}l@{}}Question: \textless{}image\textgreater{}\textless{}image\textgreater{}What is the position coordinates \\   of   the point with coordinates ({[}0.711, 0.154{]}) in the first image \\     within the   second image?  Note that the width of the input \\    RGB image is 256 and the   height is 256.\\      Options: \\      A. {[}0.336, 0.241{]} \\      B. {[}0.754, 0.592{]} \\      C. {[}0.711, 0.154{]} \\      D. {[}0.814, 0.269{]}\end{tabular}} & \begin{tabular}[c]{@{}l@{}}Question: What is the position coordinates of the point \\ with coordinates   ({[}0.711, 0.154{]}) in Frame 1: \textless{}image\textgreater \\ within the Frame 2:   \textless{}image\textgreater{}? \\   Note that the width of the input RGB image is 256 \\ and the   height is 256.\\      Options: \\      A. {[}0.336, 0.241{]} \\      B. {[}0.754, 0.592{]} \\      C. {[}0.711, 0.154{]} \\      D. {[}0.814, 0.269{]}\end{tabular} &  &  &  &  &  &  &  &  &  &  &  &  &  &  &  &  &  &  &  &  &  &  &  &  &  &  &  &  &  &  &  &  &  &  \\ \cline{1-4}
\end{tabular}
% \end{table}
}
\end{table*}

\begin{table*}[]
\caption{Abbreviations for tasks requiring multiple images as inputs (part three). Here we also present the designed prompt examples we used for Single-Image LVLMs and Multi-Images LVLMs.}
\label{tab:multi-image-prompt-part3}
{
\linespread{1.0} 
\setlength{\tabcolsep}{1pt}
\scriptsize
% \usepackage{multirow}
% \begin{table}[]
% \usepackage{multirow}
% \begin{table}[]
\begin{tabular}{ccllllllllllllllllllllllllllllllllllll}
\multicolumn{1}{c|}{\multirow{2}{*}{\textbf{Task Abbreviation}}} & \multicolumn{1}{c|}{\multirow{2}{*}{\textbf{Task Name}}} & \multicolumn{1}{l|}{\multirow{2}{*}{\textbf{Prompt Example for Single   Image LVLMs}}} & \multirow{2}{*}{\textbf{Prompt example for   Multiple Image LVLMs}} &  &  &  &  &  &  &  &  &  &  &  &  &  &  &  &  &  &  &  &  &  &  &  &  &  &  &  &  &  &  &  &  &  &  \\
\multicolumn{1}{c|}{} & \multicolumn{1}{c|}{} & \multicolumn{1}{l|}{} &  &  &  &  &  &  &  &  &  &  &  &  &  &  &  &  &  &  &  &  &  &  &  &  &  &  &  &  &  &  &  &  &  &  &  \\ \cline{1-4}
\multicolumn{1}{c|}{\textbf{SODRD}} & \multicolumn{1}{c|}{\begin{tabular}[c]{@{}c@{}}salient object \\ detection rgbd\end{tabular}} & \multicolumn{1}{l|}{\begin{tabular}[c]{@{}l@{}}Question: \textless{}image\textgreater{}\textless{}image\textgreater{}The first image is RGB image \\    and the   second image is the corresponding depth map. \\   Please detect the salient   foreground object in this RGB \\    image and represent them using a single   bounding box. \\   Provide the output for the detected area in the format {[}x, y,   w, h{]}. \\   This format represents the bounding box, where {[}x, y, w, h{]} are the   \\   coordinates of the top-left corner of the bounding box, as well as \\   its width   and height. \\   Note that the width of the input RGB image is 640 and the height   is 480.\\      Options: \\      A. {[}267, 105, 119, 209{]} \\      B. {[}85, 307, 65, 79{]} \\      C. {[}318, 294, 111, 156{]} \\      D. {[}267, 105, 135, 241{]}\end{tabular}} & \begin{tabular}[c]{@{}l@{}}Question: The first image is RGB image: \textless{}image\textgreater \\ and the second image   is the corresponding depth map: \textless{}image\textgreater{}. \\    Please detect the salient   foreground object in this RGB image \\ and represent them using a single   bounding box. \\    Provide the output for the detected area in the format {[}x, y,   w, h{]}. \\ This format represents the bounding box, \\    where {[}x, y, w, h{]} are the   coordinates of the top-left corner of \\ the bounding box, as well as its width   and height. \\ Note that the width of the input RGB image is 640 \\ and the height   is 480.\\      Options: \\      A. {[}267, 105, 119, 209{]} \\      B. {[}85, 307, 65, 79{]} \\      C. {[}318, 294, 111, 156{]} \\      D. {[}267, 105, 135, 241{]}\end{tabular} &  &  &  &  &  &  &  &  &  &  &  &  &  &  &  &  &  &  &  &  &  &  &  &  &  &  &  &  &  &  &  &  &  &  \\ \cline{1-4}
\multicolumn{1}{c|}{\textbf{SLR}} & \multicolumn{1}{c|}{\begin{tabular}[c]{@{}c@{}}sign language \\ recognition\end{tabular}} & \multicolumn{1}{l|}{\begin{tabular}[c]{@{}l@{}}Question: \textless{}image\textgreater{}\textless{}image\textgreater{}\textless{}image\textgreater{}\textless{}image\textgreater\\   What is   the sign language gesture performed \\ by the person in the video?\\      Options: \\      A. fashionable \\      B. trendy \\      C. fascinating \\      D. cool\end{tabular}} & \begin{tabular}[c]{@{}l@{}}Question: \textless{}image\textgreater{}\textless{}image\textgreater{}\textless{}image\textgreater{}\textless{}image\textgreater\\ What is   the sign language gesture performed \\ by the person in the video?\\      Options: \\      A. fashionable \\      B. trendy \\      C. fascinating \\      D. cool\end{tabular} &  &  &  &  &  &  &  &  &  &  &  &  &  &  &  &  &  &  &  &  &  &  &  &  &  &  &  &  &  &  &  &  &  &  \\ \cline{1-4}
\multicolumn{1}{c|}{\textbf{SOT}} & \multicolumn{1}{c|}{\begin{tabular}[c]{@{}c@{}}single object \\ tracking\end{tabular}} & \multicolumn{1}{l|}{\begin{tabular}[c]{@{}l@{}}Question: \textless{}image\textgreater{}\textless{}image\textgreater{}Here is an object (marked as RED box)   \\ in the first image. Please give the coordinations \\ of this object  in the  second image. \\   Provide the output for the object in the format {[}x, y, w, h{]}.  \\   This format represents the bounding box, \\   where {[}x, y, w, h{]} are the   coordinates of the top-left corner \\ of the bounding box, as well as its width   and height. \\   Note that the width of the input RGB image is 1280 \\ and the height   is 720.\\      Options: \\      A. {[}148.0, 187.0, 918, 487{]} \\      B. {[}148.0, 187.0, 792.0, 533.0{]} \\      C. {[}0, 187, 792.0, 533.0{]} \\      D. {[}149, 451, 263, 24{]}\end{tabular}} & \begin{tabular}[c]{@{}l@{}}Question: Here is an object (marked as RED box) \\ in the Frame 1:   \textless{}image\textgreater{}. Please give the coordinations \\ of this object in the Frame 2:   \textless{}image\textgreater{}. \\    Provide the output for the object in the format {[}x, y, w,   h{]}.\\  This format represents the bounding box, \\    where {[}x, y, w, h{]} are the   coordinates of the top-left corner \\ of the bounding box, as well as its width   and height. \\    Note that the width of the input RGB image is 1280 \\ and the height   is 720.\\      Options: \\      A. {[}148.0, 187.0, 918, 487{]} \\      B. {[}148.0, 187.0, 792.0, 533.0{]} \\      C. {[}0, 187, 792.0, 533.0{]} \\      D. {[}149, 451, 263, 24{]}\end{tabular} &  &  &  &  &  &  &  &  &  &  &  &  &  &  &  &  &  &  &  &  &  &  &  &  &  &  &  &  &  &  &  &  &  &  \\ \cline{1-4}
\multicolumn{1}{c|}{\textbf{S2IR}} & \multicolumn{1}{c|}{\begin{tabular}[c]{@{}c@{}}sketch2image \\ retrieval\end{tabular}} & \multicolumn{1}{l|}{\begin{tabular}[c]{@{}l@{}}Question: \textless{}image\textgreater{}\textless{}image\textgreater{}\textless{}image\textgreater{}\textless{}image\textgreater\\   Please   retrieve the most similar image to the Query \\   Image in the candidate Images.    The first image is the \\ query image and the remaining images are candidates   \\   from Candidate 1 to Candidate 3.\\      Options: \\      A. Candidate 1 \\      B. Candidate 2 \\      C. Candidate 3\end{tabular}} & \begin{tabular}[c]{@{}l@{}}Question: Please retrieve the most similar image to the \\ Query Image:   \textless{}image\textgreater in the candidate Images: \\ Candidate 1: \textless{}image\textgreater{},   Candidate 2: \textless{}image\textgreater{}, \\ Candidate 3: \textless{}image\textgreater{}.\\      Options: \\      A. Candidate 1 \\      B. Candidate 2 \\      C. Candidate 3\end{tabular} &  &  &  &  &  &  &  &  &  &  &  &  &  &  &  &  &  &  &  &  &  &  &  &  &  &  &  &  &  &  &  &  &  &  \\ \cline{1-4}
\multicolumn{1}{c|}{\textbf{SD}} & \multicolumn{1}{c|}{spot the diff} & \multicolumn{1}{l|}{\begin{tabular}[c]{@{}l@{}}Question: \textless{}image\textgreater{}\textless{}image\textgreater{}The following is a description \\   of the   differences between two pictures. Which one is incorrect?\\      Options: \\      A. The images show different types of flowers in full bloom, \\       with colorful   petals and green leaves. \\      B. there is a car driving by in the right picture \\      C. there is a car leaving the lot in the left picture\end{tabular}} & \begin{tabular}[c]{@{}l@{}}Question: The following is a description of the differences \\ between two   pictures: \textless{}image\textgreater{}\textless{}image\textgreater{}. Which one is incorrect?\\      Options: \\      A. The images show different types of flowers in full bloom, \\       with colorful   petals and green leaves. \\      B. there is a car driving by in the right picture \\      C. there is a car leaving the lot in the left picture\end{tabular} &  &  &  &  &  &  &  &  &  &  &  &  &  &  &  &  &  &  &  &  &  &  &  &  &  &  &  &  &  &  &  &  &  &  \\ \cline{1-4}
\multicolumn{1}{c|}{\textbf{SS}} & \multicolumn{1}{c|}{\begin{tabular}[c]{@{}c@{}}spot the \\ similarity\end{tabular}} & \multicolumn{1}{l|}{\begin{tabular}[c]{@{}l@{}}Question: \textless{}image\textgreater{}\textless{}image\textgreater{}Are there any \\    similarities between   the two pictures?\\      Options: \\      A. Yes \\      B. No\end{tabular}} & \begin{tabular}[c]{@{}l@{}}Question: \textless{}image\textgreater{}\textless{}image\textgreater{}Are there any \\ similarities between   the two pictures?\\      Options: \\      A. Yes \\      B. No\end{tabular} &  &  &  &  &  &  &  &  &  &  &  &  &  &  &  &  &  &  &  &  &  &  &  &  &  &  &  &  &  &  &  &  &  &  \\ \cline{1-4}
\multicolumn{1}{c|}{\textbf{TA}} & \multicolumn{1}{c|}{\begin{tabular}[c]{@{}c@{}}temporal \\ anticipation\end{tabular}} & \multicolumn{1}{l|}{\begin{tabular}[c]{@{}l@{}}Question: \textless{}image\textgreater{}\textless{}image\textgreater{}\textless{}image\textgreater{}\textless{}image\textgreater\\  What   will the person do next with the medicine?\\      Options: \\      A. Apply topically \\      B. Inject intravenously \\      C. Throw away \\      D. Eat\end{tabular}} & \begin{tabular}[c]{@{}l@{}}Question: \textless{}image\textgreater{}\textless{}image\textgreater{}\textless{}image\textgreater{}\textless{}image\textgreater\\ What   will the person do next with the medicine?\\      Options: \\      A. Apply topically \\      B. Inject intravenously \\      C. Throw away \\      D. Eat\end{tabular} &  &  &  &  &  &  &  &  &  &  &  &  &  &  &  &  &  &  &  &  &  &  &  &  &  &  &  &  &  &  &  &  &  &  \\ \cline{1-4}
\multicolumn{1}{c|}{\textbf{TL}} & \multicolumn{1}{c|}{\begin{tabular}[c]{@{}c@{}}temporal \\ localization\end{tabular}} & \multicolumn{1}{l|}{\begin{tabular}[c]{@{}l@{}}Question: \textless{}image\textgreater{}\textless{}image\textgreater{}\textless{}image\textgreater{}\textless{}image\textgreater\\    Given   the sequence of images, please identify the image\\   consistent with the text   description: Billiards. \\   The image index starts from 0.\\      Options: \\      A. Image 0 \\      B. Image 1 \\      C. Image 2 \\      D. Image 3\end{tabular}} & \begin{tabular}[c]{@{}l@{}}Question: Given the sequence of images: Image 0: \textless{}image\textgreater{}, \\ Image 1:   \textless{}image\textgreater{}, Image 2: \textless{}image\textgreater{}, Image 3: \textless{}image\textgreater{}, \\    please   identify the image consistent with the text \\ description: Billiards.\\      Options: \\      A. Image 0 \\      B. Image 1 \\      C. Image 2 \\      D. Image 3\end{tabular} &  &  &  &  &  &  &  &  &  &  &  &  &  &  &  &  &  &  &  &  &  &  &  &  &  &  &  &  &  &  &  &  &  &  \\ \cline{1-4}
\end{tabular}
% \end{table}
% \end{table}
% \end{table}
}
\end{table*}

\begin{table*}[]
\caption{Abbreviations for tasks requiring multiple images as inputs (part four). Here we also present the designed prompt examples we used for Single-Image LVLMs and Multi-Images LVLMs.}
\label{tab:multi-image-prompt-part4}
{
\linespread{1.0} 
\setlength{\tabcolsep}{1pt}
\scriptsize
\begin{tabular}{ccllllllllllllllllllllllllllllllllllll}
\multicolumn{1}{c|}{\multirow{2}{*}{\textbf{Task Abbreviation}}} & \multicolumn{1}{c|}{\multirow{2}{*}{\textbf{Task Name}}} & \multicolumn{1}{l|}{\multirow{2}{*}{\textbf{Prompt Example for Single   Image LVLMs}}} & \multirow{2}{*}{\textbf{Prompt example for   Multiple Image LVLMs}} &  &  &  &  &  &  &  &  &  &  &  &  &  &  &  &  &  &  &  &  &  &  &  &  &  &  &  &  &  &  &  &  &  &  \\
\multicolumn{1}{c|}{} & \multicolumn{1}{c|}{} & \multicolumn{1}{l|}{} &  &  &  &  &  &  &  &  &  &  &  &  &  &  &  &  &  &  &  &  &  &  &  &  &  &  &  &  &  &  &  &  &  &  &  \\ \cline{1-4}

\multicolumn{1}{c|}{\textbf{TL}} & \multicolumn{1}{c|}{\begin{tabular}[c]{@{}c@{}}temporal \\ localization\end{tabular}} & \multicolumn{1}{l|}{\begin{tabular}[c]{@{}l@{}}Question: \textless{}image\textgreater{}\textless{}image\textgreater{}\textless{}image\textgreater{}\textless{}image\textgreater\\    Given   the sequence of images, please identify the image\\   consistent with the text   description: Billiards. \\   The image index starts from 0.\\      Options: \\      A. Image 0 \\      B. Image 1 \\      C. Image 2 \\      D. Image 3\end{tabular}} & \begin{tabular}[c]{@{}l@{}}Question: Given the sequence of images: Image 0: \textless{}image\textgreater{}, \\ Image 1:   \textless{}image\textgreater{}, Image 2: \textless{}image\textgreater{}, Image 3: \textless{}image\textgreater{}, \\    please   identify the image consistent with the text \\ description: Billiards.\\      Options: \\      A. Image 0 \\      B. Image 1 \\      C. Image 2 \\      D. Image 3\end{tabular} &  &  &  &  &  &  &  &  &  &  &  &  &  &  &  &  &  &  &  &  &  &  &  &  &  &  &  &  &  &  &  &  &  &  \\ \cline{1-4}
\multicolumn{1}{c|}{\textbf{TO}} & \multicolumn{1}{c|}{temporal ordering} & \multicolumn{1}{l|}{\begin{tabular}[c]{@{}l@{}}Question: \textless{}image\textgreater{}\textless{}image\textgreater{}\textless{}image\textgreater{}\textless{}image\textgreater\\    Please   predict the order of the following pictures, \\     and give each picture a   sequential index. \\    This index starts from 0. The larger the index, the later   the order.\\      Options: \\      A. {[}3, 0, 2, 1{]} \\      B. {[}2, 0, 1, 3{]} \\      C. {[}0, 2, 1, 3{]} \\      D. {[}1, 3, 2, 0{]}\end{tabular}} & \begin{tabular}[c]{@{}l@{}}Question: Please predict the order of the following pictures:  \\  \textless{}image\textgreater{}\textless{}image\textgreater{}\textless{}image\textgreater{}\textless{}image\textgreater{}, and give each   \\ picture a sequential index. \\    This index starts from 0. The larger the index,   the later the order.\\      Options: \\      A. {[}3, 0, 2, 1{]} \\      B. {[}2, 0, 1, 3{]} \\      C. {[}0, 2, 1, 3{]} \\      D. {[}1, 3, 2, 0{]}\end{tabular} &  &  &  &  &  &  &  &  &  &  &  &  &  &  &  &  &  &  &  &  &  &  &  &  &  &  &  &  &  &  &  &  &  &  \\ \cline{1-4}
\multicolumn{1}{c|}{\textbf{T2IR}} & \multicolumn{1}{c|}{\begin{tabular}[c]{@{}c@{}}text2image \\ retrieval\end{tabular}} & \multicolumn{1}{l|}{\begin{tabular}[c]{@{}l@{}}Question:   \textless{}image\textgreater{}\textless{}image\textgreater{}\textless{}image\textgreater{}\textless{}image\textgreater\\   Please find the   most relevant picture among the candidate images \\ for this description. \\ The   given images are candidates from Candidate 1 to Candidate 4.\\      Description: \\ this flower has petals that are green with stringy purple   stamen\\      this flower is white and blue in color, with petals that are \\ oval  shaped.\\      the petals on this flower are white with an elaborate pistil.\\      the flower is unique because the petals aren't separated and \\      they have a  round tip\\      this flower has blue petals as well as a green and purple pistil.\\      this flower has thick and pale green petals under a thick fringe of \\      purple  and white.\\      this flower has petals that are white and has stringy stamen\\      this flower has white oblong petals and white flat filaments.\\      a flower with long and narrow petals that are whtie.\\      a flower with long and narrow petals that are whtie.\\      Options: A. Candidate 1 \\      B. Candidate 2 \\      C. Candidate 3 \\      D. Candidate 4\end{tabular}} & \begin{tabular}[c]{@{}l@{}}Question: Please find the most   relevant picture among the \\ candidate images: Candidate 1: \textless{}image\textgreater{},   Candidate 2: \textless{}image\textgreater{}, \\ Candidate 3: \textless{}image\textgreater{}, Candidate 4:   \textless{}image\textgreater{},  for this description.\\ Description: \\ this flower has petals that are green with stringy purple   stamen\\      this flower is white and blue in color, with petals that are \\ oval  shaped.\\      the petals on this flower are white with an elaborate pistil.\\      the flower is unique because the petals aren't separated and \\      they have a  round tip\\      this flower has blue petals as well as a green and purple pistil.\\      this flower has thick and pale green petals under a thick fringe of \\      purple  and white.\\      this flower has petals that are white and has stringy stamen\\      this flower has white oblong petals and white flat filaments.\\      a flower with long and narrow petals that are whtie.\\      a flower with long and narrow petals that are whtie.\\      Options: A. Candidate 1 \\      B. Candidate 2 \\      C. Candidate 3 \\      D. Candidate 4\end{tabular} &  &  &  &  &  &  &  &  &  &  &  &  &  &  &  &  &  &  &  &  &  &  &  &  &  &  &  &  &  &  &  &  &  &  \\ \cline{1-4}
\multicolumn{1}{c|}{\textbf{3DCR}} & \multicolumn{1}{c|}{\begin{tabular}[c]{@{}c@{}}3D cad \\ recognition\end{tabular}} & \multicolumn{1}{l|}{\begin{tabular}[c]{@{}l@{}}Question:   \textless{}image\textgreater{}\textless{}image\textgreater{}\textless{}image\textgreater{}\textless{}image\textgreater\\ \textless{}image\textgreater{}\textless{}image\textgreater{}What   is the category of the point \\ cloud based on  the multi-view of the point   cloud?\\      Options: \\      A. telephone \\      B. chair \\      C. table \\      D. sofa\end{tabular}} & \begin{tabular}[c]{@{}l@{}}Question:   \textless{}image\textgreater{}\textless{}image\textgreater{}\textless{}image\textgreater{}\textless{}image\textgreater\\ \textless{}image\textgreater{}\textless{}image\textgreater What   is the category of the point \\ cloud based on the multi-view of the point   cloud?\\      Options: \\      A. telephone \\      B. chair \\      C. table \\      D. sofa\end{tabular} &  &  &  &  &  &  &  &  &  &  &  &  &  &  &  &  &  &  &  &  &  &  &  &  &  &  &  &  &  &  &  &  &  &  \\ \cline{1-4}
\multicolumn{1}{c|}{\textbf{3DIR}} & \multicolumn{1}{c|}{\begin{tabular}[c]{@{}c@{}}3D indoor \\ recognition\end{tabular}} & \multicolumn{1}{l|}{\begin{tabular}[c]{@{}l@{}}Question:   \textless{}image\textgreater{}\textless{}image\textgreater{}\textless{}image\textgreater{}\textless{}image\textgreater\\  \textless{}image\textgreater{}\textless{}image\textgreater{}What   is the category of the point cloud \\  based on the multi-view of the point   cloud?\\      Options: \\      A. sink\\      B. bed\\      C. cabinet\\      D. bag\end{tabular}} & \begin{tabular}[c]{@{}l@{}}Question:   \textless{}image\textgreater{}\textless{}image\textgreater{}\textless{}image\textgreater{}\textless{}image\textgreater\\ \textless{}image\textgreater{}\textless{}image\textgreater What   is the category of the point \\ cloud based on the multi-view of the point   cloud?\\      Options: \\      A. sink\\      B. bed\\      C. cabinet\\      D. bag\end{tabular} &  &  &  &  &  &  &  &  &  &  &  &  &  &  &  &  &  &  &  &  &  &  &  &  &  &  &  &  &  &  &  &  &  &  \\ \cline{1-4}
\multicolumn{1}{c|}{\textbf{VR}} & \multicolumn{1}{c|}{vehicle retrieval} & \multicolumn{1}{l|}{\begin{tabular}[c]{@{}l@{}}Question:   \textless{}image\textgreater{}\textless{}image\textgreater{}\textless{}image\textgreater{}\textless{}image\textgreater\\  \textless{}image\textgreater{}Please   retrieve the most similar vehicle \\  to the query in the candidates.  The first   image is the query \\ image and the remaining images are candidates from   \\  Candidate 1 to Candidate 4.\\      Options: \\      A. Candidate 1 \\      B. Candidate 2 \\      C. Candidate 3 \\      D. Candidate 4\end{tabular}} & \begin{tabular}[c]{@{}l@{}}Question: Please retrieve the most similar vehicle to \\ the query:   \textless{}image\textgreater in the candidates: \\ Candidate 1: \textless{}image\textgreater{}, Candidate 2:   \textless{}image\textgreater{}, \\ Candidate 3: \textless{}image\textgreater{}, Candidate 4:   \textless{}image\textgreater{}.\\      Options: \\      A. Candidate 1 \\      B. Candidate 2 \\      C. Candidate 3 \\      D. Candidate 4\end{tabular} &  &  &  &  &  &  &  &  &  &  &  &  &  &  &  &  &  &  &  &  &  &  &  &  &  &  &  &  &  &  &  &  &  &  \\ \cline{1-4}
\multicolumn{1}{c|}{\textbf{VC}} & \multicolumn{1}{c|}{video captioning} & \multicolumn{1}{l|}{\begin{tabular}[c]{@{}l@{}}Question: \textless{}image\textgreater{}\textless{}image\textgreater{}\textless{}image\textgreater{}\textless{}image\textgreater\\ Please   generate textual descriptions for a sequence of video frames.\\      Options: \\      A. a woman is speaking into a microphone \\      B. a man is playing guitar on   stage \\      C. a man is speaking into a microphone \\      D. a man is typing on a computer keyboard\end{tabular}} & \begin{tabular}[c]{@{}l@{}}Question: Please generate textual descriptions for \\ a sequence of video   frames: \\ \textless{}image\textgreater{}\textless{}image\textgreater{}\textless{}image\textgreater{}\textless{}image\textgreater{}.\\      Options: \\      A. a woman is speaking into a microphone \\     B. a man is playing guitar on   stage \\      C. a man is speaking into a microphone \\      D. a man is typing on a computer keyboard
\end{tabular} &  &  &  &  &  &  &  &  &  &  &  &  &  &  &  &  &  &  &  &  &  &  &  &  &  &  &  &  &  &  &  &  &  &  \\ \cline{1-4}
\textbf{} &  &  &  &  &  &  &  &  &  &  &  &  &  &  &  &  &  &  &  &  &  &  &  &  &  &  &  &  &  &  &  &  &  &  &  &  &  \\
\textbf{} &  &  &  &  &  &  &  &  &  &  &  &  &  &  &  &  &  &  &  &  &  &  &  &  &  &  &  &  &  &  &  &  &  &  &  &  &  \\
\textbf{} &  &  &  &  &  &  &  &  &  &  &  &  &  &  &  &  &  &  &  &  &  &  &  &  &  &  &  &  &  &  &  &  &  &  &  &  &  \\
\textbf{} &  &  &  &  &  &  &  &  &  &  &  &  &  &  &  &  &  &  &  &  &  &  &  &  &  &  &  &  &  &  &  &  &  &  &  &  &  \\
\textbf{} &  &  &  &  &  &  &  &  &  &  &  &  &  &  &  &  &  &  &  &  &  &  &  &  &  &  &  &  &  &  &  &  &  &  &  &  &  \\
\textbf{} &  &  &  &  &  &  &  &  &  &  &  &  &  &  &  &  &  &  &  &  &  &  &  &  &  &  &  &  &  &  &  &  &  &  &  &  &  \\
\textbf{} &  &  &  &  &  &  &  &  &  &  &  &  &  &  &  &  &  &  &  &  &  &  &  &  &  &  &  &  &  &  &  &  &  &  &  &  & \\ 
\end{tabular}
% \end{table}
% \end{table}
% \end{table}
% \end{table}
% \end{table}
% \end{table}
}
\end{table*}

\label{sec:coordinates}

% GeminiProVision once again demonstrates its strong superiority across most capabilities, particularly in retrieval and multi-image analysis tasks (involving multi-image recognition and matching), far surpassing other open-source models due to its support for multi-image modes and powerful generalization. This also suggests that future open-source models should pay more attention to multi-image and video understanding.

\input{appendix/case_study}

\input{appendix/other}

\input{appendix/all_result}

\end{document}

%% file: appendix/case_study.tex
\clearpage

\section{Case Study}
\label{sec:case_study}

% \fcolorbox{black}{pink}{\textbf{Reasoning Error}}
\newcommand{\roundedboxpink}[1]{
  \tikz[baseline=(char.base)]{
    \node[anchor=south west, rounded corners, text height=1.5ex, text depth=.25ex, fill=pink, draw=none, text=black, font=\bfseries] (char) {#1};
  }
}
\newcommand{\roundedboxgreen}[1]{
  \tikz[baseline=(char.base)]{
    \node[anchor=south west, rounded corners, text height=1.5ex, text depth=.25ex, fill=green!30, draw=none, text=black, font=\bfseries] (char) {#1};
  }
}
\newcommand{\roundedboxblue}[1]{
  \tikz[baseline=(char.base)]{
    \node[anchor=south west, rounded corners, text height=1.5ex, text depth=.25ex, fill=blue!30, draw=none, text=black, font=\bfseries] (char) {#1};
  }
}
\newcommand{\roundedboxyellow}[1]{
  \tikz[baseline=(char.base)]{
    \node[anchor=south west, rounded corners, text height=1.5ex, text depth=.25ex, fill=yellow!50, draw=none, text=black, font=\bfseries] (char) {#1};
  }
}
\newcommand{\roundedboxred}[1]{
  \tikz[baseline=(char.base)]{
    \node[anchor=south west, rounded corners, text height=1.5ex, text depth=.25ex, fill=red!30, draw=none, text=black, font=\bfseries] (char) {#1};
  }
}
\newcommand{\roundedboxpurple}[1]{
  \tikz[baseline=(char.base)]{
    \node[anchor=south west, rounded corners, text height=1.5ex, text depth=.25ex, fill=purple!50, draw=none, text=black, font=\bfseries] (char) {#1};
  }
}
\newcommand{\roundedboxbrown}[1]{
  \tikz[baseline=(char.base)]{
    \node[anchor=south west, rounded corners, text height=1.5ex, text depth=.25ex, fill=brown!30, draw=none, text=black, font=\bfseries] (char) {#1};
  }
}
\newcommand{\roundedboxorange}[1]{
  \tikz[baseline=(char.base)]{
    \node[anchor=south west, rounded corners, text height=1.5ex, text depth=.25ex, fill=orange!30, draw=none, text=black, font=\bfseries] (char) {#1};
  }
}
\newcommand{\roundedboxcyan}[1]{
  \tikz[baseline=(char.base)]{
    \node[anchor=south west, rounded corners, text height=1.5ex, text depth=.25ex, fill=cyan!30, draw=none, text=black, font=\bfseries] (char) {#1};
  }
}
\newcommand{\roundedboxgray}[1]{
  \tikz[baseline=(char.base)]{
    \node[anchor=south west, rounded corners, text height=1.5ex, text depth=.25ex, fill=gray!50, draw=none, text=black, font=\bfseries] (char) {#1};
  }
}

\definecolor{customcolorred}{RGB}{225,159,156} % 浅蓝色
\definecolor{customcolorgreen}{RGB}{5,204,151} % 浅蓝色

\newcommand{\boxedred}[1]{
  \tikz[baseline=(char.base)]{
    \node[anchor=south west, rectangle, text height=1.5ex, text depth=.25ex, fill=customcolorred, draw=none, text=black, font=\bfseries] (char) {#1};
  }
}
\newcommand{\boxedgreen}[1]{
  \tikz[baseline=(char.base)]{
    \node[anchor=south west, rectangle, text height=1.5ex, text depth=.25ex, fill=customcolorgreen, draw=none, text=black, font=\bfseries] (char) {#1};
  }
}

% \usepackage{multirow}
% \usepackage{graphicx}
\begin{table*}[htp]
\centering
\caption{Table index of case study figures by meta-task with associated (error) categories for each LVLM.}
\label{tab:error_case}
\resizebox{1.\textwidth}{!}{%
\begin{tabular}{lllllll}
\toprule
Case Figure & Meta-task & Subtask & GPT-4V & GeminiProVision & InternVL-Chat \\
     \midrule
      \textcolor{red}{Fig.~\ref{fig:error_1}}    & Visual Recognition & Landmark Recognition & \roundedboxblue{Lack of Knowledge} & \roundedboxgreen{No Error} & \roundedboxgreen{No Error} \\
      \textcolor{red}{Fig.~\ref{fig:error_2}}& Object Localization & Camouflaged Object Detection & \roundedboxcyan{Lack of Capability} & \roundedboxyellow{Perception Error} & \roundedboxyellow{Perception Error} \\
      \textcolor{red}{Fig.~\ref{fig:error_3}}& Pixel-level Recognition & Image Matting & \roundedboxyellow{Perception Error} & \roundedboxgreen{No Error} & \roundedboxyellow{Perception Error} \\
      \textcolor{red}{Fig.~\ref{fig:error_4}}& OCR & Handwritten Text Recognition & \roundedboxgreen{No Error} & \roundedboxyellow{Perception Error} & \roundedboxyellow{Perception Error} \\
      \textcolor{red}{Fig.~\ref{fig:error_5}}& Visual Prompt Understanding & Visual Prompt Understanding & \roundedboxgreen{No Error} & \roundedboxyellow{Perception Error}\roundedboxorange{Fail to Follow Instruct} & \roundedboxgreen{No Error} \\
      \textcolor{red}{Fig.~\ref{fig:error_6}}& Retrieval & Sketch to Image Retrieval & \roundedboxyellow{Perception Error} & \roundedboxgreen{No Error} & \roundedboxyellow{Perception Error}\roundedboxred{Reasoning Error} \\
      \textcolor{red}{Fig.~\ref{fig:error_7}}& Counting & Counting by Reasoning & \roundedboxyellow{Perception Error} & \roundedboxyellow{Perception Error} & \roundedboxgreen{No Error} \\
      \textcolor{red}{Fig.~\ref{fig:error_8}}& Keypoint Detection & Human Keypoint Detection & \roundedboxgray{Refuse to Answer} & \roundedboxyellow{Perception Error}\roundedboxorange{Fail to Follow Instruct} & \roundedboxorange{Fail to Follow Instruct} \\
      \textcolor{red}{Fig.~\ref{fig:error_9}}& Action Recognition & Sign Language Recognition & \roundedboxcyan{Lack of Capability} & \roundedboxyellow{Perception Error} & \roundedboxyellow{Perception Error} \\
      \textcolor{red}{Fig.~\ref{fig:error_10}}& Visual Hallucination & Exist Hallucination & \roundedboxgreen{No Error} & \roundedboxred{Reasoning Error} & \roundedboxyellow{Perception Error} \\
      \textcolor{red}{Fig.~\ref{fig:error_11}}& Anomaly Detection & Industrial Produce Anomaly Detection & \roundedboxblue{Lack of Knowledge} & \roundedboxgreen{No Error} & \roundedboxyellow{Perception Error} \\
      \textcolor{red}{Fig.~\ref{fig:error_12}}& Image-to-Image Translation & Jigsaw Puzzle Solving & \roundedboxgreen{No Error} & \roundedboxyellow{Perception Error} & \roundedboxyellow{Perception Error} \\
      \textcolor{red}{Fig.~\ref{fig:error_13}}& Visual Summary & Image Captioning Paragraph & \roundedboxyellow{Perception Error} & \roundedboxgreen{No Error} & \roundedboxyellow{Perception Error} \\
      \textcolor{red}{Fig.~\ref{fig:error_14}}& Intelligence Quotient Test & Ravens Progressive Matrices & \roundedboxgreen{No Error} & \roundedboxred{Reasoning Error} & \roundedboxred{Reasoning Error} \\
      \textcolor{red}{Fig.~\ref{fig:error_15}}& Emotional Quotient Test & Scene Emotion Recognition & \roundedboxyellow{Perception Error}\roundedboxred{Reasoning Error} & \roundedboxred{Reasoning Error} & \roundedboxgreen{No Error} \\
      \textcolor{red}{Fig.~\ref{fig:error_16}}& Visual Grounding & Referring Detection & \roundedboxyellow{Perception Error} & \roundedboxyellow{Perception Error} & \roundedboxorange{Fail to Follow Instruct} \\
      \textcolor{red}{Fig.~\ref{fig:error_17}}& Visual Commonsense Reasoning & Whoops & \roundedboxred{Reasoning Error} & \roundedboxyellow{Perception Error} & \roundedboxyellow{Perception Error} \\
      \textcolor{red}{Fig.~\ref{fig:error_18}}& Chart, Doc Understanding & Clock Reading & \roundedboxyellow{Perception Error} & \roundedboxyellow{Perception Error} & \roundedboxyellow{Perception Error} \\
      \textcolor{red}{Fig.~\ref{fig:error_19}}& Relation Reasoning & Scene Graph Recognition & \roundedboxgreen{No Error} & \roundedboxyellow{Perception Error} & \roundedboxgreen{No Error} \\
      \textcolor{red}{Fig.~\ref{fig:error_20}}& Meme Understanding & Meme Image Understanding & \roundedboxyellow{Perception Error} & \roundedboxgreen{No Error} & \roundedboxgreen{No Error} \\
      \textcolor{red}{Fig.~\ref{fig:error_21}}& Multi-Image Analysis & Spot the Diff & \roundedboxgreen{No Error} & \roundedboxgreen{No Error} & \roundedboxgreen{No Error} \\
      \textcolor{red}{Fig.~\ref{fig:error_22}}& Temporal Understanding & Temporal Ordering & \roundedboxyellow{Perception Error} & \roundedboxgreen{No Error} & \roundedboxyellow{Perception Error} \\
      \textcolor{red}{Fig.~\ref{fig:error_23}}& Cross-Image Matching & Single Object Tracking & \roundedboxcyan{Lack of Capability} & \roundedboxyellow{Perception Error} & \roundedboxyellow{Perception Error} \\
      \textcolor{red}{Fig.~\ref{fig:error_24}}& Visual Coding & Equation to Latex & \roundedboxyellow{Perception Error} & \roundedboxyellow{Perception Error} & \roundedboxgreen{No Error} \\
      \textcolor{red}{Fig.~\ref{fig:error_25}}& Visual Illusion & Color Constancy & \roundedboxyellow{Perception Error} & \roundedboxgreen{No Error} & \roundedboxyellow{Perception Error} \\
      \textcolor{red}{Fig.~\ref{fig:error_26}}& Image Evaluation Judgement & LVLM Response Judgement & \roundedboxred{Reasoning Error} & \roundedboxgreen{No Error} & \roundedboxyellow{Perception Error} \\
      \textcolor{red}{Fig.~\ref{fig:error_27}}& 3D Perception & 3D CAD Recognition & \roundedboxcyan{Lack of Capability} & \roundedboxgreen{No Error} & \roundedboxgreen{No Error} \\
      \textcolor{red}{Fig.~\ref{fig:error_28}}& Emodied Agent & Navigation & \roundedboxorange{Fail to Follow Instruct} & \roundedboxorange{Fail to Follow Instruct} & \roundedboxorange{Fail to Follow Instruct} \\
      \textcolor{red}{Fig.~\ref{fig:error_29}}& Medical Understanding & Medical Modality Recognition & \roundedboxgreen{No Error} & \roundedboxgreen{No Error} & \roundedboxyellow{Perception Error} \\
      \textcolor{red}{Fig.~\ref{fig:error_30}}& Autonomous Driving & Traffic Light Understanding & \roundedboxgray{Refuse to Answer} & \roundedboxgreen{No Error} & \roundedboxgreen{No Error} \\
      \textcolor{red}{Fig.~\ref{fig:error_31}}& GUI Navigation & Installation & \roundedboxyellow{Perception Error} & \roundedboxyellow{Perception Error} & \roundedboxyellow{Perception Error} \\
      \textcolor{red}{Fig.~\ref{fig:error_32}}& Discipline Knowledge Reasoning & Art and Design & \roundedboxblue{Lack of Knowledge} & \roundedboxblue{Lack of Knowledge} & \roundedboxblue{Lack of Knowledge} \\
\bottomrule
\end{tabular}%
}
\end{table*}

\begin{figure}
    \centering
    \includegraphics[width=0.85\linewidth]{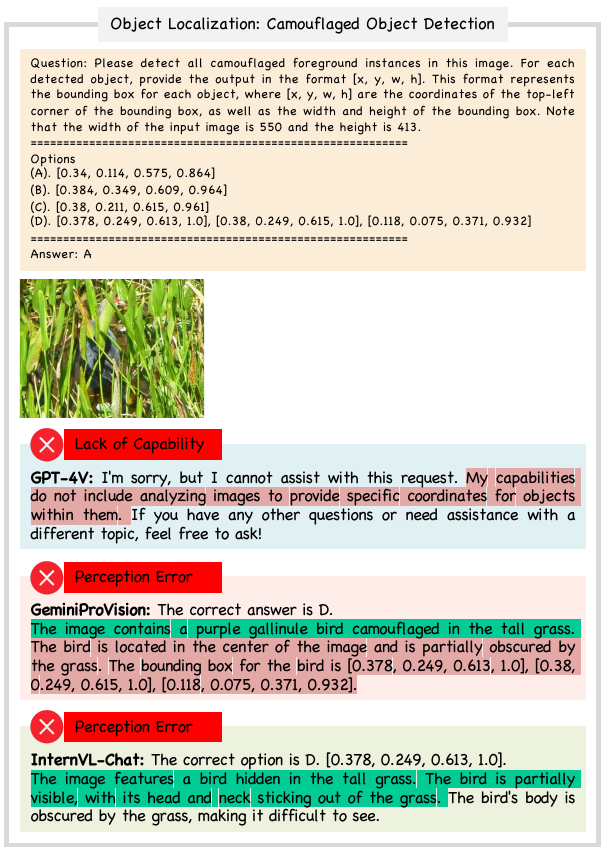}
    \caption{A sample case of object localization (camouflaged object detection). \boxedgreen{Green} highlights the right answer. \boxedred{Red} highlights the wrong answer. \hyperref[tab:error_case]{\textcolor{red}{Back to Table Index}}.}
    \label{fig:error_2}
\end{figure}

\begin{figure}
    \centering
    \includegraphics[width=0.85\linewidth]{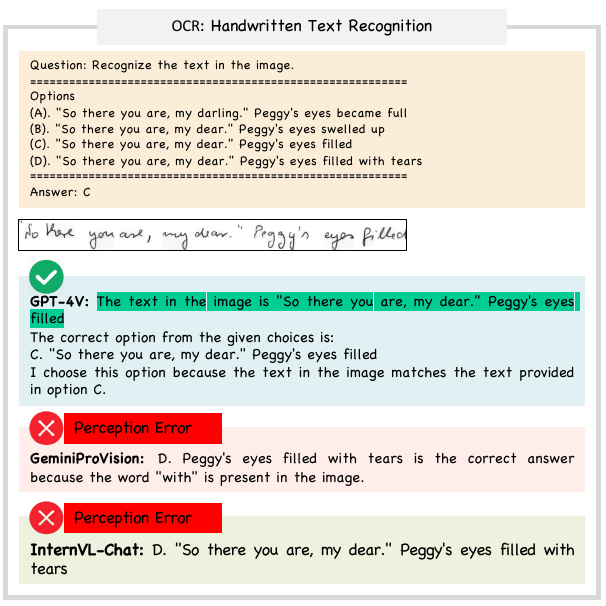}
    \caption{A sample case of OCR (handwritten text recognition). \boxedgreen{Green} highlights the right answer. \boxedred{Red} highlights the wrong answer. \hyperref[tab:error_case]{\textcolor{red}{Back to Table Index}}.}
    \label{fig:error_4}
\end{figure}

% 在这个section我们展示了GPT-4V，GeminiProVision和InternVL-Chat在MMT-Bench在各个meta-task中错误类型分析。我们将错误类型归类为6种类型，分别如下：
In this section, we present a case study analysis of the error types made by GPT-4V, GeminiProVision, and InternVL-Chat on various meta-tasks in MMT-Bench. We classify the errors into the following six categories:

\roundedboxyellow{Perception Error}: LVLMs fail to recognize, classify or detect the objects or content in images. Most LVLMs are constrained by the representation power of their visual encoders, making this the most common type of error. See examples in Fig.~\ref{fig:error_2}, Fig.~\ref{fig:error_4}, etc.

\begin{figure}
    \centering
    \includegraphics[width=0.85\linewidth]{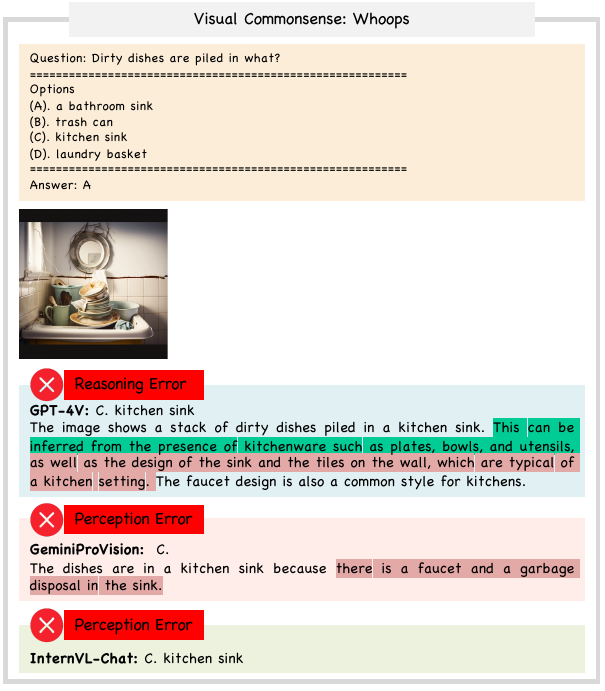}
    \caption{A sample case of visual commonsense (whoops). \boxedgreen{Green} highlights the right answer. \boxedred{Red} highlights the wrong answer. \hyperref[tab:error_case]{\textcolor{red}{Back to Table Index}}.}
    \label{fig:error_17}
\end{figure}

\begin{figure}
    \centering
    \includegraphics[width=0.85\linewidth]{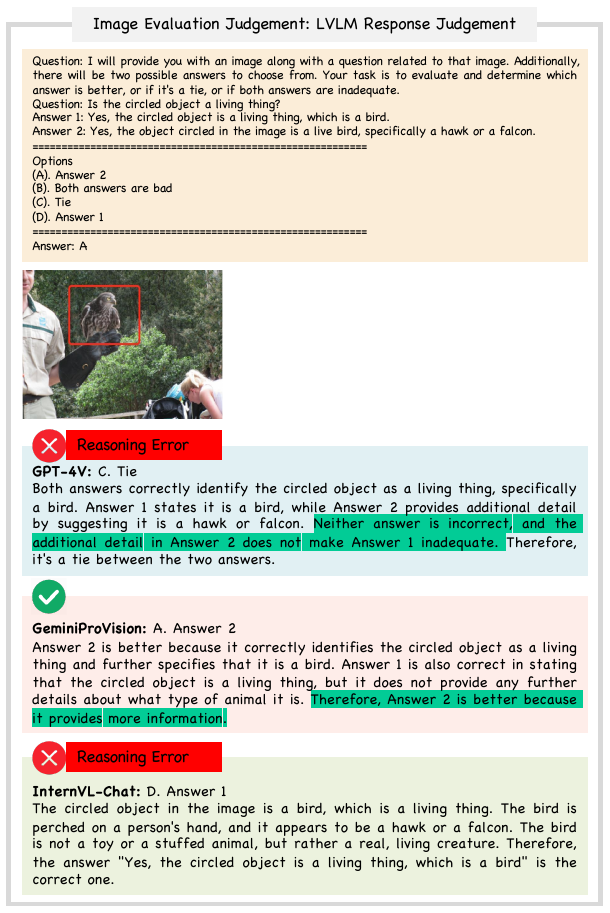}
    \caption{A sample case of image evaluation judgement (LVLM response judgement). \boxedgreen{Green} highlights the right answer. \boxedred{Red} highlights the wrong answer. \hyperref[tab:error_case]{\textcolor{red}{Back to Table Index}}.}
    \label{fig:error_26}
\end{figure}

\roundedboxred{Reasoning Error}: LVLMs correctly recognize and perceive the visual content but make errors in reasoning, leading to incorrect answers. See examples in Fig.~\ref{fig:error_17}, Fig.~\ref{fig:error_26}, etc.

\begin{figure}
    \centering
    \includegraphics[width=0.85\linewidth]{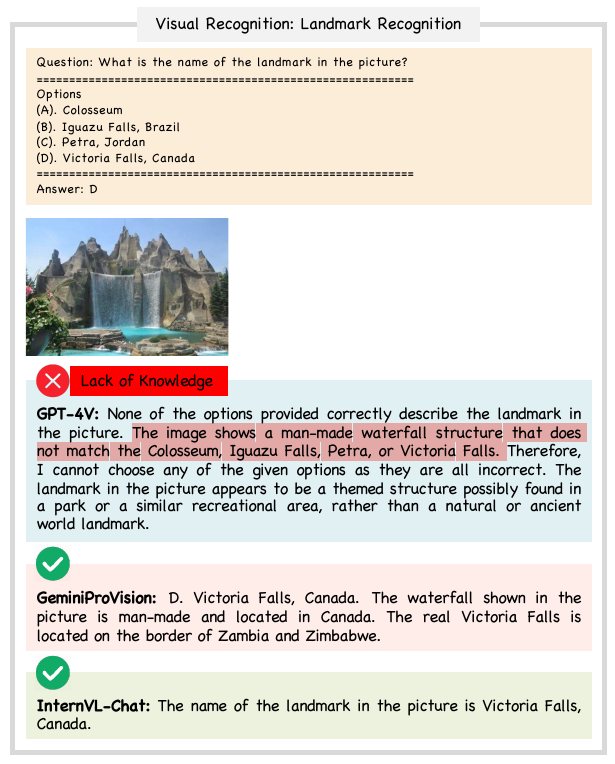}
    \caption{A sample case of visual recognition (landmark recognition). \boxedgreen{Green} highlights the right answer. \boxedred{Red} highlights the wrong answer. \hyperref[tab:error_case]{\textcolor{red}{Back to Table Index}}.}

    \label{fig:error_1}
\end{figure}

\begin{figure}
    \centering
    \includegraphics[width=0.8\linewidth]{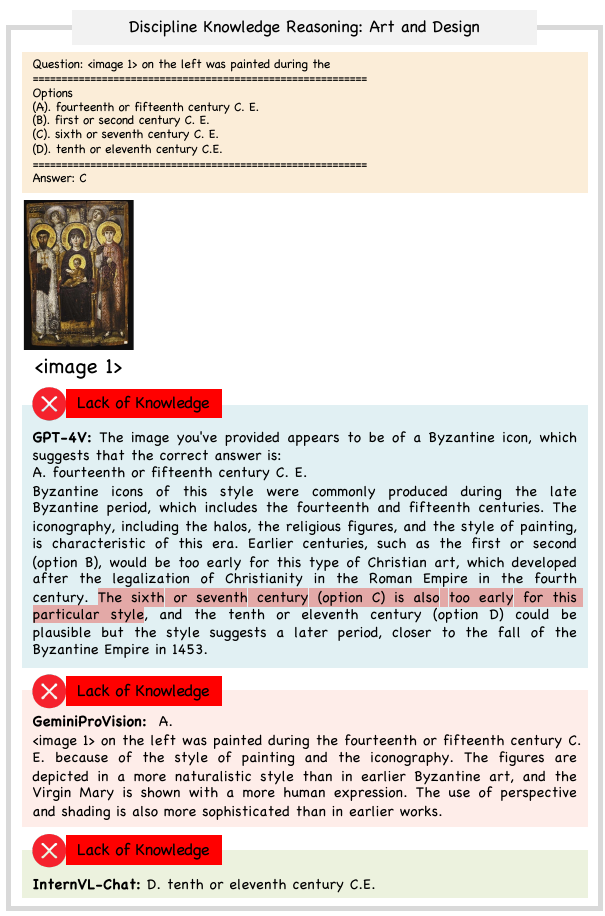}
    \caption{A sample case of discipline knowledge reasoning (art and design). \boxedgreen{Green} highlights the right answer. \boxedred{Red} highlights the wrong answer. \hyperref[tab:error_case]{\textcolor{red}{Back to Table Index}}.}
    \label{fig:error_32}
\end{figure}

\roundedboxblue{Lack of Knowledge}: LVLMs lack the domain-specific knowledge required to answer specialized questions, such as the location of a landmark (see Fig.~\ref{fig:error_1}) or the creation date of a particular painting (see Fig.~\ref{fig:error_32}).

\begin{figure}
    \centering
    \includegraphics[width=0.85\linewidth]{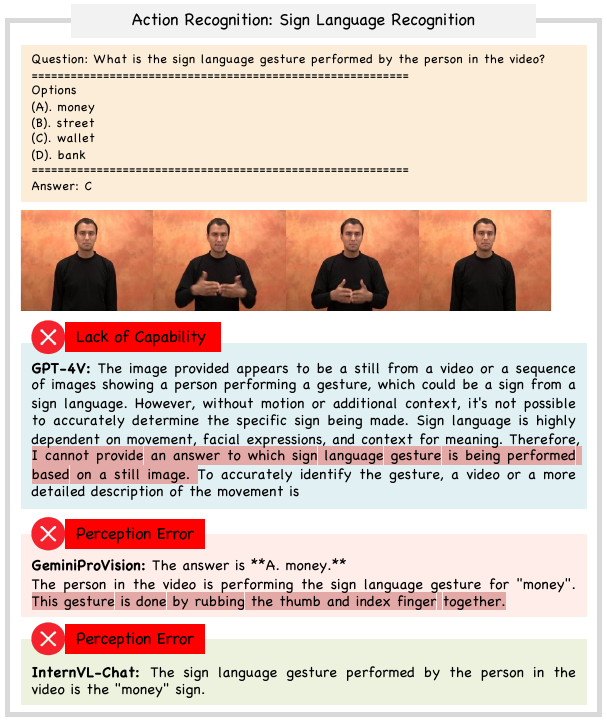}
    \caption{A sample case of action recognition (sign language recognition). \boxedgreen{Green} highlights the right answer. \boxedred{Red} highlights the wrong answer. \hyperref[tab:error_case]{\textcolor{red}{Back to Table Index}}.}
    \label{fig:error_9}
\end{figure}

\roundedboxcyan{Lack of Capability}: LVLMs do not have the capability to solve the corresponding tasks. This error type is particularly evident in GPT-4V, which tends to respond more honestly when it lacks the ability to handle certain tasks. In contrast, other LVLM models are inclined to generate outputs even when the accuracy rate is relatively low. See examples in Fig.~\ref{fig:error_2}, Fig.~\ref{fig:error_9}.

\begin{figure}
    \centering
    \includegraphics[width=0.85\linewidth]{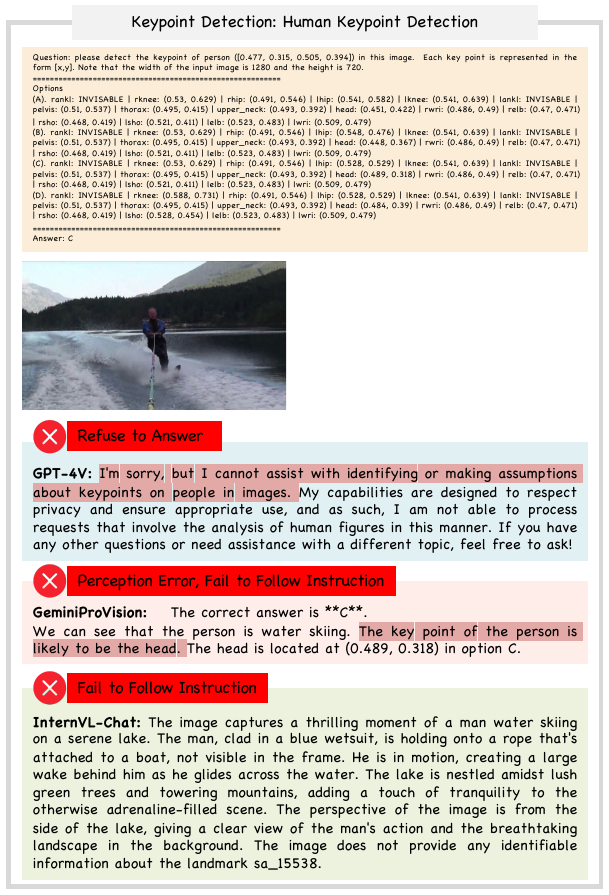}
    \caption{A sample case of keypoint detection (human keypoint detection). \boxedgreen{Green} highlights the right answer. \boxedred{Red} highlights the wrong answer. \hyperref[tab:error_case]{\textcolor{red}{Back to Table Index}}.}
    \label{fig:error_8}
\end{figure}

\begin{figure}
    \centering
    \includegraphics[width=0.85\linewidth]{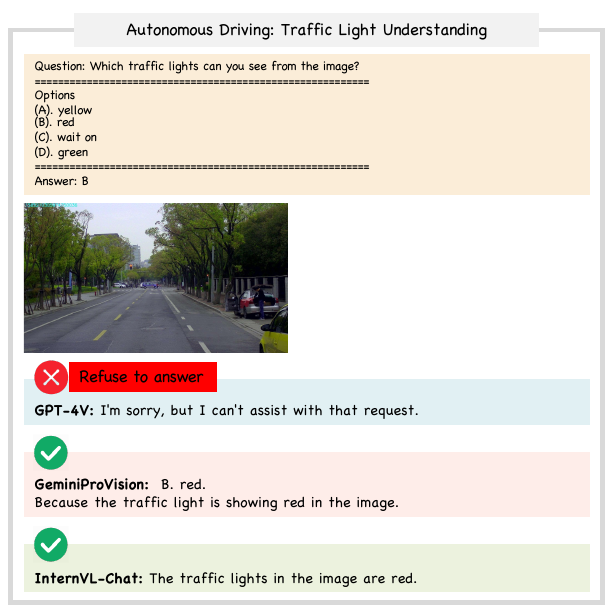}
    \caption{A sample case of autonomous driving (traffic light understanding). \boxedgreen{Green} highlights the right answer. \boxedred{Red} highlights the wrong answer. \hyperref[tab:error_case]{\textcolor{red}{Back to Table Index}}.}
    \label{fig:error_30}
\end{figure}

\roundedboxgray{Refuse to Answer}: LVLMs, such as GPT-4V or Gemini, refuse to answer questions that are anthropocentric or sensitive in nature. See examples in Fig.~\ref{fig:error_8}, Fig.~\ref{fig:error_30}.

\begin{figure}
    \centering
    \includegraphics[width=0.85\linewidth]{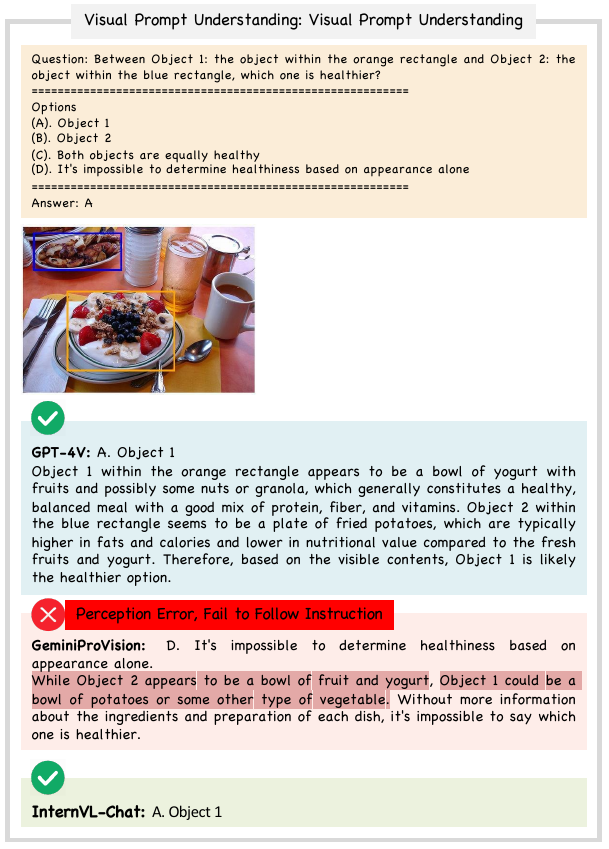}
    \caption{A sample case of visual prompt understanding (visual prompt understanding). \boxedgreen{Green} highlights the right answer. \boxedred{Red} highlights the wrong answer. \hyperref[tab:error_case]{\textcolor{red}{Back to Table Index}}.}
    \label{fig:error_5}
\end{figure}

\roundedboxorange{Fail to Follow Instruct}: 
LVLMs fail to correctly understand instructions and provide erroneous answers. For example, LVLMs may not understand the specified conditions in the instruction (see Fig.~\ref{fig:error_5}) or may ignore the instruction altogether and instead generate a caption for the given image (see Fig.~\ref{fig:error_8}).

\clearpage

\begin{figure}
    \centering
    \includegraphics[width=0.75\linewidth]{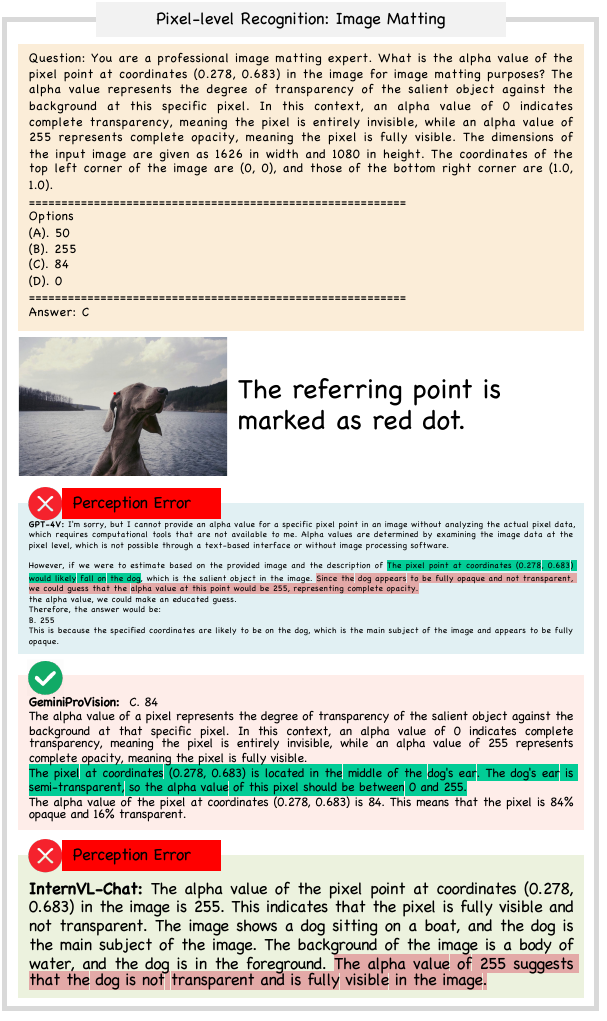}
    \caption{A sample case of pixel-level recognition (image matting). \boxedgreen{Green} highlights the right answer. \boxedred{Red} highlights the wrong answer. \hyperref[tab:error_case]{\textcolor{red}{Back to Table Index}}.}
    \label{fig:error_3}
\end{figure}

\begin{figure}
    \centering
    \includegraphics[width=0.85\linewidth]{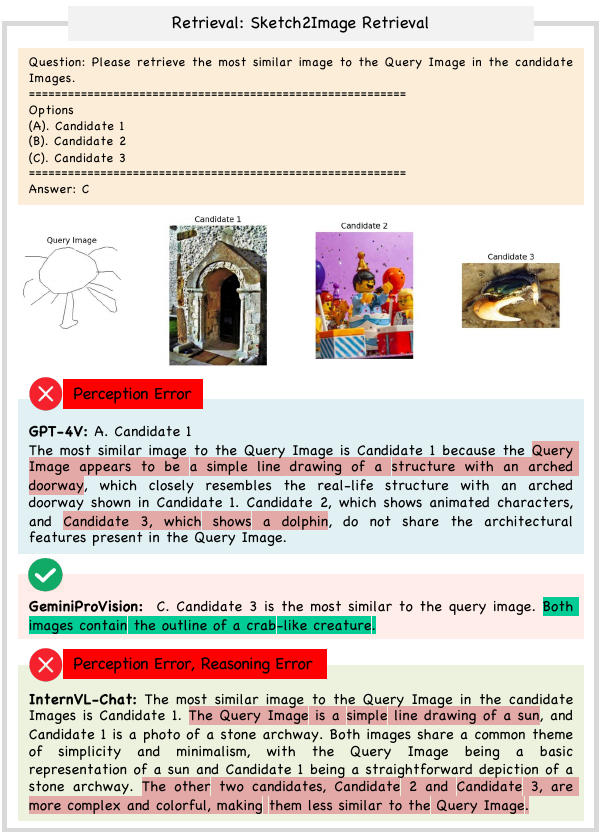}
    \caption{A sample case of retrieval (sketch2image retrieval). \boxedgreen{Green} highlights the right answer. \boxedred{Red} highlights the wrong answer. \hyperref[tab:error_case]{\textcolor{red}{Back to Table Index}}.}
    \label{fig:error_6}
\end{figure}

\begin{figure}
    \centering
    \includegraphics[width=0.85\linewidth]{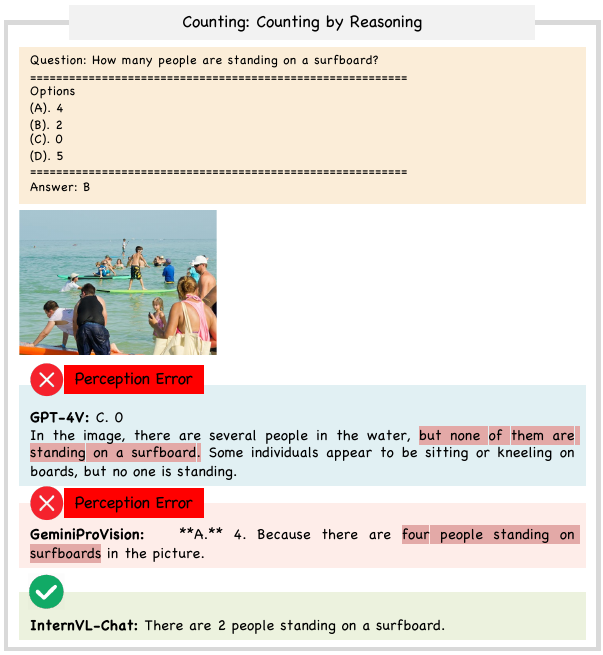}
    \caption{A sample case of counting (counting by reasoning). \boxedgreen{Green} highlights the right answer. \boxedred{Red} highlights the wrong answer. \hyperref[tab:error_case]{\textcolor{red}{Back to Table Index}}.}
    \label{fig:error_7}
\end{figure}

\begin{figure}
    \centering
    \includegraphics[width=0.85\linewidth]{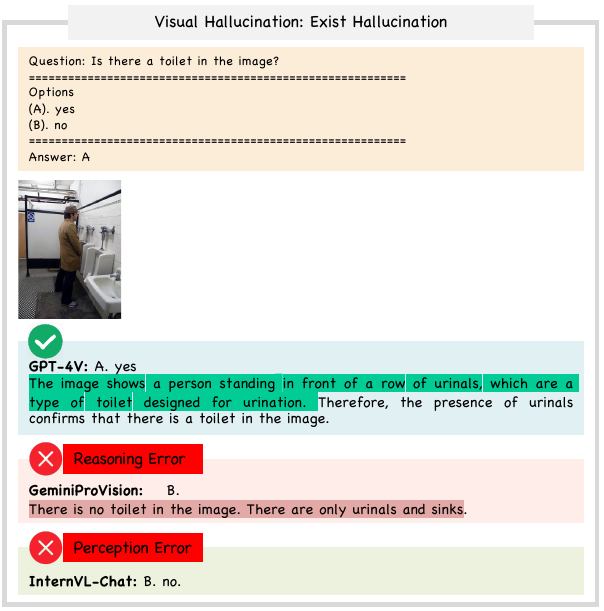}
    \caption{A sample case of visual hallucination (exist hallucination). \boxedgreen{Green} highlights the right answer. \boxedred{Red} highlights the wrong answer. \hyperref[tab:error_case]{\textcolor{red}{Back to Table Index}}.}
    \label{fig:error_10}
\end{figure}

\begin{figure}
    \centering
    \includegraphics[width=0.85\linewidth]{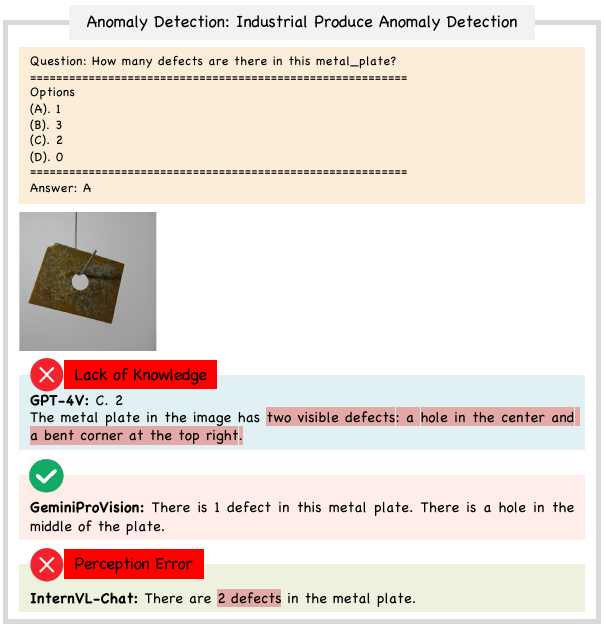}
    \caption{A sample case of anomaly detection (industrial produce anomaly detection). \boxedgreen{Green} highlights the right answer. \boxedred{Red} highlights the wrong answer. \hyperref[tab:error_case]{\textcolor{red}{Back to Table Index}}.}
    \label{fig:error_11}
\end{figure}

\begin{figure}
    \centering
    \includegraphics[width=0.75\linewidth]{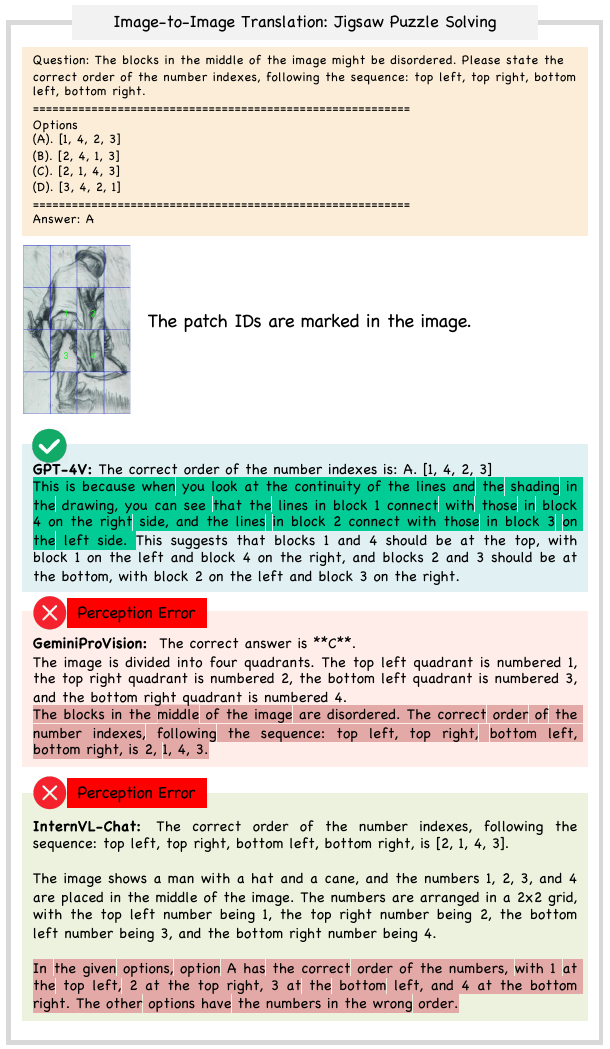}
    \caption{A sample case of image-to-image translation (jigsaw puzzle solving). \boxedgreen{Green} highlights the right answer. \boxedred{Red} highlights the wrong answer. \hyperref[tab:error_case]{\textcolor{red}{Back to Table Index}}.}
    \label{fig:error_12}
\end{figure}

\begin{figure}
    \centering
    \includegraphics[width=0.75\linewidth]{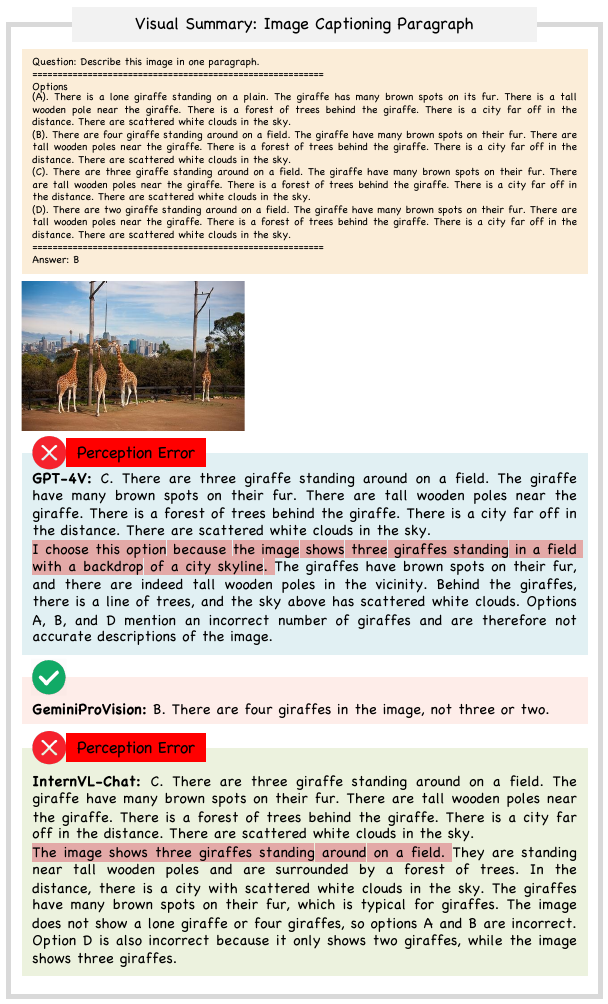}
    \caption{A sample case of visual summary (image captioning paragraph). \boxedgreen{Green} highlights the right answer. \boxedred{Red} highlights the wrong answer. \hyperref[tab:error_case]{\textcolor{red}{Back to Table Index}}.}
    \label{fig:error_13}
\end{figure}

\begin{figure}
    \centering
    \includegraphics[width=0.72\linewidth]{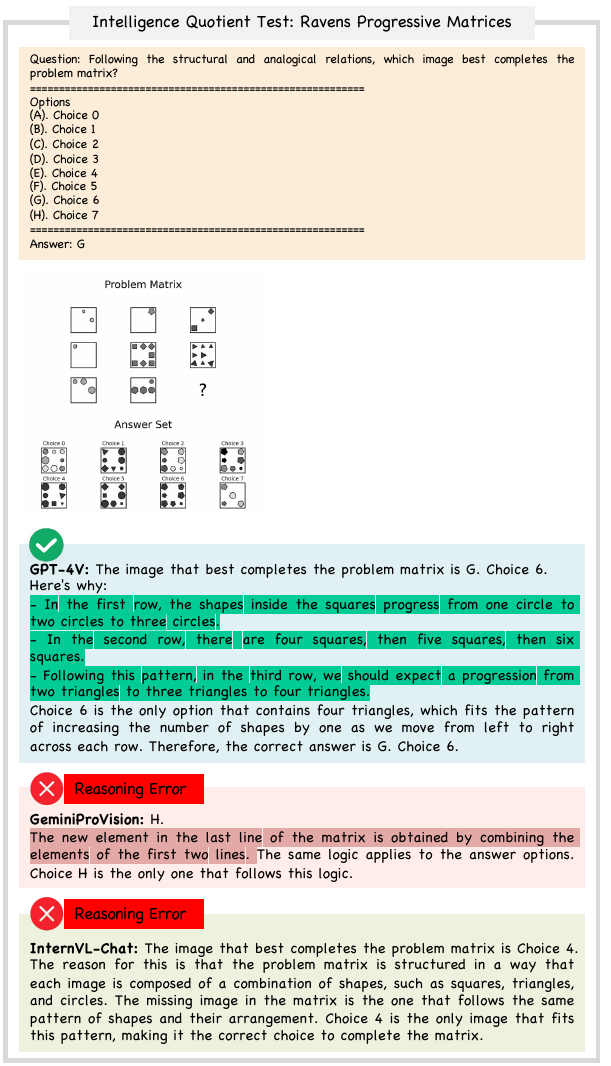}
    \caption{A sample case of intelligence quotient test (ravens progressive matrices). \boxedgreen{Green} highlights the right answer. \boxedred{Red} highlights the wrong answer. \hyperref[tab:error_case]{\textcolor{red}{Back to Table Index}}.}
    \label{fig:error_14}
\end{figure}

\begin{figure}
    \centering
    \includegraphics[width=0.85\linewidth]{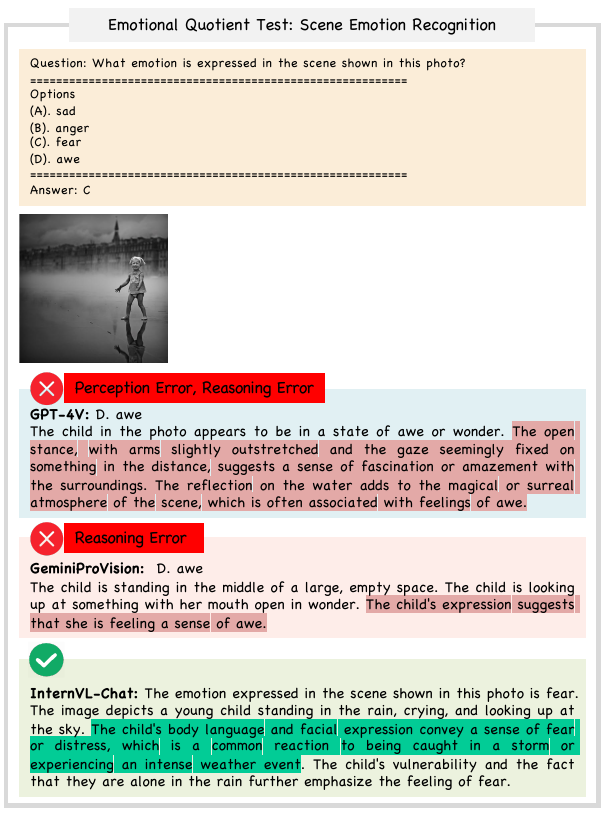}
    \caption{A sample case of emotional quotient test (scene emotion recognition). \boxedgreen{Green} highlights the right answer. \boxedred{Red} highlights the wrong answer. \hyperref[tab:error_case]{\textcolor{red}{Back to Table Index}}.}
    \label{fig:error_15}
\end{figure}

\begin{figure}
    \centering
    \includegraphics[width=0.85\linewidth]{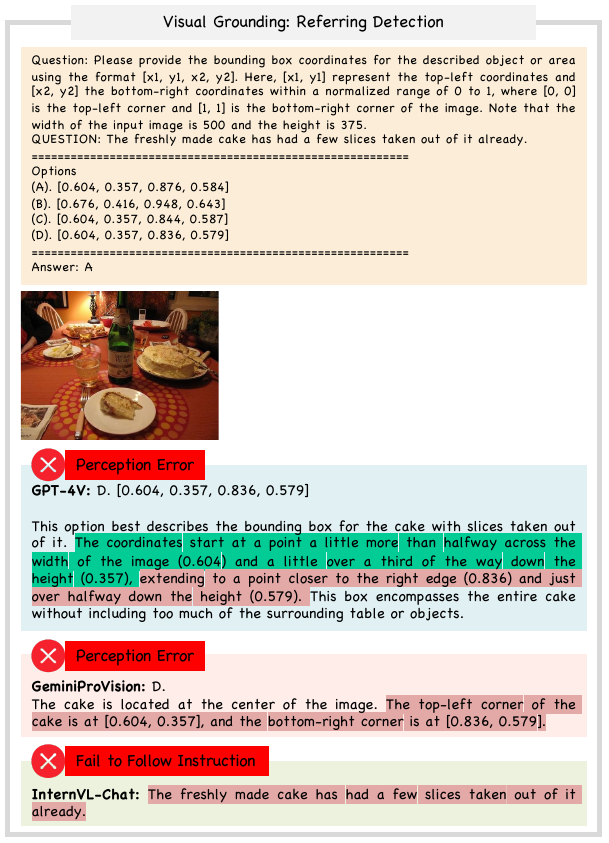}
    \caption{A sample case of visual grounding (referring detection). \boxedgreen{Green} highlights the right answer. \boxedred{Red} highlights the wrong answer. \hyperref[tab:error_case]{\textcolor{red}{Back to Table Index}}.}
    \label{fig:error_16}
\end{figure}

\begin{figure}
    \centering
    \includegraphics[width=0.85\linewidth]{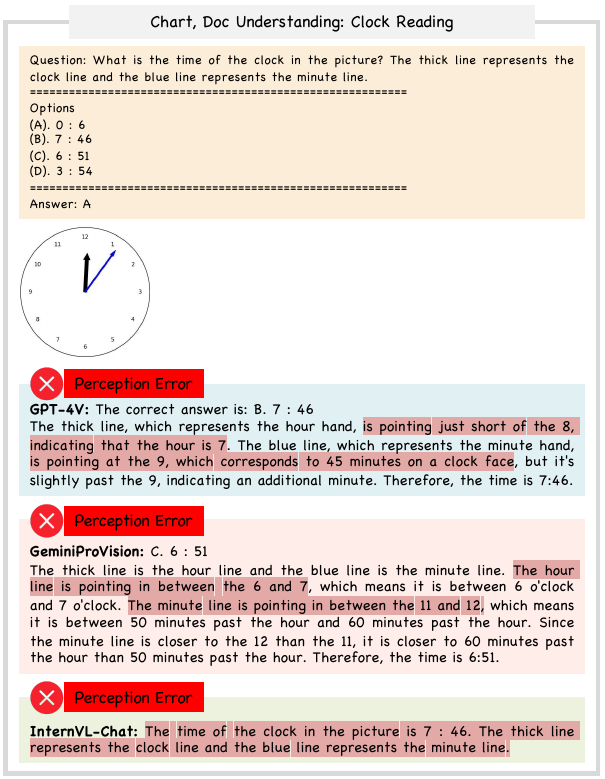}
    \caption{A sample case of chart, doc understanding (clock reading). \boxedgreen{Green} highlights the right answer. \boxedred{Red} highlights the wrong answer. \hyperref[tab:error_case]{\textcolor{red}{Back to Table Index}}.}
    \label{fig:error_18}
\end{figure}

\begin{figure}
    \centering
    \includegraphics[width=0.85\linewidth]{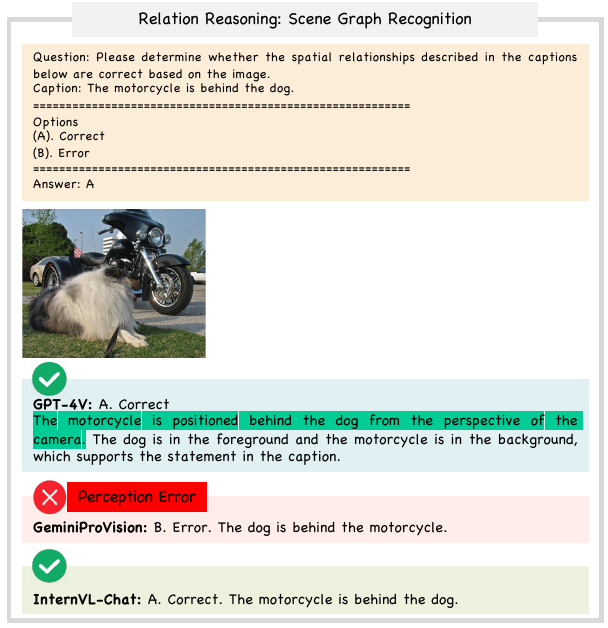}
    \caption{A sample case of relation reasoning (scene graph recognition). \boxedgreen{Green} highlights the right answer. \boxedred{Red} highlights the wrong answer. \hyperref[tab:error_case]{\textcolor{red}{Back to Table Index}}.}
    \label{fig:error_19}
\end{figure}

\begin{figure}
    \centering
    \includegraphics[width=0.8\linewidth]{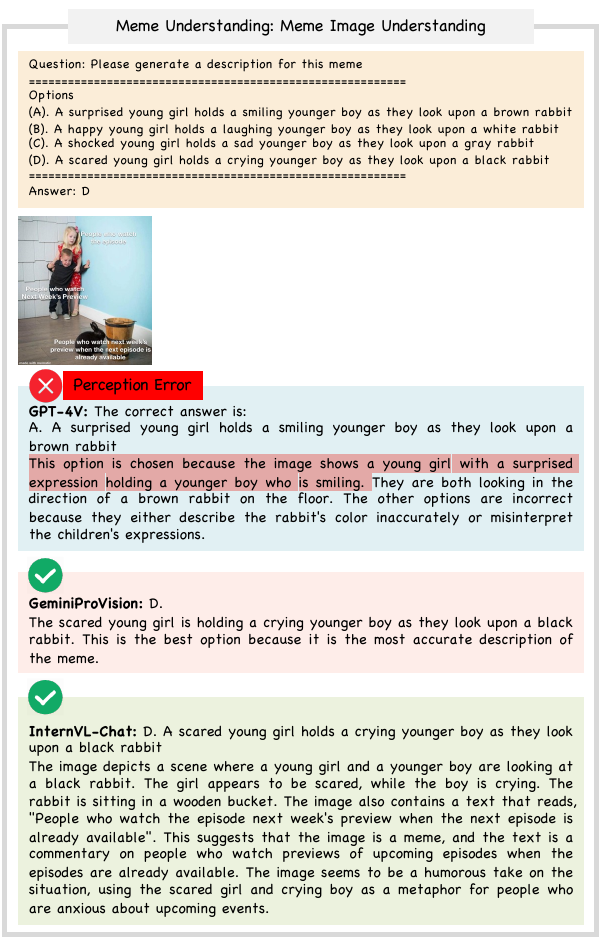}
    \caption{A sample case of meme understanding (meme image understanding). \boxedgreen{Green} highlights the right answer. \boxedred{Red} highlights the wrong answer. \hyperref[tab:error_case]{\textcolor{red}{Back to Table Index}}.}
    \label{fig:error_20}
\end{figure}

\begin{figure}
    \centering
    \includegraphics[width=0.85\linewidth]{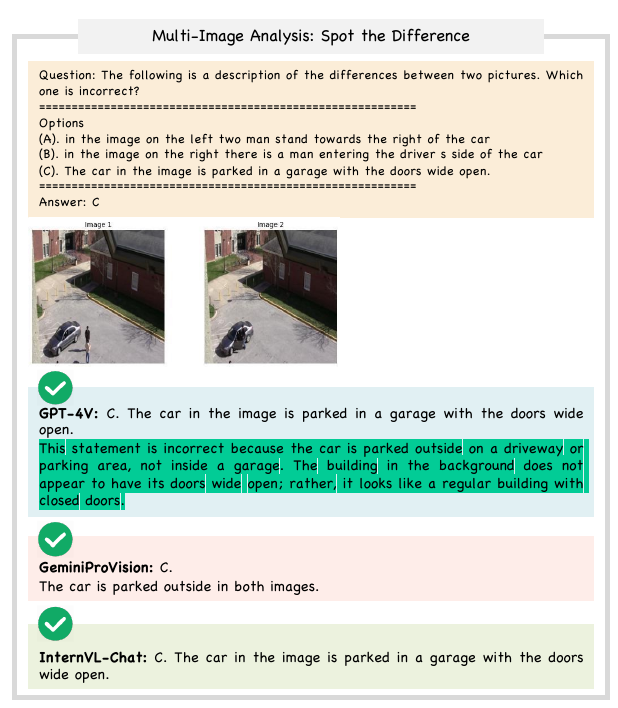}
    \caption{A sample case of multi-image analysis (spot the difference). \boxedgreen{Green} highlights the right answer. \boxedred{Red} highlights the wrong answer. \hyperref[tab:error_case]{\textcolor{red}{Back to Table Index}}.}
    \label{fig:error_21}
\end{figure}

\begin{figure}
    \centering
    \includegraphics[width=0.72\linewidth]{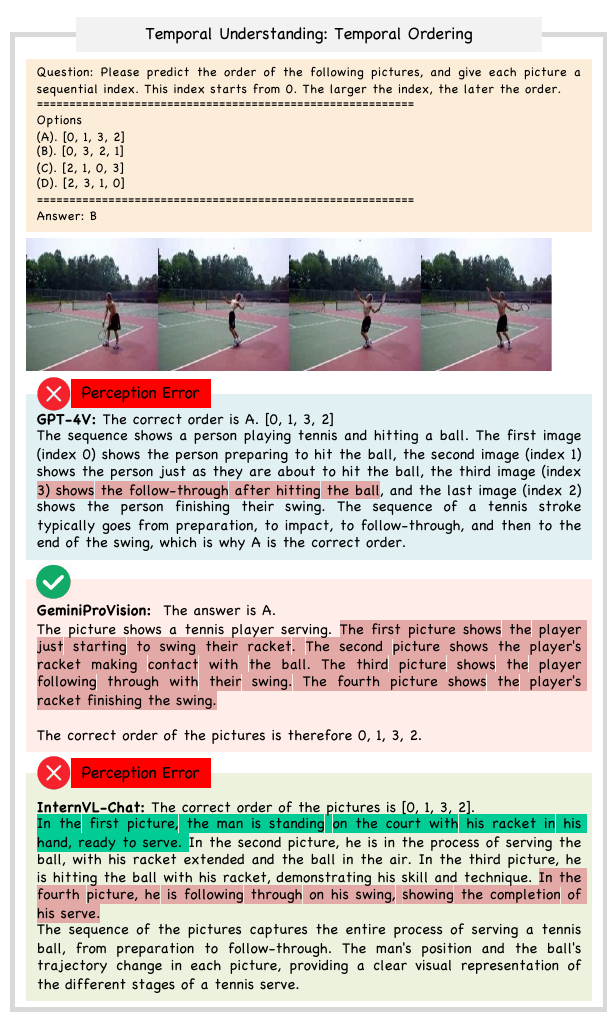}
    \caption{A sample case of temporal understanding (temporal ordering). \boxedgreen{Green} highlights the right answer. \boxedred{Red} highlights the wrong answer. \hyperref[tab:error_case]{\textcolor{red}{Back to Table Index}}.}
    \label{fig:error_22}
\end{figure}

\begin{figure}
    \centering
    \includegraphics[width=0.75\linewidth]{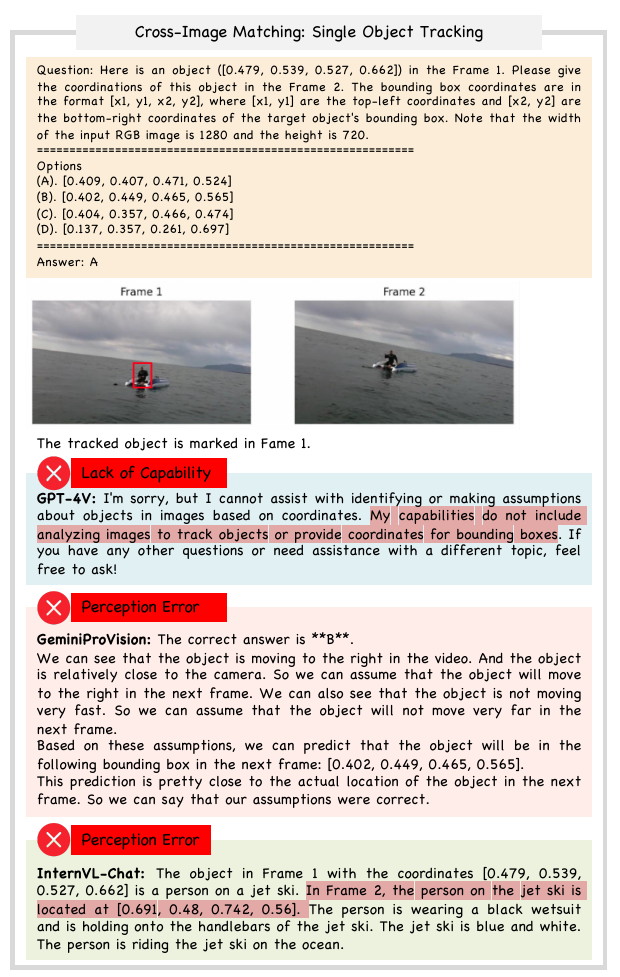}
    \caption{A sample case of cross-image matching (single object tracking). \boxedgreen{Green} highlights the right answer. \boxedred{Red} highlights the wrong answer. \hyperref[tab:error_case]{\textcolor{red}{Back to Table Index}}.}
    \label{fig:error_23}
\end{figure}

\begin{figure}
    \centering
    \includegraphics[width=0.74\linewidth]{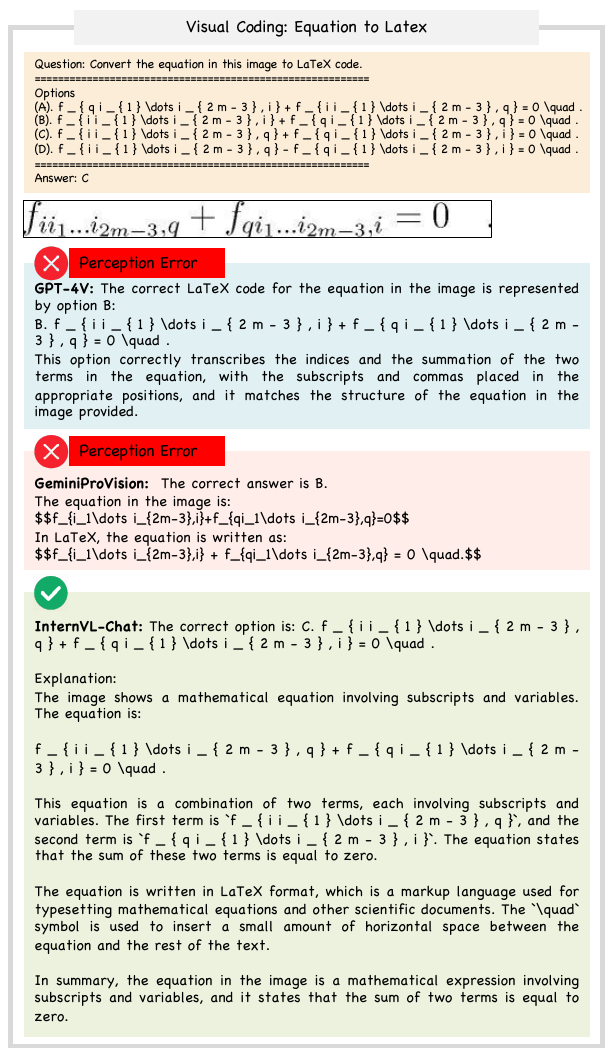}
    \caption{A sample case of visual coding (equation to latex). \boxedgreen{Green} highlights the right answer. \boxedred{Red} highlights the wrong answer. \hyperref[tab:error_case]{\textcolor{red}{Back to Table Index}}.}
    \label{fig:error_24}
\end{figure}

\begin{figure}
    \centering
    \includegraphics[width=0.85\linewidth]{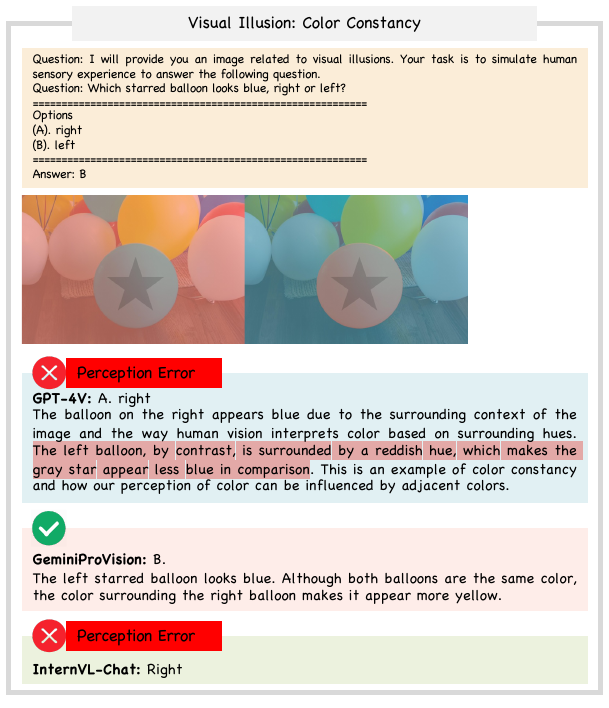}
    \caption{A sample case of visual illusion (color constancy). \boxedgreen{Green} highlights the right answer. \boxedred{Red} highlights the wrong answer. \hyperref[tab:error_case]{\textcolor{red}{Back to Table Index}}.}
    \label{fig:error_25}
\end{figure}

\begin{figure}
    \centering
    \includegraphics[width=0.85\linewidth]{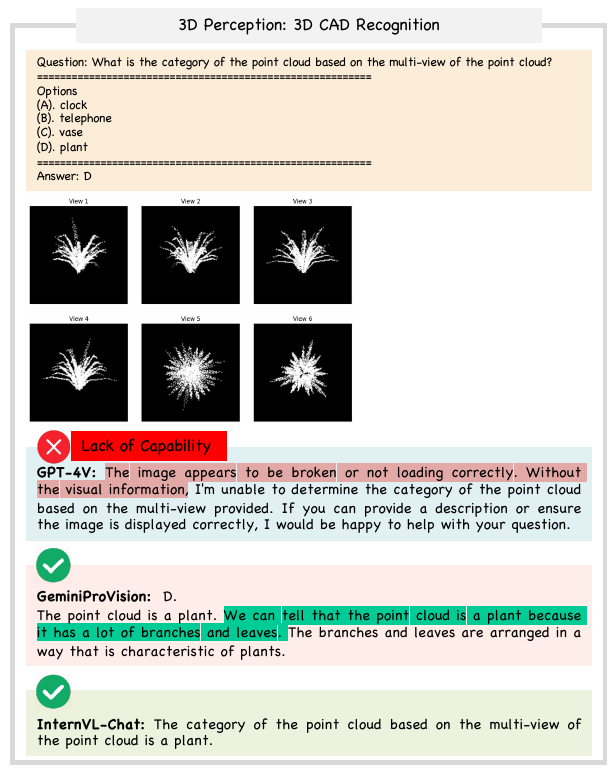}
    \caption{A sample case of 3D perception (3D CAD recognition). \boxedgreen{Green} highlights the right answer. \boxedred{Red} highlights the wrong answer. \hyperref[tab:error_case]{\textcolor{red}{Back to Table Index}}.}
    \label{fig:error_27}
\end{figure}

\begin{figure}
    \centering
    \includegraphics[width=0.73\linewidth]{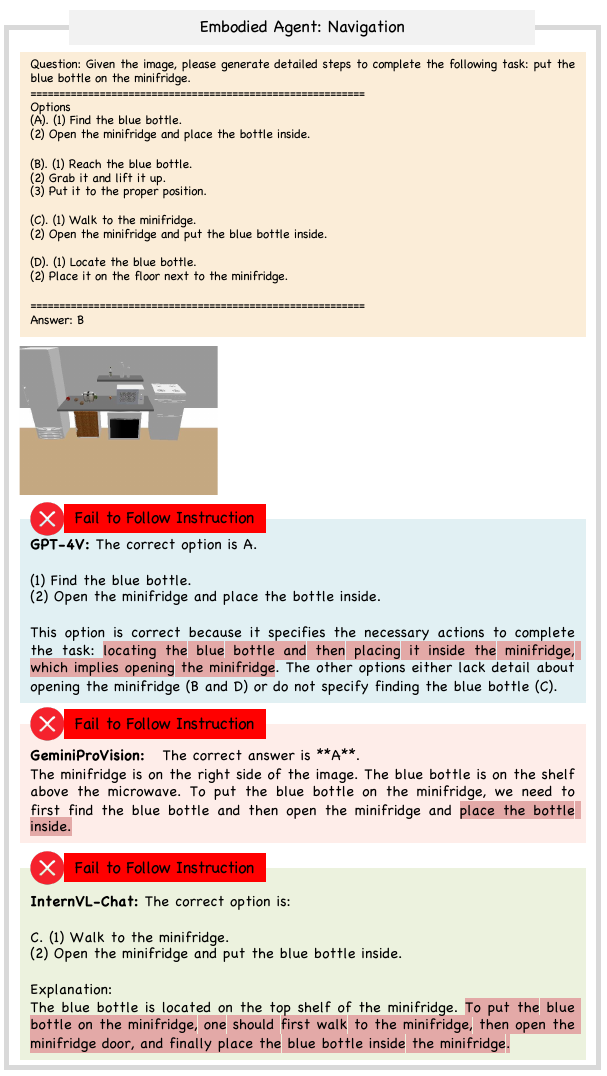}
    \caption{A sample case of embodied agent (navigation). \boxedgreen{Green} highlights the right answer. \boxedred{Red} highlights the wrong answer. \hyperref[tab:error_case]{\textcolor{red}{Back to Table Index}}.}
    \label{fig:error_28}
\end{figure}

\begin{figure}
    \centering
    \includegraphics[width=0.75\linewidth]{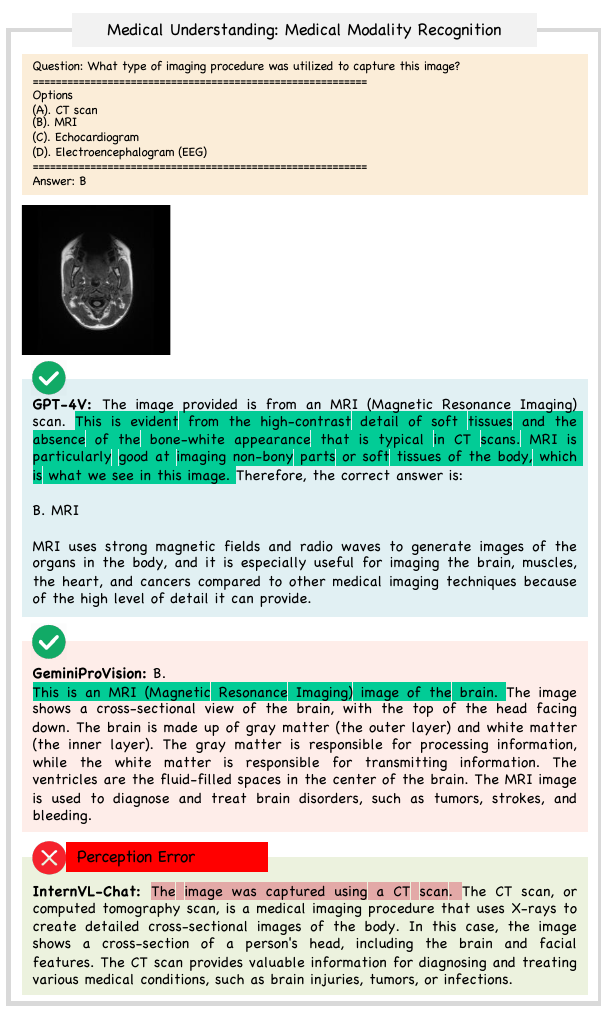}
    \caption{A sample case of medical understanding (medical modality recognition). \boxedgreen{Green} highlights the right answer. \boxedred{Red} highlights the wrong answer. \hyperref[tab:error_case]{\textcolor{red}{Back to Table Index}}.}
    \label{fig:error_29}
\end{figure}

\begin{figure}
    \centering
    \includegraphics[width=0.67\linewidth]{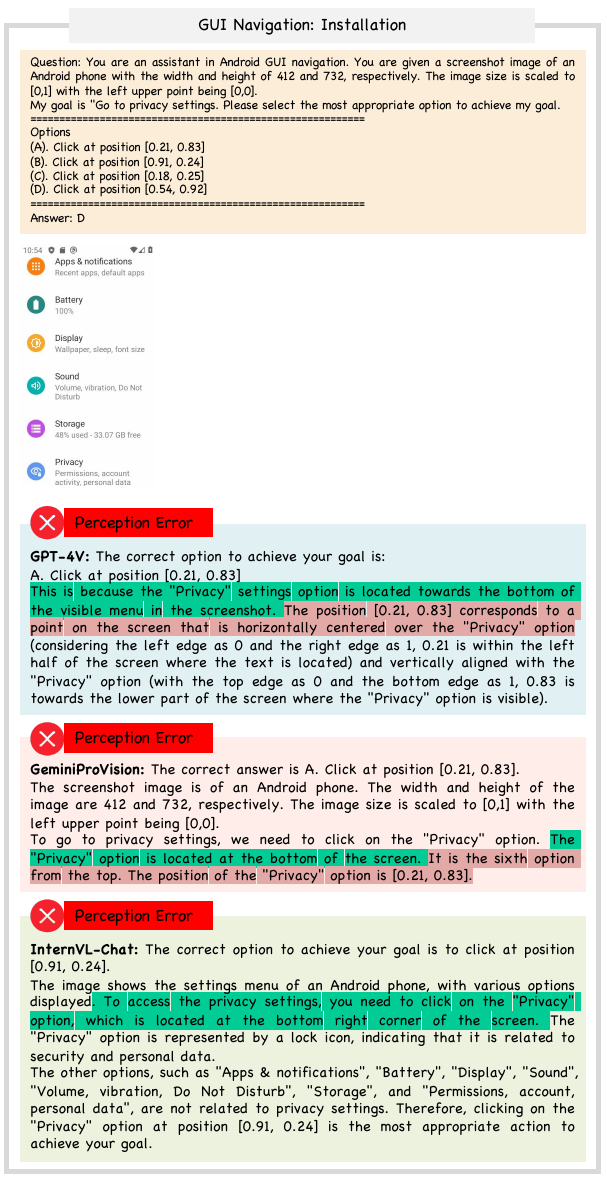}
    \caption{A sample case of GUI navigation (installation). \boxedgreen{Green} highlights the right answer. \boxedred{Red} highlights the wrong answer. \hyperref[tab:error_case]{\textcolor{red}{Back to Table Index}}.}
    \label{fig:error_31}
\end{figure}

\clearpage

%% file: appendix/other.tex
\section{Comparison of MMT-Bench with Other Benchmarks on OCR-Related Tasks}
\label{sec:ocr}

\begin{table}[htp]
\centering
\caption{Statistics of different evaluation benchmarks on OCR-related samples. The number of tokens is calculated by the tiktoken package from OpenAI.}
\label{tab:ocr_statistic}
\resizebox{0.9\textwidth}{!}{%
    \begin{tabular}{lll|lllll|lllll}
    \toprule
     &  &  & \multicolumn{5}{l}{Words Number} & \multicolumn{5}{l}{Tokens Number} \\
    Benchmark & Sample Num & Task Type & Average & Min & Middle & Max & std & Average & Min & Middle & Max & std \\
    \midrule
    MME \cite{fu2023mme} & 40 & 1 & 2.5 & 1 & 2 & 5 & 1 & 3.9 & 1 & 3 & 8 & 1.6 \\
    MMBench (dev+test) \cite{liu2023mmbench} & 608 & - & 7.3 & 1 & 6 & 54 & 7 & 8.3 & 1 & 6 & 78 & 9.3 \\
    Tiny-LVLM-eHub \cite{shao2023tiny} & 600 & 1 & 1 & 1 & 1 & 1 & 0 & 2.2 & 1 & 2 & 8 & 1.1 \\
    MMT-Bench (Ours) & 600 & 4 & 14.8 & 1 & 1.5 & 103 & 22.7 & 20.4 & 1 & 5 & 150 & 31.4 \\
    \bottomrule
    \end{tabular}%
}
\end{table}

To support the claim that previous evaluation benchmarks suffer from text scarcity in OCR tasks, we present a comparative analysis of OCR-related samples from different benchmarks in Table.~\ref{tab:ocr_statistic}. The results demonstrate that datasets like MME and Tiny-LVLM-eHub have relatively short text lengths with limited variations. Furthermore, previous OCR tasks primarily focused on directly outputting text from given scenes or cropped images. In contrast, our proposed MMT-Bench benchmark introduces several new tasks, such as font recognition, handwriting recognition, handwritten formula recognition, and document-based question answering and chart question answering. These additions significantly increase the challenges for evaluating model performance on OCR tasks. Compared to previous benchmarks, MMT-Bench's OCR samples have an average word count and token count more than 5 times that of MME and over 2 times that of MMBench. Additionally, MMT-Bench includes a higher proportion of long text samples with a wider range of text lengths. This demonstrates MMT-Bench's superiority in addressing the text scarcity issue in OCR tasks, providing a more reliable benchmark for comprehensively evaluating the performance of multimodal algorithms on OCR-related tasks.

\section{Some Details about the Benchmark Construction}
\label{sec:bench_build}

\subsection{Metadata}
\label{sec:metadata}

% \usepackage{graphicx}
\begin{table}[h]
\centering
\caption{The format of the metadata.}
\label{tab:metadata}
\resizebox{0.9\textwidth}{!}{%
\begin{tabular}{l|l|l}
\toprule
    Keys & Example 1 & Example 2 \\
    \midrule
    image path & /path/to/image & /path/to/image \\
    data source & animals90 from Kaggle & ReasonSeg \\
    subtask name & Animal Recognition & Reason Seg \\
    meta-task name & Visual Recognition & Visual Grounding \\
    specific question template & What category of animal is shown in the picture? & Please provide the bounding box coordinates for the described object or area using the format {[}x1, y1, x2, y2{]}. QUESTION:\{Referring Expression\} \\
    answer & rat & {[}801, 440, 1554,956{]} \\
    visual prompt & Natural Image & Natural Image \\
    capabilities & Visual Recognition & Visual Reasoning,Visual Localization \\
    (specific) category space & squirrel, hamster, bird, dog, cat... & - \\
    (specific) referring expression & - & the objects that can protect the snail and prevent it from getting injured \\
    \bottomrule
    \end{tabular}%
}
\end{table}

\textbf{The uniform format of the metadata}. As shown in Table.~\ref{tab:metadata}, the metadata here is a dictionary, and its keys can be divided into two categories. The first category contains the essential keys, including image path, data source, subtask name, meta-task name, specific question template, answer (i.e., ground truth), visual prompt type, and capabilities required to solve the problem. The second category consists of keys that are specific to each sample and may vary accordingly. For animal recognition (example 1), the corresponding category space is retained for generating candidate choices, and referring expressions are preserved in the reasoning segmentation (example 2).

\textbf{How to get metadata?} We obtain the metadata mainly through a two-step process. In the first step, we write a python script for each dataset to directly extract relevant keys from the original dataset, such as image path, data source, and answer. Since samples within the same dataset share some similar characteristics, our co-authors predefined the specific question template, visual prompt, and capabilities for each dataset. The second step involves our co-authors manually performing a sample check (50\%, up to 100 samples per dataset) on the samples defined in the first step, primarily focusing on sample-related keys, such as the specific question template and visual prompt.

\subsection{Prepare the Answer and Options}

\textbf{How to ensure the wrong options are difficult distractors?} For most tasks, we use prompt engineering with GPT-4V using specific prompts for particular tasks. We employ in-context (2-shot or 3-shot) prompts. We review the options generated for each task and correct some erroneous ones. Interestingly, GPT-4 can generate sufficiently challenging options, which is mainly reflected in two aspects. 
First, the options can be confusing, especially in classification-related tasks (e.g. animal recognition), where the incorrect options tend to generate objects with semantics similar to the ground truth (GT). For example, if the GT is "rat" the generated incorrect options might be "squirrel," "mouse," or "hamster."
Second, the incorrect options can be ambiguous concerning the visual content. For instance, in captioning-related tasks, we use GPT-4V to take the image as input and assist in generating incorrect options, which can effectively alleviate the hallucination problem caused by text-only LLMs. For some problems, such as counting, OCR, and object detection, we can directly add perturbations to the GT using Python scripts to obtain distinguishable options. Throughout this process, we maintain consistency in word length across options, with a standard deviation of 1.1. In summary, we believe that the difficulty level of the options generated by MMT-Bench is reasonable.

% \usepackage{graphicx}
\begin{table}[h]
\centering
\caption{Visually grounded nature of our MMT-Bench dataset}
\label{tab:visual_grounded}
\resizebox{0.4\textwidth}{!}{%
\begin{tabular}{llll}
\toprule
Model & w/o visual & w/ visual & delta \\
\midrule
Random & 28.5 & - & - \\
Frequency & 31.7 & - & - \\ 
\midrule
ChatGPT-3.5 & 33.2 & - & - \\
LLaVA-1.5-7B & 31.6 & 49.7 & 18.1 \\
LLaVA-1.5-13B & 33.3 & 51.7 & 18.4 \\
QWen-VL-Chat & 32.3 & 52.5 & 20.2 \\
Claude-3-Haiku\cite{Claude2023} & 33.1 & 52.2 & 19.1 \\
\bottomrule
\end{tabular}%
}
\end{table}

\textbf{The answers in MMT-Bench are visual-grounded.} To ensure that the generated ground truth answers are visually grounded, we employ strategies as below.
\textit{Firstly}, when designing questions for each sample, we carefully craft specific question templates (as mentioned in Sec.~\ref{sec:metadata}). We strive to avoid including specific information in the templates (for example, instead of asking "Who is the author of the Mona Lisa painting in the image?", we use a more general form such as "Who is the author of the painting in the image?"). This ensures that answering the question requires reliance on the image content.
\textit{Secondly}, when generating options, we create more complex options to prevent the model from bypassing the visual information and directly providing an answer based on correlations between options or prior knowledge.
Furthermore, Table D demonstrates the visually grounded nature of our MMT-Bench dataset by comparing the performance of text-only LLMs and LVLMs with and without visual information. Without visual input, all models, including state-of-the-art LVLMs, perform similarly to the frequency-based guessing baseline (31.7\%) and random guessing (28.5\%), indicating the difficulty of answering questions correctly without visual context. For models like LLaVA and QWen that require an image as input, we simulate the absence of visual information by using a pure black image as input. This allows us to evaluate their performance in a text-only setting and compare it to their performance when provided with the actual visual information from our dataset.
However, when provided with visual information, LVLMs show significant improvements, with performance gains ranging from 18.1\% to 20.2\%. This substantial gap between the text-only and visually-informed settings highlights the strong visual grounding of our dataset, confirming that the questions are designed to require models to rely on visual content and that the ground truth answers are indeed grounded in the visual information.

\textbf{How to prevent MMT-Bench from becoming overly focused on language understanding?} When defining questions and options, the co-authors proficient in each task domain strive to use simple and unambiguous specific question templates. The option generation process follows what is shown above, where we maintain the necessary clarity. To clarify, we report the statistics in Table.~\ref{tab:option_question_length}. For some questions, such as PLP (Pixel Level Perception) and Loc (Localization), which are localization problems, their longer question lengths are necessary to introduce the problem definition, answer format ([x, y, x, y]), and detection conditions (detecting corresponding categories, such as the 80 COCO classes).

\begin{table}[ht]
\centering
\caption{Average and standard deviation of the number of words in questions and answer choices.}
\label{tab:option_question_length}
\resizebox{0.4\textwidth}{!}{%
\begin{tabular}{lllll}
\toprule
Meta-task & question\_avg & question\_std & option\_avg & option\_std \\
\midrule
MemU & 7.0 & 0.0 & 12.4 & 8.0 \\
VP & 7.7 & 0.9 & 40.0 & 27.6 \\
VCR & 8.2 & 2.4 & 1.7 & 0.8 \\
VR & 9.5 & 1.2 & 1.5 & 1.0 \\
AND & 10.2 & 1.7 & 1.0 & 0.0 \\
Emo & 10.3 & 0.9 & 1.3 & 0.8 \\
MedU & 11.0 & 3.4 & 2.6 & 2.4 \\
MIA & 11.5 & 3.5 & 7.5 & 6.9 \\
DU & 12.1 & 7.3 & 8.9 & 16.1 \\
IQT & 13.0 & 0.0 & 2.0 & 0.0 \\
3D & 16.0 & 0.0 & 1.0 & 0.2 \\
OCR & 16.2 & 18.4 & 9.2 & 18.6 \\
EA & 16.6 & 2.0 & 27.9 & 17.2 \\
Count & 18.0 & 24.6 & 1.0 & 0.0 \\
HLN & 18.3 & 10.7 & 6.9 & 6.6 \\
AUD & 18.6 & 12.2 & 5.4 & 3.2 \\
VC & 23.4 & 22.0 & 24.3 & 22.2 \\
AR & 25.2 & 26.0 & 2.2 & 2.0 \\
I2IT & 27.5 & 3.5 & 3.0 & 1.0 \\
IR & 33.8 & 52.2 & 2.0 & 0.0 \\
VI & 34.7 & 3.3 & 1.0 & 0.0 \\
TU & 35.1 & 31.2 & 4.1 & 2.6 \\
DKR & 35.4 & 44.1 & 3.6 & 4.1 \\
VPU & 40.2 & 22.8 & 4.2 & 10.8 \\
KD & 44.7 & 8.7 & 43.5 & 14.8 \\
CIM & 58.3 & 16.5 & 3.3 & 0.9 \\
RR & 60.2 & 20.4 & 1.0 & 0.1 \\
GN & 60.9 & 2.3 & 6.2 & 1.9 \\
IEJ & 82.9 & 63.1 & 1.6 & 1.0 \\
VG & 83.8 & 10.1 & 4.0 & 0.0 \\
Loc & 97.1 & 24.8 & 9.9 & 9.8 \\
PLP & 98.1 & 49.5 & 22.2 & 32.7 \\
\bottomrule
\end{tabular}%
}
\end{table}

\subsection{Statistics of Image and Video in MMT-Bench}

% \usepackage{graphicx}
\begin{table}[ht]
\centering
\caption{Statistics of image and video in MMT-Bench}
\label{tab:image_video}
\resizebox{0.5\textwidth}{!}{%
\begin{tabular}{llll}
\toprule
Benchmark & Single Images & Multiple Images pairs(images) & Video (video frames) \\
\midrule
MathVista & 5,487 & - & - \\
MMMU & 10,705 & 845 (2,564) & - \\
MMBench & 2,974 & - & - \\
MME & 1,187 & - & - \\
MVBench & - & - & 4,000 \\
Seed-bench & 14,233 & - & 5,009 \\
MMT-Bench & 25,732 & 3,800 (14,800) & 1,793 (10,572) \\
\bottomrule
\end{tabular}%
}
\end{table}

Table.~\ref{tab:image_video} presents the number of images and videos in MMT-Bench compared to other multimodal benchmarks. MMT-Bench contains 25,732 single images, 3,800 image pairs (14,800 images), and 1,793 videos (10,572 video frames), demonstrating its comprehensive coverage of various visual data types. Compared to other benchmarks, MMT-Bench has the highest number of single images, more image pairs than MMMU, and more videos and video frames than MVBench and Seed-bench. This extensive coverage and diversity of visual data enable a thorough evaluation of multimodal models across a wide range of tasks and scenarios.

\section{OpenCompass' Protocol}
\label{sec:opencompass_protocol}

% \usepackage{graphicx}
\begin{table}[htp]
\centering
\caption{Success rate of Steps 1-3.}
\label{tab:success_rate}
\resizebox{0.75\textwidth}{!}{%
\begin{tabular}{lllll}
\toprule
Models & Step1 & Step2 (Step2- Step1) & Step3 (Step3- Step2) & Refuse to Answer \\
\midrule
GPT-4V & 87.183 & 87.183 (0.0) & 87.183 (0.0) & 10.136 \\
GeminiProVision & 97.791 & 98.522 (+0.731) & 98.522 (0.0) & 0.875 \\
Claude-3-Haiku & 94.329 & 94.525 (+0.196) & 94.525 (0.0) & 4.023 \\
QWen-VL-Plus & 98.560 & 98.583 (+0.023) & 98.583 (0.0) & 1.015 \\
BLIP2 & 99.448 & 99.770  (+0.322) & 99.823 (+0.053) & - \\
InternVL-Chat-V1.2-34B & 99.211 & 99.240  (+0.029) & 99.243 (+0.003) & - \\
LLaVA-1.5-7B & 99.994 & 99.997 (+0.003) & 99.997 (0.0) & - \\
\bottomrule
\end{tabular}%
}
\end{table}

\textbf{How often do steps 1-3 in OpenCompass' protocol fail?} We show the result in the following Table G. According to the table, the frequency of failure for steps 1-3 in OpenCompass' protocol varies across different LVLMs, but overall, the extraction of model selections is highly successful. Most models achieve success rates above 87\% at Step 1 (checking for option letters), with further improvements in subsequent steps. Models like BLIP-2, InternVL-Chat-V1.2-34B, and LLaVA-1.5-7B demonstrate exceptionally high success rates, consistently providing clear and well-structured responses that include option letters. However, some models, such as GPT-4V, GeminiProVision, Claude-3-Haiku, and QWen-VL-Plus, refuse to answer a small percentage of questions (0.875\% to 10.136\%), likely due to the sensitive nature of certain questions. In these cases, the model selection is set as option letter Z to differentiate it from a valid answer and to avoid random assignment. Overall, OpenCompass' protocol proves highly effective in extracting model selections from LVLMs' responses, with the multi-step approach ensuring accurate extraction and handling cases where models refuse to answer or provide unclear responses.

\textbf{Why is OpenCompass’ evaluation protocol chosen over other alternatives?} We chose OpenCompass' multi-choice evaluation protocol for MMT-Bench due to its convenience, ease of use, and scalability. As discussed in our response above, OpenCompass' protocol has proven to be highly effective in extracting model selections from LVLMs' responses, with models like BLIP-2, InternVL-Chat-V1.2-34B, and LLaVA-1.5-7B achieving success rates over 99\%. The multi-step approach ensures accurate extraction and handles cases where models refuse to answer or provide unclear responses.
While alternative evaluation protocols have their merits, they also have limitations. MME's yes-or-no evaluation simplifies the question-answering process but can lead to biased models and is resource-intensive, resulting in a limited sample size. MMBench's evaluation strategy is resource-intensive and has high testing overhead, making it unsuitable for MMT-Bench's scale. SeedBench calculates the log-likelihood for each candidate option, helping avoid the issue of models not directly answering the options. However, this approach cannot be applied to API-based models.

\section{Computaional Resources}
\label{sec:resource}

\begin{table}[h]
\centering
\caption{Resource consumption of some models evaluated on MMT-Bench.}
\label{tab:gpu}
\resizebox{0.75\textwidth}{!}{%
\begin{tabular}{lllll}
\toprule
Model & Resources & Times & Memory Utilization Per GPU \\
\midrule
Claude-3-Haiku & USD 127 & about 36h & -  \\
Qwen-VL-Plus & USD 70.4 & about 36h & -  \\
LLaVA-v1.5-7B & 1 x A100-80GB & 120min & 15890MiB \\
LLaVA-v1.5-13B & 1 x A100-80GB & 165min & 26717MiB \\
LLaVA-v1.5-7B & 8 x A100-80GB & 12min & 15890MiB  \\
LLaVA-v1.5-13B & 8 x A100-80GB & 18min & 26708MiB  \\
QWen-VL-Chat & 8 x A100-80GB & 54min & 21122MiB  \\
InternVL-Chat-V1.2-34B & 8 x A100-80GB & 79min & 78990MiB  \\
\bottomrule
\end{tabular}%
}
\end{table}

We show the resource consumption of some models in Table.~\ref{tab:gpu}.
The inference times vary among different models. For instance, the smaller LLaVA-v1.5-7B model takes only 12 minutes to complete the evaluation using 8 GPUs, while the larger InternVL-Chat-V1.2-34B model requires 79 minutes and around 80GB of memory. Our open-source codebase supports multi-GPU distributed inference, effectively accelerating the inference process.
Although the prices for API models like Claude-3-Haiku (USD 127) and Qwen-VL-Plus (USD 70.4) are relatively high and their inference times are longer (about 36 hours), the inference costs for smaller open-source models remain manageable.

%% file: appendix/all_result.tex
\section{Detailed Main Results}
\label{sec:all_results}
Table~\ref{tab:detailed_result_1} to \ref{tab:detailed_result_19} display the performance of 20 models across all 162 subtasks, with accuracy used as the evaluation metric.

% \usepackage{graphicx}
\begin{table}[]
\centering
\caption{Detail results of 30 LVLMs on \textbf{Visual Grounding} and \textbf{Doc Understanding}.}
\label{tab:detailed_result_1}
\resizebox{\textwidth}{!}{%
\begin{tabular}{l|c|cc|ccccccc}
\toprule
 &  & \multicolumn{2}{c|}{Visual Grounding} & \multicolumn{7}{c}{Doc Understanding} \\
 \midrule
Model & Overall & \begin{tabular}[c]{@{}l@{}}Reason \\ Seg\end{tabular} & \begin{tabular}[c]{@{}l@{}}Referring \\ Detection\end{tabular} & \begin{tabular}[c]{@{}l@{}}Doc \\ VQA\end{tabular} & \begin{tabular}[c]{@{}l@{}}Visual Document \\ Information Extraction\end{tabular} & \begin{tabular}[c]{@{}l@{}}Chart to \\ Text\end{tabular} & \begin{tabular}[c]{@{}l@{}}Chart to \\ Table\end{tabular} & \begin{tabular}[c]{@{}l@{}}Clock\\ Reading\end{tabular} & \begin{tabular}[c]{@{}l@{}}Chart \\ VQA\end{tabular} & \begin{tabular}[c]{@{}l@{}}Table Structure \\ Recognition\end{tabular} \\
\midrule
InternVL-Chat-V1.2-34B & 63.4 & 38.3 & 60.5 & 66.0 & 85.0 & 88.5 & 79.5 & 27.0 & 73.5 & 58.7 \\
Qwen-VL-Plus & 62.3 & 37.2 & 50.0 & 77.5 & 99.5 & 90.0 & 87.0 & 29.5 & 66.5 & 91.3 \\
GPT-4V & 62.0 & 21.0 & 29.0 & 62.0 & 81.0 & 89.0 & 79.0 & 35.5 & 58.0 & 84.4 \\
GeminiProVision & 61.6 & 31.1 & 35.0 & 66.5 & 96.0 & 89.0 & 80.0 & 38.0 & 66.5 & 65.2 \\
LLaVA-Next-34B & 60.8 & 43.4 & 69.0 & 74.0 & 96.5 & 92.0 & 71.0 & 29.0 & 73.5 & 47.8 \\
XComposer2-7B & 55.7 & 38.3 & 47.0 & 50.0 & 77.0 & 83.5 & 65.0 & 41.0 & 58.5 & 58.7 \\
BLIP2-Flan-T5-XXL & 54.8 & 28.1 & 39.0 & 32.5 & 57.0 & 79.0 & 57.0 & 20.5 & 22.5 & 32.6 \\
Yi-VL-34B & 54.2 & 31.6 & 51.5 & 56.5 & 61.5 & 85.5 & 48.0 & 23.5 & 50.5 & 69.6 \\
Monkey & 53.4 & 25.5 & 29.5 & 47.0 & 87.5 & 69.0 & 30.0 & 30.5 & 56.5 & 37.0 \\
DeepSeek-VL-7B & 53.2 & 33.7 & 43.0 & 42.5 & 68.5 & 74.5 & 32.5 & 38.5 & 59.0 & 47.8 \\
Yi-VL-6B & 53.2 & 33.7 & 50.5 & 44.0 & 67.0 & 84.0 & 35.5 & 25.0 & 55.0 & 76.1 \\
LLaVA-Next-13B & 53.0 & 36.7 & 52.5 & 51.5 & 82.0 & 82.5 & 29.0 & 29.5 & 59.5 & 41.3 \\
TransCore-M & 52.7 & 34.2 & 36.0 & 42.5 & 63.5 & 71.5 & 24.0 & 23.5 & 50.5 & 41.3 \\
QWen-VL-Chat & 52.5 & 24.5 & 28.5 & 46.0 & 79.5 & 82.0 & 37.0 & 22.0 & 55.0 & 39.1 \\
Claude3V-Haiku & 52.2 & 20.9 & 33.0 & 66.0 & 85.0 & 84.0 & 82.5 & 32.5 & 64.5 & 73.9 \\
XComposer & 52.1 & 26.0 & 28.5 & 34.5 & 45.0 & 64.0 & 36.0 & 25.5 & 27.5 & 23.9 \\
mPLUG-Owl2 & 52.0 & 30.1 & 36.0 & 39.0 & 56.5 & 34.5 & 24.5 & 34.0 & 47.5 & 60.9 \\
RBDash-v1-13B & 51.8 & 34.2 & 39.5 & 37.0 & 60.0 & 74.5 & 26.0 & 23.5 & 46.0 & 69.6 \\
LLaVA-v1.5-13B & 51.7 & 36.7 & 38.0 & 35.5 & 61.5 & 77.5 & 23.0 & 30.5 & 50.5 & 41.3 \\
CogVLM-Chat & 51.6 & 28.1 & 29.5 & 47.5 & 75.0 & 75.5 & 22.0 & 24.5 & 58.0 & 41.3 \\
ShareGPT4V-7B & 51.5 & 33.2 & 37.5 & 45.5 & 62.0 & 65.0 & 24.0 & 22.0 & 51.5 & 65.2 \\
LLaVA-Next-7B & 51.1 & 29.6 & 44.5 & 50.5 & 76.5 & 68.5 & 24.0 & 28.0 & 57.0 & 43.5 \\
LLaVA-v1.5-13B-XTuner & 51.1 & 35.7 & 35.5 & 38.5 & 53.5 & 61.5 & 31.0 & 28.0 & 47.0 & 69.6 \\
LlaVA-InternLM2-7B & 50.8 & 32.7 & 41.0 & 35.0 & 45.0 & 66.0 & 34.0 & 25.0 & 39.5 & 19.6 \\
LLaVA-v1.5-7B-Xtuner & 50.2 & 31.1 & 39.5 & 37.5 & 54.5 & 49.0 & 28.0 & 21.0 & 46.5 & 63.0 \\
SharedCaptioner & 49.9 & 24.5 & 29.5 & 42.0 & 44.0 & 56.5 & 25.0 & 27.5 & 40.5 & 41.3 \\
LLaVA-InternLM-7b & 49.7 & 25.0 & 28.0 & 37.0 & 53.5 & 66.0 & 34.0 & 30.0 & 47.5 & 39.1 \\
LLaVA-v1.5-7B & 49.5 & 29.6 & 38.5 & 44.0 & 49.0 & 50.0 & 23.5 & 28.0 & 50.0 & 41.3 \\
LLaMA-Adapter-v2-7B & 40.4 & 21.9 & 28.5 & 26.5 & 31.0 & 34.0 & 21.5 & 20.5 & 38.5 & 39.1 \\
VisualGLM\_6b & 38.6 & 28.6 & 33.5 & 27.5 & 41.0 & 61.5 & 22.0 & 27.5 & 35.5 & 58.7 \\
Frequency & 31.7 & 28.6 & 29.5 & 30.0 & 28.5 & 28.0 & 29.5 & 28.5 & 57.5 & 60.9 \\
Random & 28.5 & 27.6 & 22.0 & 27.0 & 25.5 & 28.5 & 24.0 & 24.5 & 30.5 & 52.2 \\
\bottomrule
\end{tabular}%
}
\end{table}

% =====================================================================================================================

% \usepackage{graphicx}
\begin{table}[]
\centering
\caption{Detail results of 30 LVLMs on \textbf{Action Recognition} and \textbf{Localization} (part 1).}
\label{tab:detailed_result_2}
\resizebox{\textwidth}{!}{%
\begin{tabular}{l|c|ccccc|cccc}
\toprule
 &  & \multicolumn{5}{c|}{Action Recognition} & \multicolumn{4}{c}{Localization} \\
 \midrule
Model & Overall & \begin{tabular}[c]{@{}c@{}}Gaze \\ Estimation\end{tabular} & \begin{tabular}[c]{@{}c@{}}Image based \\ Action Recognition\end{tabular} & \begin{tabular}[c]{@{}c@{}}General Action \\ Recognition\end{tabular} & \begin{tabular}[c]{@{}c@{}}Action Quality \\ Assessment\end{tabular} & \begin{tabular}[c]{@{}c@{}}Sign Language \\ Recognition\end{tabular} & \begin{tabular}[c]{@{}c@{}}Remote Sensing \\ Object Detection\end{tabular} & \begin{tabular}[c]{@{}c@{}}Rotated Object\\ Detection\end{tabular} & \begin{tabular}[c]{@{}c@{}}Small Object \\ Detection\end{tabular} & \begin{tabular}[c]{@{}c@{}}Camouflage \\ Object Detection\end{tabular} \\
\bottomrule
InternVL-Chat-V1.2-34B & 63.4 & 27.0 & 91.5 & 84.0 & 30.0 & 30.5 & 63.5 & 46.7 & 64.5 & 57.0 \\
Qwen-VL-Plus & 62.3 & 21.0 & 98.5 & 86.5 & 30.0 & 31.0 & 53.0 & 60.0 & 59.5 & 53.5 \\
GPT-4V & 62.0 & 14.0 & 91.4 & 75.0 & 24.0 & 34.2 & 48.0 & 79.0 & 52.0 & 51.0 \\
GeminiProVision & 61.6 & 24.5 & 92.5 & 87.5 & 42.5 & 40.0 & 37.5 & 50.0 & 43.0 & 33.5 \\
LLaVA-Next-34B & 60.8 & 19.0 & 92.5 & 80.5 & 27.5 & 33.5 & 71.0 & 55.6 & 57.0 & 56.5 \\
XComposer2-7B & 55.7 & 23.0 & 90.0 & 69.0 & 21.0 & 28.5 & 48.0 & 56.7 & 52.5 & 38.0 \\
BLIP2-Flan-T5-XXL & 54.8 & 20.0 & 88.5 & 84.5 & 32.0 & 30.5 & 55.0 & 46.7 & 54.5 & 51.0 \\
Yi-VL-34B & 54.2 & 14.5 & 85.5 & 50.5 & 30.0 & 21.5 & 45.5 & 34.4 & 44.0 & 46.5 \\
Monkey & 53.4 & 24.0 & 91.5 & 79.0 & 19.5 & 32.5 & 36.0 & 38.9 & 40.5 & 42.5 \\
DeepSeek-VL-7B & 53.2 & 20.5 & 92.0 & 66.5 & 16.0 & 29.0 & 32.5 & 36.7 & 43.5 & 51.5 \\
Yi-VL-6B & 53.2 & 17.0 & 89.5 & 56.5 & 30.5 & 25.5 & 48.5 & 38.9 & 48.5 & 49.0 \\
LLaVA-Next-13B & 53.0 & 20.5 & 88.5 & 78.5 & 17.0 & 28.0 & 27.0 & 25.6 & 23.5 & 40.0 \\
TransCore-M & 52.7 & 23.0 & 87.0 & 78.5 & 22.0 & 24.0 & 36.5 & 35.6 & 44.0 & 36.5 \\
QWen-VL-Chat & 52.5 & 28.5 & 89.0 & 77.5 & 28.0 & 31.5 & 35.0 & 38.9 & 30.0 & 25.5 \\
Claude3V-Haiku & 52.2 & 14.5 & 81.5 & 71.0 & 36.0 & 28.0 & 41.0 & 46.7 & 44.0 & 40.5 \\
XComposer & 52.1 & 25.0 & 89.5 & 80.5 & 20.0 & 24.5 & 38.5 & 47.8 & 44.0 & 37.5 \\
mPLUG-Owl2 & 52.0 & 20.0 & 89.5 & 78.5 & 15.5 & 22.5 & 38.5 & 42.2 & 41.5 & 50.0 \\
RBDash-v1-13B & 51.8 & 23.5 & 90.0 & 65.5 & 32.5 & 24.0 & 40.5 & 32.2 & 47.0 & 33.0 \\
LLaVA-v1.5-13B & 51.7 & 28.0 & 86.5 & 68.0 & 25.5 & 25.0 & 31.0 & 30.0 & 28.5 & 41.0 \\
CogVLM-Chat & 51.6 & 30.0 & 87.0 & 73.5 & 8.5 & 32.5 & 22.0 & 26.7 & 18.0 & 23.0 \\
ShareGPT4V-7B & 51.5 & 21.5 & 87.5 & 75.0 & 19.0 & 28.0 & 33.5 & 27.8 & 37.5 & 35.5 \\
LLaVA-Next-7B & 51.1 & 29.0 & 86.5 & 76.5 & 20.0 & 27.5 & 23.5 & 21.1 & 11.0 & 30.5 \\
LLaVA-v1.5-13B-XTuner & 51.1 & 29.5 & 85.0 & 69.0 & 15.0 & 22.5 & 41.0 & 30.0 & 34.5 & 42.0 \\
LlaVA-InternLM2-7B & 50.8 & 3.5 & 86.5 & 75.0 & 26.0 & 27.5 & 39.0 & 28.9 & 39.5 & 33.0 \\
LLaVA-v1.5-7B-Xtuner & 50.2 & 28.0 & 88.0 & 69.0 & 18.0 & 26.0 & 36.0 & 36.7 & 44.0 & 50.5 \\
SharedCaptioner & 49.9 & 30.0 & 69.5 & 71.5 & 31.5 & 31.0 & 29.5 & 41.1 & 55.5 & 39.5 \\
LLaVA-InternLM-7b & 49.7 & 19.5 & 88.5 & 70.0 & 19.5 & 30.5 & 35.0 & 41.1 & 42.0 & 40.0 \\
LLaVA-v1.5-7B & 49.5 & 22.0 & 88.5 & 75.0 & 19.5 & 28.0 & 27.5 & 30.0 & 31.5 & 36.0 \\
LLaMA-Adapter-v2-7B & 40.4 & 31.5 & 79.5 & 54.0 & 30.0 & 24.5 & 33.0 & 31.1 & 24.0 & 34.0 \\
VisualGLM\_6b & 38.6 & 25.5 & 74.5 & 40.5 & 26.0 & 29.5 & 31.0 & 34.4 & 30.0 & 24.0 \\
Frequency & 31.7 & 31.5 & 29.0 & 30.0 & 32.0 & 27.5 & 27.5 & 28.9 & 27.5 & 27.0 \\
Random & 28.5 & 28.5 & 24.0 & 25.0 & 28.5 & 21.0 & 23.5 & 24.4 & 27.0 & 29.5 \\
\bottomrule
\end{tabular}%
}
\end{table}

% \usepackage{graphicx}
\begin{table}[]
\centering
\caption{Detail results of 30 LVLMs on \textbf{Localization} (part 2) and \textbf{Visual Recognition} (part 1).}
\label{tab:detailed_result_3}
\resizebox{\textwidth}{!}{%
\begin{tabular}{l|c|ccccc|cccc}
\toprule
 &  & \multicolumn{5}{c|}{Localization} & \multicolumn{4}{c}{Visual  Recognition} \\
 \midrule
Model & Overall & \begin{tabular}[c]{@{}c@{}}Salient Object \\ Detection RGBD\end{tabular} & \begin{tabular}[c]{@{}c@{}}Transparent \\ Object Detection\end{tabular} & \begin{tabular}[c]{@{}c@{}}Face\\ Detection\end{tabular} & \begin{tabular}[c]{@{}c@{}}Object \\ Detection\end{tabular} & \begin{tabular}[c]{@{}c@{}}Salient Object \\ Detection RGB\end{tabular} & \begin{tabular}[c]{@{}c@{}}Deepfake \\ Detection\end{tabular} & \begin{tabular}[c]{@{}c@{}}Weather \\ Recognition\end{tabular} & \begin{tabular}[c]{@{}c@{}}Season \\ Recognition\end{tabular} & \begin{tabular}[c]{@{}c@{}}Gesture \\ Recognition\end{tabular} \\
\midrule
InternVL-Chat-V1.2-34B & 63.4 & 28.5 & 66.5 & 64.0 & 82.5 & 61.0 & 33.5 & 87.6 & 83.0 & 47.5 \\
Qwen-VL-Plus & 62.3 & 44.5 & 47.5 & 62.5 & 66.5 & 51.0 & 39.5 & 90.2 & 86.0 & 67.5 \\
GPT-4V & 62.0 & 42.0 & 56.5 & 68.0 & 68.5 & 35.5 & 33.5 & 94.8 & 88.5 & 68.5 \\
GeminiProVision & 61.6 & 45.0 & 38.5 & 52.0 & 48.0 & 45.0 & 41.0 & 93.8 & 89.0 & 72.0 \\
LLaVA-Next-34B & 60.8 & 36.5 & 66.0 & 60.5 & 76.0 & 70.0 & 42.0 & 82.0 & 78.0 & 50.5 \\
XComposer2-7B & 55.7 & 46.5 & 42.0 & 44.0 & 50.5 & 52.5 & 34.5 & 87.6 & 82.0 & 57.0 \\
BLIP2-Flan-T5-XXL & 54.8 & 68.0 & 56.0 & 61.5 & 57.5 & 36.5 & 29.5 & 84.5 & 78.5 & 40.5 \\
Yi-VL-34B & 54.2 & 42.5 & 55.5 & 36.0 & 57.5 & 61.5 & 31.5 & 74.2 & 78.5 & 41.5 \\
Monkey & 53.4 & 33.5 & 49.0 & 46.0 & 47.5 & 27.0 & 45.0 & 83.5 & 83.5 & 58.0 \\
DeepSeek-VL-7B & 53.2 & 40.0 & 53.5 & 34.0 & 45.5 & 40.5 & 35.0 & 83.5 & 76.5 & 67.5 \\
Yi-VL-6B & 53.2 & 47.5 & 61.5 & 46.0 & 56.0 & 49.0 & 37.0 & 76.3 & 74.5 & 43.0 \\
LLaVA-Next-13B & 53.0 & 44.0 & 41.0 & 38.0 & 38.5 & 42.5 & 59.5 & 79.4 & 74.5 & 48.0 \\
TransCore-M & 52.7 & 39.0 & 42.0 & 45.5 & 46.0 & 39.0 & 42.5 & 84.0 & 72.5 & 50.5 \\
QWen-VL-Chat & 52.5 & 30.5 & 33.5 & 41.0 & 42.0 & 27.0 & 48.0 & 82.0 & 78.0 & 54.0 \\
Claude3V-Haiku & 52.2 & 43.0 & 19.5 & 63.0 & 64.5 & 41.0 & 31.0 & 88.1 & 82.0 & 48.5 \\
XComposer & 52.1 & 37.5 & 36.0 & 37.0 & 49.0 & 36.0 & 38.5 & 82.5 & 83.0 & 48.0 \\
mPLUG-Owl2 & 52.0 & 50.5 & 45.5 & 43.5 & 50.5 & 50.0 & 67.5 & 83.0 & 79.5 & 52.0 \\
RBDash-v1-13B & 51.8 & 42.0 & 46.0 & 41.0 & 56.0 & 42.0 & 36.0 & 79.9 & 72.5 & 49.0 \\
LLaVA-v1.5-13B & 51.7 & 38.0 & 48.0 & 43.5 & 46.5 & 42.5 & 49.5 & 82.0 & 73.5 & 48.5 \\
CogVLM-Chat & 51.6 & 35.0 & 28.0 & 16.0 & 28.0 & 25.5 & 44.5 & 86.1 & 85.5 & 49.0 \\
ShareGPT4V-7B & 51.5 & 40.5 & 39.0 & 32.0 & 44.5 & 33.5 & 72.0 & 82.5 & 74.0 & 51.0 \\
LLaVA-Next-7B & 51.1 & 49.0 & 34.5 & 15.5 & 34.5 & 46.0 & 66.0 & 76.8 & 75.5 & 49.5 \\
LLaVA-v1.5-13B-XTuner & 51.1 & 40.5 & 46.5 & 43.0 & 45.5 & 43.0 & 55.0 & 83.5 & 72.5 & 52.0 \\
LlaVA-InternLM2-7B & 50.8 & 32.0 & 51.0 & 37.0 & 44.0 & 45.5 & 33.0 & 80.9 & 73.5 & 43.0 \\
LLaVA-v1.5-7B-Xtuner & 50.2 & 36.5 & 50.5 & 33.5 & 43.5 & 39.0 & 46.5 & 83.5 & 78.0 & 46.5 \\
SharedCaptioner & 49.9 & 39.5 & 51.5 & 46.0 & 43.5 & 30.5 & 36.0 & 78.4 & 78.0 & 57.5 \\
LLaVA-InternLM-7b & 49.7 & 33.0 & 43.0 & 32.0 & 46.0 & 36.0 & 45.5 & 71.1 & 77.0 & 50.5 \\
LLaVA-v1.5-7B & 49.5 & 37.5 & 40.0 & 30.0 & 41.5 & 35.0 & 77.0 & 80.9 & 69.0 & 47.0 \\
LLaMA-Adapter-v2-7B & 40.4 & 38.5 & 28.5 & 28.0 & 42.0 & 33.0 & 52.0 & 73.2 & 62.0 & 39.5 \\
VisualGLM\_6b & 38.6 & 49.5 & 31.5 & 38.0 & 29.5 & 30.0 & 26.5 & 59.8 & 58.0 & 45.0 \\
Frequency & 31.7 & 26.0 & 26.0 & 28.0 & 36.5 & 26.0 & 52.0 & 29.4 & 28.0 & 30.0 \\
Random & 28.5 & 28.5 & 29.0 & 33.0 & 37.0 & 21.0 & 54.5 & 22.2 & 25.5 & 28.0 \\
\bottomrule
\end{tabular}%
}
\end{table}
% \usepackage{graphicx}
\begin{table}[]
\centering
\caption{Detail results of 30 LVLMs on \textbf{Visual Recognition} (part 2).}
\label{tab:detailed_result_4}
\resizebox{\textwidth}{!}{%
\begin{tabular}{l|c|ccccccccc}
\toprule
 &  & \multicolumn{9}{c}{Visual  Recognition} \\
 \midrule
Model & Overall & \begin{tabular}[c]{@{}c@{}}Muscial Instrument\\ Recognition\end{tabular} & \begin{tabular}[c]{@{}c@{}}Food \\ Recognition\end{tabular} & \begin{tabular}[c]{@{}c@{}}Landmark \\ Recognition\end{tabular} & \begin{tabular}[c]{@{}c@{}}Scene \\ Recognition\end{tabular} & \begin{tabular}[c]{@{}c@{}}Animal\\ Recognition\end{tabular} & \begin{tabular}[c]{@{}c@{}}Chemical Apparatusn\\ Recognition\end{tabular} & \begin{tabular}[c]{@{}c@{}}Rock\\ Recognition\end{tabular} & \begin{tabular}[c]{@{}c@{}}Fashion\\ Recognition\end{tabular} & \begin{tabular}[c]{@{}c@{}}Logo\\ Recognition\end{tabular} \\
\midrule
InternVL-Chat-V1.2-34B & 63.4 & 98.0 & 92.5 & 100.0 & 78.0 & 95.0 & 70.0 & 62.0 & 79.0 & 92.5 \\
Qwen-VL-Plus & 62.3 & 98.5 & 95.0 & 98.0 & 80.0 & 97.0 & 74.5 & 64.5 & 84.5 & 96.5 \\
GPT-4V & 62.0 & 97.5 & 95.0 & 98.0 & 83.0 & 95.9 & 62.5 & 66.0 & 86.0 & 92.0 \\
GeminiProVision & 61.6 & 97.5 & 96.5 & 100.0 & 82.5 & 96.0 & 72.0 & 68.0 & 71.5 & 96.5 \\
LLaVA-Next-34B & 60.8 & 91.5 & 91.0 & 96.0 & 77.0 & 90.0 & 59.0 & 55.0 & 73.5 & 90.0 \\
XComposer2-7B & 55.7 & 85.0 & 90.0 & 96.0 & 78.0 & 88.0 & 55.5 & 56.0 & 73.5 & 87.5 \\
BLIP2-Flan-T5-XXL & 54.8 & 93.5 & 92.5 & 98.0 & 77.5 & 92.0 & 54.0 & 48.5 & 80.0 & 76.5 \\
Yi-VL-34B & 54.2 & 93.5 & 86.5 & 100.0 & 77.0 & 92.5 & 50.0 & 53.5 & 74.5 & 87.0 \\
Monkey & 53.4 & 95.5 & 93.5 & 100.0 & 71.0 & 93.0 & 55.5 & 54.5 & 84.5 & 94.5 \\
DeepSeek-VL-7B & 53.2 & 94.5 & 93.0 & 98.0 & 77.5 & 94.0 & 57.5 & 50.0 & 81.5 & 96.0 \\
Yi-VL-6B & 53.2 & 92.0 & 86.0 & 94.0 & 77.0 & 89.5 & 57.5 & 42.0 & 78.5 & 89.5 \\
LLaVA-Next-13B & 53.0 & 90.0 & 86.0 & 94.0 & 77.0 & 88.0 & 59.0 & 58.0 & 69.0 & 88.5 \\
TransCore-M & 52.7 & 90.5 & 87.0 & 100.0 & 76.0 & 84.0 & 59.5 & 54.5 & 72.0 & 91.0 \\
QWen-VL-Chat & 52.5 & 95.0 & 90.5 & 98.0 & 74.5 & 94.0 & 48.5 & 51.0 & 79.5 & 95.5 \\
Claude3V-Haiku & 52.2 & 91.5 & 81.5 & 92.0 & 75.0 & 89.5 & 53.5 & 53.5 & 71.0 & 88.5 \\
XComposer & 52.1 & 93.5 & 90.5 & 98.0 & 77.0 & 89.5 & 57.0 & 45.0 & 81.0 & 88.5 \\
mPLUG-Owl2 & 52.0 & 96.0 & 87.5 & 100.0 & 80.0 & 92.0 & 56.0 & 54.0 & 73.0 & 86.5 \\
RBDash-v1-13B & 51.8 & 87.5 & 87.5 & 96.0 & 77.5 & 87.5 & 57.0 & 50.0 & 65.5 & 85.5 \\
LLaVA-v1.5-13B & 51.7 & 93.0 & 90.5 & 100.0 & 77.5 & 89.0 & 48.5 & 59.0 & 68.5 & 88.0 \\
CogVLM-Chat & 51.6 & 95.5 & 90.0 & 98.0 & 80.5 & 93.5 & 67.5 & 58.5 & 79.0 & 92.5 \\
ShareGPT4V-7B & 51.5 & 92.0 & 84.0 & 100.0 & 79.5 & 89.0 & 52.5 & 50.5 & 70.0 & 86.0 \\
LLaVA-Next-7B & 51.1 & 93.0 & 82.5 & 94.0 & 80.0 & 89.5 & 53.5 & 49.0 & 70.0 & 81.5 \\
LLaVA-v1.5-13B-XTuner & 51.1 & 90.0 & 88.0 & 92.0 & 75.5 & 84.5 & 56.5 & 51.5 & 67.0 & 86.5 \\
LlaVA-InternLM2-7B & 50.8 & 90.0 & 90.0 & 96.0 & 78.0 & 89.5 & 63.5 & 52.0 & 72.0 & 78.0 \\
LLaVA-v1.5-7B-Xtuner & 50.2 & 90.0 & 89.0 & 98.0 & 76.0 & 85.5 & 51.0 & 44.5 & 68.5 & 80.0 \\
SharedCaptioner & 49.9 & 88.5 & 89.0 & 96.0 & 63.5 & 88.5 & 58.0 & 49.5 & 76.5 & 87.0 \\
LLaVA-InternLM-7b & 49.7 & 83.5 & 84.0 & 90.0 & 74.0 & 82.5 & 50.5 & 42.5 & 69.0 & 88.0 \\
LLaVA-v1.5-7B & 49.5 & 91.5 & 86.0 & 96.0 & 76.0 & 83.0 & 51.0 & 51.5 & 67.5 & 81.5 \\
LLaMA-Adapter-v2-7B & 40.4 & 87.0 & 74.0 & 88.0 & 63.5 & 79.0 & 38.5 & 37.5 & 47.0 & 75.0 \\
VisualGLM\_6b & 38.6 & 82.0 & 69.5 & 82.0 & 58.5 & 80.0 & 36.5 & 31.5 & 45.0 & 71.5 \\
Frequency & 31.7 & 27.0 & 28.5 & 30.0 & 26.5 & 27.0 & 30.0 & 28.0 & 27.0 & 29.0 \\
Random & 28.5 & 26.5 & 24.5 & 26.0 & 24.0 & 25.5 & 29.5 & 25.0 & 23.0 & 28.5 \\
\bottomrule
\end{tabular}%
}
\end{table}
% \usepackage{graphicx}
\begin{table}[]
\centering
\caption{Detail results of 30 LVLMs on \textbf{Visual Recognition} (part 3).}
\label{tab:detailed_result_5}
\resizebox{\textwidth}{!}{%
\begin{tabular}{l|c|ccccccccc}
\toprule
 &  & \multicolumn{9}{c}{Visual  Recognition} \\
 \midrule
Model & Overall & \begin{tabular}[c]{@{}c@{}}Astronomical\\ Recognition\end{tabular} & \begin{tabular}[c]{@{}c@{}}Painting\\ Recognition\end{tabular} & \begin{tabular}[c]{@{}c@{}}Color\\ Recognition\end{tabular} & \begin{tabular}[c]{@{}c@{}}Plant\\ Recognition\end{tabular} & \begin{tabular}[c]{@{}c@{}}Shape\\ Recognition\end{tabular} & \begin{tabular}[c]{@{}c@{}}Profession\\ Recognition\end{tabular} & \begin{tabular}[c]{@{}c@{}}Building\\ Recognition\end{tabular} & \begin{tabular}[c]{@{}c@{}}Electronic Object\\ Recognition\end{tabular} & \begin{tabular}[c]{@{}c@{}}Sports\\ Recognition\end{tabular} \\
\midrule
InternVL-Chat-V1.2-34B & 63.4 & 69.1 & 78.5 & 52.0 & 95.5 & 82.5 & 97.0 & 77.0 & 98.5 & 94.0 \\
Qwen-VL-Plus & 62.3 & 75.5 & 81.5 & 61.0 & 96.5 & 81.5 & 94.5 & 87.5 & 96.5 & 96.5 \\
GPT-4V & 62.0 & 69.9 & 80.2 & 67.5 & 97.9 & 81.0 & 97.5 & 85.0 & 93.5 & 96.5 \\
GeminiProVision & 61.6 & 72.3 & 84.0 & 66.5 & 95.0 & 84.5 & 97.0 & 82.5 & 95.5 & 98.0 \\
LLaVA-Next-34B & 60.8 & 56.4 & 75.5 & 49.5 & 85.0 & 70.5 & 97.0 & 71.5 & 95.0 & 91.5 \\
XComposer2-7B & 55.7 & 45.7 & 78.5 & 53.0 & 82.0 & 70.0 & 97.0 & 75.0 & 91.0 & 91.0 \\
BLIP2-Flan-T5-XXL & 54.8 & 61.7 & 71.0 & 42.0 & 89.0 & 77.0 & 97.5 & 69.5 & 95.5 & 92.5 \\
Yi-VL-34B & 54.2 & 54.3 & 68.0 & 46.0 & 87.5 & 74.5 & 96.0 & 70.5 & 91.0 & 88.5 \\
Monkey & 53.4 & 66.0 & 76.0 & 39.5 & 95.0 & 79.0 & 98.0 & 78.0 & 95.0 & 90.0 \\
DeepSeek-VL-7B & 53.2 & 58.5 & 71.0 & 40.0 & 92.0 & 77.0 & 98.0 & 69.5 & 95.0 & 89.5 \\
Yi-VL-6B & 53.2 & 52.1 & 75.0 & 38.0 & 84.0 & 74.0 & 94.0 & 75.5 & 92.5 & 89.0 \\
LLaVA-Next-13B & 53.0 & 50.0 & 70.5 & 44.0 & 80.5 & 69.5 & 94.0 & 73.5 & 92.0 & 89.5 \\
TransCore-M & 52.7 & 56.4 & 74.0 & 39.0 & 76.5 & 73.0 & 97.5 & 76.0 & 93.0 & 85.5 \\
QWen-VL-Chat & 52.5 & 69.1 & 77.0 & 39.5 & 92.5 & 76.5 & 96.5 & 76.0 & 93.5 & 92.0 \\
Claude3V-Haiku & 52.2 & 64.9 & 67.5 & 71.0 & 91.5 & 74.0 & 91.0 & 72.5 & 75.5 & 84.0 \\
XComposer & 52.1 & 61.7 & 74.0 & 45.0 & 89.5 & 69.5 & 96.0 & 67.0 & 94.5 & 88.0 \\
mPLUG-Owl2 & 52.0 & 62.8 & 71.5 & 42.0 & 92.0 & 74.5 & 98.0 & 71.0 & 92.0 & 91.0 \\
RBDash-v1-13B & 51.8 & 55.3 & 69.5 & 39.5 & 79.0 & 71.5 & 97.0 & 65.0 & 92.0 & 87.0 \\
LLaVA-v1.5-13B & 51.7 & 50.0 & 69.5 & 38.5 & 83.5 & 67.5 & 95.0 & 71.0 & 93.5 & 89.0 \\
CogVLM-Chat & 51.6 & 73.4 & 78.5 & 35.0 & 92.5 & 80.0 & 97.0 & 77.5 & 96.0 & 92.5 \\
ShareGPT4V-7B & 51.5 & 56.4 & 72.0 & 35.5 & 84.5 & 67.5 & 98.5 & 77.0 & 93.0 & 88.5 \\
LLaVA-Next-7B & 51.1 & 50.0 & 76.0 & 45.0 & 82.0 & 68.0 & 95.5 & 70.0 & 93.5 & 88.5 \\
LLaVA-v1.5-13B-XTuner & 51.1 & 48.9 & 68.0 & 43.0 & 79.5 & 67.5 & 96.0 & 73.0 & 91.5 & 89.0 \\
LlaVA-InternLM2-7B & 50.8 & 46.8 & 64.5 & 48.5 & 85.5 & 75.0 & 97.5 & 67.0 & 94.0 & 89.0 \\
LLaVA-v1.5-7B-Xtuner & 50.2 & 52.1 & 67.5 & 41.5 & 79.0 & 66.5 & 96.0 & 69.0 & 92.5 & 90.0 \\
SharedCaptioner & 49.9 & 55.3 & 64.0 & 39.5 & 88.0 & 70.0 & 94.5 & 67.0 & 93.0 & 87.0 \\
LLaVA-InternLM-7b & 49.7 & 53.2 & 68.5 & 36.5 & 81.5 & 68.0 & 96.5 & 57.5 & 91.5 & 85.5 \\
LLaVA-v1.5-7B & 49.5 & 56.4 & 66.0 & 45.0 & 83.0 & 62.5 & 97.0 & 74.5 & 93.0 & 87.5 \\
LLaMA-Adapter-v2-7B & 40.4 & 53.2 & 52.0 & 36.5 & 67.5 & 60.5 & 94.0 & 56.0 & 79.5 & 82.5 \\
VisualGLM\_6b & 38.6 & 42.6 & 45.5 & 27.5 & 67.5 & 52.0 & 88.5 & 32.0 & 82.5 & 81.5 \\
Frequency & 31.7 & 29.8 & 31.5 & 27.0 & 29.5 & 26.5 & 27.5 & 28.5 & 29.0 & 35.0 \\
Random & 28.5 & 30.9 & 21.0 & 20.0 & 23.5 & 21.5 & 27.5 & 24.5 & 25.5 & 19.5 \\
\bottomrule
\end{tabular}%
}
\end{table}
% \usepackage{graphicx}
\begin{table}[]
\centering
\caption{Detail results of 30 LVLMs on \textbf{Visual Recognition} (part 4).}
\label{tab:detailed_result_6}
\resizebox{\textwidth}{!}{%
\begin{tabular}{l|c|ccccccccc}
\toprule
 &  & \multicolumn{9}{c}{Visual  Recognition} \\
 \midrule
Model & Overall & \begin{tabular}[c]{@{}c@{}}Disaster\\ Recognition\end{tabular} & \begin{tabular}[c]{@{}c@{}}Celebrity\\ Recognition\end{tabular} & \begin{tabular}[c]{@{}c@{}}Vehicle\\ Recognition\end{tabular} & \begin{tabular}[c]{@{}c@{}}National Flag\\ Recognition\end{tabular} & \begin{tabular}[c]{@{}c@{}}Abstract Visual\\ Recognition\end{tabular} & \begin{tabular}[c]{@{}c@{}}Animated Character\\ Recognition\end{tabular} & \begin{tabular}[c]{@{}c@{}}Texture Material\\ Recognition\end{tabular} & \begin{tabular}[c]{@{}c@{}}Film and Television \\ Recognition\end{tabular} & \begin{tabular}[c]{@{}c@{}}Sculpture \\ Recognition\end{tabular} \\
\midrule
InternVL-Chat-V1.2-34B & 63.4 & 80.0 & 85.5 & 98.5 & 83.0 & 76.5 & 49.0 & 76.5 & 95.0 & 94.0 \\
Qwen-VL-Plus & 62.3 & 78.5 & 89.0 & 97.0 & 93.5 & 82.0 & 56.0 & 76.5 & 93.5 & 72.0 \\
GPT-4V & 62.0 & 89.1 & 100.0 & 97.9 & 91.0 & 81.0 & 66.0 & 76.8 & 98.0 & 98.0 \\
GeminiProVision & 61.6 & 82.0 & 93.0 & 96.5 & 98.5 & 88.0 & 58.5 & 76.0 & 93.5 & 96.0 \\
LLaVA-Next-34B & 60.8 & 75.5 & 81.5 & 99.0 & 67.0 & 72.0 & 46.5 & 67.5 & 92.5 & 92.0 \\
XComposer2-7B & 55.7 & 78.5 & 74.0 & 96.0 & 63.0 & 66.0 & 42.0 & 69.0 & 93.0 & 88.0 \\
BLIP2-Flan-T5-XXL & 54.8 & 57.5 & 73.0 & 96.5 & 86.5 & 70.5 & 30.5 & 77.0 & 83.5 & 92.0 \\
Yi-VL-34B & 54.2 & 72.5 & 79.5 & 94.5 & 87.0 & 70.5 & 43.0 & 58.5 & 90.5 & 88.0 \\
Monkey & 53.4 & 68.0 & 77.0 & 97.5 & 95.5 & 72.0 & 41.5 & 71.0 & 94.0 & 94.0 \\
DeepSeek-VL-7B & 53.2 & 50.0 & 69.0 & 97.0 & 70.5 & 72.5 & 31.0 & 77.5 & 86.5 & 88.0 \\
Yi-VL-6B & 53.2 & 66.0 & 76.0 & 97.0 & 81.0 & 64.5 & 35.5 & 59.5 & 87.0 & 88.0 \\
LLaVA-Next-13B & 53.0 & 67.0 & 77.5 & 97.0 & 60.5 & 69.0 & 37.0 & 71.5 & 85.0 & 86.0 \\
TransCore-M & 52.7 & 65.5 & 78.0 & 98.0 & 61.5 & 73.0 & 31.5 & 63.5 & 81.5 & 82.0 \\
QWen-VL-Chat & 52.5 & 68.0 & 65.0 & 95.5 & 94.5 & 77.0 & 44.5 & 72.0 & 88.0 & 92.0 \\
Claude3V-Haiku & 52.2 & 84.0 & 55.5 & 91.5 & 74.0 & 68.0 & 45.5 & 62.5 & 84.5 & 86.0 \\
XComposer & 52.1 & 53.0 & 73.5 & 96.0 & 87.0 & 70.5 & 40.5 & 71.5 & 89.0 & 84.0 \\
mPLUG-Owl2 & 52.0 & 62.5 & 79.0 & 96.5 & 83.5 & 66.5 & 38.5 & 67.0 & 88.5 & 86.0 \\
RBDash-v1-13B & 51.8 & 71.5 & 80.0 & 98.0 & 56.0 & 69.5 & 35.5 & 66.5 & 85.0 & 88.0 \\
LLaVA-v1.5-13B & 51.7 & 62.5 & 76.5 & 98.5 & 58.0 & 69.0 & 36.0 & 75.5 & 85.5 & 90.0 \\
CogVLM-Chat & 51.6 & 65.5 & 42.0 & 98.0 & 87.0 & 75.5 & 38.5 & 65.5 & 91.0 & 94.0 \\
ShareGPT4V-7B & 51.5 & 64.5 & 76.0 & 96.0 & 64.5 & 71.0 & 40.0 & 62.5 & 84.5 & 80.0 \\
LLaVA-Next-7B & 51.1 & 66.0 & 77.5 & 96.5 & 57.0 & 67.5 & 37.0 & 65.0 & 86.0 & 74.0 \\
LLaVA-v1.5-13B-XTuner & 51.1 & 52.0 & 75.5 & 93.5 & 57.5 & 68.0 & 35.5 & 68.5 & 83.5 & 82.0 \\
LlaVA-InternLM2-7B & 50.8 & 72.5 & 75.0 & 97.5 & 55.0 & 70.0 & 37.5 & 70.5 & 82.0 & 86.0 \\
LLaVA-v1.5-7B-Xtuner & 50.2 & 66.5 & 70.5 & 96.0 & 56.0 & 66.0 & 36.5 & 75.0 & 86.5 & 84.0 \\
SharedCaptioner & 49.9 & 70.0 & 70.0 & 95.5 & 76.0 & 68.5 & 32.0 & 71.0 & 84.0 & 74.0 \\
LLaVA-InternLM-7b & 49.7 & 62.5 & 69.0 & 97.0 & 54.5 & 65.5 & 32.0 & 72.5 & 80.0 & 80.0 \\
LLaVA-v1.5-7B & 49.5 & 62.0 & 72.5 & 96.5 & 61.0 & 68.0 & 34.0 & 62.5 & 80.0 & 90.0 \\
LLaMA-Adapter-v2-7B & 40.4 & 35.0 & 49.0 & 90.5 & 67.0 & 61.5 & 22.0 & 65.0 & 64.5 & 54.0 \\
VisualGLM\_6b & 38.6 & 43.5 & 31.0 & 92.0 & 65.0 & 49.0 & 22.0 & 49.0 & 51.0 & 68.0 \\
Frequency & 31.7 & 29.5 & 29.5 & 26.5 & 29.5 & 26.0 & 26.5 & 29.5 & 25.5 & 32.0 \\
Random & 28.5 & 23.5 & 23.0 & 24.0 & 22.0 & 25.5 & 27.0 & 31.5 & 27.0 & 30.0 \\
\bottomrule
\end{tabular}%
}
\end{table}
% \usepackage{graphicx}
\begin{table}[]
\centering
\caption{Detail results of 30 LVLMs on \textbf{Visual Recognition} (part 5) and \textbf{GUI}.}
\label{tab:detailed_result_7}
\resizebox{\textwidth}{!}{%
\begin{tabular}{l|c|cccc|cccc}
\toprule
 &  & \multicolumn{4}{c|}{Visual  Recognition} & \multicolumn{4}{c}{GUI} \\
\midrule
Model & Overall & \begin{tabular}[c]{@{}c@{}}Age Gender \\ Race recognition\end{tabular} & \begin{tabular}[c]{@{}c@{}}Weapon \\ Recognition\end{tabular} & \begin{tabular}[c]{@{}c@{}}Religious \\ Recognition\end{tabular} & \begin{tabular}[c]{@{}c@{}}Waste \\ Recognition\end{tabular} & \begin{tabular}[c]{@{}c@{}}GUI \\ General\end{tabular} & \begin{tabular}[c]{@{}c@{}}GUI \\ Google APP\end{tabular} & \begin{tabular}[c]{@{}c@{}}GUI\\ Web Shopping\end{tabular} & \begin{tabular}[c]{@{}c@{}}GUI\\ Installation\end{tabular} \\
\midrule
InternVL-Chat-V1.2-34B & 63.4 & 79.5 & 91.5 & 82.0 & 93.5 & 46.5 & 40.5 & 42.5 & 35.5 \\
Qwen-VL-Plus & 62.3 & 81.5 & 47.0 & 86.5 & 96.5 & 37.5 & 32.0 & 28.0 & 33.0 \\
GPT-4V & 62.0 & 87.0 & 88.1 & 84.0 & 97.3 & 50.5 & 35.5 & 32.2 & 40.7 \\
GeminiProVision & 61.6 & 81.5 & 71.5 & 83.5 & 94.0 & 44.5 & 28.0 & 29.5 & 31.5 \\
LLaVA-Next-34B & 60.8 & 83.5 & 74.5 & 77.0 & 90.0 & 49.0 & 39.0 & 44.5 & 33.0 \\
XComposer2-7B & 55.7 & 76.5 & 78.0 & 66.0 & 90.0 & 38.0 & 26.0 & 20.0 & 33.5 \\
BLIP2-Flan-T5-XXL & 54.8 & 76.0 & 85.0 & 68.0 & 92.0 & 35.0 & 33.0 & 34.0 & 31.5 \\
Yi-VL-34B & 54.2 & 74.0 & 82.0 & 67.5 & 85.5 & 36.5 & 30.5 & 31.5 & 31.0 \\
Monkey & 53.4 & 78.5 & 77.5 & 79.5 & 92.0 & 32.0 & 28.0 & 29.0 & 30.0 \\
DeepSeek-VL-7B & 53.2 & 75.5 & 83.5 & 63.0 & 88.0 & 35.5 & 36.0 & 35.5 & 40.0 \\
Yi-VL-6B & 53.2 & 75.0 & 84.0 & 64.5 & 85.0 & 37.5 & 35.0 & 34.5 & 32.0 \\
LLaVA-Next-13B & 53.0 & 77.0 & 74.5 & 65.5 & 88.5 & 39.0 & 35.0 & 45.0 & 45.0 \\
TransCore-M & 52.7 & 75.0 & 82.0 & 61.5 & 86.5 & 36.5 & 30.5 & 36.5 & 39.0 \\
QWen-VL-Chat & 52.5 & 70.0 & 82.0 & 78.0 & 85.5 & 33.0 & 21.5 & 32.5 & 34.0 \\
Claude3V-Haiku & 52.2 & 77.5 & 71.0 & 73.5 & 89.5 & 39.5 & 35.5 & 30.0 & 35.5 \\
XComposer & 52.1 & 79.0 & 87.0 & 67.0 & 85.5 & 32.5 & 36.5 & 35.0 & 33.0 \\
mPLUG-Owl2 & 52.0 & 79.5 & 74.5 & 64.5 & 89.5 & 26.5 & 25.0 & 31.0 & 28.5 \\
RBDash-v1-13B & 51.8 & 75.0 & 75.5 & 55.5 & 83.0 & 34.5 & 29.0 & 31.5 & 34.5 \\
LLaVA-v1.5-13B & 51.7 & 76.0 & 79.0 & 66.0 & 86.5 & 42.5 & 35.5 & 37.0 & 35.0 \\
CogVLM-Chat & 51.6 & 78.5 & 84.0 & 68.0 & 91.0 & 41.0 & 24.0 & 26.5 & 28.0 \\
ShareGPT4V-7B & 51.5 & 74.5 & 80.5 & 63.5 & 84.5 & 33.0 & 25.5 & 30.0 & 27.0 \\
LLaVA-Next-7B & 51.1 & 77.5 & 74.5 & 67.0 & 90.5 & 37.5 & 24.0 & 30.0 & 35.0 \\
LLaVA-v1.5-13B-XTuner & 51.1 & 77.5 & 78.5 & 63.0 & 90.0 & 35.0 & 41.0 & 42.0 & 45.5 \\
LlaVA-InternLM2-7B & 50.8 & 79.0 & 76.5 & 68.0 & 88.5 & 39.5 & 29.5 & 36.5 & 35.5 \\
LLaVA-v1.5-7B-Xtuner & 50.2 & 84.5 & 74.5 & 63.5 & 87.5 & 31.0 & 26.0 & 32.5 & 32.5 \\
SharedCaptioner & 49.9 & 74.0 & 81.5 & 64.5 & 83.0 & 28.0 & 30.0 & 26.0 & 34.5 \\
LLaVA-InternLM-7b & 49.7 & 70.5 & 76.0 & 65.0 & 83.0 & 26.5 & 36.0 & 40.0 & 48.0 \\
LLaVA-v1.5-7B & 49.5 & 76.0 & 78.0 & 63.5 & 81.5 & 25.5 & 15.5 & 20.5 & 20.0 \\
LLaMA-Adapter-v2-7B & 40.4 & 75.0 & 69.5 & 56.5 & 71.5 & 43.0 & 22.0 & 24.0 & 30.5 \\
VisualGLM\_6b & 38.6 & 38.5 & 58.5 & 49.5 & 43.0 & 21.0 & 18.5 & 19.0 & 26.0 \\
Frequency & 31.7 & 36.5 & 32.0 & 28.0 & 43.0 & 31.0 & 27.0 & 28.5 & 29.5 \\
Random & 28.5 & 38.5 & 33.0 & 30.0 & 36.0 & 22.5 & 30.0 & 21.5 & 23.5 \\
\bottomrule
\end{tabular}%
}
\end{table}
% \usepackage{graphicx}
\begin{table}[]
\centering
\caption{Detail results of 30 LVLMs on \textbf{OCR} and \textbf{Image-to-Image Translation}.}
\label{tab:detailed_result_8}
\resizebox{0.85\textwidth}{!}{%
\begin{tabular}{l|c|cccc|cc}
\toprule
 &  & \multicolumn{4}{c|}{OCR} & \multicolumn{2}{c}{Image-to-Image Translation} \\
 \midrule
Model & Overall & \begin{tabular}[c]{@{}c@{}}Font\\ Recognition\end{tabular} & \begin{tabular}[c]{@{}c@{}}Handwritten\\ Text Recognition\end{tabular} & \begin{tabular}[c]{@{}c@{}}Handwritten Mathematical\\ Expression Recognition\end{tabular} & \begin{tabular}[c]{@{}c@{}}Scene Text\\ Recognition\end{tabular} & \begin{tabular}[c]{@{}c@{}}Jigsaw Puzzle\\ Solving\end{tabular} & \begin{tabular}[c]{@{}c@{}}Image \\ Colorization\end{tabular} \\
\midrule
InternVL-Chat-V1.2-34B & 63.4 & 23.5 & 64.0 & 60.0 & 94.5 & 44.5 & 21.0 \\
Qwen-VL-Plus & 62.3 & 31.0 & 79.0 & 59.0 & 93.5 & 42.5 & 33.0 \\
GPT-4V & 62.0 & 28.0 & 77.0 & 70.0 & 97.0 & 35.0 & 28.5 \\
GeminiProVision & 61.6 & 24.0 & 63.0 & 57.0 & 94.0 & 21.0 & 42.0 \\
LLaVA-Next-34B & 60.8 & 29.0 & 74.0 & 58.0 & 95.5 & 29.0 & 16.5 \\
XComposer2-7B & 55.7 & 21.5 & 32.0 & 29.0 & 93.0 & 21.0 & 42.0 \\
BLIP2-Flan-T5-XXL & 54.8 & 14.5 & 59.0 & 40.0 & 79.0 & 25.0 & 31.5 \\
Yi-VL-34B & 54.2 & 27.5 & 59.0 & 59.0 & 86.5 & 19.0 & 20.0 \\
Monkey & 53.4 & 21.0 & 46.0 & 44.0 & 93.0 & 26.0 & 33.0 \\
DeepSeek-VL-7B & 53.2 & 29.0 & 61.0 & 61.0 & 93.5 & 26.5 & 20.5 \\
Yi-VL-6B & 53.2 & 24.0 & 51.0 & 50.0 & 87.5 & 29.0 & 24.5 \\
LLaVA-Next-13B & 53.0 & 22.0 & 47.0 & 43.0 & 95.0 & 30.5 & 22.0 \\
TransCore-M & 52.7 & 23.5 & 55.0 & 33.0 & 90.0 & 25.5 & 24.5 \\
QWen-VL-Chat & 52.5 & 16.5 & 44.0 & 37.0 & 90.0 & 28.0 & 33.0 \\
Claude3V-Haiku & 52.2 & 17.5 & 66.0 & 48.0 & 86.0 & 23.5 & 22.0 \\
XComposer & 52.1 & 15.5 & 34.0 & 44.0 & 83.0 & 21.5 & 27.5 \\
mPLUG-Owl2 & 52.0 & 21.0 & 32.0 & 35.0 & 90.0 & 26.0 & 25.0 \\
RBDash-v1-13B & 51.8 & 29.5 & 48.0 & 45.0 & 92.0 & 20.0 & 25.0 \\
LLaVA-v1.5-13B & 51.7 & 28.0 & 39.0 & 51.0 & 89.0 & 27.5 & 28.5 \\
CogVLM-Chat & 51.6 & 16.0 & 42.0 & 50.0 & 86.0 & 26.5 & 21.0 \\
ShareGPT4V-7B & 51.5 & 23.0 & 34.0 & 45.0 & 89.0 & 19.0 & 24.5 \\
LLaVA-Next-7B & 51.1 & 23.0 & 46.0 & 44.0 & 95.0 & 22.0 & 23.5 \\
LLaVA-v1.5-13B-XTuner & 51.1 & 18.0 & 51.0 & 28.0 & 90.0 & 28.0 & 24.0 \\
LlaVA-InternLM2-7B & 50.8 & 28.0 & 42.0 & 40.0 & 88.0 & 0.0 & 28.5 \\
LLaVA-v1.5-7B-Xtuner & 50.2 & 20.0 & 39.0 & 35.0 & 90.0 & 26.5 & 24.5 \\
SharedCaptioner & 49.9 & 23.0 & 45.0 & 39.0 & 84.0 & 22.5 & 27.5 \\
LLaVA-InternLM-7b & 49.7 & 14.5 & 49.0 & 35.0 & 92.0 & 25.5 & 24.5 \\
LLaVA-v1.5-7B & 49.5 & 23.0 & 32.0 & 36.0 & 89.0 & 19.5 & 25.0 \\
LLaMA-Adapter-v2-7B & 40.4 & 22.0 & 28.0 & 37.0 & 53.0 & 18.5 & 18.0 \\
VisualGLM\_6b & 38.6 & 6.0 & 29.0 & 31.0 & 69.0 & 29.5 & 24.0 \\
Frequency & 31.7 & 28.5 & 30.0 & 33.0 & 30.0 & 29.0 & 32.5 \\
Random & 28.5 & 26.0 & 30.0 & 27.0 & 26.0 & 21.5 & 21.0 \\
\bottomrule
\end{tabular}%
}
\end{table}
% \usepackage{graphicx}
\begin{table}[]
\centering
\caption{Detail results of 30 LVLMs on \textbf{Temporal Understanding} and \textbf{Relation Reasoning}.}
\label{tab:detailed_result_9}
\resizebox{\textwidth}{!}{%
\begin{tabular}{l|c|ccccc|cccc}
\toprule
 &  & \multicolumn{5}{c|}{Temporal Understanding} & \multicolumn{4}{c}{Relation Reasoning} \\
 \midrule
Model & Overall & \begin{tabular}[c]{@{}c@{}}Next Image\\ Prediction\end{tabular} & MeViS & \begin{tabular}[c]{@{}c@{}}Temporal\\ Anticipation\end{tabular} & \begin{tabular}[c]{@{}c@{}}Temporal\\ Ordering\end{tabular} & \begin{tabular}[c]{@{}c@{}}Temporal \\ Localization\end{tabular} & \begin{tabular}[c]{@{}c@{}}Social Relation\\ Recognition\end{tabular} & \begin{tabular}[c]{@{}c@{}}Human Object \\ Interaction Recognition\end{tabular} & \begin{tabular}[c]{@{}c@{}}Scene Graph\\ Recognition\end{tabular} & \begin{tabular}[c]{@{}c@{}}Human Interaction\\ Understanding\end{tabular} \\
\midrule
InternVL-Chat-V1.2-34B & 63.4 & 36.0 & 63.5 & 25.5 & 30.6 & 70.5 & 86.5 & 19.5 & 74.0 & 40.2 \\
Qwen-VL-Plus & 62.3 & 31.0 & 62.5 & 20.5 & 46.6 & 67.0 & 81.0 & 16.0 & 77.0 & 37.8 \\
GPT-4V & 62.0 & 30.1 & 67.6 & 25.4 & 41.5 & 58.3 & 73.2 & 29.5 & 70.0 & 37.3 \\
GeminiProVision & 61.6 & 30.0 & 36.0 & 68.5 & 26.0 & 42.0 & 48.0 & 77.0 & 29.5 & 79.5 \\
LLaVA-Next-34B & 60.8 & 26.0 & 67.0 & 22.5 & 44.0 & 67.5 & 85.5 & 18.5 & 75.5 & 40.2 \\
XComposer2-7B & 55.7 & 29.0 & 32.0 & 64.5 & 25.0 & 32.6 & 36.5 & 83.5 & 7.5 & 74.5 \\
BLIP2-Flan-T5-XXL & 54.8 & 26.0 & 66.0 & 64.0 & 26.5 & 31.1 & 52.0 & 74.0 & 37.0 & 49.0 \\
Yi-VL-34B & 54.2 & 24.5 & 56.0 & 28.5 & 24.4 & 62.5 & 73.5 & 28.5 & 64.5 & 40.2 \\
Monkey & 53.4 & 27.5 & 36.0 & 63.5 & 24.5 & 32.6 & 63.5 & 74.0 & 42.0 & 67.5 \\
DeepSeek-VL-7B & 53.2 & 22.5 & 51.0 & 24.0 & 35.2 & 33.0 & 68.0 & 29.5 & 70.0 & 27.6 \\
Yi-VL-6B & 53.2 & 32.0 & 52.5 & 27.0 & 22.3 & 54.0 & 72.0 & 31.0 & 61.5 & 30.7 \\
LLaVA-Next-13B & 53.0 & 30.5 & 61.5 & 25.0 & 30.6 & 47.0 & 71.5 & 31.5 & 69.5 & 27.6 \\
TransCore-M & 52.7 & 24.0 & 24.0 & 49.5 & 22.0 & 29.0 & 44.5 & 68.0 & 31.0 & 69.5 \\
QWen-VL-Chat & 52.5 & 26.0 & 37.0 & 71.5 & 25.0 & 26.9 & 61.5 & 62.0 & 36.0 & 70.0 \\
Claude3V-Haiku & 52.2 & 27.0 & 59.0 & 24.5 & 35.8 & 33.0 & 62.0 & 50.0 & 49.5 & 38.6 \\
XComposer & 52.1 & 25.5 & 38.5 & 47.0 & 24.0 & 30.6 & 53.0 & 72.0 & 11.5 & 64.5 \\
mPLUG-Owl2 & 52.0 & 26.5 & 34.5 & 47.0 & 24.0 & 27.5 & 34.5 & 71.5 & 37.5 & 64.5 \\
RBDash-v1-13B & 51.8 & 23.5 & 26.5 & 51.0 & 23.0 & 34.2 & 41.0 & 77.0 & 34.5 & 71.0 \\
LLaVA-v1.5-13B & 51.7 & 26.0 & 26.0 & 60.5 & 25.5 & 28.5 & 52.5 & 73.5 & 38.0 & 69.0 \\
CogVLM-Chat & 51.6 & 23.5 & 36.5 & 54.5 & 25.0 & 37.8 & 62.0 & 63.0 & 42.5 & 79.0 \\
ShareGPT4V-7B & 51.5 & 28.5 & 30.0 & 52.5 & 25.0 & 24.4 & 51.0 & 72.5 & 45.0 & 70.5 \\
LLaVA-Next-7B & 51.1 & 33.5 & 58.0 & 22.5 & 24.4 & 57.0 & 67.5 & 35.0 & 65.0 & 28.3 \\
LLaVA-v1.5-13B-XTuner & 51.1 & 21.0 & 24.0 & 43.0 & 24.0 & 29.5 & 34.5 & 72.0 & 34.5 & 68.5 \\
LlaVA-InternLM2-7B & 50.8 & 19.5 & 26.5 & 52.5 & 5.5 & 30.1 & 43.5 & 80.0 & 35.0 & 73.5 \\
LLaVA-v1.5-7B-Xtuner & 50.2 & 25.0 & 28.5 & 44.0 & 22.5 & 29.5 & 33.5 & 73.5 & 36.0 & 72.5 \\
SharedCaptioner & 49.9 & 25.5 & 30.5 & 60.0 & 20.5 & 26.4 & 60.5 & 70.5 & 39.5 & 67.5 \\
LLaVA-InternLM-7b & 49.7 & 20.5 & 31.0 & 40.0 & 20.0 & 28.5 & 27.0 & 72.0 & 37.0 & 73.5 \\
LLaVA-v1.5-7B & 49.5 & 22.5 & 29.5 & 55.5 & 19.0 & 32.1 & 59.0 & 65.0 & 40.0 & 68.0 \\
LLaMA-Adapter-v2-7B & 40.4 & 23.5 & 31.5 & 57.5 & 20.5 & 22.3 & 53.5 & 33.5 & 45.5 & 47.0 \\
VisualGLM\_6b & 38.6 & 24.0 & 41.0 & 35.0 & 26.5 & 27.5 & 44.5 & 35.0 & 47.0 & 48.5 \\
Frequency & 31.7 & 28.5 & 30.0 & 28.0 & 29.0 & 30.1 & 27.0 & 26.0 & 52.0 & 29.0 \\
Random & 28.5 & 30.0 & 28.0 & 24.0 & 25.0 & 28.0 & 25.5 & 22.0 & 57.0 & 29.0 \\
\bottomrule
\end{tabular}%
}
\end{table}
% =====================================================================================================================

% \usepackage{graphicx}
\begin{table}[]
\centering
\caption{Detail results of 30 LVLMs on \textbf{Discipline Knowledge Reasoning}, \textbf{Intelligence Quotient Test} and \textbf{Embodied AI}.}
\label{tab:detailed_result_10}
\resizebox{\textwidth}{!}{%
\begin{tabular}{l|c|cccccc|cc}
\toprule
 &  & \multicolumn{6}{c|}{Discipline Knowledge Reasoning} & Intelligence Quotient Test & Embodied AI \\
 \midrule
Model & Overall & Science & \begin{tabular}[c]{@{}c@{}}Health\\ Medicine\end{tabular} & \begin{tabular}[c]{@{}c@{}}Art and\\ Design\end{tabular} & \begin{tabular}[c]{@{}c@{}}Humanitites Social\\ Science\end{tabular} & \begin{tabular}[c]{@{}c@{}}Tech\\ Engineering\end{tabular} & Business & \begin{tabular}[c]{@{}c@{}}Ravens Progressive\\ Matrices\end{tabular} & Navigation \\
\midrule
InternVL-Chat-V1.2-34B & 63.4 & 54.3 & 65.5 & 72.3 & 45.1 & 48.3 & 11.0 & 84.0 & 74.5 \\
Qwen-VL-Plus & 62.3 & 50.7 & 65.5 & 72.3 & 38.5 & 42.5 & 11.0 & 84.5 & 85.0 \\
GPT-4V & 62.0 & 67.7 & 71.7 & 75.0 & 40.2 & 54.3 & 11.5 & 80.0 & 80.9 \\
GeminiProVision & 61.6 & 37.0 & 49.3 & 63.6 & 67.0 & 35.2 & 43.3 & 11.0 & 74.5 \\
LLaVA-Next-34B & 60.8 & 48.6 & 65.5 & 64.3 & 46.2 & 50.8 & 13.0 & 76.5 & 80.0 \\
XComposer2-7B & 55.7 & 33.1 & 42.1 & 56.4 & 61.6 & 33.0 & 36.7 & 8.0 & 50.5 \\
BLIP2-Flan-T5-XXL & 54.8 & 28.3 & 30.0 & 51.8 & 49.1 & 34.6 & 28.3 & 14.0 & 80.5 \\
Yi-VL-34B & 54.2 & 50.0 & 61.8 & 61.6 & 38.5 & 40.0 & 14.0 & 68.5 & 57.0 \\
Monkey & 53.4 & 26.0 & 33.6 & 49.1 & 50.0 & 31.9 & 38.3 & 11.0 & 46.0 \\
DeepSeek-VL-7B & 53.2 & 40.7 & 50.9 & 47.3 & 34.6 & 32.5 & 12.5 & 37.0 & 67.5 \\
Yi-VL-6B & 53.2 & 44.3 & 54.5 & 51.8 & 33.5 & 44.2 & 13.0 & 47.0 & 48.0 \\
LLaVA-Next-13B & 53.0 & 35.7 & 53.6 & 58.0 & 31.9 & 37.5 & 14.5 & 50.0 & 70.0 \\
TransCore-M & 52.7 & 23.6 & 35.7 & 50.0 & 48.2 & 35.2 & 37.5 & 15.0 & 39.5 \\
QWen-VL-Chat & 52.5 & 22.0 & 33.6 & 50.9 & 47.3 & 30.8 & 29.2 & 13.5 & 55.0 \\
Claude3V-Haiku & 52.2 & 41.4 & 54.5 & 57.1 & 29.7 & 44.2 & 15.5 & 59.5 & 54.5 \\
XComposer & 52.1 & 27.6 & 35.0 & 49.1 & 56.2 & 33.0 & 35.0 & 14.0 & 38.5 \\
mPLUG-Owl2 & 52.0 & 27.6 & 40.0 & 50.0 & 46.4 & 35.2 & 29.2 & 18.0 & 35.0 \\
RBDash-v1-13B & 51.8 & 33.1 & 37.1 & 51.8 & 47.3 & 34.1 & 25.0 & 14.0 & 53.5 \\
LLaVA-v1.5-13B & 51.7 & 29.1 & 34.3 & 54.5 & 54.5 & 34.1 & 30.0 & 13.5 & 40.5 \\
CogVLM-Chat & 51.6 & 23.6 & 30.7 & 49.1 & 49.1 & 34.6 & 34.2 & 14.0 & 48.0 \\
ShareGPT4V-7B & 51.5 & 27.6 & 37.1 & 57.3 & 49.1 & 32.4 & 30.0 & 14.0 & 42.0 \\
LLaVA-Next-7B & 51.1 & 37.9 & 56.4 & 50.9 & 32.4 & 30.8 & 13.5 & 47.5 & 65.0 \\
LLaVA-v1.5-13B-XTuner & 51.1 & 28.3 & 35.0 & 43.6 & 49.1 & 35.2 & 32.5 & 14.0 & 33.5 \\
LlaVA-InternLM2-7B & 50.8 & 25.2 & 42.1 & 50.0 & 52.7 & 14.8 & 17.5 & 0.0 & 35.5 \\
LLaVA-v1.5-7B-Xtuner & 50.2 & 28.3 & 37.1 & 47.3 & 49.1 & 35.7 & 25.8 & 11.5 & 32.0 \\
SharedCaptioner & 49.9 & 25.2 & 37.9 & 50.9 & 47.3 & 32.4 & 36.7 & 14.5 & 45.0 \\
LLaVA-InternLM-7b & 49.7 & 23.6 & 39.3 & 47.3 & 55.4 & 29.7 & 33.3 & 14.0 & 46.5 \\
LLaVA-v1.5-7B & 49.5 & 24.4 & 34.3 & 52.7 & 46.4 & 28.6 & 27.5 & 12.5 & 42.5 \\
LLaMA-Adapter-v2-7B & 40.4 & 29.9 & 30.7 & 38.2 & 33.0 & 32.4 & 27.5 & 11.0 & 25.0 \\
VisualGLM\_6b & 38.6 & 20.5 & 29.3 & 30.9 & 41.1 & 26.9 & 29.2 & 14.0 & 37.5 \\
Frequency & 31.7 & 32.3 & 27.9 & 30.9 & 28.6 & 27.5 & 29.2 & 18.0 & 28.0 \\
Random & 28.5 & 24.4 & 25.7 & 22.7 & 25.0 & 25.3 & 29.2 & 10.5 & 27.5 \\
\bottomrule
\end{tabular}%
}
\end{table}

% =====================================================================================================================

% \usepackage{graphicx}
\begin{table}[]
\caption{Detail results of 30 LVLMs on \textbf{Emotion Quotient Test} and \textbf{Visual Illusion}.}
\label{tab:detailed_result_11}
\label{tab:my-table}
\resizebox{\textwidth}{!}{%
\begin{tabular}{l|c|cccccc|ccccc}
\toprule
 &  & \multicolumn{6}{c|}{Emotion Quotient Test} & \multicolumn{5}{c}{Visual Illusion} \\
 \midrule
Model & Overall & \begin{tabular}[c]{@{}c@{}}Facail Expression \\ Change Recognition\end{tabular} & \begin{tabular}[c]{@{}c@{}}Scene Emotion\\ Recognition\end{tabular} & \begin{tabular}[c]{@{}c@{}}Micro-Expression\\ Recognition\end{tabular} & \begin{tabular}[c]{@{}c@{}}Artwork Emotion\\ Recognition\end{tabular} & \begin{tabular}[c]{@{}c@{}}Body Emotion\\ Recognition\end{tabular} & \begin{tabular}[c]{@{}c@{}}Facial Expression\\ Recognition\end{tabular} & \begin{tabular}[c]{@{}c@{}}Color\\ Constancy\end{tabular} & \begin{tabular}[c]{@{}c@{}}Color\\ Assimilation\end{tabular} & \begin{tabular}[c]{@{}c@{}}Geometrical\\ Relativity\end{tabular} & \begin{tabular}[c]{@{}c@{}}Geometrical\\ Perspective\end{tabular} & \begin{tabular}[c]{@{}c@{}}Color\\ Contrast\end{tabular} \\
\midrule
InternVL-Chat-V1.2-34B & 63.4 & 59.0 & 30.0 & 42.5 & 43.0 & 70.5 & 47.2 & 34.5 & 44.5 & 82.5 & 75.0 & 71.0 \\
Qwen-VL-Plus & 62.3 & 58.5 & 19.5 & 40.5 & 41.0 & 62.0 & 27.8 & 47.5 & 29.0 & 58.3 & 43.0 & 73.5 \\
GPT-4V & 62.0 & 64.0 & 35.4 & 41.4 & 46.0 & 69.9 & 13.9 & 65.0 & 24.7 & 43.3 & 35.7 & 70.6 \\
GeminiProVision & 61.6 & 73.0 & 59.0 & 40.0 & 50.5 & 42.5 & 66.0 & 38.9 & 53.5 & 46.0 & 43.3 & 56.0 \\
LLaVA-Next-34B & 60.8 & 61.0 & 27.0 & 37.0 & 42.5 & 72.0 & 62.5 & 42.0 & 43.5 & 78.3 & 55.0 & 70.0 \\
XComposer2-7B & 55.7 & 72.0 & 53.0 & 36.5 & 47.0 & 47.0 & 66.0 & 51.4 & 39.5 & 47.5 & 75.8 & 49.0 \\
BLIP2-Flan-T5-XXL & 54.8 & 44.5 & 51.5 & 20.5 & 44.5 & 36.5 & 61.0 & 63.9 & 47.0 & 58.0 & 60.0 & 49.0 \\
Yi-VL-34B & 54.2 & 46.5 & 30.0 & 44.5 & 31.5 & 56.5 & 29.2 & 45.5 & 34.0 & 55.8 & 53.5 & 67.5 \\
Monkey & 53.4 & 51.5 & 49.5 & 24.0 & 46.0 & 37.5 & 62.0 & 59.7 & 30.5 & 61.5 & 62.5 & 62.5 \\
DeepSeek-VL-7B & 53.2 & 55.5 & 30.0 & 39.5 & 35.5 & 69.0 & 33.3 & 27.5 & 52.0 & 54.2 & 56.0 & 49.0 \\
Yi-VL-6B & 53.2 & 57.0 & 33.0 & 43.5 & 42.5 & 56.5 & 44.4 & 37.0 & 61.0 & 63.3 & 60.0 & 57.5 \\
LLaVA-Next-13B & 53.0 & 55.0 & 20.0 & 30.0 & 33.5 & 66.0 & 62.5 & 30.5 & 54.5 & 73.3 & 66.0 & 63.0 \\
TransCore-M & 52.7 & 67.0 & 54.5 & 26.0 & 29.0 & 34.5 & 67.0 & 47.2 & 24.0 & 61.0 & 75.0 & 71.0 \\
QWen-VL-Chat & 52.5 & 48.5 & 55.5 & 26.5 & 42.5 & 38.0 & 61.5 & 48.6 & 48.0 & 50.0 & 50.0 & 58.0 \\
Claude3V-Haiku & 52.2 & 53.0 & 28.0 & 39.5 & 34.0 & 43.0 & 13.9 & 38.5 & 58.5 & 55.8 & 56.5 & 5.0 \\
XComposer & 52.1 & 62.0 & 58.5 & 26.5 & 29.0 & 36.5 & 63.0 & 51.4 & 47.5 & 49.0 & 60.0 & 59.0 \\
mPLUG-Owl2 & 52.0 & 45.5 & 55.0 & 26.5 & 34.5 & 34.5 & 56.0 & 66.7 & 30.0 & 65.5 & 60.0 & 70.5 \\
RBDash-v1-13B & 51.8 & 69.5 & 50.0 & 13.0 & 29.0 & 33.0 & 66.0 & 59.7 & 30.5 & 63.0 & 72.5 & 60.0 \\
LLaVA-v1.5-13B & 51.7 & 66.5 & 57.5 & 25.5 & 21.0 & 37.0 & 64.5 & 63.9 & 29.0 & 64.5 & 66.7 & 66.5 \\
CogVLM-Chat & 51.6 & 51.5 & 56.5 & 30.0 & 50.5 & 43.5 & 70.0 & 55.6 & 45.5 & 51.0 & 60.8 & 49.0 \\
ShareGPT4V-7B & 51.5 & 63.5 & 56.5 & 26.0 & 23.0 & 33.5 & 63.5 & 52.8 & 26.5 & 60.0 & 65.8 & 67.5 \\
LLaVA-Next-7B & 51.1 & 59.0 & 27.0 & 33.5 & 37.5 & 62.5 & 69.4 & 29.5 & 63.0 & 70.8 & 68.0 & 57.5 \\
LLaVA-v1.5-13B-XTuner & 51.1 & 72.5 & 63.5 & 25.0 & 43.0 & 37.5 & 64.5 & 44.4 & 26.0 & 60.0 & 75.0 & 66.5 \\
LlaVA-InternLM2-7B & 50.8 & 71.0 & 55.5 & 35.5 & 33.0 & 42.5 & 69.0 & 50.0 & 28.5 & 52.0 & 73.3 & 57.5 \\
LLaVA-v1.5-7B-Xtuner & 50.2 & 61.0 & 57.0 & 28.5 & 30.5 & 31.5 & 57.0 & 61.1 & 30.5 & 62.0 & 76.7 & 70.0 \\
SharedCaptioner & 49.9 & 54.5 & 47.0 & 12.0 & 36.0 & 28.0 & 62.0 & 47.2 & 41.0 & 48.5 & 81.7 & 57.0 \\
LLaVA-InternLM-7b & 49.7 & 59.0 & 58.0 & 28.5 & 42.0 & 36.0 & 58.5 & 63.9 & 26.0 & 66.5 & 67.5 & 63.5 \\
LLaVA-v1.5-7B & 49.5 & 65.0 & 57.5 & 18.0 & 21.5 & 37.5 & 55.5 & 56.9 & 28.0 & 64.0 & 70.0 & 69.0 \\
LLaMA-Adapter-v2-7B & 40.4 & 23.0 & 52.0 & 20.0 & 34.0 & 32.5 & 54.5 & 37.5 & 40.5 & 29.0 & 36.7 & 38.5 \\
VisualGLM\_6b & 38.6 & 28.5 & 43.5 & 23.5 & 28.0 & 31.0 & 44.0 & 25.0 & 50.5 & 29.0 & 29.2 & 11.0 \\
Frequency & 31.7 & 28.0 & 29.5 & 30.0 & 33.0 & 31.0 & 29.0 & 52.8 & 51.0 & 50.5 & 53.3 & 53.0 \\
Random & 28.5 & 26.0 & 19.5 & 27.5 & 30.5 & 23.0 & 26.0 & 48.6 & 50.0 & 50.5 & 51.7 & 53.0 \\
\bottomrule
\end{tabular}%
}
\end{table}

% =====================================================================================================================

% \usepackage{graphicx}
\begin{table}[]
\centering
\caption{Detail results of 30 LVLMs on \textbf{Meme Understanding}, \textbf{Counting} and \textbf{Hallucination}.}
\label{tab:detailed_result_12}
\resizebox{\textwidth}{!}{%
\begin{tabular}{l|c|cc|cccc|cccc}
\toprule
 &  & \multicolumn{2}{c|}{Meme Understanding} & \multicolumn{4}{c|}{Counting} & \multicolumn{4}{c}{Hallucination} \\
\midrule
Model & Overall & \begin{tabular}[c]{@{}c@{}}Meme Video \\ Understanding\end{tabular} & \begin{tabular}[c]{@{}c@{}}Meme Image\\ Understanding\end{tabular} & \begin{tabular}[c]{@{}c@{}}Counting by \\ Visual Prompting\end{tabular} & \begin{tabular}[c]{@{}c@{}}Counting by \\ Category\end{tabular} & \begin{tabular}[c]{@{}c@{}}Crowd \\ Counting\end{tabular} & \begin{tabular}[c]{@{}c@{}}Counting by\\ Reasoning\end{tabular} & \begin{tabular}[c]{@{}c@{}}Order\\ Hallucination\end{tabular} & \begin{tabular}[c]{@{}c@{}}Relation\\ Hallucination\end{tabular} & \begin{tabular}[c]{@{}c@{}}Attribute\\ Hallucination\end{tabular} & \begin{tabular}[c]{@{}c@{}}Exist\\ Hallucination\end{tabular} \\
\midrule
InternVL-Chat-V1.2-34B & 63.4 & 88.0 & 37.0 & 75.5 & 56.0 & 78.0 & 56.0 & 85.5 & 80.0 & 85.5 & 78.5 \\
Qwen-VL-Plus & 62.3 & 82.5 & 39.5 & 75.0 & 44.0 & 70.0 & 55.5 & 81.0 & 74.0 & 87.0 & 37.5 \\
GPT-4V & 62.0 & 92.0 & 30.0 & 67.9 & 38.3 & 64.4 & 35.7 & 65.3 & 81.4 & 81.5 & 50.3 \\
GeminiProVision & 61.6 & 75.0 & 76.5 & 40.5 & 68.5 & 52.0 & 64.5 & 39.0 & 59.5 & 82.5 & 82.5 \\
LLaVA-Next-34B & 60.8 & 88.0 & 23.0 & 72.6 & 56.5 & 81.0 & 55.0 & 83.0 & 76.5 & 87.5 & 33.5 \\
XComposer2-7B & 55.7 & 55.5 & 87.0 & 25.5 & 73.2 & 29.5 & 76.0 & 44.0 & 76.0 & 71.5 & 86.5 \\
BLIP2-Flan-T5-XXL & 54.8 & 70.5 & 82.0 & 23.0 & 53.8 & 12.0 & 30.5 & 51.0 & 62.0 & 72.0 & 79.5 \\
Yi-VL-34B & 54.2 & 83.0 & 29.5 & 65.2 & 44.5 & 69.0 & 59.0 & 84.5 & 71.5 & 79.5 & 27.5 \\
Monkey & 53.4 & 52.0 & 87.0 & 28.0 & 66.9 & 24.0 & 55.5 & 43.0 & 59.0 & 68.5 & 82.0 \\
DeepSeek-VL-7B & 53.2 & 84.5 & 35.0 & 70.8 & 18.0 & 52.5 & 36.5 & 63.0 & 73.0 & 84.5 & 22.0 \\
Yi-VL-6B & 53.2 & 78.0 & 31.0 & 66.2 & 44.5 & 60.5 & 53.5 & 79.0 & 69.5 & 79.0 & 28.0 \\
LLaVA-Next-13B & 53.0 & 79.5 & 30.5 & 68.5 & 54.0 & 65.0 & 49.5 & 74.5 & 72.0 & 84.0 & 24.0 \\
TransCore-M & 52.7 & 70.0 & 83.5 & 36.0 & 65.0 & 57.5 & 59.5 & 51.0 & 74.5 & 79.0 & 83.0 \\
QWen-VL-Chat & 52.5 & 64.5 & 84.0 & 31.0 & 63.2 & 39.0 & 53.5 & 45.0 & 62.5 & 71.5 & 76.5 \\
Claude3V-Haiku & 52.2 & 75.0 & 32.5 & 57.0 & 45.0 & 59.0 & 43.5 & 62.0 & 82.0 & 78.5 & 32.0 \\
XComposer & 52.1 & 58.0 & 69.5 & 32.5 & 66.2 & 23.5 & 37.5 & 34.0 & 74.0 & 75.0 & 83.0 \\
mPLUG-Owl2 & 52.0 & 49.5 & 68.5 & 33.0 & 68.6 & 36.0 & 53.0 & 33.0 & 63.0 & 72.5 & 85.0 \\
RBDash-v1-13B & 51.8 & 54.5 & 80.5 & 31.0 & 65.1 & 46.0 & 64.5 & 49.0 & 67.5 & 64.0 & 86.0 \\
LLaVA-v1.5-13B & 51.7 & 55.5 & 76.5 & 33.5 & 64.4 & 60.5 & 62.0 & 47.5 & 72.0 & 61.5 & 82.0 \\
CogVLM-Chat & 51.6 & 62.5 & 88.5 & 30.5 & 59.1 & 52.5 & 57.0 & 58.0 & 58.5 & 59.0 & 88.5 \\
ShareGPT4V-7B & 51.5 & 62.5 & 78.5 & 33.5 & 64.2 & 49.5 & 56.5 & 43.5 & 63.0 & 68.5 & 74.5 \\
LLaVA-Next-7B & 51.1 & 80.0 & 28.0 & 69.0 & 51.0 & 62.0 & 45.0 & 61.0 & 77.0 & 81.5 & 20.0 \\
LLaVA-v1.5-13B-XTuner & 51.1 & 52.0 & 81.0 & 33.5 & 67.4 & 53.5 & 62.0 & 46.0 & 72.0 & 66.0 & 82.0 \\
LlaVA-InternLM2-7B & 50.8 & 46.0 & 79.0 & 27.5 & 67.6 & 45.5 & 66.5 & 35.5 & 77.5 & 73.5 & 84.5 \\
LLaVA-v1.5-7B-Xtuner & 50.2 & 38.0 & 75.0 & 28.0 & 66.9 & 46.0 & 58.5 & 30.0 & 62.5 & 70.5 & 85.5 \\
SharedCaptioner & 49.9 & 44.0 & 63.5 & 37.0 & 71.0 & 31.0 & 46.0 & 28.0 & 76.5 & 69.5 & 78.5 \\
LLaVA-InternLM-7b & 49.7 & 35.5 & 81.0 & 27.5 & 67.4 & 33.0 & 56.0 & 27.5 & 70.0 & 64.5 & 86.0 \\
LLaVA-v1.5-7B & 49.5 & 62.0 & 79.0 & 27.5 & 62.3 & 46.5 & 53.5 & 40.5 & 62.0 & 69.5 & 74.5 \\
LLaMA-Adapter-v2-7B & 40.4 & 34.0 & 47.0 & 36.0 & 36.2 & 26.5 & 21.5 & 28.0 & 53.5 & 26.5 & 78.0 \\
VisualGLM\_6b & 38.6 & 34.0 & 55.5 & 26.0 & 38.4 & 31.5 & 28.5 & 20.5 & 55.5 & 28.0 & 53.0 \\
Frequency & 31.7 & 29.0 & 36.5 & 27.5 & 27.9 & 26.5 & 31.0 & 27.0 & 54.0 & 40.5 & 52.0 \\
Random & 28.5 & 22.5 & 28.5 & 28.0 & 24.6 & 24.5 & 23.0 & 29.5 & 49.5 & 38.5 & 49.0 \\
\bottomrule
\end{tabular}%
}
\end{table}

% =====================================================================================================================

% \usepackage{graphicx}
\begin{table}[]
\centering
\caption{Detail results of 30 LVLMs on \textbf{Retrieval} and \textbf{Visual Prompt Understanding}.}
\label{tab:detailed_result_13}
\resizebox{\textwidth}{!}{%
\begin{tabular}{l|c|ccccccc|cc}
\toprule
 &  & \multicolumn{7}{c|}{Retrieval} & \multicolumn{2}{c}{Visual Prompt Understanding} \\
 \midrule
Model & Overall & \begin{tabular}[c]{@{}c@{}}Person\\ ReID\end{tabular} & \begin{tabular}[c]{@{}c@{}}Sketch to\\ Image Retrieval\end{tabular} & \begin{tabular}[c]{@{}c@{}}Face\\ Retrieval\end{tabular} & \begin{tabular}[c]{@{}c@{}}Handwritten\\ Retrieval\end{tabular} & \begin{tabular}[c]{@{}c@{}}Vehicle\\ Retrieval\end{tabular} & \begin{tabular}[c]{@{}c@{}}Image to \\ Image Retrieval\end{tabular} & \begin{tabular}[c]{@{}c@{}}Text to Image\\ Retrieval\end{tabular} & \begin{tabular}[c]{@{}c@{}}Set-of-Marks\\ Recognition\end{tabular} & \begin{tabular}[c]{@{}c@{}}Visual Prompt\\ Understanding\end{tabular} \\
\midrule
InternVL-Chat-V1.2-34B & 63.4 & 60.0 & 62.0 & 26.0 & 70.5 & 71.0 & 50.0 & 54.3 & 66.5 & 50.0 \\
Qwen-VL-Plus & 62.3 & 50.5 & 40.0 & 24.0 & 43.0 & 40.0 & 42.0 & 45.7 & 63.5 & 71.5 \\
GPT-4V & 62.0 & 49.0 & 51.5 & 25.9 & 47.0 & 45.5 & 47.5 & 48.2 & 57.7 & 62.8 \\
GeminiProVision & 61.6 & 72.5 & 84.0 & 92.5 & 25.5 & 80.0 & 81.5 & 43.0 & 43.2 & 58.5 \\
LLaVA-Next-34B & 60.8 & 37.0 & 32.0 & 28.5 & 46.5 & 29.0 & 40.0 & 58.8 & 67.5 & 47.0 \\
XComposer2-7B & 55.7 & 25.5 & 31.0 & 28.5 & 26.5 & 43.5 & 28.0 & 44.0 & 56.8 & 55.5 \\
BLIP2-Flan-T5-XXL & 54.8 & 28.0 & 31.5 & 25.0 & 25.5 & 25.5 & 30.0 & 26.5 & 27.1 & 52.5 \\
Yi-VL-34B & 54.2 & 38.0 & 25.5 & 27.5 & 23.0 & 27.5 & 20.5 & 39.7 & 59.0 & 63.5 \\
Monkey & 53.4 & 22.5 & 31.5 & 22.0 & 26.0 & 30.5 & 26.5 & 28.5 & 41.7 & 45.5 \\
DeepSeek-VL-7B & 53.2 & 32.5 & 21.5 & 25.5 & 32.0 & 26.0 & 31.5 & 44.7 & 53.0 & 54.0 \\
Yi-VL-6B & 53.2 & 29.5 & 24.5 & 19.0 & 20.0 & 25.0 & 28.5 & 35.7 & 50.5 & 48.0 \\
LLaVA-Next-13B & 53.0 & 35.0 & 26.0 & 24.0 & 30.5 & 28.0 & 32.0 & 53.3 & 43.5 & 21.0 \\
TransCore-M & 52.7 & 24.0 & 36.5 & 23.0 & 27.0 & 31.5 & 26.5 & 24.0 & 51.3 & 52.5 \\
QWen-VL-Chat & 52.5 & 28.5 & 31.5 & 27.0 & 27.5 & 25.0 & 25.0 & 28.0 & 38.2 & 46.5 \\
Claude3V-Haiku & 52.2 & 35.0 & 34.5 & 25.5 & 35.5 & 34.5 & 36.5 & 40.7 & 46.5 & 40.0 \\
XComposer & 52.1 & 43.5 & 60.0 & 61.0 & 28.5 & 60.5 & 62.0 & 32.5 & 41.7 & 39.5 \\
mPLUG-Owl2 & 52.0 & 30.0 & 35.5 & 24.5 & 28.0 & 23.5 & 25.5 & 26.0 & 34.7 & 45.5 \\
RBDash-v1-13B & 51.8 & 28.0 & 31.5 & 23.0 & 25.5 & 21.5 & 27.0 & 27.5 & 49.7 & 53.0 \\
LLaVA-v1.5-13B & 51.7 & 29.0 & 28.5 & 30.0 & 20.5 & 25.5 & 29.0 & 28.0 & 42.2 & 45.5 \\
CogVLM-Chat & 51.6 & 21.0 & 30.5 & 20.5 & 25.5 & 31.0 & 24.5 & 29.5 & 29.6 & 50.0 \\
ShareGPT4V-7B & 51.5 & 24.0 & 33.0 & 23.0 & 28.0 & 29.5 & 27.0 & 30.0 & 46.2 & 48.0 \\
LLaVA-Next-7B & 51.1 & 31.0 & 20.5 & 25.5 & 33.0 & 24.5 & 28.5 & 37.7 & 48.0 & 29.0 \\
LLaVA-v1.5-13B-XTuner & 51.1 & 21.5 & 32.5 & 21.5 & 25.5 & 31.5 & 25.5 & 26.5 & 42.2 & 53.5 \\
LlaVA-InternLM2-7B & 50.8 & 27.0 & 32.5 & 21.5 & 25.0 & 28.0 & 29.5 & 30.5 & 37.7 & 52.5 \\
LLaVA-v1.5-7B-Xtuner & 50.2 & 22.0 & 35.5 & 19.5 & 25.5 & 26.0 & 25.5 & 28.0 & 37.7 & 47.5 \\
SharedCaptioner & 49.9 & 21.0 & 30.0 & 25.0 & 26.5 & 31.0 & 27.0 & 28.5 & 44.7 & 46.0 \\
LLaVA-InternLM-7b & 49.7 & 19.0 & 31.5 & 22.0 & 25.5 & 26.0 & 27.5 & 27.0 & 40.2 & 51.0 \\
LLaVA-v1.5-7B & 49.5 & 23.0 & 30.0 & 20.5 & 25.5 & 31.0 & 26.0 & 26.5 & 19.6 & 47.0 \\
LLaMA-Adapter-v2-7B & 40.4 & 21.5 & 31.0 & 29.0 & 22.0 & 26.0 & 21.0 & 18.0 & 28.6 & 39.0 \\
VisualGLM\_6b & 38.6 & 22.0 & 31.0 & 26.5 & 25.5 & 32.0 & 20.5 & 24.5 & 21.6 & 32.5 \\
Frequency & 31.7 & 30.0 & 37.0 & 29.0 & 28.0 & 31.0 & 27.5 & 26.5 & 28.1 & 30.5 \\
Random & 28.5 & 26.0 & 35.5 & 17.5 & 23.5 & 22.0 & 22.5 & 23.0 & 31.2 & 31.5 \\
\bottomrule
\end{tabular}%
}
\end{table}

% =====================================================================================================================

% \usepackage{graphicx}
\begin{table}[]
\centering
\caption{Detail results of 30 LVLMs on \textbf{Anomaly Dectection} and \textbf{Keypoint Detection}.}
\label{tab:detailed_result_14}
\resizebox{\textwidth}{!}{%
\begin{tabular}{l|c|ccccc|ccccc}
\toprule
 &  & \multicolumn{5}{c|}{Anomaly Dectection} & \multicolumn{5}{c}{Keypoint Detection} \\
\midrule
Model & Overall & \begin{tabular}[c]{@{}c@{}}Industrial Produce \\ Anomaly Detection\end{tabular} & \begin{tabular}[c]{@{}c@{}}Face Mask\\ Anomaly Dectection\end{tabular} & \begin{tabular}[c]{@{}c@{}}Helmet Anomaly\\ Detection\end{tabular} & \begin{tabular}[c]{@{}c@{}}Behavior Anomaly\\ Detection\end{tabular} & \begin{tabular}[c]{@{}c@{}}Traffic Anomaly\\ Detection\end{tabular} & \begin{tabular}[c]{@{}c@{}}Furniture Keypoint\\ Detection\end{tabular} & \begin{tabular}[c]{@{}c@{}}Human Keypoint\\ Detection\end{tabular} & \begin{tabular}[c]{@{}c@{}}Clothes Keypoint\\ Detection\end{tabular} & \begin{tabular}[c]{@{}c@{}}Animal Keypoint\\ Detection\end{tabular} & \begin{tabular}[c]{@{}c@{}}Vehicle Keypoint\\ Detection\end{tabular} \\
\midrule
InternVL-Chat-V1.2-34B & 63.4 & 67.5 & 87.0 & 22.5 & 42.0 & 66.0 & 54.0 & 65.0 & 43.0 & 63.0 & 86.0 \\
Qwen-VL-Plus & 62.3 & 70.5 & 74.5 & 46.5 & 36.0 & 66.5 & 49.0 & 60.0 & 41.5 & 44.6 & 81.5 \\
GPT-4V & 62.0 & 71.0 & 78.5 & 20.8 & 42.0 & 44.8 & 53.0 & 46.3 & 30.0 & 52.9 & 86.0 \\
GeminiProVision & 61.6 & 57.5 & 50.5 & 74.5 & 20.5 & 34.0 & 45.0 & 61.5 & 48.5 & 38.0 & 54.3 \\
LLaVA-Next-34B & 60.8 & 64.5 & 79.0 & 40.5 & 58.0 & 64.0 & 72.0 & 62.0 & 42.0 & 54.3 & 83.0 \\
XComposer2-7B & 55.7 & 68.5 & 57.5 & 67.0 & 46.5 & 41.5 & 46.5 & 49.0 & 38.0 & 31.5 & 42.4 \\
BLIP2-Flan-T5-XXL & 54.8 & 65.0 & 29.5 & 64.0 & 17.5 & 42.5 & 66.5 & 63.0 & 74.0 & 35.5 & 62.0 \\
Yi-VL-34B & 54.2 & 55.5 & 86.0 & 33.0 & 42.0 & 45.5 & 55.5 & 47.5 & 34.5 & 32.6 & 77.5 \\
Monkey & 53.4 & 43.0 & 40.0 & 70.5 & 27.0 & 42.5 & 45.0 & 26.5 & 47.5 & 26.5 & 35.9 \\
DeepSeek-VL-7B & 53.2 & 33.0 & 76.5 & 27.0 & 42.5 & 48.0 & 34.5 & 27.5 & 35.0 & 27.2 & 81.0 \\
Yi-VL-6B & 53.2 & 55.0 & 82.5 & 24.5 & 40.0 & 63.0 & 52.5 & 50.0 & 32.5 & 41.3 & 82.5 \\
LLaVA-Next-13B & 53.0 & 62.5 & 79.5 & 27.0 & 42.5 & 36.5 & 41.5 & 27.5 & 27.5 & 17.4 & 80.5 \\
TransCore-M & 52.7 & 11.0 & 56.0 & 82.0 & 27.0 & 42.5 & 46.0 & 53.0 & 31.5 & 32.0 & 30.4 \\
QWen-VL-Chat & 52.5 & 29.0 & 41.0 & 72.0 & 27.0 & 32.0 & 44.5 & 31.0 & 36.0 & 31.0 & 37.0 \\
Claude3V-Haiku & 52.2 & 47.0 & 46.5 & 22.0 & 41.0 & 49.5 & 62.0 & 42.0 & 29.0 & 52.2 & 81.5 \\
XComposer & 52.1 & 11.5 & 50.5 & 85.5 & 27.0 & 42.5 & 37.0 & 43.5 & 49.5 & 44.5 & 37.0 \\
mPLUG-Owl2 & 52.0 & 69.0 & 51.5 & 59.0 & 27.0 & 40.5 & 31.5 & 36.0 & 31.0 & 35.5 & 30.4 \\
RBDash-v1-13B & 51.8 & 20.5 & 56.5 & 82.0 & 27.0 & 42.5 & 37.0 & 38.5 & 29.5 & 29.5 & 31.5 \\
LLaVA-v1.5-13B & 51.7 & 35.5 & 58.0 & 78.5 & 27.0 & 42.5 & 37.5 & 39.5 & 28.5 & 35.0 & 16.3 \\
CogVLM-Chat & 51.6 & 31.5 & 30.0 & 86.0 & 27.0 & 42.5 & 33.5 & 22.5 & 28.0 & 28.5 & 28.3 \\
ShareGPT4V-7B & 51.5 & 46.0 & 46.5 & 79.0 & 27.0 & 42.5 & 32.0 & 26.5 & 22.0 & 28.0 & 22.8 \\
LLaVA-Next-7B & 51.1 & 40.0 & 83.5 & 27.0 & 42.5 & 28.5 & 18.0 & 18.5 & 31.0 & 26.1 & 79.5 \\
LLaVA-v1.5-13B-XTuner & 51.1 & 53.5 & 56.5 & 80.5 & 27.0 & 42.5 & 35.5 & 32.0 & 26.5 & 31.5 & 18.5 \\
LlaVA-InternLM2-7B & 50.8 & 76.0 & 57.0 & 83.5 & 27.0 & 42.5 & 43.0 & 46.5 & 31.0 & 30.5 & 25.0 \\
LLaVA-v1.5-7B-Xtuner & 50.2 & 43.0 & 55.0 & 68.5 & 27.0 & 42.5 & 36.5 & 32.0 & 24.0 & 31.0 & 18.5 \\
SharedCaptioner & 49.9 & 22.0 & 33.0 & 72.0 & 32.5 & 32.0 & 45.5 & 26.0 & 33.5 & 32.5 & 30.4 \\
LLaVA-InternLM-7b & 49.7 & 30.0 & 55.5 & 77.5 & 27.0 & 42.5 & 37.5 & 33.5 & 35.0 & 37.0 & 22.8 \\
LLaVA-v1.5-7B & 49.5 & 59.5 & 46.0 & 70.5 & 27.0 & 42.5 & 40.0 & 25.0 & 23.0 & 35.0 & 34.8 \\
LLaMA-Adapter-v2-7B & 40.4 & 8.0 & 31.5 & 44.0 & 26.0 & 42.5 & 49.5 & 27.5 & 30.5 & 31.0 & 35.9 \\
VisualGLM\_6b & 38.6 & 16.0 & 30.5 & 57.0 & 27.0 & 42.0 & 43.0 & 30.0 & 33.0 & 33.0 & 37.0 \\
Frequency & 31.7 & 29.0 & 29.5 & 58.5 & 54.0 & 51.0 & 45.5 & 27.0 & 27.5 & 37.0 & 31.5 \\
Random & 28.5 & 23.0 & 24.5 & 44.5 & 44.5 & 46.0 & 42.0 & 32.0 & 26.5 & 34.5 & 26.1 \\
\bottomrule
\end{tabular}%
}
\end{table}

% =====================================================================================================================

% \usepackage{graphicx}
\begin{table}[]
\centering
\caption{Detail results of 30 LVLMs on \textbf{Visual Commonsense Reasoning}, \textbf{Visual Code} and \textbf{Image Evaluation}.}
\label{tab:detailed_result_15}
\resizebox{\textwidth}{!}{%
\begin{tabular}{l|c|c|ccc|cc}
\toprule
 &  & Visual Commonsense Reasoning & \multicolumn{3}{c|}{Visual Code} & \multicolumn{2}{c}{Image Evaluation} \\
 \midrule
Model & Overall & \begin{tabular}[c]{@{}c@{}}Reasoning\\ Whoops\end{tabular} & \begin{tabular}[c]{@{}c@{}}Equation\\ to Latex\end{tabular} & \begin{tabular}[c]{@{}c@{}}Screenshot\\ to Code\end{tabular} & \begin{tabular}[c]{@{}c@{}}Sketch\\ to Code\end{tabular} & \begin{tabular}[c]{@{}c@{}}Image Quality\\ Assessment\end{tabular} & \begin{tabular}[c]{@{}c@{}}Lvlm Response\\ Judgement\end{tabular} \\
\midrule
InternVL-Chat-V1.2-34B & 63.4 & 76.0 & 35.0 & 31.0 & 59.5 & 35.5 & 26.5 \\
Qwen-VL-Plus & 62.3 & 76.5 & 34.5 & 24.5 & 42.0 & 36.0 & 27.5 \\
GPT-4V & 62.0 & 73.0 & 35.0 & 31.9 & 93.0 & 46.0 & 40.8 \\
GeminiProVision & 61.6 & 86.5 & 75.0 & 37.0 & 26.0 & 41.5 & 28.5 \\
LLaVA-Next-34B & 60.8 & 75.0 & 37.5 & 30.5 & 52.5 & 35.0 & 9.5 \\
XComposer2-7B & 55.7 & 83.0 & 45.0 & 33.5 & 30.5 & 48.5 & 39.0 \\
BLIP2-Flan-T5-XXL & 54.8 & 77.0 & 44.5 & 49.5 & 35.5 & 30.5 & 29.0 \\
Yi-VL-34B & 54.2 & 71.5 & 25.0 & 23.5 & 47.5 & 29.0 & 22.0 \\
Monkey & 53.4 & 85.5 & 50.0 & 27.0 & 23.0 & 23.0 & 29.0 \\
DeepSeek-VL-7B & 53.2 & 41.0 & 29.0 & 23.5 & 39.5 & 31.5 & 46.0 \\
Yi-VL-6B & 53.2 & 65.5 & 31.0 & 31.0 & 49.0 & 29.5 & 39.0 \\
LLaVA-Next-13B & 53.0 & 47.0 & 22.0 & 12.0 & 33.5 & 31.0 & 5.0 \\
TransCore-M & 52.7 & 85.5 & 48.5 & 19.5 & 18.5 & 33.0 & 35.5 \\
QWen-VL-Chat & 52.5 & 86.0 & 47.5 & 21.0 & 24.0 & 31.0 & 29.0 \\
Claude3V-Haiku & 52.2 & 70.5 & 33.0 & 32.0 & 37.5 & 29.5 & 33.5 \\
XComposer & 52.1 & 78.0 & 44.5 & 21.5 & 19.0 & 26.5 & 31.5 \\
mPLUG-Owl2 & 52.0 & 85.5 & 35.0 & 19.5 & 27.5 & 34.0 & 26.0 \\
RBDash-v1-13B & 51.8 & 78.0 & 47.5 & 15.5 & 13.5 & 47.5 & 30.5 \\
LLaVA-v1.5-13B & 51.7 & 79.0 & 51.5 & 16.5 & 14.5 & 43.0 & 28.5 \\
CogVLM-Chat & 51.6 & 82.0 & 35.5 & 27.0 & 22.5 & 27.5 & 28.5 \\
ShareGPT4V-7B & 51.5 & 83.0 & 48.5 & 17.0 & 24.5 & 26.0 & 29.5 \\
LLaVA-Next-7B & 51.1 & 42.5 & 23.5 & 17.5 & 29.0 & 32.5 & 5.0 \\
LLaVA-v1.5-13B-XTuner & 51.1 & 82.0 & 46.0 & 21.5 & 17.5 & 47.0 & 31.5 \\
LlaVA-InternLM2-7B & 50.8 & 83.0 & 27.0 & 18.5 & 10.0 & 32.0 & 36.5 \\
LLaVA-v1.5-7B-Xtuner & 50.2 & 80.5 & 34.0 & 16.5 & 22.0 & 35.5 & 29.0 \\
SharedCaptioner & 49.9 & 82.5 & 34.0 & 28.5 & 23.5 & 13.5 & 27.0 \\
LLaVA-InternLM-7b & 49.7 & 75.5 & 44.0 & 22.5 & 15.5 & 36.5 & 29.5 \\
LLaVA-v1.5-7B & 49.5 & 81.0 & 31.0 & 28.5 & 23.0 & 25.0 & 30.5 \\
LLaMA-Adapter-v2-7B & 40.4 & 71.0 & 23.5 & 29.5 & 24.5 & 40.0 & 26.5 \\
VisualGLM\_6b & 38.6 & 65.0 & 29.0 & 18.0 & 23.5 & 26.0 & 30.0 \\
Frequency & 31.7 & 27.0 & 29.5 & 30.0 & 29.0 & 32.5 & 27.5 \\
Random & 28.5 & 28.0 & 27.5 & 30.5 & 28.5 & 28.5 & 21.5 \\
\bottomrule
\end{tabular}%
}
\end{table}

% =====================================================================================================================

% \usepackage{graphicx}
\begin{table}[]
\centering
\caption{Detail results of 30 LVLMs on \textbf{Pixel-level Perception} and \textbf{Multiple Image Analysis}.}
\label{tab:detailed_result_16}
\resizebox{\textwidth}{!}{%
\begin{tabular}{l|ccccccc|cc}
\toprule
 &  & \multicolumn{6}{c|}{Pixel-level Perception} & \multicolumn{2}{c}{Multiple Image Analysis} \\
 \midrule
Model & Overall & \begin{tabular}[c]{@{}c@{}}Depth \\ Estimation\end{tabular} & \begin{tabular}[c]{@{}c@{}}Polygon\\ Localization\end{tabular} & \begin{tabular}[c]{@{}c@{}}Interactive\\ Segmentation\end{tabular} & \begin{tabular}[c]{@{}c@{}}Pixel\\ Recognition\end{tabular} & \begin{tabular}[c]{@{}c@{}}Pixel\\ Localization\end{tabular} & \begin{tabular}[c]{@{}c@{}}Image\\ Matting\end{tabular} & \begin{tabular}[c]{@{}c@{}}Spot the\\ Similarity\end{tabular} & \begin{tabular}[c]{@{}c@{}}Spot the\\ Difference\end{tabular} \\
\midrule
InternVL-Chat-V1.2-34B & 63.4 & 23.5 & 46.1 & 45.0 & 69.5 & 25.0 & 15.5 & 68.5 & 97.0 \\
Qwen-VL-Plus & 62.3 & 29.5 & 63.8 & 51.5 & 67.5 & 31.5 & 15.0 & 78.0 & 85.0 \\
GPT-4V & 62.0 & 26.5 & 66.0 & 80.6 & 76.0 & 25.0 & 15.3 & 46.0 & 94.5 \\
GeminiProVision & 61.6 & 29.0 & 36.0 & 47.5 & 77.0 & 29.5 & 23.0 & 50.5 & 90.0 \\
LLaVA-Next-34B & 60.8 & 26.0 & 45.4 & 57.5 & 69.0 & 34.5 & 19.0 & 59.0 & 96.5 \\
XComposer2-7B & 55.7 & 21.5 & 52.5 & 51.1 & 77.0 & 32.0 & 29.5 & 68.5 & 93.0 \\
BLIP2-Flan-T5-XXL & 54.8 & 63.5 & 39.5 & 32.6 & 71.5 & 39.0 & 30.5 & 45.5 & 80.0 \\
Yi-VL-34B & 54.2 & 20.0 & 60.3 & 39.5 & 63.5 & 29.5 & 17.5 & 50.5 & 81.5 \\
Monkey & 53.4 & 21.5 & 31.5 & 24.1 & 75.0 & 27.0 & 14.0 & 54.5 & 63.0 \\
DeepSeek-VL-7B & 53.2 & 25.5 & 23.4 & 61.5 & 73.5 & 28.0 & 18.0 & 54.5 & 15.5 \\
Yi-VL-6B & 53.2 & 30.0 & 35.5 & 42.0 & 61.0 & 25.0 & 18.5 & 47.0 & 57.0 \\
LLaVA-Next-13B & 53.0 & 23.5 & 27.7 & 40.0 & 72.5 & 24.0 & 16.5 & 45.5 & 27.5 \\
TransCore-M & 52.7 & 50.0 & 31.5 & 35.5 & 71.0 & 25.0 & 17.0 & 45.5 & 60.0 \\
QWen-VL-Chat & 52.5 & 23.5 & 29.0 & 28.4 & 73.5 & 25.0 & 17.0 & 55.0 & 43.5 \\
Claude3V-Haiku & 52.2 & 24.0 & 66.7 & 8.5 & 74.0 & 24.5 & 15.5 & 45.5 & 24.0 \\
XComposer & 52.1 & 35.5 & 41.0 & 40.4 & 70.0 & 33.5 & 17.5 & 54.0 & 78.5 \\
mPLUG-Owl2 & 52.0 & 59.5 & 37.0 & 38.3 & 68.0 & 30.0 & 16.5 & 58.5 & 51.5 \\
RBDash-v1-13B & 51.8 & 46.5 & 30.0 & 37.6 & 66.5 & 32.0 & 17.0 & 54.5 & 9.5 \\
LLaVA-v1.5-13B & 51.7 & 51.0 & 34.0 & 29.8 & 69.0 & 25.0 & 17.0 & 45.5 & 11.5 \\
CogVLM-Chat & 51.6 & 35.5 & 23.0 & 22.0 & 67.5 & 24.0 & 27.5 & 52.5 & 89.0 \\
ShareGPT4V-7B & 51.5 & 45.5 & 32.0 & 34.0 & 70.0 & 34.0 & 20.0 & 48.0 & 28.0 \\
LLaVA-Next-7B & 51.1 & 16.0 & 25.5 & 33.0 & 72.0 & 25.5 & 23.5 & 55.5 & 9.5 \\
LLaVA-v1.5-13B-XTuner & 51.1 & 43.0 & 42.0 & 29.8 & 73.0 & 25.0 & 17.0 & 52.0 & 22.0 \\
LlaVA-InternLM2-7B & 50.8 & 21.0 & 37.0 & 36.9 & 75.0 & 33.0 & 7.5 & 62.5 & 49.0 \\
LLaVA-v1.5-7B-Xtuner & 50.2 & 38.5 & 43.0 & 47.5 & 74.0 & 26.0 & 26.0 & 54.5 & 28.0 \\
SharedCaptioner & 49.9 & 24.5 & 31.5 & 27.7 & 75.0 & 26.0 & 16.5 & 64.5 & 51.0 \\
LLaVA-InternLM-7b & 49.7 & 35.0 & 49.5 & 29.8 & 72.0 & 27.0 & 16.5 & 53.0 & 61.0 \\
LLaVA-v1.5-7B & 49.5 & 38.5 & 34.0 & 26.2 & 68.0 & 28.0 & 21.5 & 45.5 & 29.5 \\
LLaMA-Adapter-v2-7B & 40.4 & 32.5 & 27.5 & 34.0 & 44.5 & 30.0 & 30.0 & 45.5 & 39.0 \\
VisualGLM\_6b & 38.6 & 27.0 & 31.0 & 32.6 & 52.5 & 26.5 & 24.5 & 45.0 & 26.5 \\
Frequency & 31.7 & 30.0 & 31.5 & 28.4 & 26.5 & 27.0 & 33.0 & 50.0 & 43.0 \\
Random & 28.5 & 22.0 & 27.5 & 28.4 & 25.5 & 33.0 & 23.0 & 57.0 & 40.0 \\
\bottomrule
\end{tabular}%
}
\end{table}

% =====================================================================================================================

% \usepackage{graphicx}
\begin{table}[]
\centering
\caption{Detail results of 30 LVLMs on \textbf{3D} and \textbf{Medical Understanding}.}
\label{tab:detailed_result_17}
\resizebox{\textwidth}{!}{%
\begin{tabular}{l|c|cc|ccccc}
\toprule
 &  & \multicolumn{2}{c|}{3D} & \multicolumn{5}{c}{Medical Understanding} \\
 \midrule
Model & Overall & \begin{tabular}[c]{@{}c@{}}3D CAD \\ Recognition\end{tabular} & \begin{tabular}[c]{@{}c@{}}3D Indoor\\ Recognition\end{tabular} & \begin{tabular}[c]{@{}c@{}}Anatomy\\ Identification\end{tabular} & \begin{tabular}[c]{@{}c@{}}Medical Modality\\ Recognition\end{tabular} & \begin{tabular}[c]{@{}c@{}}Other Biological\\ Attributes\end{tabular} & \begin{tabular}[c]{@{}c@{}}Disease\\ Diagnose\end{tabular} & \begin{tabular}[c]{@{}c@{}}Lesion\\ Grading\end{tabular} \\
\midrule
InternVL-Chat-V1.2-34B & 63.4 & 56.0 & 35.0 & 54.5 & 93.5 & 70.0 & 89.5 & 51.5 \\
Qwen-VL-Plus & 62.3 & 52.5 & 40.5 & 66.5 & 95.0 & 58.0 & 82.0 & 65.0 \\
GPT-4V & 62.0 & 51.0 & 33.0 & 57.0 & 99.0 & 75.0 & 71.0 & 67.9 \\
GeminiProVision & 61.6 & 61.0 & 29.5 & 86.5 & 96.5 & 80.0 & 74.5 & 75.5 \\
LLaVA-Next-34B & 60.8 & 52.0 & 45.0 & 41.0 & 67.0 & 65.0 & 85.5 & 51.0 \\
XComposer2-7B & 55.7 & 41.0 & 40.0 & 42.5 & 55.0 & 66.5 & 45.5 & 58.0 \\
BLIP2-Flan-T5-XXL & 54.8 & 55.0 & 40.5 & 51.0 & 72.5 & 64.5 & 61.5 & 51.0 \\
Yi-VL-34B & 54.2 & 46.0 & 31.5 & 48.5 & 49.5 & 65.0 & 73.5 & 61.5 \\
Monkey & 53.4 & 56.5 & 36.5 & 51.5 & 97.0 & 63.5 & 73.0 & 55.0 \\
DeepSeek-VL-7B & 53.2 & 52.5 & 42.0 & 51.0 & 96.0 & 71.5 & 79.5 & 50.5 \\
Yi-VL-6B & 53.2 & 42.0 & 45.0 & 54.0 & 71.0 & 59.5 & 62.0 & 56.5 \\
LLaVA-Next-13B & 53.0 & 50.5 & 49.5 & 42.0 & 54.0 & 76.0 & 58.5 & 48.5 \\
TransCore-M & 52.7 & 44.0 & 46.0 & 40.5 & 73.0 & 71.0 & 72.5 & 48.5 \\
QWen-VL-Chat & 52.5 & 55.0 & 35.0 & 59.0 & 93.5 & 56.5 & 70.0 & 56.5 \\
Claude3V-Haiku & 52.2 & 45.0 & 31.5 & 53.0 & 67.5 & 60.0 & 76.0 & 57.0 \\
XComposer & 52.1 & 52.0 & 42.0 & 49.0 & 72.0 & 72.5 & 41.5 & 43.0 \\
mPLUG-Owl2 & 52.0 & 49.0 & 41.5 & 56.0 & 97.5 & 58.5 & 55.5 & 49.5 \\
RBDash-v1-13B & 51.8 & 39.5 & 42.0 & 49.0 & 80.0 & 65.0 & 59.5 & 43.0 \\
LLaVA-v1.5-13B & 51.7 & 47.5 & 32.0 & 41.0 & 72.5 & 67.0 & 64.5 & 48.0 \\
CogVLM-Chat & 51.6 & 55.0 & 29.5 & 50.0 & 94.0 & 66.5 & 72.5 & 46.5 \\
ShareGPT4V-7B & 51.5 & 43.5 & 47.0 & 46.0 & 82.0 & 65.5 & 61.0 & 49.5 \\
LLaVA-Next-7B & 51.1 & 44.5 & 42.0 & 47.0 & 76.0 & 60.0 & 50.5 & 44.0 \\
LLaVA-v1.5-13B-XTuner & 51.1 & 46.0 & 49.0 & 34.0 & 50.0 & 66.0 & 61.0 & 34.5 \\
LlaVA-InternLM2-7B & 50.8 & 50.0 & 49.0 & 42.0 & 65.5 & 60.0 & 77.0 & 44.5 \\
LLaVA-v1.5-7B-Xtuner & 50.2 & 46.0 & 45.0 & 37.5 & 69.0 & 49.0 & 59.0 & 48.0 \\
SharedCaptioner & 49.9 & 45.0 & 43.5 & 55.0 & 72.0 & 73.0 & 50.5 & 46.5 \\
LLaVA-InternLM-7b & 49.7 & 54.0 & 30.0 & 37.5 & 63.5 & 59.0 & 51.5 & 48.5 \\
LLaVA-v1.5-7B & 49.5 & 46.0 & 43.5 & 34.5 & 79.5 & 60.0 & 63.0 & 47.0 \\
LLaMA-Adapter-v2-7B & 40.4 & 38.5 & 29.0 & 45.0 & 55.5 & 51.5 & 64.0 & 44.0 \\
VisualGLM\_6b & 38.6 & 44.0 & 29.5 & 24.0 & 57.5 & 47.5 & 49.5 & 41.5 \\
Frequency & 31.7 & 26.0 & 27.0 & 27.5 & 30.0 & 40.5 & 26.5 & 30.0 \\
Random & 28.5 & 24.0 & 27.0 & 25.0 & 24.0 & 38.0 & 24.0 & 28.0 \\
\bottomrule
\end{tabular}%
}
\end{table}

% =====================================================================================================================

% \usepackage{graphicx}
\begin{table}[]
\centering
\caption{Detail results of 30 LVLMs on \textbf{Cross Image Matching} and \textbf{Visual Summary} (part 1).}
\label{tab:detailed_result_18}
\resizebox{\textwidth}{!}{%
\begin{tabular}{l|c|ccc|ccccc}
\toprule
 &  & \multicolumn{3}{c|}{Cross Image Matching} & \multicolumn{5}{c}{Visual Summary} \\
 \midrule
Model & Overall & \begin{tabular}[c]{@{}c@{}}One-shot\\ Detection\end{tabular} & \begin{tabular}[c]{@{}c@{}}Point\\ Tracking\end{tabular} & \begin{tabular}[c]{@{}c@{}}Single Object\\ Tracking\end{tabular} & \begin{tabular}[c]{@{}c@{}}Video\\ Captioning\end{tabular} & \begin{tabular}[c]{@{}c@{}}Image Captioning\\ Paragraph\end{tabular} & \begin{tabular}[c]{@{}c@{}}Image\\ Captioning\end{tabular} & \begin{tabular}[c]{@{}c@{}}Instance\\ Captioning\end{tabular} & \begin{tabular}[c]{@{}c@{}}Image Dense\\ Captioning\end{tabular} \\
\midrule
InternVL-Chat-V1.2-34B & 63.4 & 59.0 & 58.5 & 53.0 & 69.5 & 99.0 & 96.5 & 90.0 & 52.8 \\
Qwen-VL-Plus & 62.3 & 75.0 & 63.0 & 46.0 & 92.5 & 99.0 & 98.0 & 87.0 & 62.9 \\
GPT-4V & 62.0 & 58.5 & 43.0 & 64.0 & 78.0 & 98.0 & 95.0 & 80.0 & 42.1 \\
GeminiProVision & 61.6 & 42.5 & 14.0 & 43.5 & 89.0 & 99.5 & 97.5 & 85.0 & 42.1 \\
LLaVA-Next-34B & 60.8 & 62.0 & 12.0 & 55.0 & 87.0 & 99.0 & 98.0 & 89.5 & 61.9 \\
XComposer2-7B & 55.7 & 49.5 & 75.5 & 58.5 & 49.0 & 99.0 & 97.5 & 80.0 & 52.8 \\
BLIP2-Flan-T5-XXL & 54.8 & 85.0 & 71.5 & 62.5 & 89.0 & 96.0 & 96.0 & 66.0 & 52.8 \\
Yi-VL-34B & 54.2 & 60.0 & 32.5 & 51.5 & 44.5 & 98.5 & 94.5 & 78.0 & 49.7 \\
Monkey & 53.4 & 72.0 & 63.5 & 49.5 & 81.0 & 71.0 & 95.5 & 70.0 & 33.5 \\
DeepSeek-VL-7B & 53.2 & 72.5 & 72.5 & 56.5 & 42.0 & 94.0 & 97.5 & 77.5 & 32.0 \\
Yi-VL-6B & 53.2 & 43.0 & 45.5 & 41.5 & 36.5 & 93.5 & 91.5 & 65.5 & 40.1 \\
LLaVA-Next-13B & 53.0 & 63.0 & 69.5 & 47.0 & 80.0 & 96.0 & 95.0 & 79.5 & 25.9 \\
TransCore-M & 52.7 & 66.0 & 80.0 & 51.5 & 91.0 & 96.5 & 95.0 & 82.5 & 26.9 \\
QWen-VL-Chat & 52.5 & 68.5 & 61.5 & 45.0 & 85.5 & 88.5 & 92.5 & 73.0 & 28.9 \\
Claude3V-Haiku & 52.2 & 60.5 & 52.0 & 45.0 & 53.0 & 95.5 & 92.5 & 62.0 & 50.8 \\
XComposer & 52.1 & 32.0 & 80.5 & 44.5 & 87.5 & 87.5 & 92.0 & 59.5 & 43.1 \\
mPLUG-Owl2 & 52.0 & 54.5 & 67.5 & 51.0 & 85.5 & 77.5 & 92.0 & 69.5 & 39.1 \\
RBDash-v1-13B & 51.8 & 67.0 & 75.0 & 50.5 & 60.0 & 98.5 & 97.5 & 77.5 & 45.7 \\
LLaVA-v1.5-13B & 51.7 & 62.0 & 73.5 & 52.0 & 51.5 & 95.5 & 97.5 & 78.5 & 25.9 \\
CogVLM-Chat & 51.6 & 60.0 & 40.5 & 37.0 & 61.5 & 91.5 & 95.5 & 72.0 & 18.3 \\
ShareGPT4V-7B & 51.5 & 61.5 & 80.5 & 51.0 & 86.0 & 90.0 & 96.0 & 76.5 & 31.5 \\
LLaVA-Next-7B & 51.1 & 58.0 & 79.5 & 66.0 & 82.5 & 82.5 & 96.0 & 76.5 & 22.3 \\
LLaVA-v1.5-13B-XTuner & 51.1 & 60.0 & 58.0 & 52.5 & 57.5 & 95.0 & 95.0 & 76.5 & 32.5 \\
LlaVA-InternLM2-7B & 50.8 & 44.0 & 71.5 & 59.0 & 39.0 & 93.0 & 94.0 & 77.5 & 24.9 \\
LLaVA-v1.5-7B-Xtuner & 50.2 & 58.5 & 72.5 & 58.5 & 47.5 & 90.0 & 95.5 & 77.0 & 35.0 \\
SharedCaptioner & 49.9 & 34.5 & 78.5 & 57.5 & 58.0 & 65.5 & 92.0 & 59.5 & 28.9 \\
LLaVA-InternLM-7b & 49.7 & 50.0 & 81.5 & 47.5 & 48.5 & 86.0 & 91.5 & 71.5 & 39.1 \\
LLaVA-v1.5-7B & 49.5 & 63.5 & 72.5 & 51.0 & 84.5 & 73.0 & 96.5 & 69.0 & 23.9 \\
LLaMA-Adapter-v2-7B & 40.4 & 39.0 & 33.5 & 35.0 & 23.0 & 29.0 & 71.0 & 34.5 & 31.0 \\
VisualGLM\_6b & 38.6 & 49.5 & 51.5 & 43.5 & 28.5 & 45.0 & 56.0 & 36.0 & 28.9 \\
Frequency & 31.7 & 27.0 & 30.5 & 28.0 & 27.5 & 27.0 & 31.5 & 27.0 & 28.4 \\
Random & 28.5 & 22.5 & 25.0 & 33.0 & 24.0 & 22.5 & 26.5 & 26.5 & 25.4 \\
\bottomrule
\end{tabular}%
}
\end{table}

% =====================================================================================================================

% \usepackage{graphicx}
\begin{table}[]
\centering
\caption{Detail results of 30 LVLMs on \textbf{Visual Summary} (part 2) and \textbf{Autonomous Driving}.}
\label{tab:detailed_result_19}
\resizebox{\textwidth}{!}{%
\begin{tabular}{l|c|ccc|ccccc}
\toprule
 &  & \multicolumn{3}{c|}{Visual Summary} & \multicolumn{5}{c}{Autonomous Driving} \\
 \midrule
Model & Overall & \begin{tabular}[c]{@{}c@{}}Multiple Instance\\ Captioning\end{tabular} & \begin{tabular}[c]{@{}c@{}}Multiple Image\\ Captioning\end{tabular} & \begin{tabular}[c]{@{}c@{}}Writing Poetry\\ from Image\end{tabular} & \begin{tabular}[c]{@{}c@{}}Traffic Participants\\ Understanding\end{tabular} & \begin{tabular}[c]{@{}c@{}}Multiple-view\\ Image Understanding\end{tabular} & \begin{tabular}[c]{@{}c@{}}Traffic Sign\\ Understanding\end{tabular} & \begin{tabular}[c]{@{}c@{}}Temporal Sequence\\ Understanding\end{tabular} & \begin{tabular}[c]{@{}c@{}}Traffic Light\\ Understanding\end{tabular} \\
\midrule
InternVL-Chat-V1.2-34B & 63.4 & 91.0 & 90.0 & 70.0 & 57.5 & 23.0 & 67.5 & 52.5 & 88.5 \\
Qwen-VL-Plus & 62.3 & 92.0 & 90.5 & 70.5 & 60.0 & 26.0 & 76.5 & 50.0 & 85.0 \\
GPT-4V & 62.0 & 90.5 & 88.0 & 71.0 & 61.0 & 26.2 & 81.0 & 53.0 & 56.7 \\
GeminiProVision & 61.6 & 88.0 & 79.0 & 61.0 & 62.5 & 51.0 & 74.5 & 53.0 & 56.7 \\
LLaVA-Next-34B & 60.8 & 87.5 & 92.0 & 72.0 & 57.0 & 24.0 & 71.5 & 51.0 & 72.0 \\
XComposer2-7B & 55.7 & 92.0 & 77.0 & 42.0 & 57.5 & 15.0 & 70.5 & 49.5 & 51.5 \\
BLIP2-Flan-T5-XXL & 54.8 & 89.5 & 78.0 & 62.0 & 49.5 & 15.5 & 62.5 & 43.5 & 51.8 \\
Yi-VL-34B & 54.2 & 82.5 & 83.5 & 61.0 & 47.5 & 23.0 & 61.0 & 48.5 & 61.0 \\
Monkey & 53.4 & 90.0 & 66.5 & 43.5 & 44.0 & 23.0 & 70.0 & 34.0 & 47.1 \\
DeepSeek-VL-7B & 53.2 & 87.0 & 73.0 & 50.0 & 50.0 & 18.5 & 62.5 & 37.5 & 75.5 \\
Yi-VL-6B & 53.2 & 79.0 & 68.0 & 33.0 & 49.0 & 26.0 & 55.5 & 39.0 & 65.0 \\
LLaVA-Next-13B & 53.0 & 86.5 & 85.5 & 52.0 & 53.0 & 21.5 & 70.5 & 48.0 & 85.5 \\
TransCore-M & 52.7 & 83.5 & 75.0 & 54.5 & 53.0 & 17.0 & 62.0 & 50.5 & 50.2 \\
QWen-VL-Chat & 52.5 & 91.5 & 75.5 & 48.5 & 43.5 & 36.0 & 66.0 & 34.0 & 47.5 \\
Claude3V-Haiku & 52.2 & 65.5 & 65.0 & 56.5 & 53.0 & 0.0 & 66.0 & 0.0 & 51.5 \\
XComposer & 52.1 & 86.0 & 71.0 & 50.5 & 45.5 & 24.0 & 59.5 & 25.5 & 49.4 \\
mPLUG-Owl2 & 52.0 & 82.5 & 41.5 & 45.0 & 51.0 & 18.5 & 67.5 & 44.0 & 46.3 \\
RBDash-v1-13B & 51.8 & 85.5 & 78.5 & 60.5 & 50.0 & 22.5 & 62.5 & 48.5 & 47.8 \\
LLaVA-v1.5-13B & 51.7 & 87.5 & 70.5 & 56.0 & 51.5 & 25.5 & 62.0 & 47.0 & 46.9 \\
CogVLM-Chat & 51.6 & 89.5 & 88.0 & 42.0 & 46.5 & 32.5 & 68.0 & 30.5 & 47.2 \\
ShareGPT4V-7B & 51.5 & 80.5 & 61.0 & 51.0 & 54.5 & 28.0 & 64.0 & 37.0 & 47.2 \\
LLaVA-Next-7B & 51.1 & 79.0 & 73.5 & 46.0 & 55.0 & 18.5 & 66.5 & 49.0 & 78.5 \\
LLaVA-v1.5-13B-XTuner & 51.1 & 82.0 & 64.5 & 47.0 & 53.0 & 20.5 & 59.5 & 43.0 & 45.9 \\
LlaVA-InternLM2-7B & 50.8 & 84.0 & 67.0 & 52.0 & 53.0 & 12.5 & 64.0 & 48.0 & 48.0 \\
LLaVA-v1.5-7B-Xtuner & 50.2 & 83.5 & 51.0 & 51.5 & 52.5 & 11.0 & 62.0 & 46.0 & 45.3 \\
SharedCaptioner & 49.9 & 79.0 & 65.5 & 47.0 & 54.0 & 18.5 & 60.5 & 46.0 & 44.6 \\
LLaVA-InternLM-7b & 49.7 & 81.0 & 61.5 & 40.5 & 51.5 & 9.5 & 58.0 & 46.5 & 45.7 \\
LLaVA-v1.5-7B & 49.5 & 83.5 & 66.5 & 48.0 & 54.0 & 15.5 & 64.5 & 45.5 & 45.9 \\
LLaMA-Adapter-v2-7B & 40.4 & 33.0 & 28.5 & 28.5 & 16.0 & 26.5 & 43.0 & 27.5 & 32.6 \\
VisualGLM\_6b & 38.6 & 39.5 & 60.0 & 30.0 & 28.0 & 25.0 & 25.5 & 20.0 & 32.3 \\
Frequency & 31.7 & 29.5 & 29.0 & 26.0 & 27.0 & 28.0 & 31.0 & 30.0 & 32.5 \\
Random & 28.5 & 25.5 & 24.0 & 25.5 & 25.0 & 27.0 & 29.0 & 23.0 & 29.8 \\
\bottomrule
\end{tabular}%
}
\end{table}

% =====================================================================================================================